\documentclass{article}
\usepackage{microtype,hyperref,url,booktabs,graphicx,amsmath,amssymb,mathtools}
\usepackage{array,tabularx,longtable,multirow,makecell,colortbl,xspace,enumitem,pifont}
\usepackage{tikz,pgfplots,iftex}
\usepackage{placeins}
\usetikzlibrary{arrows.meta,backgrounds,calc,fit,positioning,shapes.geometric}
\pgfplotsset{compat=1.18}
\usepackage{nexcua-tiffany}
\setcitestyle{citesep={,}}
\ifXeTeX
  \usepackage{xeCJK}
  \setCJKsansfont{FandolHei-Regular.otf}[BoldFont=FandolHei-Bold.otf]
\fi
\usepackage{needspace}
\newcommand{\sectionfloatboundary}{\FloatBarrier}
\pretocmd{\section}{\sectionfloatboundary\Needspace{7\baselineskip}}{}{}
\apptocmd{\thebibliography}{\interlinepenalty=10000}{}{}

\newcolumntype{P}[1]{>{\raggedright\arraybackslash}p{#1}}
\newcolumntype{C}[1]{>{\centering\arraybackslash}p{#1}}
\ifXeTeX\fi
\newcommand{\name}{\textsf{NorthCUA}\xspace}
\newcommand{\northcua}{\textsf{NorthCUA}\xspace}

\newcommand{\casespec}{\textsf{CaseSpec}\xspace}
\newcommand{\experienceflywheel}{\textsf{Experience Flywheel}\xspace}

\newcommand{\paperfigure}[3]{\begin{figure}[htbp]\centering\includegraphics[width=\linewidth]{figures/generated/#1.pdf}\caption{#3}\label{#2}\end{figure}}
\title{\textcolor{TiffanyDeep}{NorthCUA} Technical Report}
\subtitle{\fontsize{13.5}{19}\selectfont Turbocharging Computer-Use Agents through Experience at Scale}
\slogan{Worlds in mind. Wonders at your fingertips.}
\hypersetup{pdftitle={NorthCUA Technical Report: Turbocharging Computer-Use Agents through Experience at Scale},pdfauthor={NorthCUA Team}}
\author{{\ifPDFTeX\bfseries\else\addfontfeatures{FakeBold=0.5}\fi NorthCUA Team}{\color{NexSubtitle}, Shanghai Qiji Zhifeng Co., Ltd.}}
\begin{document}
\begingroup
\setlength{\parskip}{2pt}
\patchcmd{\nexliquidtitle}{\vspace{15pt}}{\vspace{3pt}}{}{}
\patchcmd{\nexliquidtitle}{\vspace{11pt}}{\vspace{4pt}}{}{}
\patchcmd{\nexliquidtitle}{\vspace{13pt}}{\vspace{3pt}}{}{}
\patchcmd{\nexliquidtitle}{\vspace{13pt}}{\vspace{3pt}}{}{}
\patchcmd{\nexliquidtitle}{\vspace{12pt}}{\vspace{3pt}}{}{}
\patchcmd{\abstract}{\fontsize{10}{13}}{\fontsize{10}{12.7}}{}{}
\maketitle
\begin{abstract}
Scaling computer-use agents from individual actions to complex software work
requires experience that connects decisions to complete outcomes. We present
\name, an integrated system for turbocharging computer-use agents through
experience at scale. Its \experienceflywheel supplies diverse executable
tasks to the \northcua Harness, which sustains long interactions, records
decisions, and checks task outcomes. The Flywheel uses these records to
curate verified trajectories.
For unsolved tasks, recursive self-improvement evolves reusable skills and
preserves verified milestones so later attempts build on earlier progress.
Staged Learning combines this experience with grounding data from the
Flywheel to develop GUI competence through mid-training and long-horizon
supervised fine-tuning.
The policy then learns from fresh attempts in the same Harness through joint
desktop--web reinforcement learning, closing the loop between experience
production and policy improvement. The resulting Nano, Mini, and Pro family
achieves state-of-the-art results among open-source models across GUI
grounding, desktop execution, and Web interaction under the reported
evaluation settings (Figure~\ref{fig:capability-overview}). For more responsive
Web workflows, \name-Ultrafast pairs direct action selection with open-ended
generation on pre-RL Nano, offering a measured quality--latency tradeoff.
Together, these advances establish \name as a high-performing model family
and a reusable foundation for agents that carry complex software work from
intent to completion.
\end{abstract}
\begingroup
\enlargethispage{8pt}
\sffamily\fontsize{9.2}{13.4}\selectfont
\renewcommand{\arraystretch}{1.06}
\definecolor{ResourceTeal}{HTML}{00838A}
\newcommand{\resourceicon}[2]{\raisebox{-2pt}[0pt][0pt]{\includegraphics[width=#2,height=#2,keepaspectratio]{assets/resource-icons/#1.pdf}}}
\newcommand{\resourcelabel}[1]{\textcolor{NexInk}{\bfseries #1}}
\newcommand{\resourcelink}[3]{\href{#1}{\textcolor{ResourceTeal}{#2#3}}}
\noindent\hspace*{11pt}\begin{tabular}{@{}>{\centering\arraybackslash}p{10.5pt}@{\hspace{7.5pt}}p{35pt}@{\hspace{10pt}}l@{}}
\resourceicon{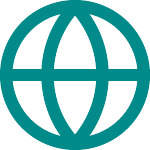}{10pt} & \resourcelabel{Page} & \resourcelink{https://beipo.top/blog/northcua/}{beipo.top}{/blog/northcua/}\\
\resourceicon{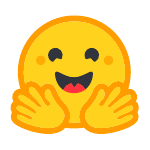}{10.8pt} & \resourcelabel{Models} & \resourcelink{https://huggingface.co/beipo-ai/NorthCUA-Pro}{huggingface.co}{/beipo-ai/NorthCUA-Pro}\\
\resourceicon{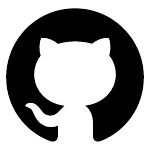}{9.7pt} & \resourcelabel{Code} & \resourcelink{https://github.com/beipo-ai/NorthCUA}{github.com}{/beipo-ai/NorthCUA}
\end{tabular}\par
\endgroup
\setlength{\textfloatsep}{6pt}
\setlength{\intextsep}{13pt}
\begin{figure}[!ht]
\centering
\captionsetup{skip=6pt}
\includegraphics[width=\linewidth]{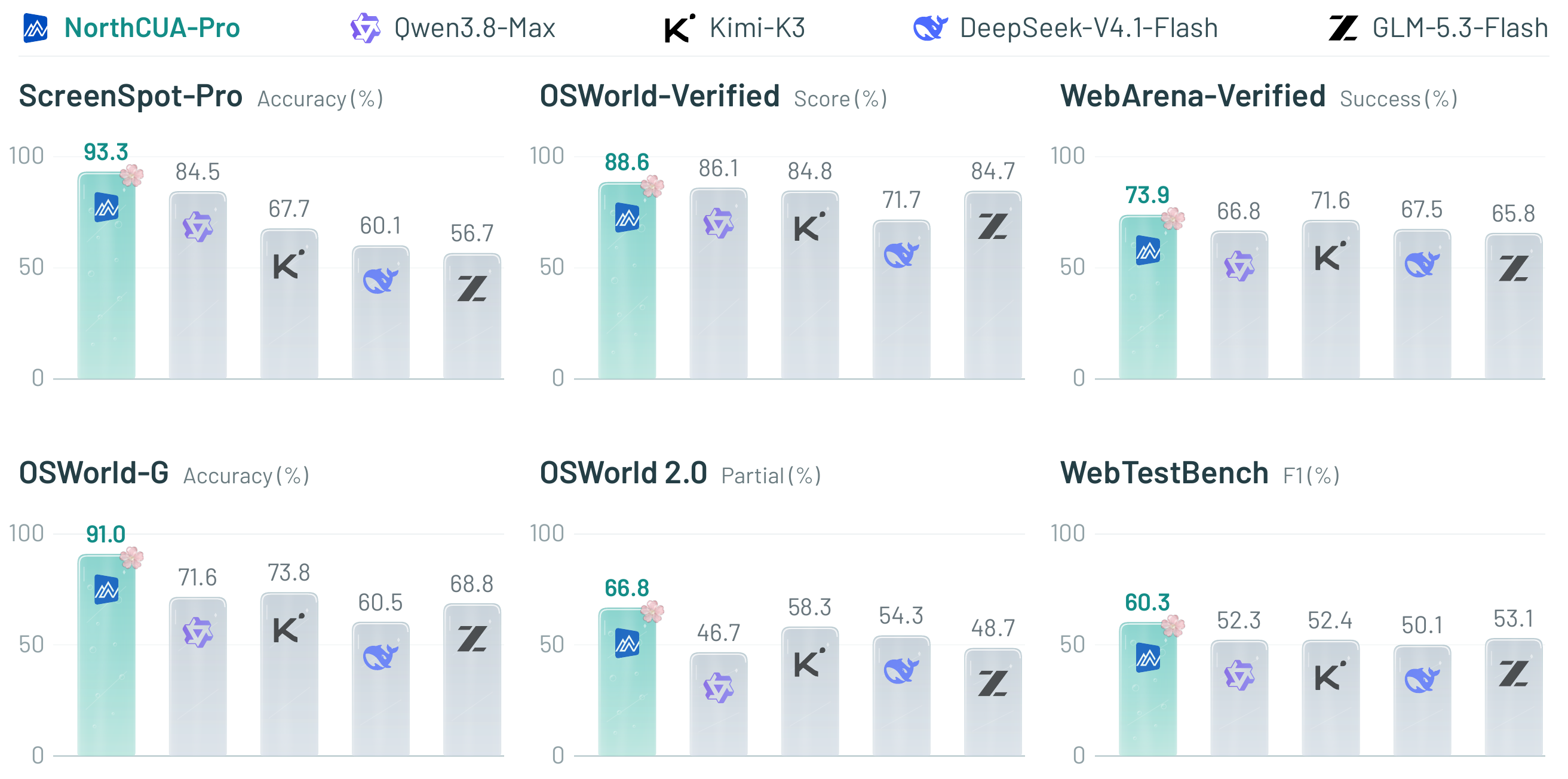}
\caption{\textbf{NorthCUA-Pro performance compared with leading models.}}
\label{fig:capability-overview}
\end{figure}

\clearpage
\endgroup
\section{Introduction}
\label{sec:introduction}

Computer-use agents (CUAs) extend the capabilities of large language
models from generating responses to operating computers and automating
software tasks \citep{zhang2024large,wang2024large}.
They can invoke software functions through application programming
interfaces (APIs) and command-line interfaces (CLIs).
Graphical user interfaces (GUIs) extend this access to functions that
applications do not expose programmatically, providing a critical last
mile for software automation \citep{zhang2025api,zhang2025ufo}.
Through GUI interaction, agents interpret the screen and translate
natural-language goals into mouse and keyboard actions.
Advances in visual grounding and action generation have expanded the range of operations
these agents can perform \citep{qin2025uitars,wang2025opencua,zhou2025maiui}.
As agents combine these operations into longer workflows, earlier actions
shape the software state on which later decisions depend. Reliable completion
requires agents to preserve intermediate progress and the user's
constraints, adapt to changing interfaces, and recover from errors while
keeping the overall goal in view \citep{zhang2025ufo2,lu2026qwencua}.

Training agents to coordinate such workflows therefore requires experience that
develops precise interface skills and connects successive decisions to
complete task outcomes. Obtaining this experience at scale begins with
constructing varied tasks, each of which needs a request grounded in available software and
content, a usable initial environment, and a reliable way to judge success
\citep{ge2026syntheticcomputers,zhou2026qwenuiagent}. Once these tasks are
available, collecting experience from them requires an agent framework that
sustains long interactions and records decisions and outcomes across desktop
and web environments. The resulting demonstrations can teach the policy how to
complete workflows, but independent execution may lead it into states
absent from those demonstrations. Continued improvement therefore requires
fresh interaction and outcome feedback from the policy's own attempts
\citep{wang2025opencua,jung2026procuasft,lu2026qwencua}. Meeting these linked
needs calls for co-designing how tasks are constructed, how agents execute
them, and how the resulting experience is used for learning.

\begin{figure}[!t]
\centering
\includegraphics[width=\linewidth]{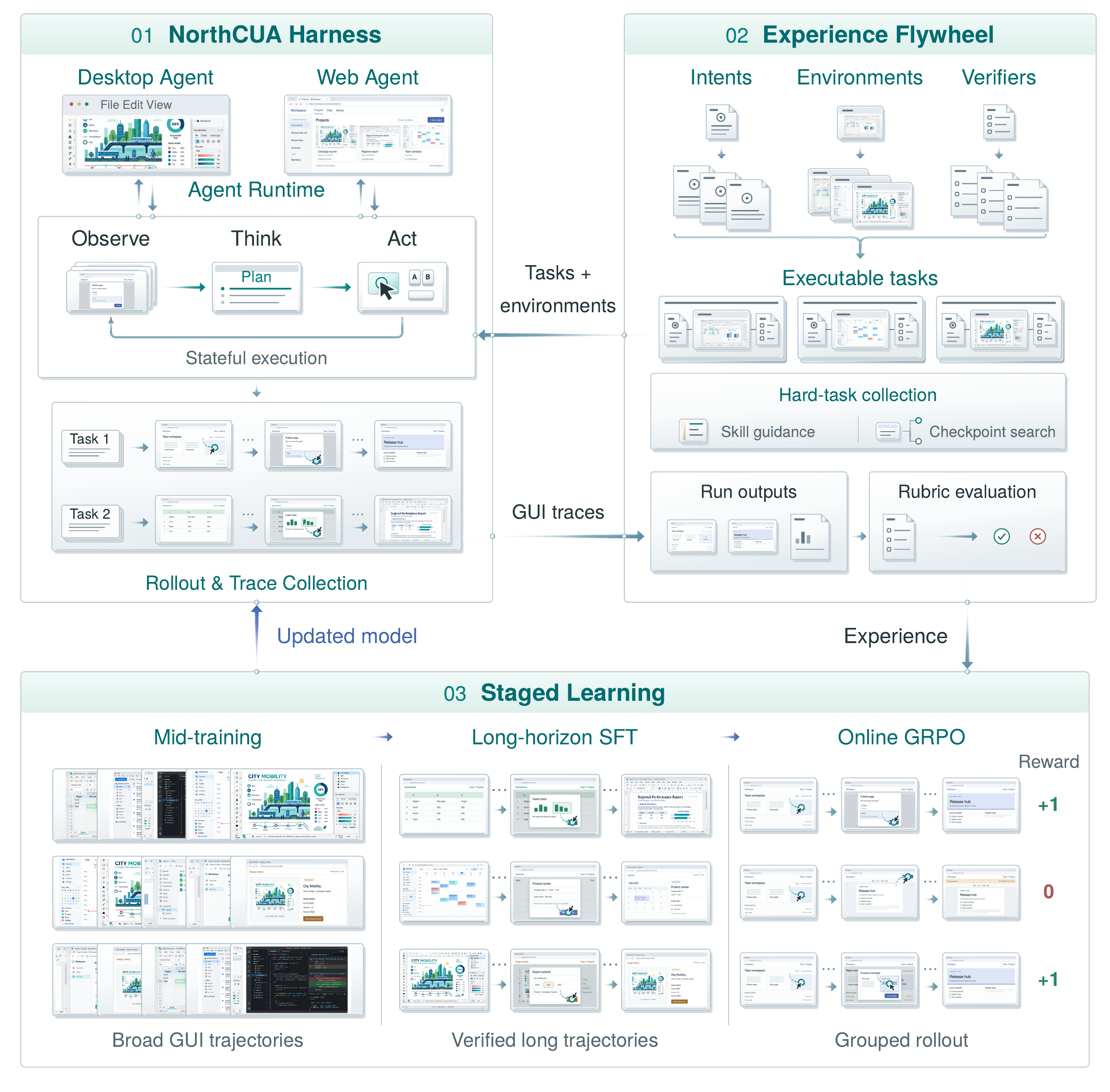}
\caption{\textbf{The NorthCUA experience loop.}
\textbf{01} The harness executes desktop and web tasks, records their
trajectories, and retains workspaces through required live inspection.
\textbf{02} The Experience Flywheel expands intents, environments, and verifiers
into executable tasks, then curates assessed trajectories.
The hard-task panel denotes skill-guided and checkpoint-assisted
collection through the harness (Section~\ref{sec:hardexperience}).
\textbf{03} Staged Learning converts complementary experience into an updated
policy for subsequent execution. GUI states, cursor cues, and rewards are
schematic. Trajectory rows omit intermediate actions, and GRPO retains valid
failures alongside successes.}
\label{fig:concept}
\end{figure}

We present \name, an integrated system for \emph{turbocharging computer-use
agents through experience at scale}. Our central idea is to co-design the
production and use of experience, making software interaction a continuing
source of learning. We connect executable tasks to policy
updates through three components (Figure~\ref{fig:concept}), focusing on
GUI-based software operation. The \textbf{\mbox{\experienceflywheel}}
constructs executable tasks for the \textbf{\mbox{\northcua Harness}} to
run and assess. The recorded trajectories and outcome evidence guide
curation, supplying \textbf{\mbox{Staged Learning}} with demonstrations
that complement the Flywheel's separately constructed grounding examples.
This mixture develops interface understanding and workflow completion. Shared execution
and outcome interfaces also let the learner collect fresh attempts from
the policy itself, extending the same system from demonstration collection
to continued policy improvement.

\paragraph{NorthCUA Harness: full-lifecycle infrastructure.}
We design the \northcua Harness as \emph{full-lifecycle infrastructure} that
executes software tasks and turns the resulting attempts into inspectable
experience for learning. This requires keeping the agent's decisions
connected to its workspace and outcomes, from environment preparation
through execution and evaluation to record preservation.
Within the shared runtime, visual context and task memory help agents
build on progress across long workflows
\citep{zhang2025ufo2,zhang2025ufo3}. An~evaluator uses the execution record and
live workspace to check task completion and execution quality, supplementing task-specific verifiers where their
coverage is incomplete \citep{zhuge2024agentasajudgeevaluateagentsagents}.
To~scale~this~process, the Batch Rollout Service coordinates parallel
workers and overlaps lifecycle stages, supporting efficient,
high-concurrency experience collection (Section~\ref{sec:harness}).

\Needspace{5\baselineskip}
\paragraph{Experience Flywheel: scaling grounding and agentic experience.}
With a runtime for collecting and checking software experience, the next
challenge is to supply enough varied, useful work for the agent to learn.
The \experienceflywheel addresses two levels of supervision: single-step
grounding examples for locating controls and assessed agent trajectories
for completing tasks. Grounding pairs screenshots with queries and target
locations. Visual and structural annotations provide broad coverage
\citep{lu2024omniparser,li2025autogui}, while application documentation helps
explain a control's function when appearance alone is insufficient.
At the trajectory level, three complementary construction routes supply
tasks to the harness. \emph{Scaling Intents} derives requests from available
software and content, while \emph{Scaling Environments} builds the workspaces
needed by sourced requests or reconstructs reference websites.
For tasks with executable checks, \emph{Scaling Verifiers} creates further
tasks by adapting requests, initial states, and verifiers together
\citep{ge2026syntheticcomputers,lu2026qwencua,lv2026scalecua}.
When a valid task remains unsolved in ordinary collection, we use
recursive self-improvement (RSI) for hard-task collection.
Skill-guided exploration retains tested procedures for later attempts,
while milestone-guided search preserves software states and agent context
at verified milestones to explore alternative ways of completing the remaining work. Task outcomes and
execution quality determine which trajectories become demonstrations for
learning (Section~\ref{sec:data}).

\paragraph{Staged Learning: from experience to capability.}
A model's ability to coordinate a long workflow builds on familiarity with
the interfaces and operations it uses. Staged Learning follows this
progression, starting with \emph{Broad GUI mid-training} on grounding
examples and desktop and web trajectories judged to complete their tasks
\citep{wang2025opencua,jung2026procuasft}. The broad trajectory corpus spans
over four million GUI interaction steps across dozens of applications.
Building on this operational foundation, \emph{Verified long-horizon
supervised fine-tuning (SFT)}
concentrates on carefully selected, complete workflows whose outcomes pass
executable task verifiers and whose intermediate results support later decisions.
Independent execution can still lead the policy into states absent from
its demonstrations. \emph{Joint desktop--web agentic reinforcement learning}
therefore lets one policy practice in both domains and learn from fresh
successes and failures. Task verifiers score these attempts, and
group-relative policy optimization (GRPO, \citealp{shao2024deepseekmath})
uses their outcomes to update the policy (Section~\ref{sec:training}).

Shared task, execution, and assessment interfaces connect these learning
stages to experience collection (Figure~\ref{fig:concept}). An updated
policy can return to the same Harness, where its new attempts are evaluated
and can be curated into subsequent training experience. This return path
lets advances in model capability support further collection, giving
\name a reusable foundation for continued learning across desktop and web
environments.

The resulting \name family comprises Nano (9B), Mini (35B), and
Pro (397B), all built on Qwen3.5 \citep{qwen2026qwen35}.
We evaluate this family on six benchmarks spanning GUI grounding, desktop
execution, and Web interaction. The suite follows the agent from locating
controls to completing long workflows and checking whether Web applications
behave as intended. \name establishes state-of-the-art results among
open-source models in all three domains (Figure~\ref{fig:capability-overview}).
With 397B parameters, Pro outperforms trillion-parameter Qwen3.8-Max and
Kimi-K3 on both grounding benchmarks, OSWorld-Verified, and
WebArena-Verified \citep{qwen2026qwen38,kimiteam2026kimik3}.
The 35B Mini also surpasses Kimi-K3 on OSWorld-Verified and Qwen3.8-Max
on WebArena-Verified. Stage comparisons show desktop and Web gains across
all three model sizes after long-horizon SFT and joint RL
(Section~\ref{sec:experiments}).

Long workflows also accumulate the latency of successive model decisions.
To reduce this cost, \name-Ultrafast adapts pre-RL Nano with a
Jev-inspired choice head, selecting finite actions directly while retaining
generation for open-ended outputs \citep{almeida2026jev}. Its
Head + LoRA variant completes matched successful WebArena-Verified
tasks $2.3\times$ faster on average than full generation
(Section~\ref{sec:fastweb}). We release all three models, the full
\northcua Harness code, and this technical report, providing an open
foundation for developing agents that carry software work from user intent
to completed outcomes.

\section{Scope and System Overview}
\label{sec:overview}

\subsection{Task Scope and Characteristics}

\begin{figure}[!ht]
\centering
\includegraphics[width=\linewidth]{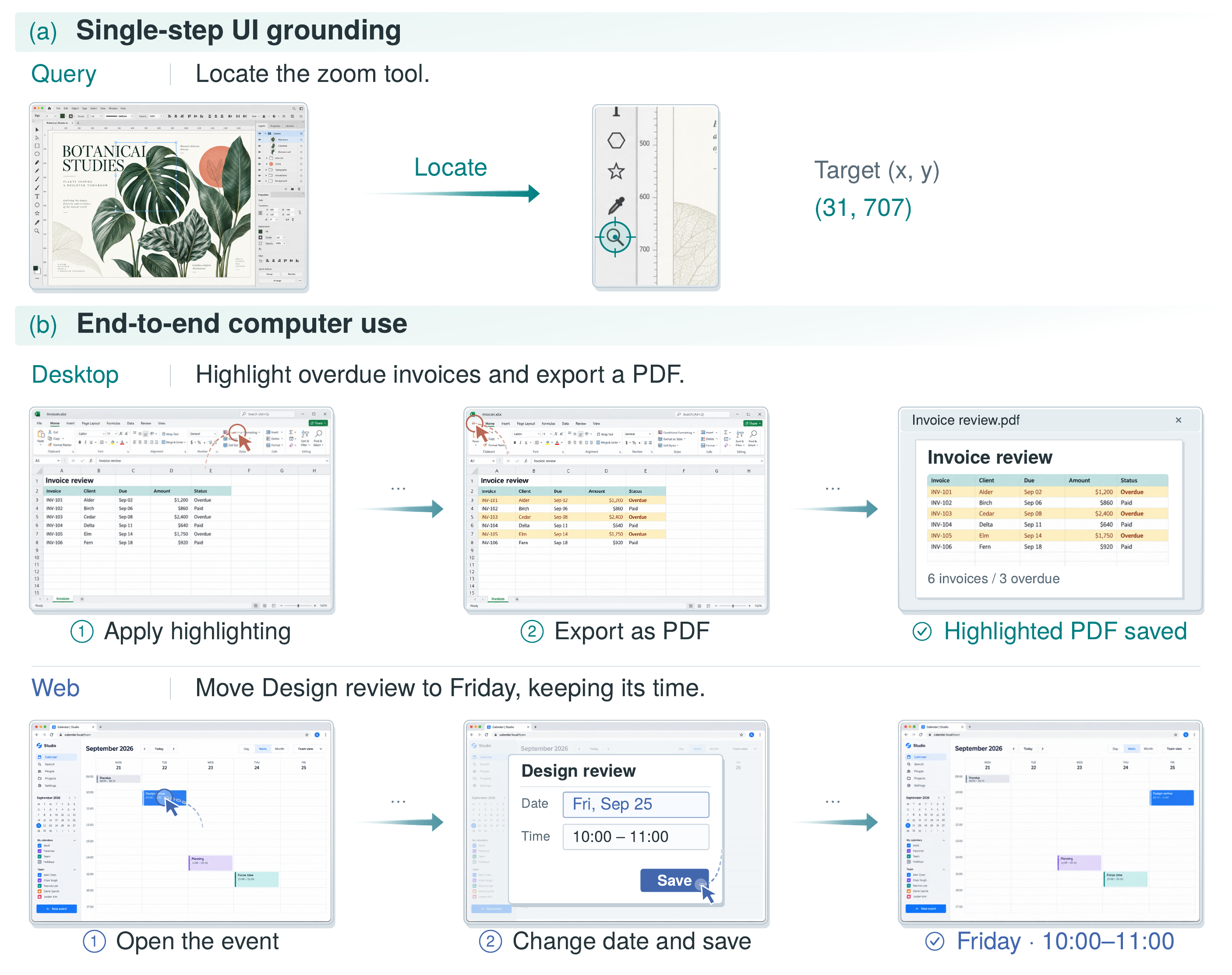}
\caption{\textbf{Task scope and interaction.}
(a) Single-step grounding predicts a target location in a fixed screenshot.
The enlarged output view marks the zoom tool, with absolute pixel coordinates
in the original image.
(b) Desktop and web agents change software state through successive
actions, with completion assessed from the resulting artifacts or state.
Web observations also include DOM information.
Generated GUI scenes, action overlays, and outcomes are illustrative.
The displayed transitions omit intermediate interactions.}
\label{fig:learning-targets}
\end{figure}

We study two capability targets in desktop and web environments
(Figure~\ref{fig:learning-targets}): locating a requested control in a
screenshot and completing a task through software interaction. We
evaluate these targets separately.

\paragraph{Single-step UI grounding.}
Given a screenshot and a natural-language query, the model predicts the
coordinates \((x,y)\) of the target UI element in the original image
\citep{cheng2024seeclick,li2025screenspotpro}. The model completes the task
with one prediction over a fixed screenshot. The query can describe the
element's appearance or function, such as locating the zoom tool in
Figure~\ref{fig:learning-targets}(a).

\Needspace{6\baselineskip}
\paragraph{End-to-end computer use.}
An end-to-end task combines a user request, an initial software environment,
and conditions for judging completion. The environment provides the
applications, files, and website records needed for the request. The agent
observes the current state, takes an action, and uses the resulting feedback
to decide what to do next. Later decisions can use information and results
from earlier steps. The request and completion conditions stay fixed
throughout the attempt, so the final files or software state can be checked
against the original goal \citep{zhou2023webarena,xie2024osworld}.
Figure~\ref{fig:learning-targets}(b) illustrates the two interaction settings:

\begin{itemize}
\item \textbf{Desktop agents} observe desktop screenshots and operate
applications through mouse and keyboard actions, including clicking, typing,
scrolling, and dragging. Tasks may span multiple applications and files.
\item \textbf{Web agents} observe webpage screenshots and document object
model (DOM) information, and interact through browser tools for navigation,
element-based or coordinate-based actions, text input, and tab management.
\end{itemize}
Both settings follow the same observation, action, and feedback loop,
using their respective desktop or browser interfaces. The acting policy
uses these interfaces to perform the task. Preparation, annotation, and
evaluation can use additional programmatic tools with separate permissions.

\subsection{System Organization and Experience Lifecycle}

The experience loop in Figure~\ref{fig:concept} connects task construction
to execution and learning. Its handoffs specify how an executable task
becomes an assessed record and then a training sample.

The \experienceflywheel packages each executable task as a
\mbox{\casespec}, a task specification
containing the request, input materials, environment configuration, and
setup instructions, together with requirements for assessing the result.
For the invoice task in Figure~\ref{fig:learning-targets}(b), this means
providing the workbook and spreadsheet application and specifying that
the exported PDF should highlight overdue invoices.

Using this specification, the \northcua Harness prepares the environment,
uploads input files, and runs setup and launch commands before starting
the agent on the request. Observations, actions, and feedback are recorded
throughout the attempt so that its outcome can be assessed in the context
of the work that produced it. Task-specific checks, an evaluator, or both
inspect the result against the declared requirements. For the invoice task,
this means checking the saved PDF and its highlighted contents. These
outcome and execution-quality assessments give the \experienceflywheel
the evidence needed to select demonstrations. Valid tasks that remain
unsolved can instead enter the hard-task collection branch, where
skill-guided exploration or milestone-guided search produces further
attempts for the same assessment process (Section~\ref{sec:hardexperience}).

Staged Learning turns this collected experience into capability in three
steps. Broad GUI mid-training combines diverse task trajectories with
the Flywheel's separate screenshot--query--target examples, building a
foundation in interface understanding and software operation. Grounding
supervision is used only in this stage. Verified long-horizon SFT then
focuses on high-quality, complete workflows with checked outcomes, teaching
the policy to carry information and intermediate results across dependent
steps. This prepares the model to learn further from its own interaction.

For joint desktop--web agentic reinforcement learning, we first select
tasks on which the SFT policy's trial attempts have different outcomes.
This focuses initial collection on tasks with useful outcome differences
for comparing executions. The Harness collects fresh groups of attempts, executable
verifiers score their outcomes, and GRPO uses comparisons within each task
group to update one shared policy (Section~\ref{sec:rl}). The updated policy
returns to the Harness for subsequent attempts under the same task and
outcome interfaces.

The following sections explain these components in detail.
Section~\ref{sec:harness} presents the execution and assessment framework,
Section~\ref{sec:data} describes task construction and experience curation,
and Section~\ref{sec:training} develops the staged learning recipe.

\providecommand{\harnessfigureplaceholder}[2]{%
  \begingroup
  \setlength{\fboxsep}{8pt}%
  \noindent\fbox{%
    \begin{minipage}{\dimexpr\linewidth-2\fboxsep-2\fboxrule\relax}
      \small\raggedright
      \textbf{Figure placeholder: #1}\par\medskip
      #2
    \end{minipage}%
  }%
  \endgroup
}

\section{NorthCUA Harness: Full-Lifecycle Infrastructure for Computer Use}
\label{sec:harness}

The \northcua Harness provides \emph{full-lifecycle infrastructure} for
executing desktop and web tasks and turning each attempt into inspectable
experience for learning. During execution, a persistent workspace and
retained task context let the agent build on earlier progress.
We retain its execution record and resulting artifacts for outcome
assessment, keeping the original environment available when live checks
need to inspect the state the agent produced. Resources are released after
the required inspection and record preservation are complete. By carrying
each attempt through assessment and preservation, this
lifecycle keeps decisions, outputs, and judgments connected to the same
task attempt (Section~\ref{sec:overview}).

\begin{figure}[!t]
\centering
\includegraphics[width=\linewidth]{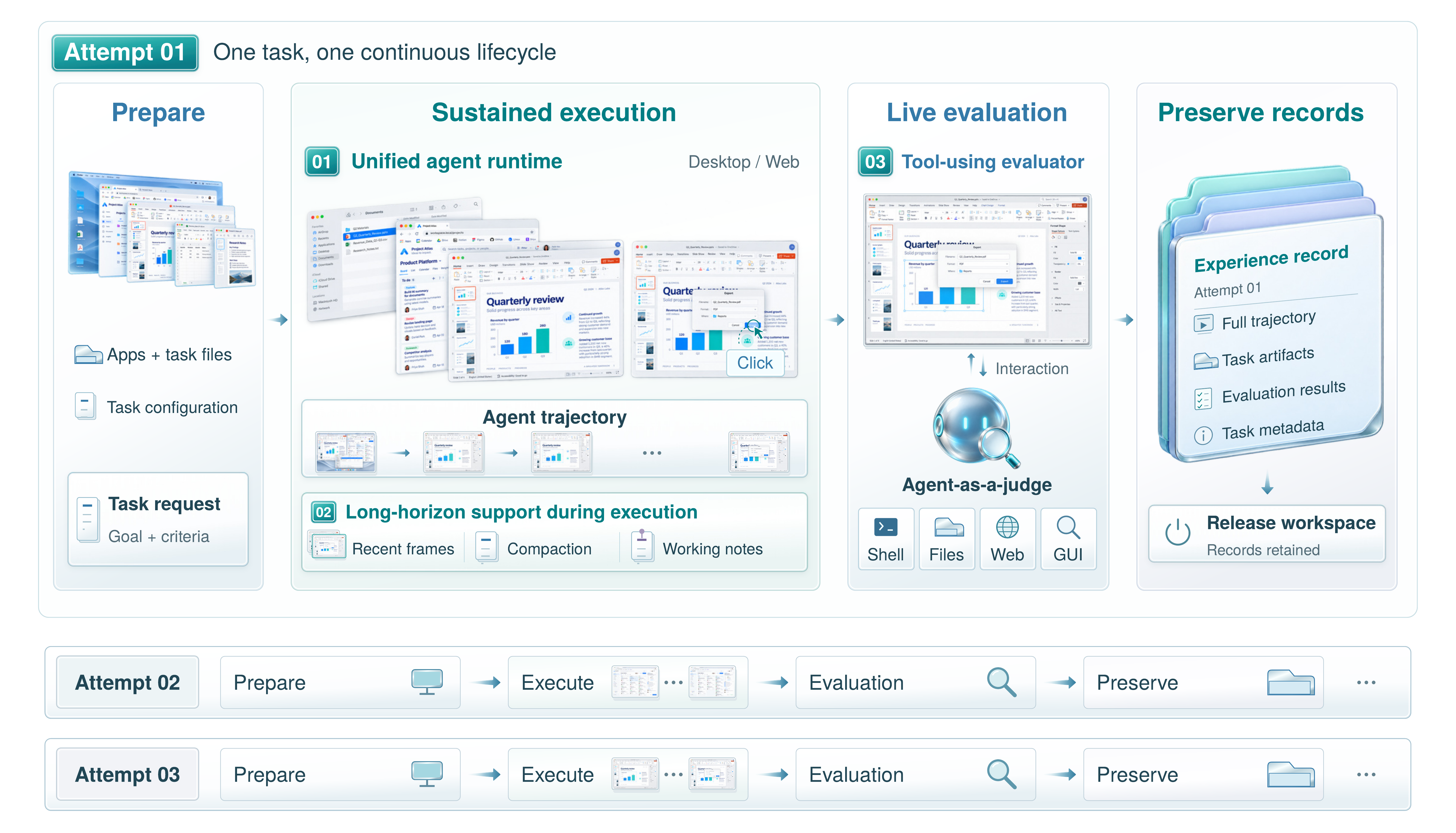}
\caption{\textbf{NorthCUA Harness lifecycle.}
Within the expanded attempt, the shared runtime and long-horizon support
sustain GUI interaction while retaining the full trajectory. The evaluator
inspects recorded behavior and the live environment. Trajectory, task artifacts,
evaluation results, and task metadata are saved in an experience record before
the workspace is released. The lower lanes show parallel instances of this
lifecycle, coordinated by the batch collection service. GUI scenes, agent
imagery, and trajectories are illustrative.}
\label{fig:harness}
\end{figure}

Figure~\ref{fig:harness} organizes this lifecycle at two levels. Within one
task attempt, the unified agent runtime (\textbf{01}) supplies the shared
interaction loop and platform tool interfaces. Long-horizon support
(\textbf{02}) operates within this execution, keeping the next decision
informed while retaining the complete trajectory. After acting ends, a
tool-using evaluator (\textbf{03}) checks the declared requirements using
that trajectory and the live workspace. The Batch Rollout Service
coordinates the parallel task lifecycles shown below. Each
isolated environment proceeds through preparation, execution, and assessment.
Centralized control and recording manage these attempts, keeping each workspace
available through inspection and output capture. This distinction separates
the mechanisms that make one attempt inspectable from the service that turns
them into sustained experience production. Online RL reuses the execution
and outcome interfaces under training-side lifecycle management
(Section~\ref{sec:rl}), extending the same foundation from standalone
collection to policy learning.

\Needspace{12\baselineskip}
\subsection{Unified Execution Across Desktop and Web}
\label{sec:harness-execution}

Desktop and web agents follow the same basic loop: observe the software,
choose an action, execute it, and use the feedback to continue. What differs
is how each platform exposes its state and controls. Building on shared-runtime
approaches to computer use \citep{zhang2025ufo2,yan2025stepgui}, we design
\emph{Agent Kit} to separate this common loop from platform-specific
observation and action tools. Long-horizon support and evaluation reuse this same
execution foundation.

\subsubsection{Shared Agent Runtime}
We use a shared runtime to organize interaction as a ReAct-style loop with
visual feedback \citep{yao2023react}. It
assembles the request, observation, and relevant history into the next model
input, invokes the policy, and executes the resulting tool calls. Returned
observations and action results then
inform the next decision. Common runtime services manage progress, budgets,
and termination, while environment-specific implementations supply the
observations and operations. To build an agent for another platform,
developers supply its perception functions, action tools, and observation
processing hooks. The shared runtime then carries these components through
the complete decision and recording loop, without rebuilding the agent
flow. Section~\ref{sec:platformagents} details the desktop and web instances.

CUA execution also involves several timescales. Application and browser
sessions persist across actions, a user request can require many decisions,
and each decision produces feedback needed by the next. We represent these
lifetimes through three nested scopes:
\begin{itemize}
\item \textbf{Session} maintains the agent's execution state and associated
software resources, such as a browser session, across a sequence of user
requests.
\item \textbf{Round} tracks the processing of one user request, including
its multi-step execution, budget, and termination status.
\item \textbf{Step} records one policy decision and the resulting tool
execution and feedback. Each step can contain multiple tool calls.
\end{itemize}
These runtime scopes organize interaction, while the task attempt supplies the
collection-level identity used to associate actions, artifacts, and
assessment. Each recorded action retains its request and session context,
supporting task-level inspection and collection recovery. Session resources
are managed within the broader environment lifecycle coordinated by the
collection or training system. Appendix~\ref{app:executioninterfaces}
gives the execution-scope details.

\subsubsection{Tool Integration and Dispatch}
Desktop input, browser interaction, and workspace inspection expose
different operations, whose implementations may run beside the agent or
near a remotely hosted application. To make these capabilities available
to the shared decision loop, we use a common tool interface that allows
their implementations to evolve independently.

Tool definitions describe operations and their arguments, while dispatch
connects each call to its implementation and returns the result to the
requesting step. The Agent Kit integrates local functions and tools exposed
through Model Context Protocol (MCP), whose standard tool descriptions and
invocation messages support capability discovery and execution
\citep{mcp2025tools}. Both paths feed the same interaction history and
recording services. Platform-specific controls can therefore be added through
tool bindings while reusing task progression and feedback handling.

To reduce the repeated inference overhead of long GUI workflows, we allow
the model to issue multiple action calls in a single response. The runtime
executes these calls serially in the specified order, allowing actions that
can be planned together to share one model invocation.

For computer use, tool feedback also needs to describe the state after an
operation. Post-action observation hooks attach fresh screenshots or page
observations to state-changing tool results, giving the next decision a
current view of the interface. Section~\ref{sec:visualcontext} develops this action-aware
feedback design. The same tool separation supports centralized agents
operating remote devices, whose collection lifecycle is described in
Section~\ref{sec:batch}.

\subsubsection{Desktop and Web Instantiations}
\label{sec:platformagents}
We build desktop and web agents on this runtime by supplying the observations,
tools, and execution context appropriate to each platform.
Table~\ref{tab:adapters} compares the two instances along these interfaces:
\begin{itemize}
\item \textbf{Desktop agent.} The agent uses desktop screenshots to observe
applications and mouse and keyboard tools to click, type, scroll, and drag.
Its platform adapter manages the active
application, foreground window, and input focus so that an action reaches
the intended control. This supports workflows across application windows
and files, with screenshots, action results, and output artifacts retained
for later assessment. Appendix~\ref{app:desktopactions} lists its actions
and parameter schemas. Appendix~\ref{app:os2terminal} follows a recorded
animation task in which these same desktop actions open a terminal,
enter scripts and return to the presentation UI for playback checks.
\item \textbf{Web agent.} The agent observes page screenshots together with
DOM-derived information. Browser tools support navigation, element-reference
or coordinate-based actions, form entry, and tab operations. Its adapter
maintains the active tab, page state, and element references as navigation
changes the interface. The resulting record includes page observations,
action results, and browser artifacts. Appendix~\ref{app:webactions}
specifies the corresponding browser tools and argument constraints.
\end{itemize}
The adapters handle these platform differences, while both agents reuse
model invocation, request progression, context management, and trajectory
recording.

\begin{table}[htbp]
\centering
\caption{Desktop and web interfaces used in this report. Platform adapters
provide observations and action execution, while the Agent Kit supplies the
shared runtime and trajectory-recording services.}
\label{tab:adapters}
\small
\renewcommand{\arraystretch}{1.12}
\begin{tabularx}{\textwidth}{@{}>{\raggedright\arraybackslash}p{0.18\textwidth}
                                  >{\raggedright\arraybackslash}X
                                  >{\raggedright\arraybackslash}X@{}}
\hline
\textbf{Interface} & \textbf{Desktop agent} & \textbf{Web agent} \\
\hline
Interaction surface
& Applications, windows, and system interfaces
& Pages, tabs, and browser state \\
\hline
Policy observations
& Desktop screenshots
& Page screenshots and DOM-derived information \\
\hline
Environment-facing actions
& Mouse-and-keyboard GUI operations
& Navigation, element-reference or coordinate actions, keyboard input,
  forms, and tab operations \\
\hline
Execution context
& Active application, foreground window, and input focus
& Active tab, page state, and element references \\
\hline
Retained evidence
& Screenshots, action results, and application artifacts
& Screenshots, page snapshots, action results, and browser artifacts \\
\hline
\end{tabularx}
\end{table}

The same separation provides a path to additional platforms. Mobile
extensions, for example, would supply screen observations, touch operations,
and application/session handling around the shared runtime. Across the
desktop and web instantiations, this reusable execution
foundation also exposes a common challenge: carrying earlier constraints
and intermediate results into decisions made much later in a workflow.

\subsection{Sustaining Long-Horizon Interaction}
\label{sec:longruntime}

Long-horizon execution is a central challenge in practical computer use.
Real software work can require many fine-grained GUI operations across
applications, files, and intermediate artifacts, with later actions depending
on earlier discoveries and decisions \citep{yuan2026osworld2}. Editing a
document may still require a value read from a spreadsheet
and an output requirement encountered much earlier. Meanwhile, repeated
visual observations and tool feedback enlarge the history supplied to each
decision \citep{lu2026qwencua,lv2026scalecua}. Sustaining these workflows
requires controlling context growth while keeping current state, earlier
progress, and task-relevant knowledge available for continued action.

\begin{figure}[!t]
\centering
\includegraphics[width=\linewidth]{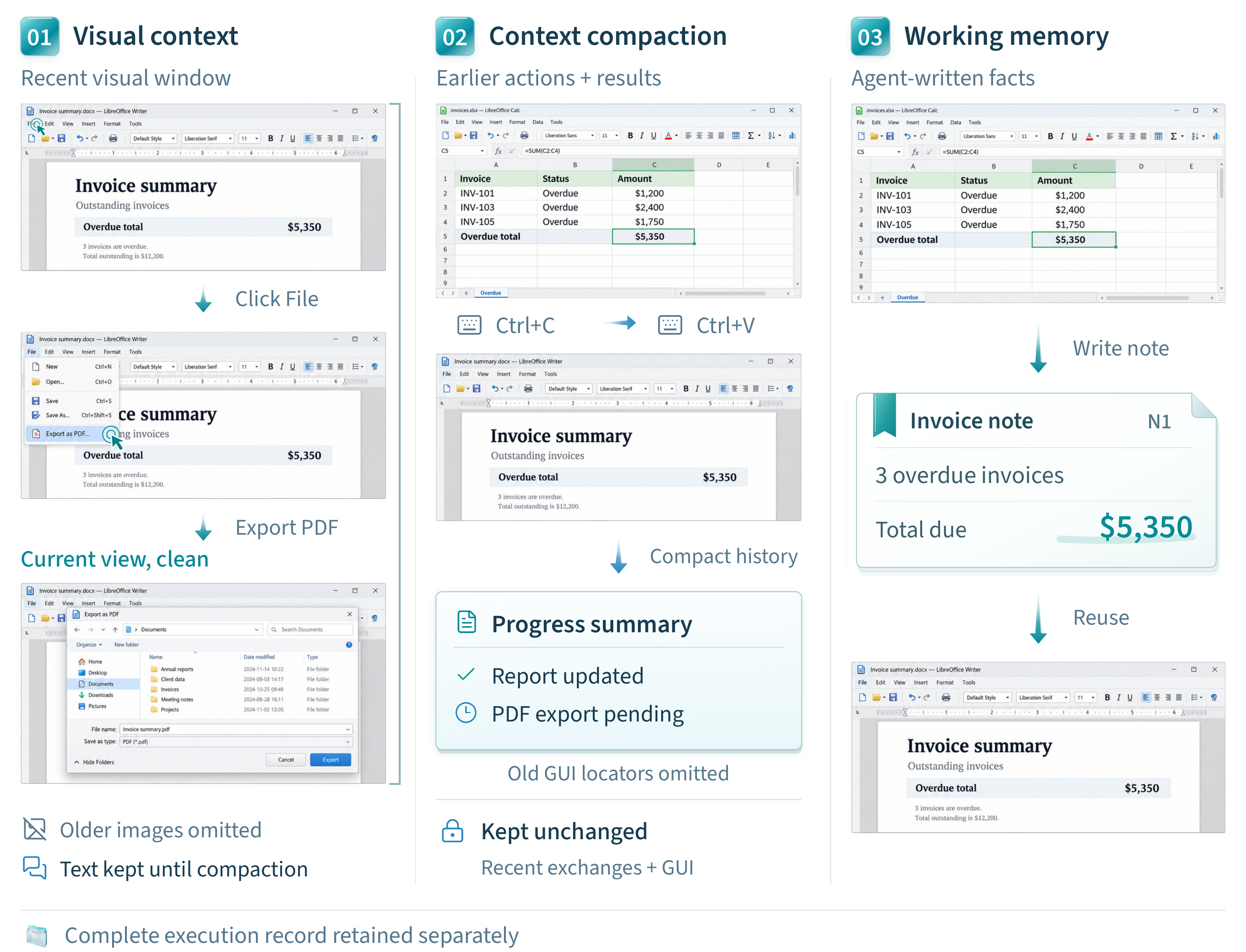}
\caption{\textbf{Context for long-horizon execution.}
The complete trajectory is retained separately from the next-action context.
The visual window contains historical pre-action overlays and the latest clean
tool-result image. Textual action and result history can outlast image eviction
until compaction. Compaction summarizes earlier actions and results, omits old
GUI locators, and preserves recent complete exchanges and the current GUI.
An active agent-written note carries the overdue total from workbook to report.
PDF export remains unconfirmed. GUI scenes and the displayed visual-window size
are illustrative, with intermediate interactions omitted.}
\label{fig:longhorizon}
\end{figure}

We separate the \emph{decision context} presented to the model from the
\emph{complete execution record} retained for inspection and learning.
To keep later decisions informed within a limited context budget, we design
three complementary mechanisms: \emph{action-aware visual context} for
recent states and operations, \emph{task-preserving compaction} for earlier
progress and remaining work, and \emph{structured working memory} for facts
and decisions the agent chooses to retain. Figure~\ref{fig:longhorizon}
follows one invoice workflow across these mechanisms. The visual window
shows the steps leading to export, the summary keeps export marked as
pending, and an agent-written note carries the overdue total into the report.
Each view supplies information needed for continued action without replacing
the full execution record.

\Needspace{11\baselineskip}
\subsubsection{Action-Aware Visual Context}
\label{sec:visualcontext}

Repeated GUI interaction produces an image-heavy history whose size grows
with the workflow \citep{lu2026qwencua}. Its visual content serves two related
purposes: the current screen guides the next action, and recent states help
the agent interpret what it just did. The resulting screen alone can leave the
location of a preceding click or the path of a drag unclear, particularly
after a window or page changes. Visual context therefore needs to preserve
both current state and the spatial meaning of recent operations within a
limited image budget.

We use a sliding window to provide a \textbf{bounded visual history} of up
to $k$ recent frames, including the latest clean screenshot. Textual action
and tool-result history survives image eviction, retaining a longer account
of the interaction until the text-history policy calls for compaction. To
keep the newest frame current, post-action observation hooks supply
\textbf{fresh post-action feedback} alongside state-changing tool results.
The next policy invocation receives this screenshot directly, avoiding
a dedicated model turn solely to request the resulting screen.

\begin{figure}[!t]
\centering
\includegraphics[width=\linewidth]{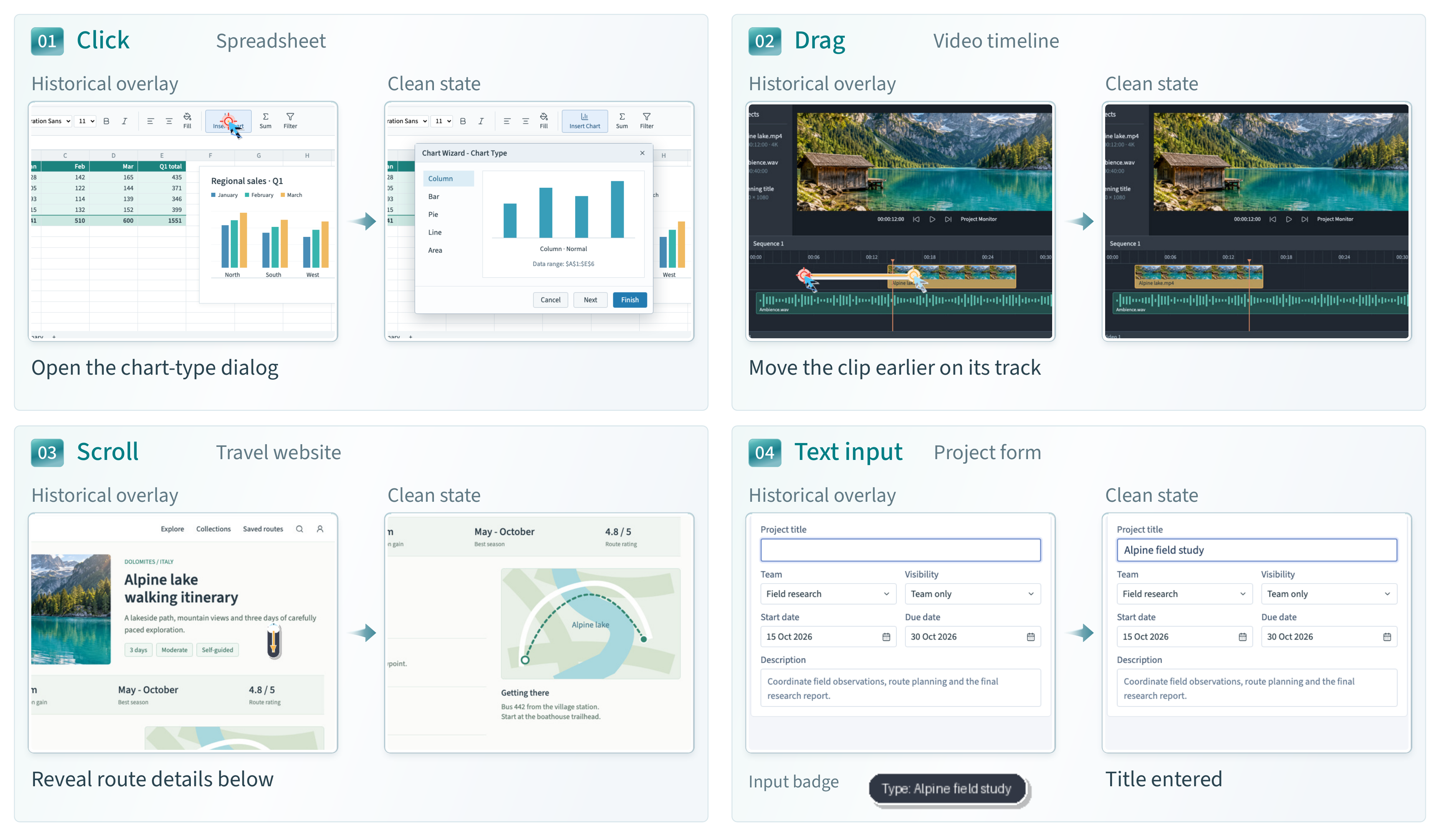}
\caption{\textbf{Action-aware visual feedback.}
Matched crops show the historical pre-action view with its NorthCUA overlay
(left) and the clean resulting state (right).
\textbf{01} A click opens the chart dialog.
\textbf{02} A drag moves a video clip earlier on its track.
\textbf{03} Scrolling reveals route details below.
\textbf{04} A text badge records the payload entered into the focused field.
Screenshots come from interactive GUI mockups with generated scenery.
Overlays use the unmodified runtime renderer. The input badge is enlarged
below its historical view.}
\label{fig:actionoverlays}
\end{figure}

To help the agent connect its previous operation with the change now visible
on screen, we design a complementary action view in the historical frames.
Visual prompting work shows how spatial marks can make image regions easier
to refer to \citep{yang2023setofmarkpromptingunleashesextraordinary}.
Our \textbf{historical action overlays} mark the corresponding
\emph{pre-action} image, using different cues for different operations.
As shown in Figure~\ref{fig:actionoverlays}, a click marker locates the control
that opens a dialog, a drag path links the clip's old and new positions,
and a scroll cue identifies the direction in which the page moved.
Text and key badges record operations that are less clear from a spatial
marker alone. Figure~\ref{fig:actionoverlays} pairs each annotated pre-action
view with its clean resulting state to make the operation and observed
change easy to compare. In the runtime's decision context, historical
overlays identify recent operations, while the latest clean screenshot
provides the current interface for the next decision.
Replacing an earlier clean frame with an overlay changes the image history.
The training-segment rule in Section~\ref{sec:rlcontext} preserves the input
seen before that change.

This bounded visual history preserves both the spatial meaning of recent
actions and the current interface state.
The textual record continues to accumulate across these visual windows,
creating the next context-management challenge.

\subsubsection{Task-Preserving Context Compaction}
\label{sec:contextcompaction}

Bounding images leaves another source of context growth: assistant messages, reasoning, 
tool feedback, and intermediate task information accumulate throughout the
workflow. Earlier constraints and discoveries may still be needed when this
text approaches the model's usable input budget. Context-management methods
consolidate long histories to retain useful information within finite prompts
\citep{ye2025agentfoldlonghorizonwebagents}. We adapt this principle to
computer use by designing task-preserving compaction around three priorities.
When older history needs to be summarized, we preserve the immediate
action--feedback context, retain task progress, and avoid carrying stale
GUI locators into later decisions:
\begin{itemize}
\item \textbf{Protect current interactions.} The task request, latest GUI
observation, and latest few steps remain protected. Assistant actions stay
grouped with their tool results, and open exchanges remain
intact until their responses arrive. This preserves the immediate
action--feedback context, including the current GUI observation whether it
appears in a user message or a tool result.
\item \textbf{Summarize task progress.} The summary organizes goals,
constraints, supported facts, decisions, completed and failed work, current
state, open items, and artifact references around task continuation.
Desktop and web profiles instruct the summarizer to retain semantic software
state and observed effects while condensing routine navigation. For example,
an extracted spreadsheet value, its source file, and a pending export requirement can
therefore remain available after the navigation history is compressed.
The instructions also distinguish issued actions, observed effects, and
unresolved completion claims: an export command and an observed saved file
represent different evidence about progress.
\item \textbf{Handle transient GUI locators.} Coordinates and element
references can become stale after a page update or application switch.
The summary-input projection filters these volatile fields, and the summary
instructions exclude old locators from current executable state. The retained
current observation remains available for grounding subsequent actions.
\end{itemize}

Later compaction updates the existing summary with newly eligible history,
with source references available for inspection. This gives continued work
a compact account of its dependencies while keeping the full execution record
available for outcome assessment. Alongside this budget-triggered
consolidation, the agent can
actively preserve important information as it is discovered.

\subsubsection{Structured Working Memory}
\label{sec:workingmemory}

During a long task, an agent encounters information with clear future value:
a figure read from another file, a requirement for the final deliverable,
an unsuccessful approach, or a plan for remaining work. Assistant messages
can mention these details in the chronological history, where their later
visibility depends on history retention and summarization. Giving selected
information an explicit working-memory representation lets the agent decide
what to retain and revise during execution, following the broader principle
of agent-controlled memory management
\citep{packer2024memgptllmsoperatingsystems}.

To let the agent preserve useful information when it discovers it, we design
a dedicated text-only \texttt{note} tool. The agent can call this tool
proactively to record a spreadsheet value before switching to a document,
or to preserve a correction while revising an unsuccessful approach.
The right-hand panel of Figure~\ref{fig:longhorizon} shows this transfer:
note N1 records the overdue total from the workbook for later use in the report.
Notes are stored
independently of the history summary and can be presented in subsequent
decision contexts. This complements compaction: a summary consolidates
accumulated work under budget pressure, while a note explicitly preserves
information the agent expects to use again.

Notes serve three purposes: (i)~retaining reusable knowledge and choices
through \emph{facts}, \emph{constraints}, and \emph{decisions},
(ii)~recording lessons and risks through \emph{corrections} and
\emph{precautions}, and (iii)~tracking progress and unfinished work through
\emph{checkpoints}. This organization encourages concise entries with a clear
role in later decisions.

As work progresses, replacement notes can supersede outdated entries. The
memory service maintains them within the authorized task and session, then
supplies a budgeted view of active entries to prompt construction. A
corrected value or revised plan can therefore replace the corresponding
obsolete item in working context, keeping remembered information aligned
with the agent's evolving understanding of the task.

\Needspace{6\baselineskip}
Together, these mechanisms keep the current screen, accumulated progress,
and reusable knowledge available for continued action. The complete execution record
connects these decisions to the resulting workspace, providing the evidence
for outcome evaluation and subsequent training. Appendix~\ref{app:runtimeconfig}
gives the runtime settings, and Appendix~\ref{app:supervisedrecipe} and
Section~\ref{sec:rlcontext} describe their training interfaces.

Appendix~\ref{app:os2context} follows a photo-delivery task in which a payment
constraint remains available from reconciliation through the final email.

\subsection{Agentic Outcome Evaluation}
\label{sec:evaluator}

\begin{figure}[!htbp]
\centering
\includegraphics[width=\linewidth]{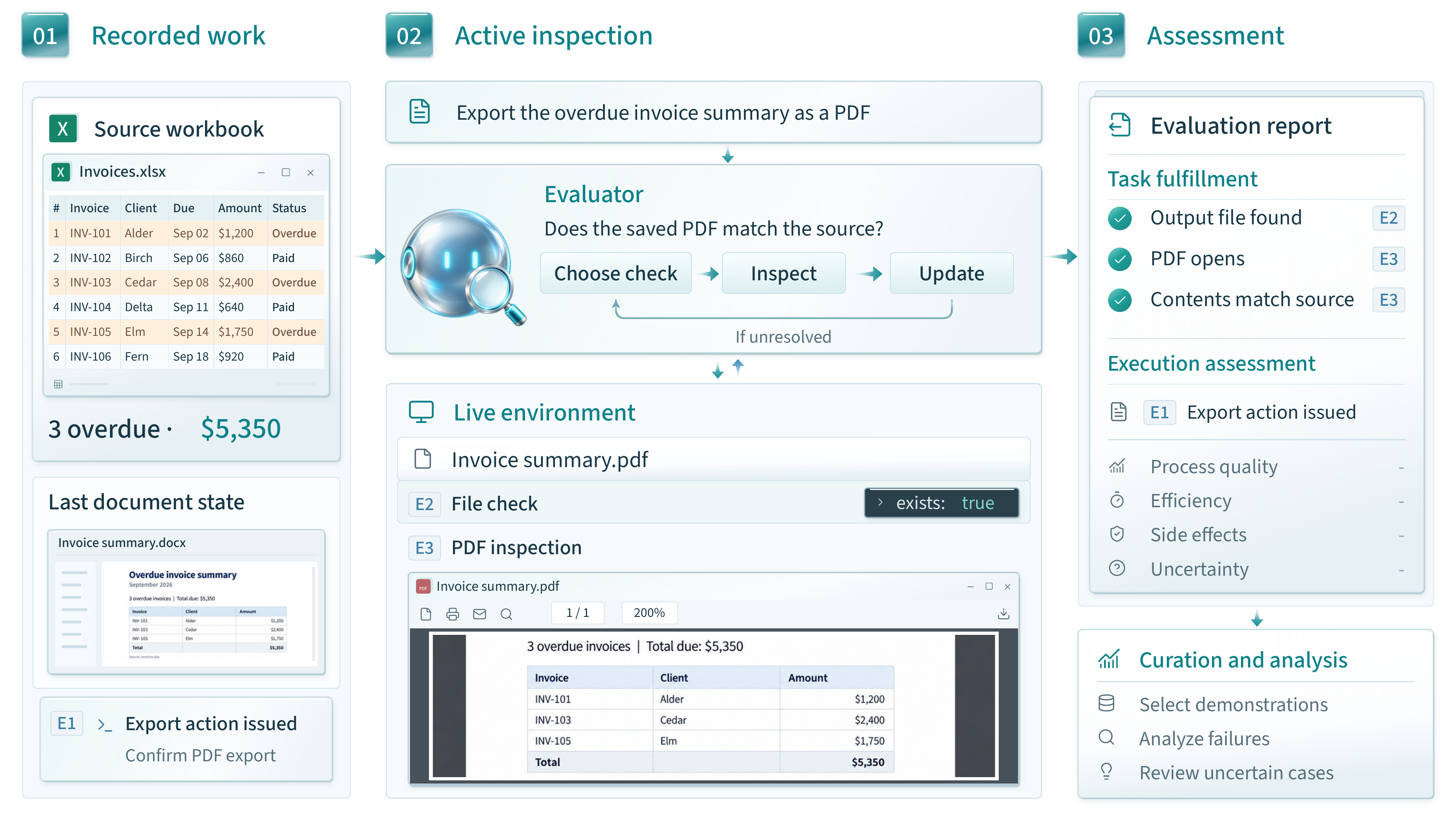}
\caption{\textbf{Agentic outcome evaluation.}
Open questions guide selective retrieval and live inspection in the task
environment. E1 records an issued export, E2 establishes file presence, and
E3 exposes PDF contents for comparison with the source workbook. Unresolved
questions prompt further checks and can remain unresolved in the final report.
Evidence-linked findings distinguish task fulfillment from execution assessment.
GUI views, tool results, actions, and findings are illustrative. Generated
document views are cropped, intermediate actor actions are omitted, and
selected quality fields are shown without ratings.}
\label{fig:evaluator}
\end{figure}

Selecting GUI trajectories for learning requires accurate outcome judgments
across a growing variety of tasks. Dedicated verifiers become difficult to
write and validate when correctness depends on content, presentation, or
application-specific state \citep{rosset2026verifiers}.
We therefore use an adaptive evaluator that interprets the declared task
requirements and gathers the evidence needed to assess the work.
Figure~\ref{fig:evaluator} shows how recorded work and active inspection
together support an evidence-linked assessment.

We build the evaluator as an \emph{agent-as-a-judge}
\citep{zhuge2024agentasajudgeevaluateagentsagents}, reusing Agent Kit tools.
After the acting phase ends, it selectively retrieves saved records and artifacts and can
inspect the original workspace using authorized shell,
GUI, and browser tools. It adapts its inspection to each requirement and
seeks further evidence when the available observations leave the outcome
unclear. This combines reusable inspection capabilities with task-specific
judgment, extending assessment to requirements that lack dedicated executable
checks. Task-grounded rubrics guide this process, and multidimensional reports
connect the findings to experience selection and failure analysis.
In Figure~\ref{fig:evaluator}'s export example, the evaluator checks that
the PDF exists and opens it to compare its contents with the source.
This artifact inspection supplies the evidence for its final report.
We require this report to associate each criterion-level judgment with
its evidence, uncertainty, and evaluator operations.

\subsubsection{Artifact and Live-State Inspection}

The evaluator receives the task request, assessment criteria, and a compact
view of the execution records. Requirements may concern an earlier action,
a final file, or a state that is visible only inside an application. To reach
these different forms of evidence, it uses two complementary tool groups:
\begin{itemize}
\item \textbf{Recorded evidence.} Retrieval tools expose selected steps,
screenshots, tool results, and retained artifacts. They let the evaluator
inspect relevant parts of a long trajectory as questions arise, while the
full record remains available outside its current context.
\item \textbf{Live task state.} Live tools inspect the original device
or sandbox through fresh screenshots, read-only shell commands, and
GUI or browser operations. Web probes can reuse the actor's browser session,
including its tabs and page state. This access makes environment-resident
results available for checking while the workspace is still alive.
\end{itemize}
Appendix~\ref{app:evaluatoractions} lists the evaluator's retrieval tools,
authorized live interfaces, and report schema.

The evaluator chooses between these sources according to the requirement
under review, continuing its investigation when an observation leaves an
important question open. This environment-aware approach is also studied in
\citet{shi2026ajbenchbenchmarkingagentasajudgeenvironmentaware}. For a document
export, for example, a recorded action can show that export was attempted,
a file check can establish whether an output exists, and an
application inspection can reveal its rendered contents. Comparing these
contents with the task inputs supports judgments about their correctness.
Each observation answers a different part of the acceptance question. Selective retrieval
and live access thus connect the history of an operation to its actual result.

\begin{table}[!t]
\centering
\caption{Evaluator reporting dimensions and their uses in trajectory
curation and failure analysis.}
\label{tab:evaluator-dimensions}
\small
\renewcommand{\arraystretch}{1.12}
\begin{tabularx}{\textwidth}{@{}>{\raggedright\arraybackslash}p{0.18\textwidth}
                                  >{\raggedright\arraybackslash}X
                                  >{\raggedright\arraybackslash}X@{}}
\hline
\textbf{Dimension} & \textbf{Assessment} & \textbf{Analysis and selection use} \\
\hline
Task completion
& Criterion fulfillment, overall verdict, and completion rating
& Identify supported completion examples and unresolved requirements \\
\hline
Process quality
& Suitability of actions, tool use, and avoidable mistakes
& Distinguish reliable demonstrations from fragile paths to the same result \\
\hline
Efficiency
& Repetition, detours, unnecessary steps, and recovery behavior
& Compare successful paths and apply cohort-specific efficiency criteria \\
\hline
Safety and side effects
& Sensitive operations, unsolicited changes, and observed risks
& Review unsafe or undesired behavior separately from completion \\
\hline
Failure diagnosis
& Failure mechanism, controllability, and supported divergence location
& Analyze capability gaps and distinguish them from environment blockers \\
\hline
Judgment uncertainty
& Confidence, missing information, and uncertain verdicts
& Route unresolved cases for further checking \\
\hline
\end{tabularx}
\end{table}

To preserve attribution, the assessment protocol pairs pre-inspection
terminal evidence with a separate record of evaluator operations. Each
judgment is linked to the submitted work and the observations used to assess
it, making the investigation traceable to the original attempt. The rubric
then connects this evidence to task fulfillment and execution quality.
Appendix~\ref{app:evaluatorprotocol} details the inspection and attribution
protocol.

\subsubsection{Task-Grounded, Multidimensional Rubrics}
\label{sec:evaluator-rubric}
The value of a GUI trajectory depends on both its result and the behavior
it demonstrates. Successful attempts can contain unnecessary detours or
unintended changes, while an attempt blocked late by the environment can
contain useful completed subtasks. Data curation therefore needs answers to
two complementary questions: (i)~\emph{which task requirements were fulfilled},
to establish what the trajectory accomplished, and (ii)~\emph{how the agent
performed the work}, to assess the behavior available for learning.
We therefore design the rubric to combine task-grounded acceptance criteria
with multidimensional execution assessment \citep{rosset2026verifiers}.

For \textbf{task fulfillment}, the request and declared acceptance conditions
define the intended outcome. The rubric expresses these obligations as
specific, individually checkable questions. Export tasks may require the
right contents, destination, and format. Explicitly required procedures
also form part of acceptance. Request-derived criteria are generated before
scoring the trajectory, with the original task requirements governing their
interpretation. Assessing these obligations individually identifies complete,
partial, and unresolved fulfillment, including which requested subtasks have
supporting evidence.

For \textbf{execution assessment}, the report records process quality,
efficiency, side effects, and failure diagnosis alongside judgment
uncertainty. These dimensions explain the path taken to the result: repeated
ineffective actions may reduce its value as a demonstration, while a successful
recovery can provide useful behavior to study. Failure diagnosis distinguishes
agent errors from environment blockers, and uncertainty identifies judgments
that need further checking. Findings are linked to the relevant requirements
and recorded steps so that their scope can be inspected.
Table~\ref{tab:evaluator-dimensions} summarizes the dimensions and their uses.

Together, the rubric and execution report describe both the result achieved
and the behavior that produced it. Their links to specific requirements and
recorded steps make the judgments useful for selecting complete demonstrations,
reviewing local segments, and analyzing recurring errors.

\subsubsection{From Evaluation to Experience Selection}

We use the report's separate rubric scores to select data for different
learning goals. Task fulfillment identifies completed work, while process
quality, efficiency, and side-effect assessments let us adjust selection
thresholds for each data cohort. For example, a cohort of concise successful
demonstrations can apply stricter efficiency criteria, while a cohort meant
to teach recovery can retain useful corrections and exploration within
successful attempts.

At the \textbf{trajectory level}, these criteria select complete
demonstrations. At the \textbf{segment level}, requirement-linked evidence
can guide review of completed subtasks or recovery sequences, with the
original decision context and local outcome kept alongside the behavior.
Figure~\ref{fig:evaluator}'s report separates completion from execution
quality so that these dimensions can be used together in selection.
Section~\ref{sec:experience-selection} applies the reports to trajectory
curation, and Section~\ref{sec:longsft} retains complete episodes for
long-horizon SFT.

\subsection{Reliable Batch Rollout at Scale}
\label{sec:batch}

The preceding mechanisms turn one task attempt into assessed experience.
Producing a large training corpus requires many such attempts to run in
parallel, each with its own software state. Their preparation, GUI execution,
evaluation, and recording take different amounts of time and consume
different resources \citep{xie2024osworld,yuan2026osworld2}. Simply launching
more agents can therefore create a backlog at another stage, while an
interruption after execution can leave useful work without a complete record.

\begin{figure}[!htbp]
\centering
\includegraphics[width=\linewidth]{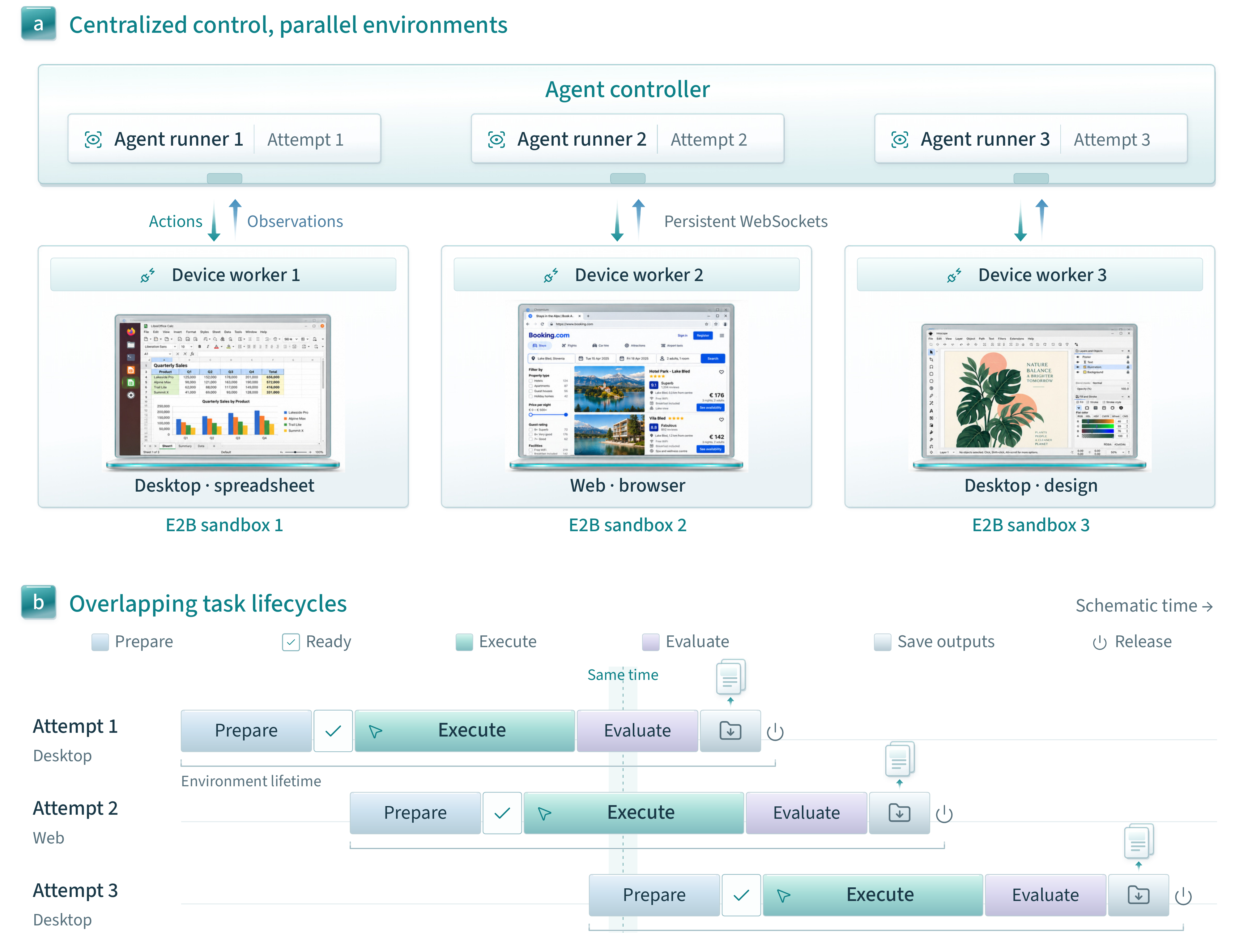}
\caption{\textbf{Batch rollout at scale.}
(a) A centralized controller manages agent runners connected to device workers
in independent E2B sandboxes over long-lived WebSocket connections. Each worker
shares its sandbox's lifecycle. (b) Staggered lifecycles overlap preparation,
execution, and evaluation. Ready intervals mark prepared environments awaiting
dispatch. Sandboxes remain available through live checks and output capture.
Record icons mark captured outputs that persist after release. The vertical
time slice marks concurrent stages. Timings and concurrency are schematic,
and GUI scenes are generated illustrations.}
\label{fig:rollout}
\label{fig:remote}
\end{figure}

To support efficient large-scale collection, we design a \textbf{Batch
Rollout Service} that coordinates the entire task lifecycle. Following the
environment-lifecycle perspective also explored in \citet{peng2026orchard},
we combine centralized agent control with isolated workspaces, prepare
upcoming tasks alongside ongoing execution, and manage load and recovery
through record delivery. Figure~\ref{fig:rollout} shows the two relationships
behind this design. Panel (a) connects one controller to independent device
workers. Panel (b) shows how their lifecycles overlap: the same-time marker
crosses evaluation, execution, and preparation in three different attempts.
The service coordinates this concurrent work so that completed attempts
produce usable experience records.

\subsubsection{Remote Execution with Centralized Agent Control}
As collection grows, prompts, model settings, and agent logic continue to
change. Placing that logic inside every task environment would require
updating those environments whenever the agent changes. We therefore
\textbf{separate centralized agent control from device execution}, as shown
in Figure~\ref{fig:rollout}(a):
\begin{itemize}
\item \textbf{Centralized controller.} One service manages multiple agent
runners and their device workers. It maintains prompts and agent
configuration, assembles context, invokes the model, and dispatches actions.
\item \textbf{Device worker.} Each worker runs inside one task's isolated
sandbox. It executes tool calls and captures observations from that
environment. The worker starts with its sandbox and ends when the sandbox
is released.
\end{itemize}

The controller maintains a long-lived WebSocket connection to each worker,
sending actions and receiving observations over the same connection.
Related distributed device orchestration appears in \citet{zhang2025ufo3}.
Here, each runner communicates with the worker assigned to its task,
keeping the concurrent attempts separate.

This division lets us update compatible prompt and model configurations
centrally for later batches while reusing the same device-worker
implementation and software templates. The device layer remains focused on observation and tool
execution, so routine agent changes do not require changing that layer.
We save each batch's configuration for resumed collection and retain its
trajectories on the controller side, where they survive sandbox release.
With control and recording shared, scaling execution then depends on
preparing enough independent task environments.
\FloatBarrier

\subsubsection{Isolated Environments with Pipelined Preparation}

A new attempt needs the files and software state specified by its task.
Reusing a workspace changed by an earlier attempt could alter those starting
conditions \citep{xie2024osworld}. We therefore give each attempt its own
initialized sandbox in our fresh-start collection mode. For Linux desktop
and browser tasks, we choose E2B to create and manage these isolated
environments \citep{e2b_runtime}, with its desktop extension providing the
graphical workspace \citep{e2b_desktop}. Custom templates supply the operating
system and shared applications, while per-task setup uploads files,
launches applications, and prepares the required state. This lets tasks
reuse a software template without inheriting another attempt's changes.

Creating a fresh environment takes time, so preparing tasks only when an
agent is ready would leave collection waiting on setup. We address this
with a \textbf{task-bound preparation queue}: upcoming environments are
prepared while earlier attempts are executing or being evaluated.
Figure~\ref{fig:rollout}(b) makes this overlap visible in the offset
Prepare blocks. Each ready workspace is assigned once to its matching
task, provided enough time remains for execution and cleanup. It stays
available through the required outcome checks and output capture, then
is released. Preparation can thus run ahead without sacrificing fresh
starting conditions or losing the evidence needed for assessment.

\subsubsection{Completion-Oriented Scheduling and Recovery}
Preparation, execution, and evaluation progress at different rates.
Independent admission can leave prepared environments waiting for execution
or completed attempts waiting for evaluation. We coordinate admission with
two controls. \textbf{Stage-specific
concurrency limits} bound the number of attempts admitted to each stage.
Because resource demand can vary within these limits,
\textbf{controller-pressure feedback} slows new admissions when the shared
controller remains busy.
The first control limits local demand, while the second responds to load
across the concurrent attempts.

When a connection is lost or a downstream process fails, we first check
whether the affected attempt is still running or has completed. Completed
execution records can support retries of record-based evaluation or
artifact processing, preserving the work already done. If the task itself
needs to be rerun, we start it in a fresh environment.

Before delivery, we check that the required trajectory, observations, and
assessments are complete and linked to the same attempt.
Appendix~\ref{app:batchprotocol} gives the configuration and recovery details.

Together, these designs carry the same runtime from individual task execution
to sustained delivery of complete, inspectable experience. The \experienceflywheel
in Section~\ref{sec:data} builds on this foundation to construct tasks and
curate the resulting records for model training.

\section{Experience Flywheel: Scaling Grounding and Agentic Experience}
\label{sec:data}

\providecommand{\datafigureplaceholder}[2]{%
  \harnessfigureplaceholder{#1}{#2}%
}

The \northcua Harness provides the execution infrastructure, but training a
computer-use model also requires a large and varied supply of experience.
The model needs examples of which control a request refers to, as well as
trajectories that show how successive decisions achieve a complete goal.
We design the \experienceflywheel to construct and curate these two levels
of supervision. \textbf{Single-step grounding} pairs a screenshot and a
query with the location of one visible control. \textbf{End-to-end task
construction} supplies user goals, prepared workspaces, and outcome checks
from which the harness can collect multi-step trajectories.

These levels require different construction methods. For grounding, we
start from visible controls and generate language that identifies them.
For complete tasks, we connect available software and content, sourced
requests, or verified seed tasks to the environments and assessments they
need. This gives three complementary routes for scaling intents,
environments, and verifiers. Executing the resulting tasks supplies
experience for learning, although successful trajectories can remain scarce
for the hardest workflows. We address that remaining gap through recursive
exploration that retains useful skills and verified progress across
attempts. Together, these methods supply localization examples and curated
trajectories for the staged learning in Section~\ref{sec:training}.

\subsection{Single-Step UI Grounding}
\label{sec:grounding}

Single-step grounding takes a screenshot and a natural-language query, then
predicts the location of the requested control
\citep{cheng2024seeclick}. Each training example contains a screenshot, a
query that identifies a visible control such as a
button or toolbar icon, and the control's coordinate in the original image.
The main construction
challenge is to write queries that identify the same control across different
visual states and ways of describing an operation, especially in specialized
applications \citep{lu2024omniparser,li2025autogui}.

We construct these examples through two complementary routes:
\begin{enumerate}
\item \textbf{Broad visual and structural coverage.} We use screenshot
evidence and available interface metadata to describe visible controls
across diverse applications and states (Figure~\ref{fig:grounding}).
\item \textbf{Documentation-grounded functional localization.} We retrieve
application knowledge when a control's appearance leaves its function
unclear, then use the matched function to generate queries
(Figure~\ref{fig:documentationgrounding}).
\end{enumerate}
The first route expands interface and language coverage. The second fills
gaps that visual evidence alone leaves unresolved. Both preserve the target's
original location and produce the same screenshot--query--coordinate format,
so their candidates can pass through shared quality and coverage checks.

\subsubsection{Broad Visual and Structural Coverage}

\begin{figure}[!t]\centering
\includegraphics[width=\linewidth]{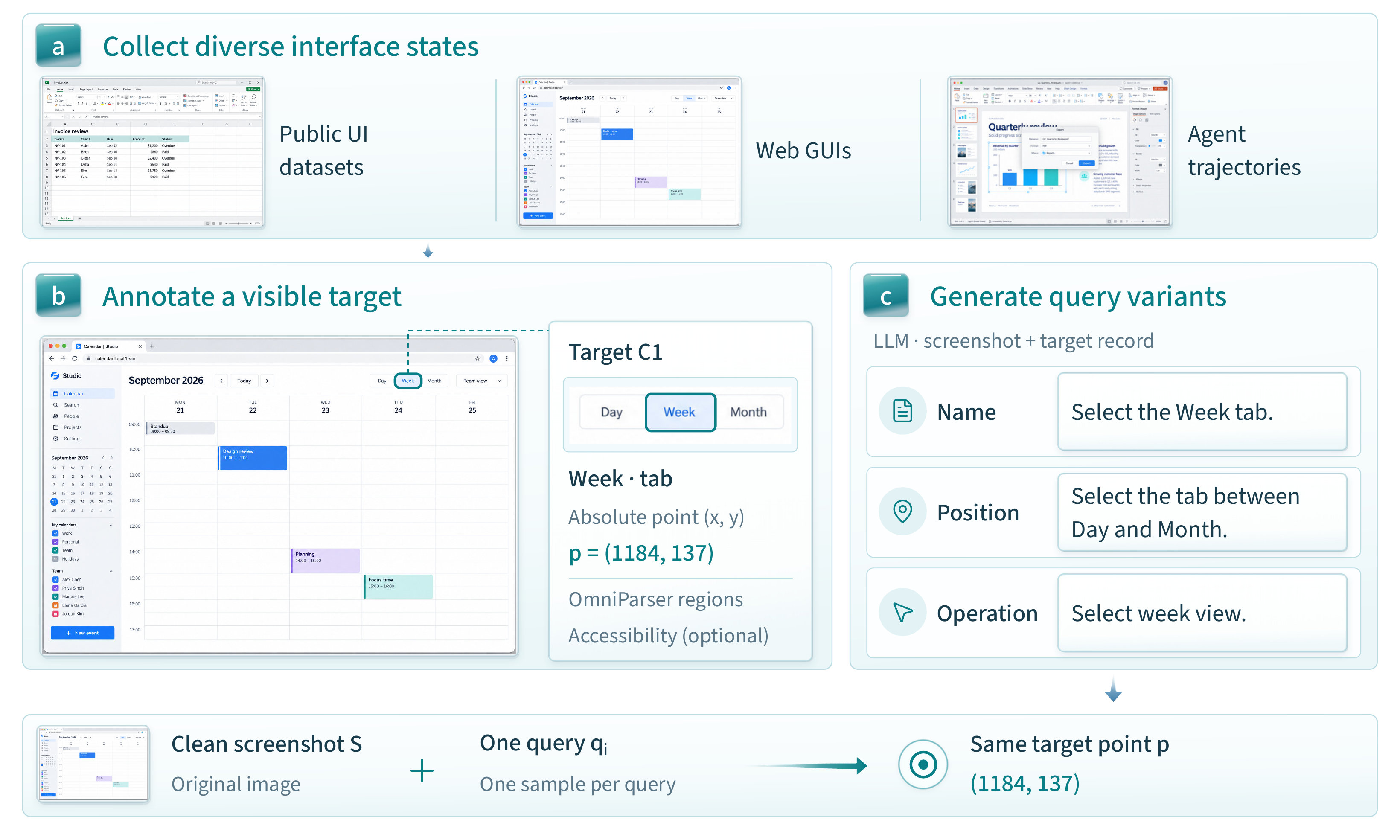}
\caption{\textbf{Constructing GUI grounding examples.}
(a) Screenshots cover public UI datasets, web-collected interfaces, and
states encountered in agent trajectories. (b) Visual annotations and available
accessibility metadata describe one target in its original screenshot.
(c) Name-, position-, and operation-based queries each form a separate
candidate example with the same clean screenshot and absolute target point.
The enlarged Day--Week--Month control group supplies the local evidence for
these descriptions. GUI assets are illustrative. Target outlines explain
construction, while training uses the unmarked $1536\times1024$ source image.}
\label{fig:grounding}
\end{figure}

To obtain broad grounding coverage, we first need screenshots of varied
interfaces, then a way to turn their visible controls into language targets.
We use the three-step procedure in Figure~\ref{fig:grounding}. Panel (a)
collects interface states, including states encountered during agent work.
Panel (b) associates one visible control with its appearance and original
location. This target record gives the query generator in (c) a fixed
referent: it can vary how the control is described while keeping the
location label unchanged.

To collect these states, we combine public UI datasets, web-collected GUI screenshots, and
screenshots from agent trajectories, including real Linux desktop
interaction. Trajectories contribute intermediate states such as open
dialogs, expanded menus, and partially completed forms.
We then use OmniParser, a screen-parsing system that detects interactive
regions and describes icons, to obtain candidate control boxes and
descriptions \citep{lu2024omniparser}. Accessibility annotations,
when available for the same state, add control names and roles. These sources
provide complementary evidence for identifying a target. Visual parsing
supports screenshots collected without accessibility metadata, while names
and roles help connect visible controls to the language used in requests.

For each target, we consolidate this evidence into one control record that
keeps the crop and original image position together with the available
name, role, color, appearance, and surrounding context
\citep{zhang2025ufo2}. For the calendar example in
Figure~\ref{fig:grounding}(b), the record identifies the blue \emph{Week}
tab between \emph{Day} and \emph{Month}, together with a point inside its
clickable region. The full screenshot preserves the neighboring controls
needed to make a description specific.

\begin{samepage}
With this record in place, we use an LLM to generate several queries from
the record and full screenshot. The descriptions refer to the control's name, appearance, position,
or an operation supported by the available evidence. In the example,
\emph{Select the Week tab} names the control directly, while
\emph{Select week view} expresses its purpose.
Each query becomes a separate candidate example paired with the clean
original screenshot and the same coordinate. This variation broadens the
language supervision without changing which control the model learns to
locate \citep{chen2025uiins}.\par
\end{samepage}

This route supplies broad coverage of interfaces and visual states. When a
control's operation depends on application-specific knowledge, the
documentation route provides the missing functional description.

\subsubsection{Documentation-Grounded Functional Localization}
Some controls belong to specialized applications whose functions are hard
to infer from a screenshot alone. Parsing may locate a toolbar icon while a
general-purpose LLM remains unsure what the icon does. This distinction
matters when several visually similar controls perform different operations,
as in a layout editor with alignment controls for different parts of a document
\citep{wang2025opencua}.

\begin{figure}[!t]\centering
\includegraphics[width=\linewidth]{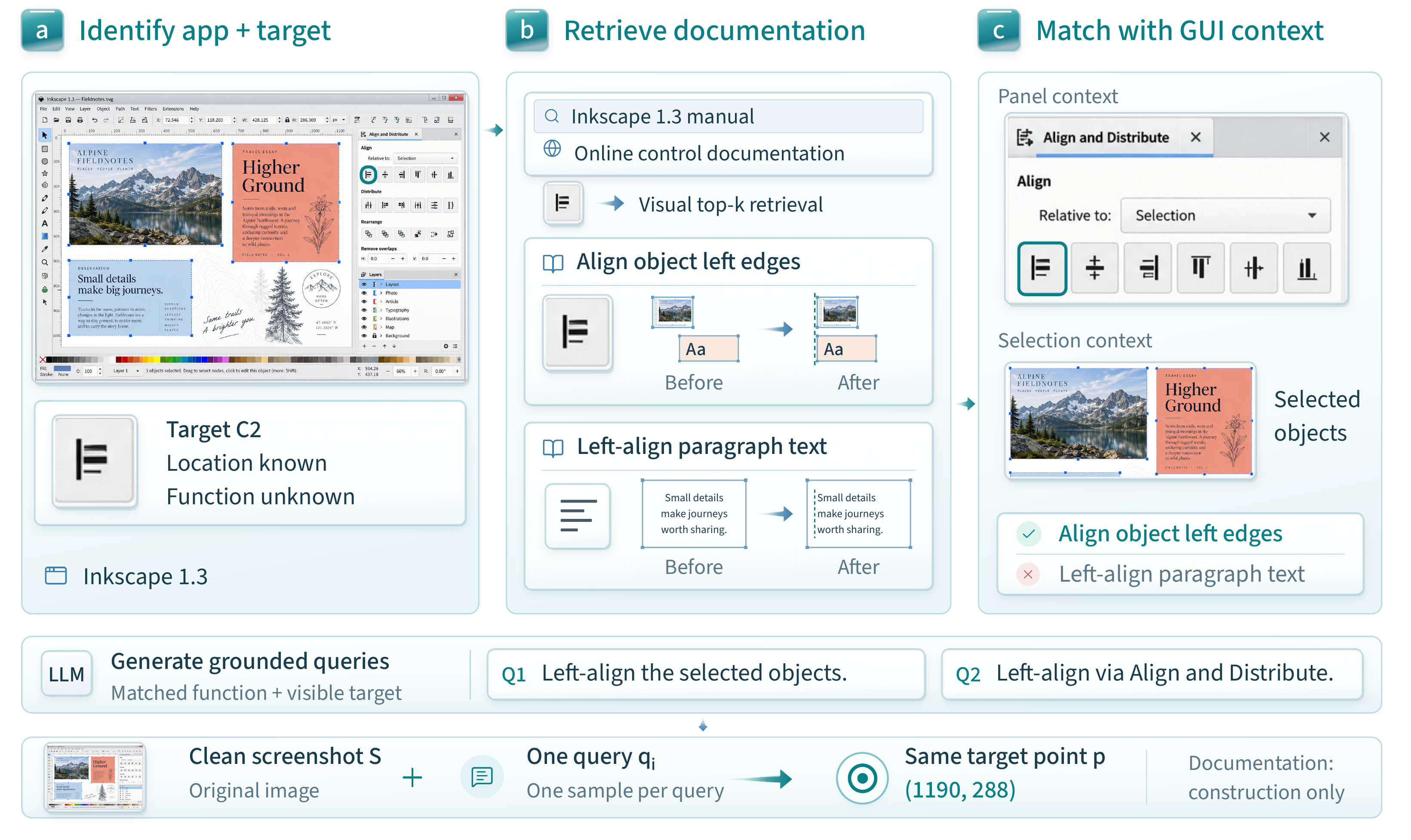}

\caption{\textbf{Documentation-guided control disambiguation.}
(a) The screenshot supplies the application, version, and located target.
(b) Online manual search retrieves control images and descriptions for visual
top-$k$ candidates. One aligns the left edges of separate objects, while the
other aligns paragraph text within a fixed text object. (c) Panel and
selection context determine the function match, which guides query generation.
Each query retains the original screenshot and target point.
The GUI, documentation cards, and before/after operation diagrams are
illustrative. Documentation supports sample construction.}
\label{fig:documentationgrounding}
\end{figure}

\begingroup\interlinepenalty=10000
We design a data-construction agent to supply the missing application
knowledge before query generation. Figure~\ref{fig:documentationgrounding}
illustrates three steps. The agent identifies the application and target,
then searches online documentation for candidate functions. It uses the
surrounding GUI to select the function that fits the visible context.
\par\endgroup

In the alignment example, the
same-looking alignment symbol could refer to moving objects or formatting
text. Matching its documented function resolves what a query can reliably
ask the agent to do with the already located control.

\paragraph{Identify the application and target.}
This route begins with the same OmniParser-based screen parsing, which
provides candidate control boxes and crops while preserving their locations
in the original screenshot \citep{lu2024omniparser}. Given the screenshot
and a target crop, the construction agent identifies the application and
any visible version information, such as Inkscape 1.3 in panel (a).
These details constrain the documentation search to the software in which
the control appears, while the crop provides the visual target to match.

\paragraph{Retrieve documented functions.}
Using that application identity, the agent searches online user guides and
manuals and extracts illustrated controls with their function descriptions.
We keep each illustration with its function description and page context,
so that a visually similar icon retains the information needed to interpret
its operation. Image-embedding similarity
then narrows these references to top-$k$ candidates
\citep{radford2021clip}. Panel (b) shows both parts of this step: the online
manual search and the candidate operations retrieved for the target icon.

\paragraph{Match with GUI context and generate queries.}
Visual similarity supplies candidates but can leave the operation ambiguous.
The agent therefore compares their descriptions with the full screenshot,
neighboring controls, and active tool group.
This contextual check matters because similar icons can denote different
operations.

In Figure~\ref{fig:documentationgrounding}, object alignment and
paragraph alignment provide plausible visual candidates. Object alignment
moves selected canvas objects to align their left edges, whereas paragraph
alignment left-aligns the text within a text object. The visible
\emph{Align and Distribute} panel identifies the relevant tool group, and
the selected canvas objects provide complementary evidence for the
object-alignment interpretation. The accepted match therefore links a
specific function description to the control already located in the screenshot.

Once the function is resolved, we give the accepted description, target crop,
and full screenshot to an LLM to generate query variants. One can request the documented
operation, such as aligning selected objects along their left edges.
Another can identify the same control through its tool group and appearance.
All variants inherit the original target coordinate. Thus, application
knowledge enriches the query while preserving the common grounding format.
The documentation and retrieval candidates remain construction-side evidence,
and the resulting samples enter the shared quality checks below.

\subsubsection{Quality Control and Coverage}
Both routes produce the same candidate format: a clean screenshot, a query,
and a target coordinate. We use two complementary checks before adding a
sample to the corpus:
\begin{itemize}
\item \textbf{Correctness check.} Geometry checks confirm that the point lies
inside the annotated control. Multimodal review then checks that the
control supports the described operation, the query refers to it, and
the description is unambiguous among nearby controls. Documentation-derived examples also carry
their accepted function match into this review.
\item \textbf{Coverage check.} Deduplication removes repeated screenshots
and near-identical descriptions. Selection retains useful variation across
applications, UI states, control types, and query styles, so that different
ways of referring to a control remain represented.
\end{itemize}
These checks address different reasons a larger candidate pool may fail to
provide useful supervision: incorrect target labels and repetitive coverage.
Geometry or identity failures return to construction when they can be
repaired, while unresolved cases are excluded. Selection then retains
accepted examples across interfaces and language styles.
Appendix~\ref{app:grounding}
records the annotation, matching, and curation settings.

The accepted examples enter mid-training as direct localization supervision.
Screenshots acquired during multi-step collection can also enrich this pool
with states encountered in actual software work. The next subsection extends
the same construction logic from identifying one control to carrying out a
complete user goal.

\subsection{Scaling Intents, Environments, and Verifiers}
\label{sec:threeaxes}

\begingroup\interlinepenalty=10000
Grounding examples teach the model to locate a requested control. Learning
to plan and act across a complete workflow also requires interaction
trajectories \citep{wang2024large,jung2026procuasft,wang2026cuagym}.
To collect them, we must turn each user goal into an executable task by
providing the workspace it needs and a reliable assessment of its outcome.
\par\endgroup

\begin{figure}[!t]\centering
\includegraphics[width=\linewidth]{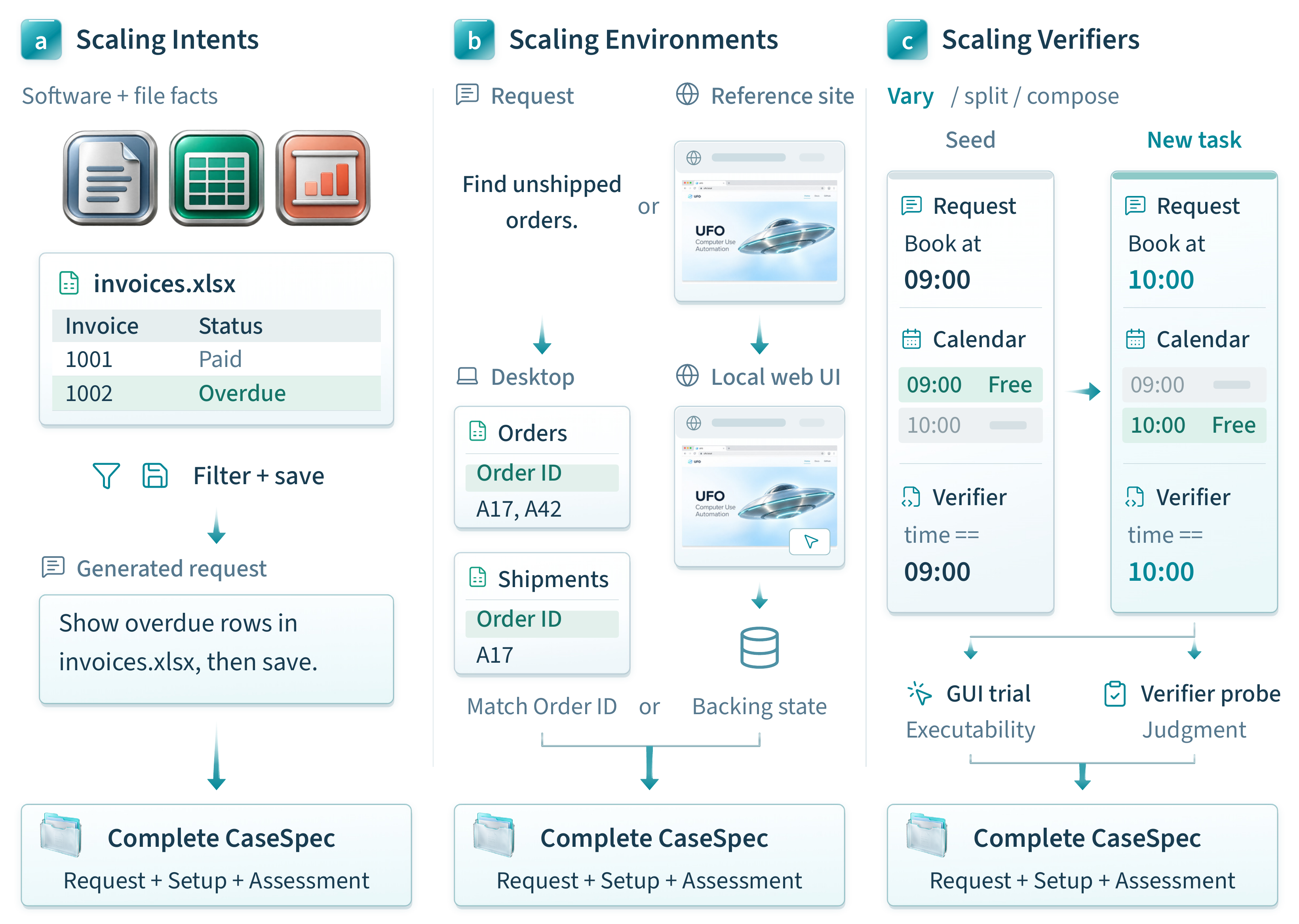}
\caption{\textbf{Three task-construction routes.}
Scaling Intents binds functions to file or website content before writing
requests. Scaling Environments prepares files and applications for sourced
requests or reconstructs website interfaces and backing state from references.
Scaling Verifiers jointly adapts requests, setup, and executable checks, then tests
tasks and verifiers. Each alternative route produces a complete task binding
a request, environment setup, and outcome assessment. GUI trials test completion
from the initial state, while result probes test acceptance of correct outcomes
and rejection of incorrect ones.
GUI assets and task packages are illustrative.}
\label{fig:threeaxesoverview}
\end{figure}

Figure~\ref{fig:threeaxesoverview} compares three construction routes by
what they start with and what they generate:
\begin{itemize}
\item \textbf{Scaling Intents} starts from applications, files, or website
content and generates requests supported by their functions and facts.
\item \textbf{Scaling Environments} starts from sourced requests or
reference websites and builds the files, applications, or local website
needed to execute tasks in that setting.
\item \textbf{Scaling Verifiers} starts from tasks whose outcomes can be
checked by reliable scripts. It generates related tasks by adapting the
request, setup, and executable verifier together.
\end{itemize}

\begingroup\interlinepenalty=10000
Each column ends with a complete task: a request, workspace
initialization, and outcome assessment. The routes thus expand task supply
from different resources while retaining the same requirements for execution
and checking. Depending on the task, assessment uses an executable verifier
or evaluator inspection of the declared acceptance conditions.
\par\endgroup

\begin{figure}[!t]
\centering
\includegraphics[width=\linewidth]{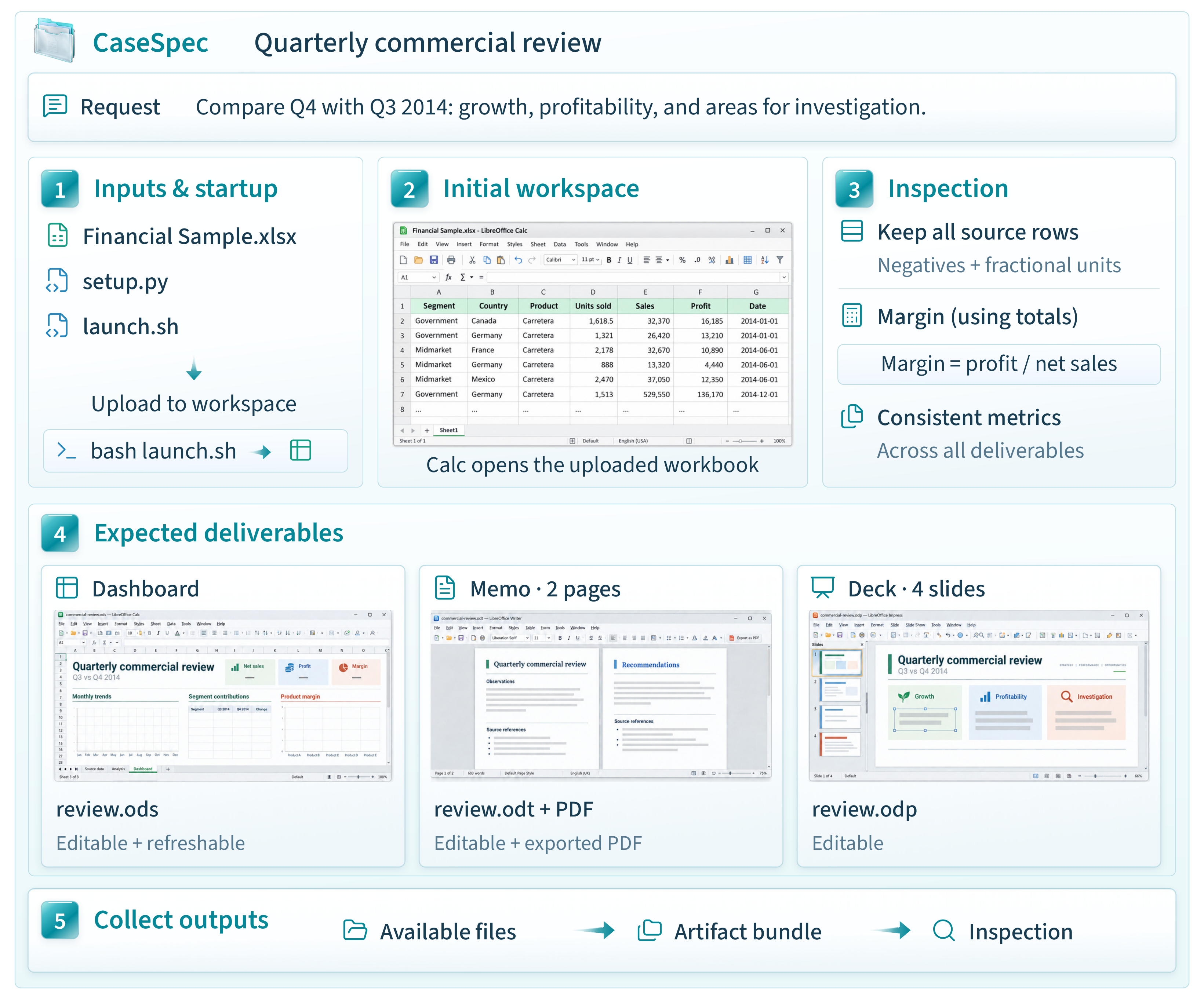}

\caption{\textbf{A concrete executable task contract.}
The saved commercial-review CaseSpec binds a request, input files, startup,
and output collection. The initial-workspace illustration redraws
representative source records in Calc. Requested deliverables comprise a
refreshable dashboard, an editable two-page memo with PDF export, and an
editable four-slide review deck. Inspection requirements summarize the
query, while post-run collection gathers available files for assessment.
GUI views, requested output layouts, abbreviated filenames, and the launch
command are illustrative.}
\label{fig:casespec}
\end{figure}

We package the output of each route in the same \casespec{} contract
introduced in Section~\ref{sec:overview}. Figure~\ref{fig:casespec} shows
one saved commercial-review task: its request asks for a quarterly comparison,
its workbook and startup instructions establish the initial Calc workspace,
and its requirements specify the deliverables and how they should agree.

\begingroup\interlinepenalty=10000
The dashboard, memo, and review deck in Figure~\ref{fig:casespec} are outputs the agent is asked to
create. The post-run hook gathers available files for inspection against
those requirements. This example shows what all three routes need to
establish, even though they start from different resources.
\par\endgroup

\subsubsection{Scaling Intents}
\label{sec:scaleintents}
Applications and working materials are readily available from software
templates, public file collections, and websites. However, these resources
do not come with a varied set of natural user requests grounded in their
contents and capabilities. We design Scaling Intents to generate such requests
from the operations the software supports and the content already present.
This grounding gives the requested work concrete subjects and feasible
actions, while variation in goals and phrasing broadens the collection.
We first describe desktop base tasks, then extend the construction to
dependent long-horizon workflows and web content.

\begin{figure}[!t]\centering
\includegraphics[width=\linewidth]{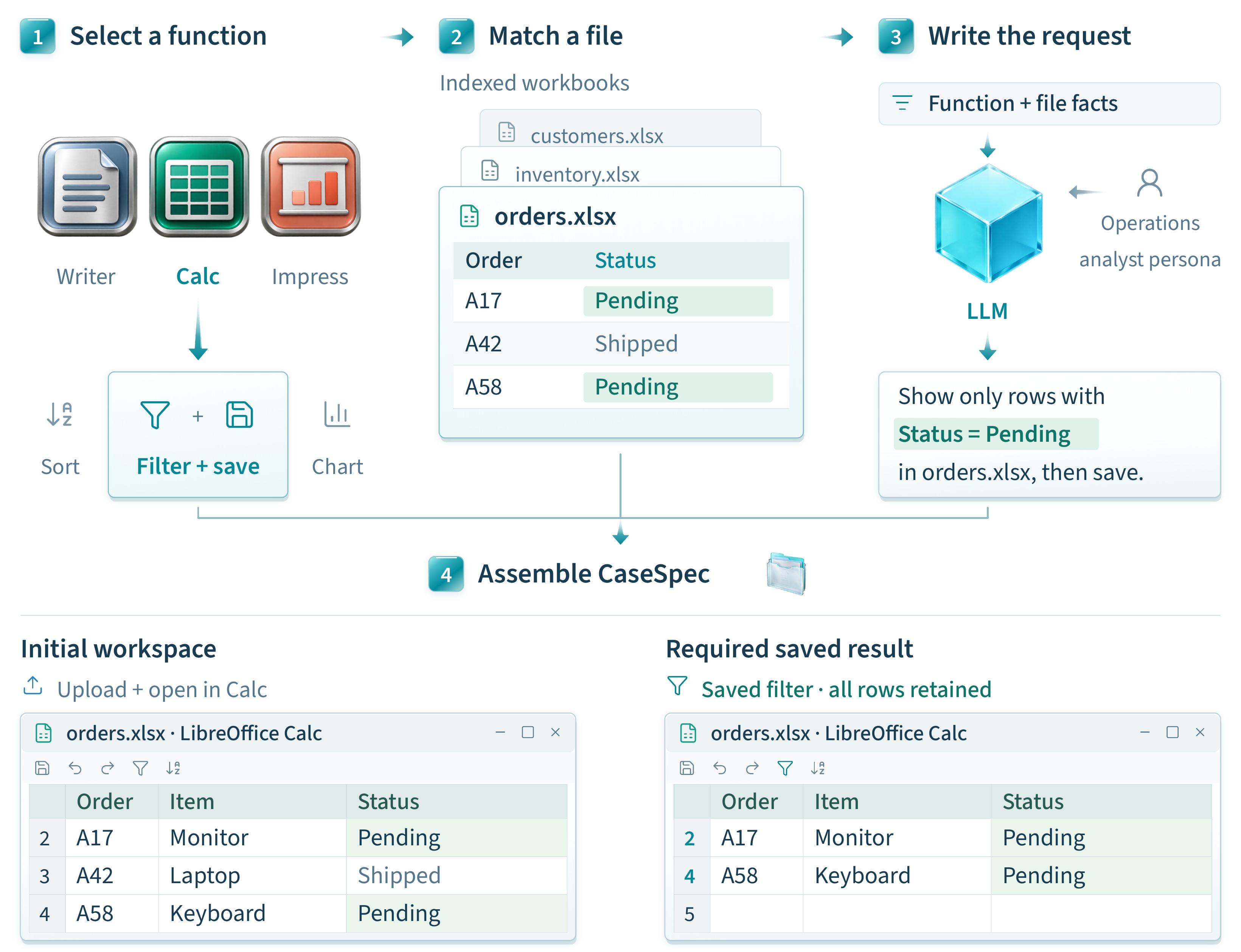}
\caption{\textbf{Constructing a desktop base task.}
The selected Calc function narrows compatible workbooks.
The \texttt{orders.xlsx} summary grounds the \texttt{Status = Pending}
condition, while the persona guides request wording.
CaseSpec binds this request and file to startup instructions and the
required saved filter. The required view retains A17 and A58 and hides A42
without deleting source rows. Software and GUI assets, records, and setup
are illustrative. The right-hand view specifies a requirement, not an
observed rollout.}
\label{fig:intentconstruction}
\end{figure}

\paragraph{Desktop base tasks.}
For a desktop base task, we need to connect a supported software operation
to material on which it makes sense. Figure~\ref{fig:intentconstruction}
illustrates the three steps used to generate the request. We first choose
an application and one of its supported functions. That choice establishes
the requirements for selecting a compatible source file. The selected
file's content then supplies the facts from which an LLM writes a grounded
request. Once the request is defined, we prepare its starting workspace
and assessment to form the complete task shown in the lower panel.

\smallskip\noindent\textbf{Select a function.}
We begin with E2B templates that provide installed applications and their
dependencies for office work, media, file management, browsing, and system
settings. Each application has a profile of supported operations and
readable file formats. We choose the target application and an eligible
operation according to the desired coverage of software and functions.
Selecting Calc and filtering, for example, determines both what the eventual
request will ask for and what kind of input it needs.

\smallskip\noindent\textbf{Match a file.}
To supply these inputs, we collect tens of thousands of files from public
and open resources and supplement them with agent-generated materials
\citep{ge2026syntheticcomputers}. We index their formats and extract bounded
content summaries, including spreadsheet fields and sample values or
document text excerpts. With the function selected, we use its format and
material requirements to find usable candidates. A spreadsheet-filtering
task needs a readable workbook whose fields support a meaningful selection
condition. Sampling favors files used less often within the construction
run, and the selected files' summaries become inputs to request generation.
For settings tasks, the relevant material is the application preference
state instead of a source file.

\smallskip\noindent\textbf{Write the request.}
We give an LLM the selected function and file summaries, together with a
persona describing the user's role and motivation \citep{jung2026procuasft}.
The function fixes the operation, and the summaries identify the concrete
subjects, fields, and values to which it can be applied.
For example, a workbook summary that identifies a \texttt{Status} column
containing \texttt{Pending} supports a request such as
\emph{``Filter the open spreadsheet to show only the pending records and
save it.''} The filtering condition comes from the workbook, and the
persona guides how the user expresses the need.

\smallskip\noindent\textbf{Assemble \casespec{}.}
The request now describes feasible work, but the executing agent still
needs the corresponding files and application state. To establish this
starting point, we specify where the harness should place
the selected files in the sandbox, which software template and dependencies
to use, and which application or document to open before the first action.
In the example, opening the workbook in Calc makes ``the open spreadsheet''
an unambiguous reference. Tasks involving several files instead identify
them by name or workspace-relative path.

The same requested change determines the completion criteria. The workbook
should be saved with the pending-record filter applied and all source rows
retained.
Figure~\ref{fig:intentconstruction} uses matched views of the initial workbook
and the required saved filter, which hides nonmatching rows without deleting them.
The request, workbook, startup instructions, and required result form one
\casespec{}. The same selection therefore
supplies the content of the request, the file the agent operates on, and
the result to inspect.

\paragraph{Desktop long-horizon tasks.}
Base-task generation connects a function to suitable content. Long-horizon
construction extends these local matches into a coherent workflow, especially
across applications, by defining how later stages use information or
results produced earlier
\citep{yuan2026osworld2,ge2026syntheticcomputers}.

\begin{figure}[!t]\centering
\includegraphics[width=\linewidth]{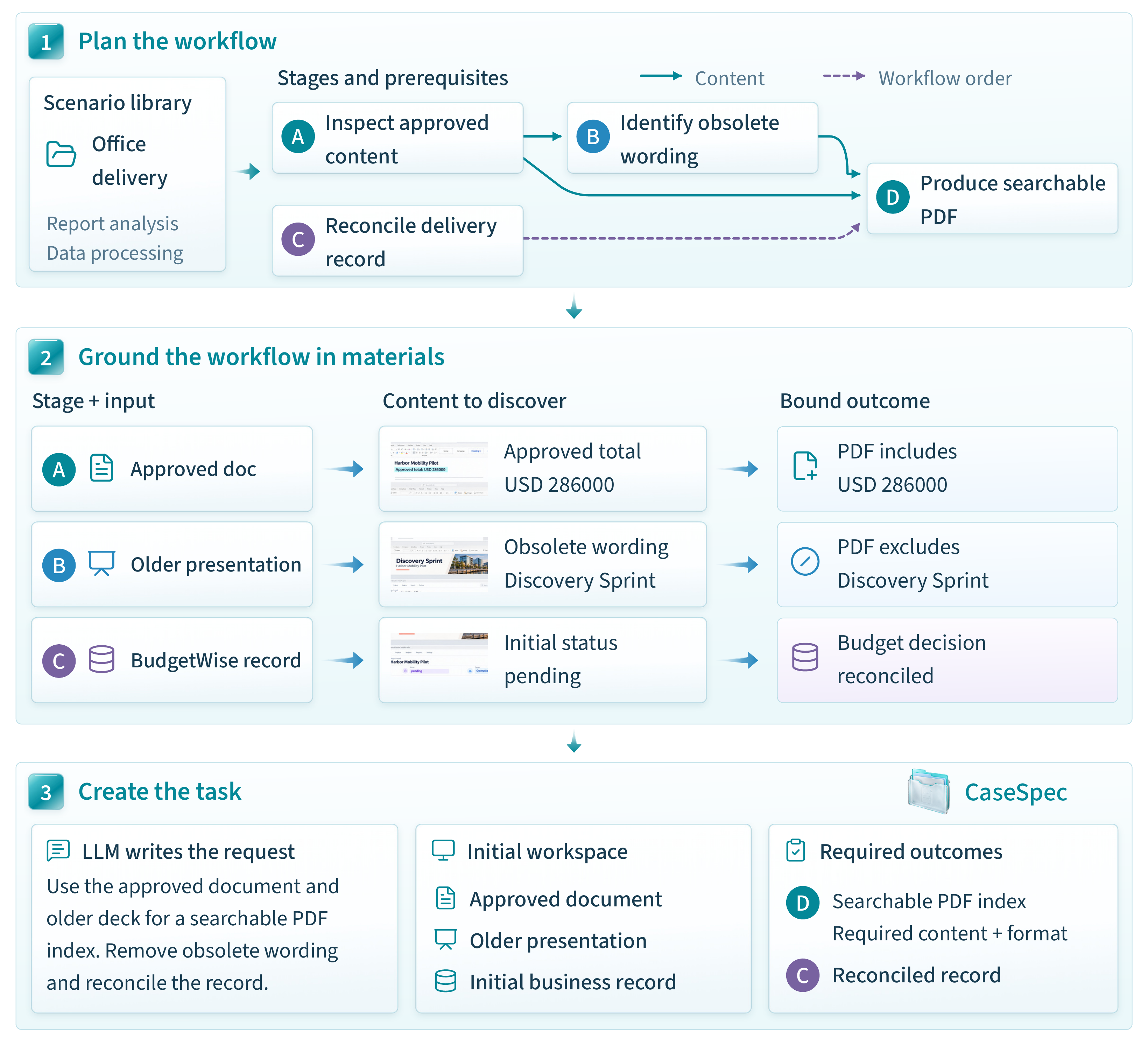}
\caption{\textbf{Constructing long-horizon tasks from blueprints.}
(1) Blueprint selection assembles work stages and prerequisites. Solid edges
carry content dependencies. The dashed C-to-D edge sets workflow order.
A node can require several GUI actions. (2) Materials bind the approved amount and
obsolete wording to inclusion and exclusion conditions for D's PDF. C binds
an initial record to the required budget-decision update. (3) The grounded
plan guides LLM request writing and supplies the workspace and outcome
requirements in one CaseSpec. Its two outputs are the PDF and updated record.
Discovery facts remain in the prepared files. The structure follows a saved
plan. GUI details, the shortened request, and prospective outcomes are
illustrative.}
\label{fig:longhorizonconstruction}
\end{figure}

\begingroup\interlinepenalty=10000
We therefore plan the workflow before choosing its concrete materials and
writing the request. The procedure in Figure~\ref{fig:longhorizonconstruction}
has three steps. First, we assemble work stages into a dependency graph so
that they contribute to a shared goal. We then prepare the files and records
that make each stage executable and its required outcome specific. Finally,
we use this grounded plan to write the request and package the starting
workspace and outcome requirements. The resulting task asks for a coherent
workflow whose operations are connected by a shared purpose and explicit
prerequisites.
\par\endgroup

\Needspace{5\baselineskip}
\smallskip
\noindent\textbf{Plan the workflow.}
We encode recurring workflows as \emph{scenario blueprints},
each defining a final deliverable, candidate stages, their applications and
materials, and their prerequisites. For the office scenario, a source review,
a delivery-record update, and a final document all contribute to preparing
the same delivery package.\par

\begin{samepage}
For each instance, target task length, application coverage, and desired
interaction challenges guide the choice of a blueprint and a compatible
stage combination. We favor less-used scenarios and components, then use
seeded sampling among eligible combinations to vary the work. A combination
is retained when its prerequisites are present and every selected stage
contributes, through its dependencies, to the overall workflow goal. Operation
variants further specify the requested result, such as requiring the PDF's
text to remain searchable.\par
\end{samepage}

We represent the selected workflow as a directed acyclic graph (DAG),
a planning abstraction also used for cross-device orchestration in UFO3
\citep{zhang2025ufo3}. Here the graph specifies the task-construction plan.
Each node is a work stage, such as inspecting a document for approved
content, and may require several GUI actions. Edges can express use of earlier
information or artifacts, or a required execution order. We retain the
blueprint's dependencies and, when constructing longer workflows, add
precedence between selected stages that use shared facts.\par

Figure~\ref{fig:longhorizonconstruction}(1) shows an office-delivery example
of this planning process. The shared goal is to produce a current delivery
package and reconcile its business record. We organize that work into four
stages. Source inspection (A) establishes the approved content, which guides
the search for obsolete wording in an older deck (B). Record reconciliation
(C) can begin independently. The final PDF stage (D) uses the findings from
A and B. The blueprint also places the record update before PDF production,
shown by the dashed C-to-D edge. This edge specifies workflow order, while
the updated record remains a separate deliverable with its own check.

Before preparing the materials, we check the assembled plan for cycles,
missing prerequisites, and inconsistent production or use of files.
The checked plan associates each stage with its application, required
inputs, and expected output, so material preparation can follow the
dependencies of the workflow.\par

\Needspace{5\baselineskip}
\smallskip
\noindent\textbf{Ground the workflow in materials.}
The DAG specifies what each stage should accomplish, but its nodes still
need concrete information to operate on. We ground each node by selecting
or generating the files and service records required by that subtask.
Existing-file selection follows the same principles as base-task selection,
matching application support, file format, and content. Long-horizon
construction also considers whether the selected files belong to a
shared project or directory and whether their contents support the
relationships between stages. File summaries supply the names, values,
and other details used to specify stage goals and acceptance conditions.

When the workflow calls for generated materials, we instead choose scenario
facts and place them in the files or records the agent will inspect. The
office sample follows this path in Figure~\ref{fig:longhorizonconstruction}(2).
The approved document supplies a value that the final PDF should include,
while the older deck contains obsolete wording that the PDF should exclude.
The business record supplies the initial state for a reconciliation update.
Each material therefore has a role in both the work stage and its outcome
condition. The example requires two results: the PDF and the updated record.

We check that the required information is available through the selected
applications and that generated files contain their assigned content.
Linking each requirement to its source and its later use keeps the
workflow grounded across stages. The output is a task plan with concrete
input files, specified initial record values, stage goals, and corresponding
outcome criteria.

\Needspace{5\baselineskip}
\smallskip
\noindent\textbf{Create the executable task.}
With the work and its materials fixed, we use an LLM to write the user
request. It receives the user-visible goal, references to the available
materials, and required constraints, together with a selected persona
and expression style. These choices vary how users describe their work
while preserving the constructed task. Facts designated for discovery
remain in the workspace for the executing agent to find. The request can
ask for a reconciled delivery record and a searchable PDF index without
revealing the values or obsolete terms the agent needs to discover.
Query checks compare the wording with the task plan, return inconsistencies
for revision, and filter semantically duplicate requests.

Figure~\ref{fig:longhorizonconstruction}(3) brings the request-writing,
workspace, and assessment branches of the same grounded plan together.
As in base-task construction, initialization instructions place the files
and open the required applications. Here they also prepare the business
record used by the workflow. Outcome assessment checks the requested PDF,
record changes, and source preservation. Together with the request, these
components form the \casespec{} executed by the harness. The DAG organizes
task construction, while the executing agent plans its own GUI actions.

The shared task plan keeps the goal, workspace, and
assessment aligned as the collection expands to longer, multi-application
workflows.

\paragraph{Web tasks.}
An existing website supplies records and pages that can be read or changed
through interface controls. We generate requests from this material in
three steps. Examining the content and controls first establishes what work
the site supports. We then select a goal involving that content and ask an
LLM to express it as a user request. The chosen goal determines the initial
records to prepare and the outcome to assess, completing the web task.

We first identify the information visible on the relevant pages and the
controls used to access or modify it. On an illustrative task-management
site, these include task names, due dates, and status values, together
with controls for filtering the task list, opening a task, and changing
its status. This combination supports requests to find tasks with
particular properties and apply a corresponding update.

Using this account of the site, we select conditions that match records in
the starting data and an operation available through the interface. We then
use an LLM to express the grounded goal as a user request, for example:
\emph{``Mark tasks due before 1 September 2026 that are still In progress
as Needs follow-up.''} The due dates and current statuses determine which
tasks qualify, and the status control provides the requested operation.
Other requests use different items, conditions, and available actions
to cover information retrieval, comparison, and form filling.

The generated request determines which initial state to prepare and which
result to assess. We record its entry page, startup instructions to open
that page in the browser, and any setup needed to make the selected records
available.
In the example, the starting records establish which task identifiers
satisfy the date and status conditions. Outcome assessment checks the
updated status of those same tasks after execution. Keeping their
identities tied to the starting data lets the check follow the requested
items even after their status changes.

Across desktop and web, these constructions turn available software and
content into user requests with defined starting states and outcomes to
assess. The resulting \casespec binds the goal and requirements to the
selected resources and initialization instructions.
The next route starts with an existing request or reference website and
builds the workspace needed to support it.

\subsubsection{Scaling Environments}
\label{sec:scaleenvironments}

Scaling Intents uses the workspaces already available to generate new
requests. Sourced requests offer a complementary resource: they describe
useful work, but often omit the files, dependencies, or application state
needed to execute it. Reference websites and software issues likewise
describe interactions without necessarily providing an isolated environment
for collection. We design Scaling Environments to build these missing
execution conditions. For desktop requests, this means preparing the files
and applications assumed by the user. For Web tasks, we collect open-source
frontends and reconstruct workflows from public URLs to create local
environments that we can initialize and reuse. We also develop repository-based frontend testing as
a distinct web-agent application, where a runnable project lets browser
interaction test a candidate code repair.

\paragraph{Desktop environments.}
Forums contain user-reported software problems, while tutorials and help
resources describe practical workflows and their intended results
\citep{wang2024large}. These sources provide goals grounded in software use,
but their descriptions often assume attachments or workspace state that
the collector does not have. We collect tens of thousands of such requests
and task seeds from forums, wikiHow, and help resources. The construction
problem is then to supply the inputs and startup conditions each seed
assumes while preserving the work it asks the agent to do.

\begin{figure}[!t]
\centering
\includegraphics[width=\linewidth]{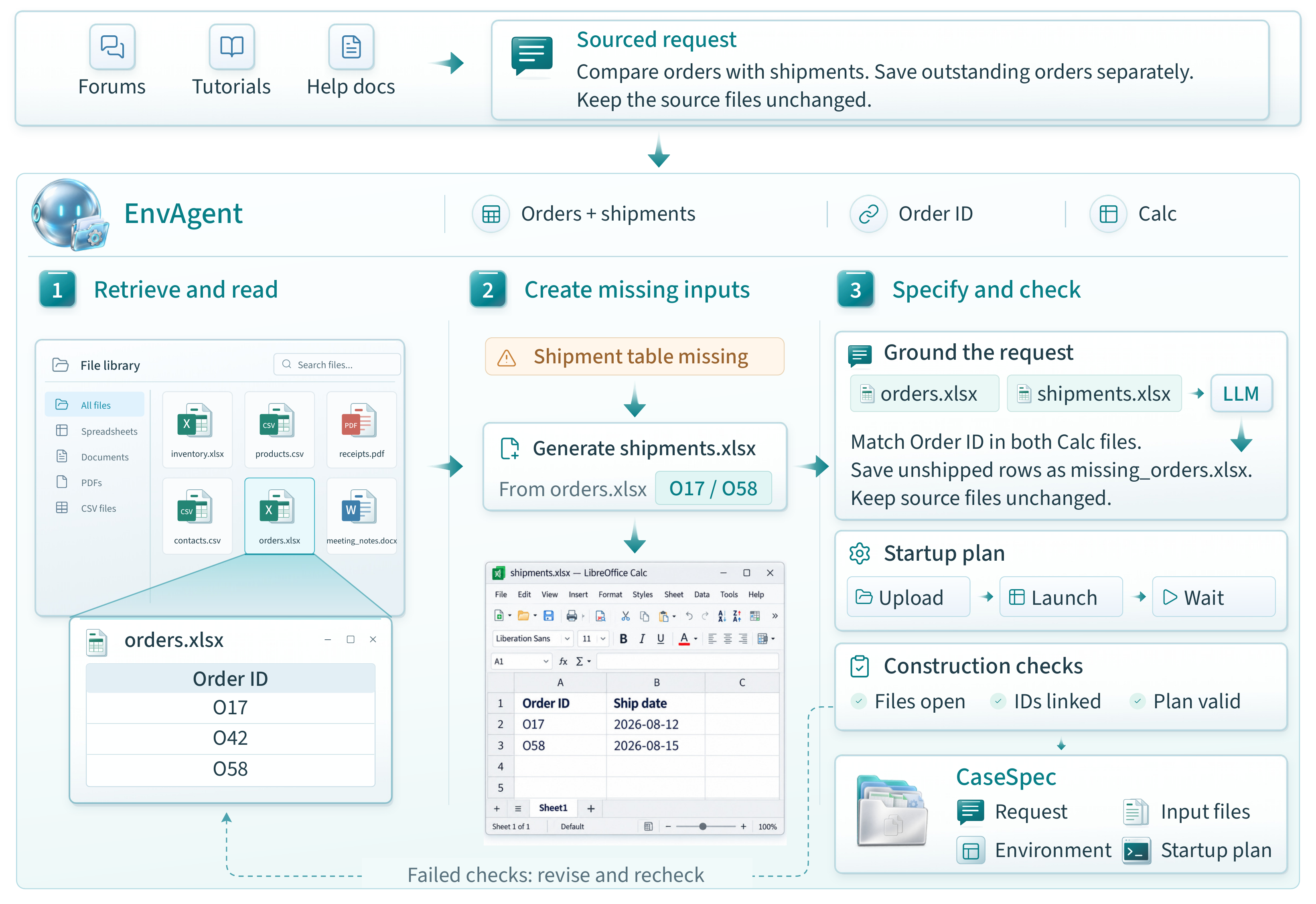}
\caption{\textbf{Request-driven desktop environment construction.}
EnvAgent identifies the required resources, retrieves an order table, and
uses its identifiers to construct a missing shipment table. An LLM grounds
the sourced request in the prepared filenames and their shared Order ID
field. EnvAgent specifies startup and checks the materials and preparation
plan, revising inputs or instructions when checks fail before producing a
\casespec{}.
The GUI agent subsequently performs the requested comparison.
GUI views and records are illustrative.}
\label{fig:desktopenvironment}
\end{figure}

We therefore design an environment-building agent, EnvAgent, to interpret
each request and prepare its input files and startup plan.
Figure~\ref{fig:desktopenvironment} illustrates three construction steps.
The agent first retrieves and reads candidate files to establish which
requirements are already supported. Any missing inputs then guide file
generation, using the retrieved material to keep related records consistent.
With the inputs available, the agent specifies the request and startup
plan and checks that they agree. A failed check returns the affected input
or instruction for revision before the package becomes a \casespec{}.
The order-comparison example below shows how the retrieved and generated
materials contribute to the same task.

\Needspace{5\baselineskip}
\smallskip
\noindent\textbf{Prepare the input materials.}
EnvAgent uses the request and selected software template to identify the
applications and inputs the task needs. Comparing orders and shipments,
for example, requires two tables that refer to the same orders, with a
shared identifier through which their records can be matched. A request
restricted to a particular month also requires the corresponding date
information. We reuse the indexed file formats and content summaries from
Scaling Intents to locate candidates, using project and date information
to find related tables. EnvAgent then reads their records to confirm that
the identifiers refer to the same orders and that the files contain
both shipped and outstanding orders for the comparison. If the retrieved
materials meet the input requirements, the agent proceeds to request and
startup specification. Otherwise, the unmet requirements guide generation.

Figure~\ref{fig:desktopenvironment}(2) shows an example of this step.
Given a suitable order table but no matching shipment table, EnvAgent uses
file tools and reusable construction
skills to create a shipment table with matching identifiers and business
context. The resulting input set gives the requested comparison concrete
files to operate on.\par

\Needspace{5\baselineskip}
\smallskip
\noindent\textbf{Specify and check the task environment.}
The files establish which names and fields the request can refer to.
We use the original request and these file details to guide an LLM's
replacement of general references with concrete ones, preserving the goal
and constraints. In the example, the request now names the two workbooks
and their shared \texttt{Order ID} field. It asks for unmatched orders in
a separate file and preserves any original source-protection or time
restriction.
The GUI agent still needs to identify the outstanding orders and create
the result file.

The refined request also determines how execution should begin.
EnvAgent specifies that both inputs be placed in a project directory,
Calc be launched, and the order table be opened. The startup plan waits
for the application window to be ready, with the shipment table available
alongside the order table. These instructions give the harness
the sequence needed to present the files and application assumed by
the request.

Before handoff, EnvAgent checks the prepared materials and startup plan.
File tools establish whether the inputs exist and have valid formats,
while the agent reviews whether their contents support the requested comparison,
whether the template provides the required applications, and whether
startup actions place the files before opening them. Findings such as a
missing identifier field or an incomplete startup action lead to revisions
of the affected material or instruction. After a file changes, EnvAgent
submits a new inspection report before finalization, so the report
describes the materials that will actually be delivered.

The checked request, files, environment configuration, and startup plan
form the \casespec{} from which the harness creates the E2B instance and
establishes the starting application state. With the input files in place,
the GUI agent performs the comparison and saves \path{missing_orders.xlsx},
whose contents are then checked against the required orders alongside
preservation of the source files. Construction checks thus establish the
starting conditions, while outcome assessment evaluates the work performed
from those conditions.

\paragraph{Web environments.}
We collect Web frontends from open-source websites as starting points for
local task environments. To extend this pool, we also reconstruct selected
workflows from public reference URLs \citep{chae2026safescalablewebagent}.
These references provide realistic pages and interactions, while local
control lets us initialize records, repeat attempts, and construct multiple
tasks within the same site.

\begin{figure}[!t]
\centering
\includegraphics[width=\linewidth]{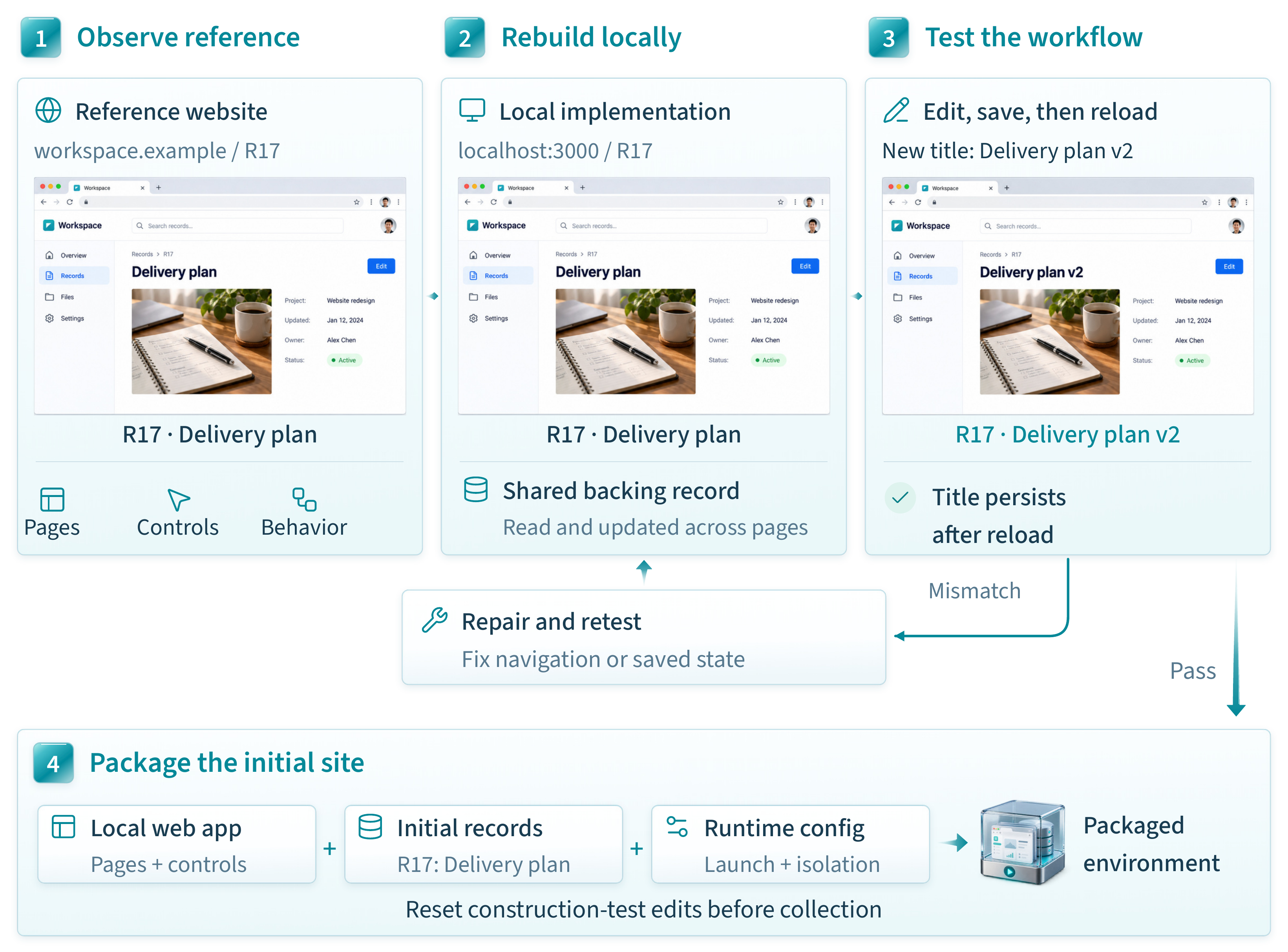}
\caption{\textbf{Reconstructing interactive websites from references.}
The aligned GUI views follow record R17 from the reference site to the local
implementation and a reload test. Observed pages and interactions guide the
interface and shared backing record. The test edits and saves the title, then
reloads the record to check persistence. Discrepancies return to implementation
for repair and retesting. The checked site is packaged with its declared initial
records and runtime configuration, separately from construction-test edits.
GUI states and URLs are illustrative.}
\label{fig:webenvironment}
\end{figure}

Our URL-to-environment construction builds on WebAnything
\citep{xie2026webanything} and follows the four steps in
Figure~\ref{fig:webenvironment}.
We first explore the reference to identify the required pages and behavior.
Those observations guide construction of a local interface connected to backing state.
Once it runs, browser tests check the interactions and return discrepancies
for repair. We then restore the intended starting data and package the
checked site for task collection. These steps turn an external example
into an interactive environment with controlled initial conditions.

The builder begins at the reference URL and explores the workflows selected
for reconstruction, retrieving relevant pages and accessible frontend
assets. For record editing, this covers the list page, detail view, editing
controls, and navigation between them. An existing request can further
focus the exploration. The resulting observations establish which pages
the local site needs and what each interaction should do.\par

We use these observations and accessible frontend assets to build the local
pages. Their interactions also specify the backend behavior to reproduce.
When the reference's backing data or implementation is unavailable, the
builder creates a local schema, initial records, and update logic for the
observed workflow. In the record-editing example, the list and detail views
refer to the same record, and saving an edit updates that record for
subsequent visits. Linking the reconstructed controls to these state changes
makes the workflow available for actual browser interaction. If construction
raises questions about a page or control, the builder returns to the
reference site for targeted inspection before continuing the implementation.

Once the local workflow is runnable, a browser testing agent checks its
visual correspondence with the reference and the effects of its controls.
For the illustrated check, it opens a record, edits and saves it,
and refreshes the page to inspect the stored value. An unreachable detail
page leads to a navigation repair, while a lost update leads to a correction
of the backend state change. After revision, the affected interactions are
tested again. Figure~\ref{fig:webenvironment}(3) shows a persistence test
on record \texttt{R17}. The agent changes \emph{Delivery plan} to
\emph{Delivery plan v2}, saves, and reloads. Seeing the edited title after
reload provides the evidence for this particular check.

\begingroup\interlinepenalty=10000
These tests modify the records that future tasks will use. Before collection,
we therefore restore the declared initial records and entry-page state,
keeping construction-test edits separate from task setup. We package this
starting state with the checked site for isolated E2B execution. Its content and tested functionality
also support Scaling Intents, allowing the same environment to supply
multiple user tasks.
\par\endgroup

\Needspace{5\baselineskip}
\paragraph{Repository-based frontend testing.}
Frontend testing is an important application of web agents because code
that builds successfully can still behave incorrectly when a user
interacts with it \citep{yang2024swebenchmultimodal}. We therefore treat
issue-driven testing as a distinct setting for environment construction and
experience collection. A repository supplies the application, and its issue
report specifies behavior that browser interaction can test.

\begin{figure}[!t]
\centering
\includegraphics[width=\linewidth]{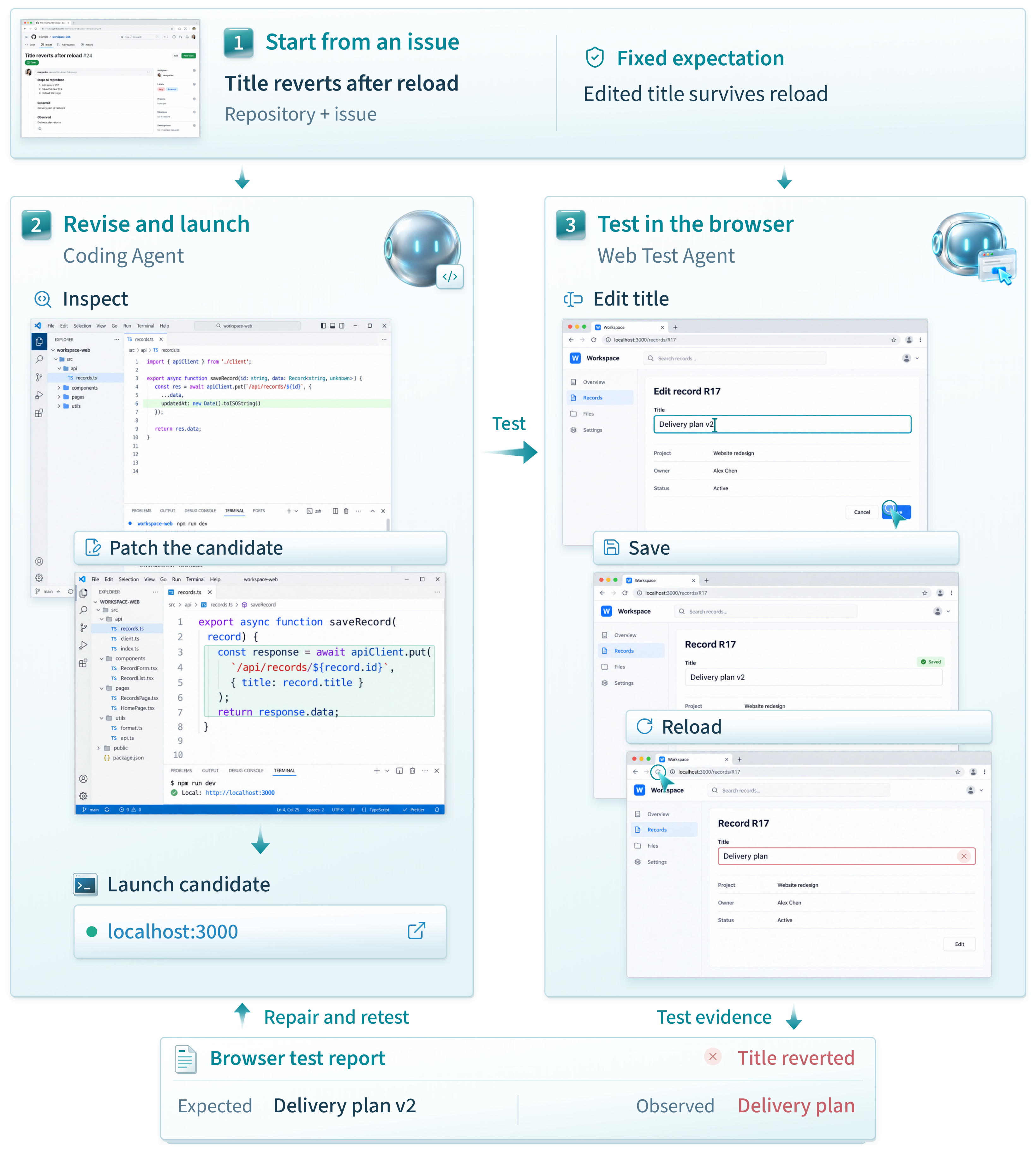}
\caption{\textbf{Browser interaction guides iterative code repair.}
(1) The issue fixes the expected behavior. (2) The Coding Agent inspects and
edits the relevant implementation, then launches a candidate application.
(3) The Web Test Agent edits R17, saves the new title, and reloads the page.
The stacked views show the new title after saving and the old title after
reload, with cursor overlays marking interactions. A test report returns this
discrepancy, recorded actions, and screenshots to the Coding Agent for repair
and retesting under the unchanged expectation. Generated GUI views, code,
and test feedback are illustrative.}
\label{fig:repositoryenvironment}
\end{figure}

Figure~\ref{fig:repositoryenvironment} shows how browser testing guides
successive code revisions.
We first prepare a runnable project and turn the issue into a fixed test
expectation. A coding agent then revises and launches the application,
making a candidate repair available to the web agent. The web agent
replays the reported interactions and compares their effects with the
expectation. Finally, its action trace, screenshots, and verdict return
to the coding agent in a test report. This evidence directs the next
repair while keeping every iteration focused on the same issue.

\begingroup\interlinepenalty=10000
Environment preparation begins with GitHub frontend repositories whose
issues describe browser-reproducible behavior. The builder installs
dependencies, starts the project, and prepares the pages and data involved
in the issue. This makes the reported problem available for direct
interaction and supplies the workspace in which repairs will be tested.\par
\endgroup

The issue then defines what to check in this workspace. Its expected
behavior supplies the coding agent's repair objective and a test list of
browser actions and expected results for the web agent. For example, an
issue may report that an edited page title appears to save but disappears
after reload. The corresponding test changes the title, saves it, reloads
the page, and checks that the new title remains. Both the code change and
its assessment therefore address the same user-reported problem.\par

The coding agent inspects the relevant implementation using the issue and
any feedback from the previous browser test. It then changes the code and
launches the revised application. These actions, illustrated by the editor
views in Figure~\ref{fig:repositoryenvironment}(2), supply a runnable
candidate whose behavior still needs to be tested.

The web agent performs the test actions against that candidate and captures
the resulting states. A visible save confirmation establishes what the
interface showed immediately after saving. Reloading tests whether the
change persists. In the illustrated failure, the edited title appears
after save but the old title returns after reload.

\begin{figure}[!t]
\centering
\includegraphics[width=\linewidth]{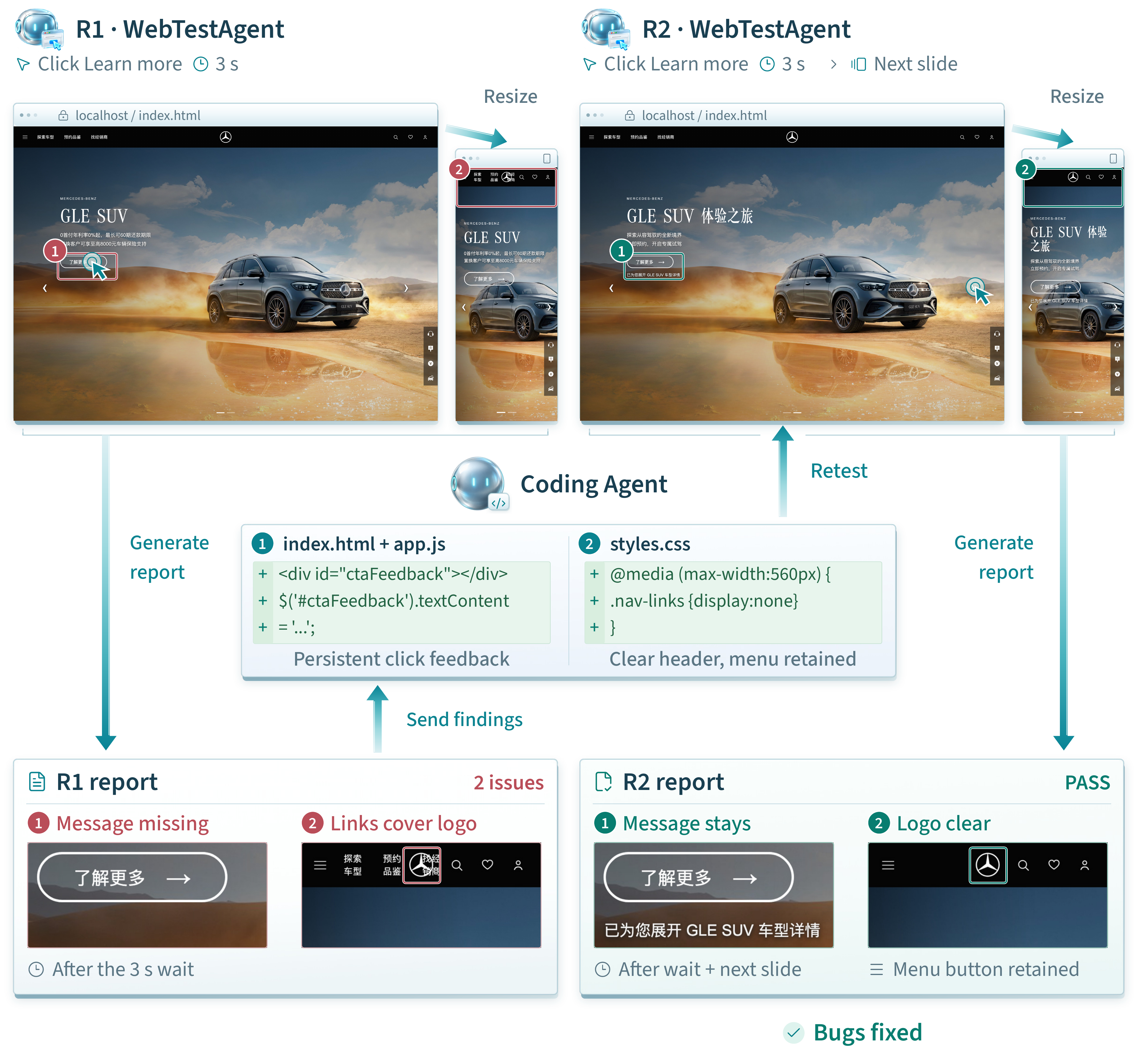}
\caption{\textbf{Bug-guided repair and targeted retesting.}
Desktop and narrow-screen tests produce one report, which guides both code
changes before retesting. Screenshots are later replays of archived code,
with subtitle line breaks cleaned for presentation. Overlays, report excerpts,
and abbreviated patches are explanatory. R2 passed three selected checks, including homepage
appearance, without repeating all five R1 items.}
\label{fig:webtestcase}
\end{figure}

The browser test report links these states to the actions that produced
them and states the discrepancy directly. The stacked views in
Figure~\ref{fig:repositoryenvironment}(3) follow the same R17 record through
editing, saving, and reloading. Because the edited title should persist,
the observed old title identifies a discrepancy for the coding agent to
investigate. This feedback guides its next revision, which the web agent
tests again through the running interface.
The issue's expected behavior remains fixed across iterations. This loop supplies GUI
interaction trajectories and outcome feedback tied to real development
requests, with evidence qualification and final acceptance detailed in
Appendix~\ref{sec:webscenario}.
\paragraph{Case study of a responsive homepage repair.}
Figure~\ref{fig:webtestcase} follows a local Mercedes-Benz GLE SUV homepage
built from supplied materials. The request called for a responsive page
with usable controls and navigation. Two defects emerged when the
WebTestAgent followed a button click over time and inspected the page
at a narrow width.

In the first browser test (R1), the WebTestAgent clicked \emph{Learn more}
and waited. The transient notification disappeared, leaving no stable
confirmation. Resizing the page to 320 px then exposed a second defect,
with text navigation covering the centered logo. The R1 report on the
left links both findings to the actions and screenshots that exposed them,
giving the Coding Agent two specific behaviors to repair.

The central patch excerpts show how this feedback changed the code.
For the first finding, the Coding Agent added an HTML confirmation element
below the button and updated its text through the JavaScript click handler,
so the response could remain visible. For the second, a CSS rule hid
the text links at widths up to 560 px while retaining the hamburger menu.
This freed space around the logo while preserving navigation access.
The matching numbers connect each reported defect to its corresponding patch.

With the revised page running, the Coding Agent requested a second browser
test (R2). The WebTestAgent clicked the button again, waited three seconds,
and changed the carousel slide. The inline confirmation remained visible.
At 320 px, the links no longer covered the logo, the menu opened and closed,
and the page had no horizontal overflow. Together with a homepage visual
check, these observations supported PASS on three selected items from
R1's five-item checklist. The two reported defects were resolved in this
targeted retest.

Together, these routes expand environment supply from user requests,
reference websites, and runnable software projects. Their prepared
workspaces support experience collection, and reconstructed sites also feed
new task generation through Scaling Intents. Tasks equipped with reliable
executable verifiers offer a further route to expansion: the next section
uses their goals and outcome checks as seeds for additional task construction.

\subsubsection{Scaling Verifiers}
\label{sec:verifiers}

Task datasets and manually authored cases provide user goals already paired
with executable verifiers. A reliable verifier
makes the intended outcome explicit and supplies a repeatable check on
collected trajectories. Creating additional checks is costly, however,
because each needs to match its task and distinguish correct results from
plausible mistakes. We design Scaling Verifiers to expand this resource by
adapting existing task--verifier pairs \citep{wang2026cuagym,lv2026scalecua}.

\begin{figure}[!t]
\centering
\includegraphics[width=\linewidth]{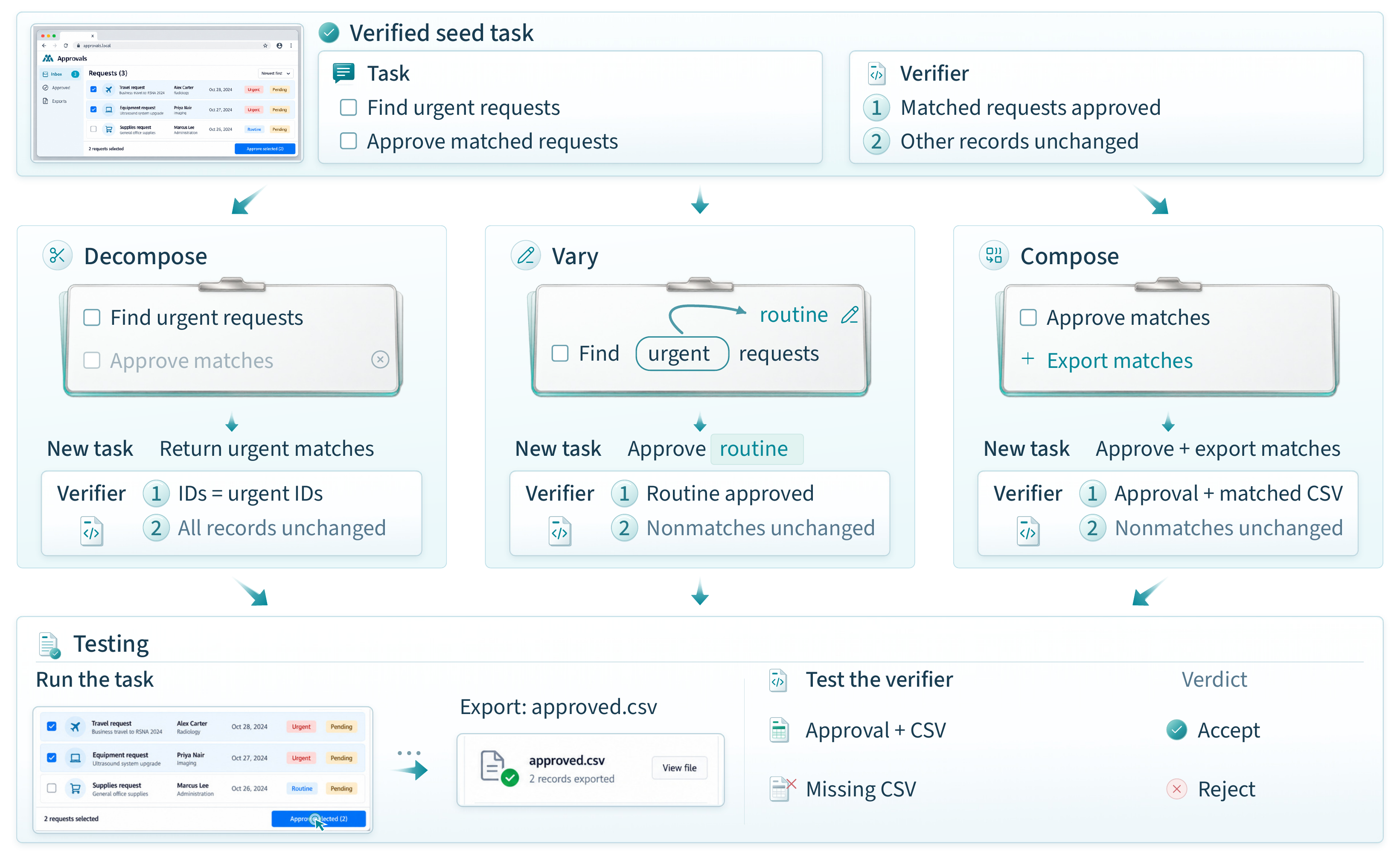}

\caption{\textbf{Expanding tasks and their verifiers together.}
The agent extracts a subtask, changes a selection criterion, or joins approval
with export, adapting completion and preservation checks to the new task.
The lower composition example separates a GUI execution trial from verifier
tests on prepared outcomes. The verdict column gives the requirement-derived
response for each probe. Testing checks whether the verifier returns it.
The task lists, generated GUI views, and hypothetical judgments illustrate
the construction procedure. The ellipsis omits intervening GUI actions, including
export. Initial-state preparation and repair iterations are not shown.}
\label{fig:tasks}
\end{figure}

Figure~\ref{fig:tasks} illustrates the construction and validation phases.
We first use a seed's goal and success conditions to construct related
tasks, changing the request, initial state, and verifier together. The new
pair then undergoes a GUI execution trial and tests of the verifier's
judgments. Their findings guide repairs before the pair enters collection.

This joint treatment matters because changing a request's selection
criterion also changes which records the verifier should accept as updated
and which it should require to remain intact.

\paragraph{Expanding the work and its success conditions.}
The construction agent reads the seed request, initialization, and verifier
to identify useful subgoals, their prerequisites, and the observable changes
that establish success. A subgoal can involve several GUI actions, such as
editing a document and exporting the result. We use these subgoals in
three different ways. Decomposition extracts a meaningful part of the seed
task. Variation changes the condition or operation while retaining its
basic structure. Composition connects subgoals into a longer workflow,
whose later work uses earlier results. The approval example in
Figure~\ref{fig:tasks} shows the effect of each choice on the requested work
and the corresponding outcome checks.

\begin{itemize}
\item \begingroup\interlinepenalty=10000
\textbf{Decomposition} turns a meaningful part of the workflow into a
task of its own. Extracting record selection yields a request to identify
the qualifying records, an initial state containing the required data,
and a verifier for the returned selection and preservation of all records.
\par\endgroup
\item \textbf{Variation} changes the materials, parameters, or operations.
A new selection criterion leads to a new target set derived from the
corresponding input records. The request and verifier are updated to
express and assess that change.
\item \textbf{Composition} joins compatible subgoals through their inputs
and outputs. Adding a summary export to an approval task makes the approved
identifiers the contents of the later file. The adapted verifier checks the
updates, the exported summary, and preservation of the records the request
protects. The added dependency can produce a more demanding multi-stage task.
\end{itemize}

Across these changes, the agent resolves the actual identifiers and file
references used by the request, environment, and verifier. In the composition
example, the targets established during selection remain the targets during
assessment, even when the update changes a field used for selection.
Desktop compositions carry the same relationship through documents and
intermediate files. Requirements that matter during a stage are checked
there, while final-state requirements are assessed at completion.\par

\paragraph{Testing the task and its verifier.}
Each rewritten task undergoes two complementary checks before collection.
A GUI
trial examines whether an agent can carry out the request from the prepared
state. Verifier probes examine whether the checker judges known outcomes
as the task requires. The lower part of Figure~\ref{fig:tasks} keeps these
two sources of evidence separate.

For the execution trial, we run a CUA from the declared initial state with the rewritten
request. Its interaction and resulting state reveal problems such as
missing inputs, unreachable controls, or instructions inconsistent with
the workspace. These findings identify which setup or request needs repair.

To test the verifier itself, we construct outcome probes whose expected verdicts follow from the request
and starting data, then compare those verdicts with the verifier's outputs.
In the approval-and-export example, a complete result should pass, whereas
an otherwise correct update without the required export should fail.
Additional probes isolate incomplete updates, changes to protected records,
and allowed alternative outputs. A disagreement identifies a checker rule
to revise, even when the GUI task itself can be completed.

We collect trial failures and verifier disagreements into a construction
report. Setup and instruction failures return to task construction, while
incorrect verdicts return to the verifier. After the
revision, we rerun the affected tests against the same intended requirements
before admitting the pair to collection. Requirements that need contextual
judgment use the evaluator in Section~\ref{sec:evaluator}. This feedback loop
keeps task expansion connected to executable conditions and meaningful
outcome assessment.

The three construction routes supply tasks to the shared collection service
(Section~\ref{sec:batch}). Executing these tasks produces trajectories and
outcome assessments. Obtaining successful experience from the hardest
tasks introduces the next challenge.

\subsection{Recursive Self-Improvement for Hard-Task Collection}
\label{sec:hardexperience}

The routes in Section~\ref{sec:threeaxes} supply executable tasks whose
rollouts provide experience for learning. During collection, however, we
encounter long and difficult tasks that remain unsolved within the ordinary
budget. Collecting demonstrations of sustained planning and recovery from
these tasks requires addressing two recurring obstacles. Repeated attempts
can lose useful discoveries and repeat ineffective procedures. A late
failure can also force a new attempt to rebuild substantial earlier work.

We design two recursive self-improvement (RSI) pipelines for collecting
experience from these tasks. Each carries forward the information needed
to address one of these obstacles.
\begin{itemize}
\item \textbf{Knowledge accumulation.} Skill-guided recursive exploration
turns trial-and-error findings into tested procedures that guide
subsequent attempts (Figure~\ref{fig:skillexploration}).
\item \textbf{Progress accumulation.} Milestone-guided trajectory search
preserves completed work and resumes exploration from the corresponding
software state and agent context (Figure~\ref{fig:milestonesearch}).
\end{itemize}
Both use feedback from earlier trials to shape
later exploration \citep{zhu2026rsiagentautonomousexplorationrecursive}.
They target tasks that pass construction checks but remain unsuccessful
in ordinary collection. The original acceptance conditions stay fixed.
We check the completed task before admitting an assisted trajectory as a
full-task demonstration. Milestone-guided collection can also retain verified
subtask traces from attempts that leave the full task unfinished.

\subsubsection{Skill-Guided Recursive Exploration}
\label{sec:skill-exploration}

Some GUI tasks require a procedure that the agent has not discovered.
The difficulty may be an unfamiliar application command, a hidden
prerequisite, or a sequence whose useful effect becomes clear only after
trying it and inspecting the result. Repeated attempts can make partial
discoveries, but those discoveries remain scattered across their histories
unless the collector carries them forward. We therefore need a way to turn
trial-and-error experience into reusable operating guidance.

\begin{figure}[!t]
\centering
\includegraphics[width=\linewidth]{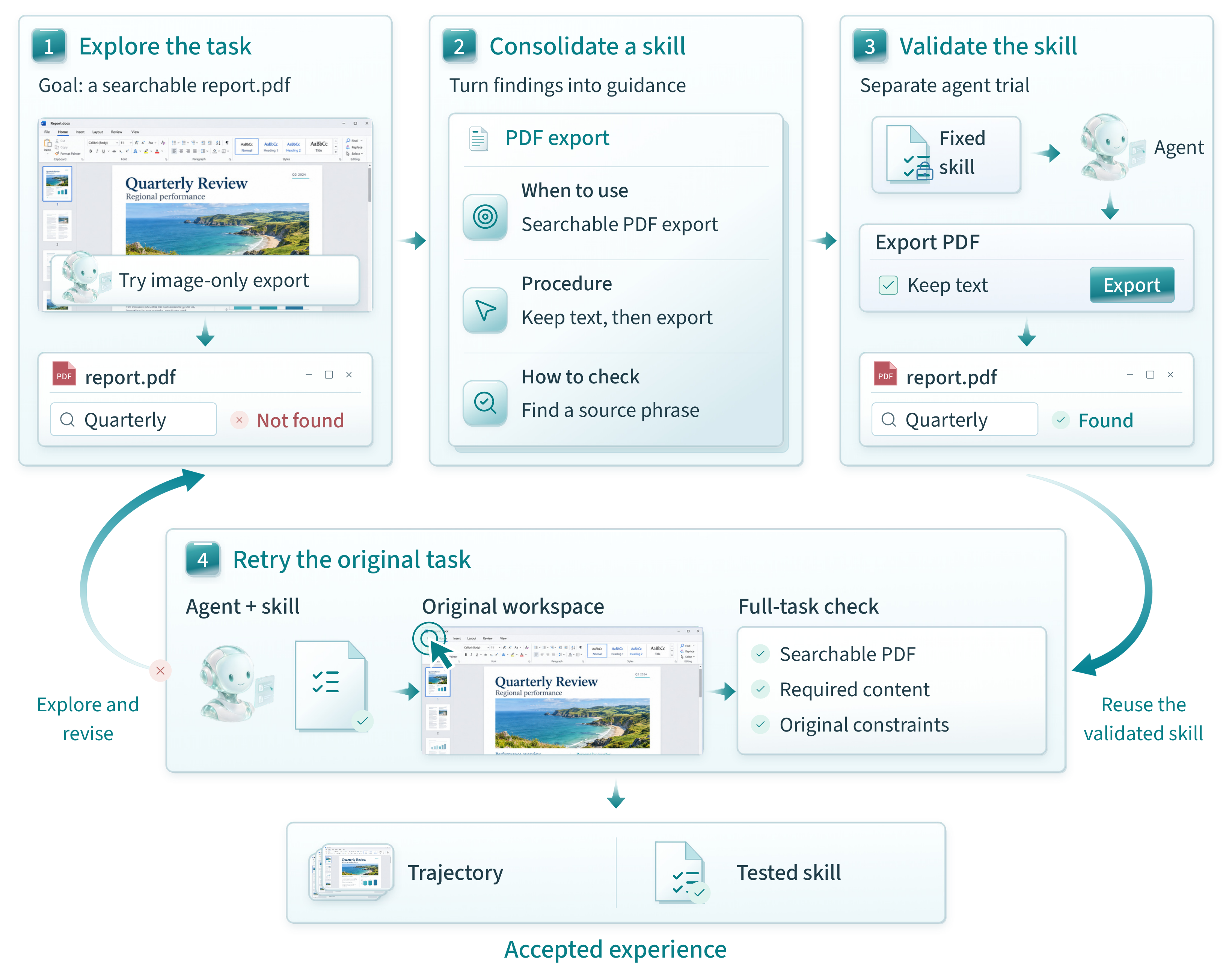}
\caption{\textbf{Accumulating operating knowledge through exploration.}
The source report contains ``Quarterly,'' but an image-only export makes
the phrase unsearchable in the PDF. Further exploration yields a
text-preserving procedure. An agent follows its fixed instructions in a
separate trial and searches the resulting PDF. The validated skill then
guides a new attempt in the original workspace, where full-task checks
determine trajectory acceptance. The return arrow depicts a failed task
attempt, while failed validation also prompts revision. Accepted trajectories retain the supplied skill
in context. GUI scenes, controls, actions, and outcomes are illustrative,
with intermediate exploratory trials omitted.}
\label{fig:skillexploration}
\end{figure}

We design a skill-guided loop with four stages, shown in
Figure~\ref{fig:skillexploration}. Task-directed exploration first produces
observations about useful operations and unresolved obstacles. A skill
creator then reviews that experience and consolidates actionable findings
into written guidance. A separate agent tests the candidate instructions
before they are supplied to a new task attempt. That attempt can use the
validated procedure to make further progress, and its new discoveries feed
the next round of skill refinement. The loop accumulates tested knowledge
across attempts while retaining full-task verification for the final result.

\paragraph{Explore the task and identify obstacles.}
We use a computer-use agent to attempt the original query in its prepared
workspace. It inspects the relevant applications and files, tries controls,
and adjusts its action sequence as it observes their effects.
An unsuccessful attempt can still reveal an important prerequisite, an
ineffective action to avoid, or a way to reach a previously inaccessible
state. Successful operations likewise show procedures worth retaining.
We examine these findings in the trajectory and result, using evaluator
observations of the live state to help identify the remaining obstacle
(Section~\ref{sec:evaluator}) \citep{NEURIPS2023_1b44b878}.

In the representative report workflow in Figure~\ref{fig:skillexploration},
exploration may reveal several export options. One trial produces a PDF with all the
required pages but represents their text as an image. The searchability
check then focuses further trials on which export settings preserve text.
The initial exploration has thus turned a broad completion goal into a
specific procedural question that can be tested in the application.

\paragraph{Consolidate a candidate skill.}
The skill creator reviews the accumulated trajectories and their outcomes.
When the useful procedure is still unclear, further exploration or focused
practice tests alternative operations. The creator turns the resulting
findings into textual guidance that states when the procedure applies,
which actions to take, and how to check its intended effect
\citep{wang2023voyageropenendedembodiedagent,wang2026webxskill}.
For the PDF example, this means recording the text-preserving export
settings and a search check, together with the image-only choice to avoid.
The candidate gives later attempts a concrete procedure and the relevant
pitfall.

\paragraph{Validate the skill independently.}
Guidance inferred from exploratory trials may omit a prerequisite or
misidentify why an operation worked. We therefore freeze the candidate
and give it to a separate agent in an independent validation trial.
The agent applies the instructions and checks their claimed effect in the
environment.

In the PDF example, it exports using the specified settings
and searches for a phrase from the source document. Finding the phrase
establishes the intended effect of this skill, while a failed check returns
the instructions to exploration and revision. The revised candidate is
tested again before admission for reuse.

\begin{samepage}
\paragraph{Reuse the skill and continue exploration.}
We provide the validated guidance to a new attempt at the original task.
The agent can now use the tested procedure when it reaches the relevant
operation. If it encounters another obstacle, or discovers a useful way
to complete a later stage, that experience feeds the same consolidation
and validation loop. Previously validated skills remain available, so each
round can build on what earlier attempts established.
\par\end{samepage}

The process continues until an attempt satisfies the original task's full
acceptance conditions or the exploration budget is exhausted. Accepted
attempts supply trajectories with the actual guidance retained in their
decision context. The skill library also makes a tested procedure available
to later tasks that meet its applicability conditions. This accumulation
is designed to improve hard-task collection by carrying useful knowledge
between attempts. The TERRA dashboard case in
Section~\ref{sec:rsi-cases} follows how a charting problem becomes revised
guidance that a fresh attempt applies (Figure~\ref{fig:skill-case}).

\begingroup\interlinepenalty=10000
For a long workflow, however, even a better-informed
attempt can spend substantial effort repeating earlier work. The next
pipeline addresses that remaining cost by retaining execution progress.
\par\endgroup

\subsubsection{Milestone-Guided Trajectory Search}
\label{sec:checkpoint-search}

A long task can be difficult to finish in a single attempt even when its
individual stages are within the agent's reach. When a late failure ends an
attempt, restarting from the initial state repeats work already completed.
This motivates a second RSI strategy in which collection
focuses on smaller, verifiable goals and retains the work that reaches them.

\begin{figure}[!t]
\centering
\includegraphics[width=\linewidth]{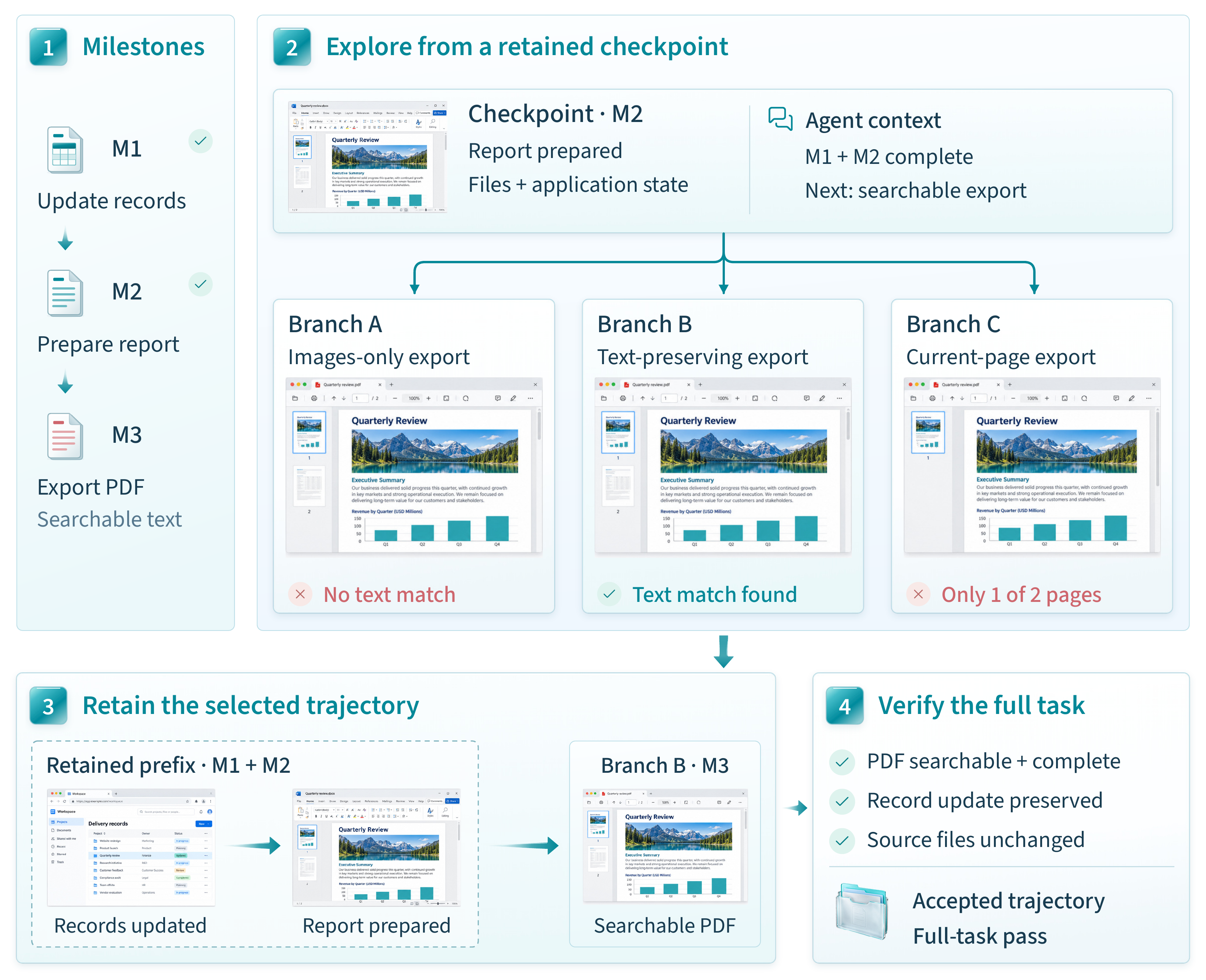}
\caption{\textbf{Retaining progress for branching trajectory search.}
Checkpoint M2 pairs the prepared report and software state with the agent
context. Independent continuations restore this checkpoint to try different
exports. The example contrasts image-only, text-preserving, and incomplete
single-page exports. The lower panel retains the common prefix and the chosen
continuation as one candidate path. The original task verifier checks the PDF,
earlier record update, and source preservation before accepting the trajectory.
Failed continuations leave the parent available for further attempts.
GUI states and diagnostic outcomes are illustrative.}
\label{fig:milestonesearch}
\end{figure}

We design milestone-guided trajectory search to implement this strategy.
We first decompose the original task into dependent milestones and their
checks. GUI attempts then work toward one milestone at a time, saving the
workspace and agent context when a check passes. Further attempts restore
that checkpoint to explore the next stage, following the principle of
returning to useful states for exploration \citep{ecoffet2021returnexplore}.
After the final milestone, the original verifier checks the complete result
before the selected path becomes a full-task demonstration.
Figure~\ref{fig:milestonesearch} illustrates how this process builds
successful continuations on retained progress. It reduces the work each
trial needs to repeat while keeping acceptance tied to the original task.
A construction case in Section~\ref{sec:rsi-cases} follows this accumulation of
progress from an empty Blender scene to a complete collected trajectory
(Figure~\ref{fig:milestone-case}).

\paragraph{Define milestones and their checks.}
We use a planning agent to read the user request and the original verifier
code, identify dependencies, and propose meaningful intermediate goals.
Each milestone pairs a subtask with a check on its observable result.
The goal should represent useful progress and leave a state from which
later work can continue. Reviewing the pairs against the original
requirements ensures that local checks support the intended full task.
This uses the decomposition principle in Section~\ref{sec:verifiers},
here to guide collection within one long task.

For the office workflow in Figure~\ref{fig:milestonesearch}, the original
goal includes updating a business record and producing a searchable PDF.
We organize it into three milestones: complete the record update (M1),
prepare the report (M2), and export the searchable PDF (M3). Their checks
inspect the corresponding records, document content, and output file.
The executing agent receives the permitted subtask guidance, while private
verifier constants remain in the checking context.

\paragraph{Preserve progress and explore continuations.}
The harness supplies the first subtask to a computer-use agent in the
initial workspace and runs repeated attempts within the collection budget.
When an attempt passes the milestone check, we save an E2B workspace
snapshot together with the agent context about completed work, useful
findings, and remaining goals. Restoring both gives the next attempt a
consistent starting point for the following milestone
(Section~\ref{sec:longruntime}). We retain the full interaction prefix
alongside the bounded decision context and can keep several passing states
to support different continuations.

Retained checkpoints form the nodes of a search tree, and GUI action
sequences toward the next milestone form its branches. Each continuation
restores its parent checkpoint in an isolated instance, observes the
interface, and attempts the next stage. Passing results become checkpoints
for the next milestone. If a continuation stalls or fails its check, its
parent remains available so that a new attempt can try a different sequence
without replaying the earlier stages.

We can also retain selected recoverable failed states for further exploration.
These states keep their unresolved milestone labels. A continuation attempts
the necessary correction before advancing to later goals, allowing the
collected path to include genuine recovery from an earlier mistake.
Thus the tree can support both alternative attempts from verified progress
and recovery attempts from a useful but incomplete state.

In Figure~\ref{fig:milestonesearch}, branches A and C fail the export
requirement, while B produces a searchable PDF. All three begin at M2,
retaining the same earlier record update and report preparation. A failed
export can also trigger the skill exploration described above, whose
validated guidance can be used by a new continuation. The saved checkpoint
preserves the work already done, and the skill helps the agent choose how
to perform the unresolved operation.

\paragraph{Verify outcomes and retain useful traces.}
After the final milestone, the original task verifier assesses candidate
completions against all requirements, including those established in earlier stages. For the office
task, it checks the PDF, the required record update, and source preservation.
Search continues until a branch passes this final check or the budget is
exhausted. A passing branch yields a \emph{search-assisted trajectory}:
the actual parent--child path from the initial state to completion,
combining its retained prefix with the selected continuations. It keeps the path's
observations, actions, and supplied guidance, including mistakes followed
by recovery. Failed sibling branches remain exploration records and are
not spliced into the accepted trajectory.

If an attempt completes only part of the milestone sequence, we retain only
the trace portions associated with passed milestones as subtask-level
demonstrations, preserving their original decision context. This recovers
useful supervision from unfinished attempts and makes fuller use of the
collection budget.

\subsubsection{Case Studies of Knowledge and Progress Accumulation}
\label{sec:rsi-cases}

We use two case studies to explain how the RSI strategies collect experience
on difficult tasks. The first follows skill refinement while building a live
retail dashboard in Calc. The second follows milestone accumulation while
constructing a greenhouse cafe in Blender. In each case, the recorded attempts
show what goes wrong and how retaining knowledge or progress helps subsequent
attempts complete the task.

\begin{figure}[!t]
\centering
\includegraphics[width=\linewidth]{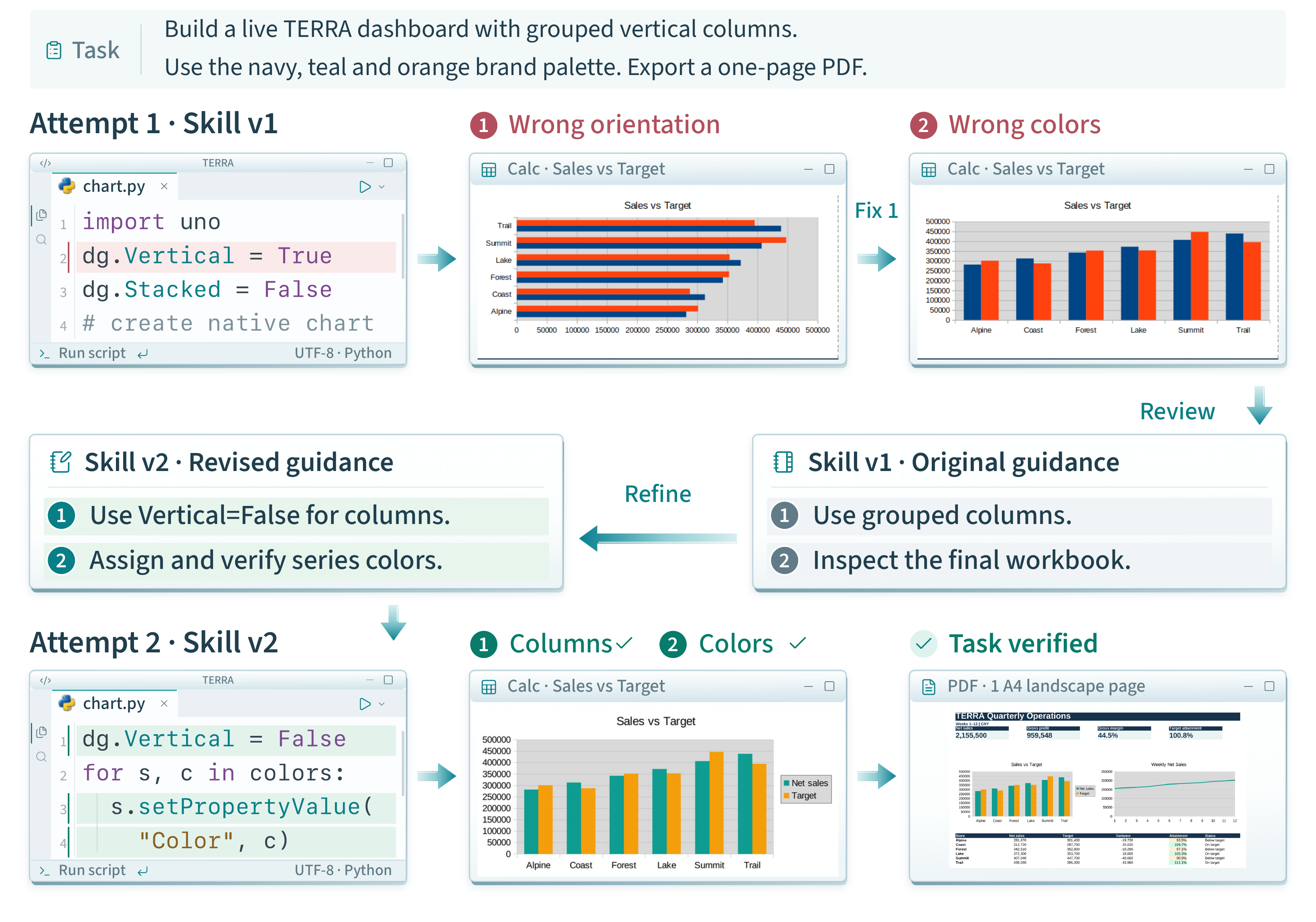}
\caption{\textbf{Skill refinement in the TERRA dashboard case.}
Numbered annotations track chart orientation (1) and series colors (2)
through exploration, revised guidance, and a fresh attempt. Attempt 1
repairs the orientation during execution but fails the final palette check.
Attempt 2 applies skill v2 and passes all 26 final checks. Chart crops and
the final PDF come from recorded artifacts. Window frames and equivalent
Python excerpts are composed for presentation.}
\label{fig:skill-case}
\end{figure}

\paragraph{Knowledge accumulation in a dashboard task.}
Figure~\ref{fig:skill-case} starts with a request to build a quarterly
operations dashboard for TERRA from sales records and store targets.
The deliverables are an editable workbook with live calculations and charts,
and a matching one-page PDF. The request specifies grouped vertical columns
and a navy, teal, and orange palette. The difficulty is translating these
visual requirements into Calc's orientation and color settings, whose effects
need to be checked in the saved workbook and exported PDF.

In Attempt 1, the agent follows skill v1, which asks for grouped columns
and final document inspection. These instructions leave the relevant API
settings unspecified. The agent sets \texttt{Vertical=True} through the
legacy chart interface, producing the horizontal bars shown in the top row.
It discovers the correct setting and fixes the direction during this
attempt, but the charts retain the default blue and red series colors.
The final palette check therefore rejects the delivery.

Reviewing this trajectory identifies the two changes shown in the middle
of the figure. Skill v2 records \texttt{Vertical=False} as the setting for
vertical columns and adds explicit series-color assignment through
\texttt{chart2}, followed by property readback and visual inspection.
Engineering probes check these procedures before the revised skill is fixed
in a new attempt's context. Attempt 2 then starts from the original input
in a fresh workspace. It applies the orientation setting and assigns teal
to net sales and orange to the target series before exporting the dashboard.
The bottom row shows the resulting columns, palette, and PDF. The completed
task passes all 26 final checks, including input-change tests of the live
formulas and charts. Earlier trial-and-error has thus become precise operating
guidance that the next attempt can apply to produce a verified demonstration.

\begin{figure}[!t]
\centering
\includegraphics[width=\linewidth]{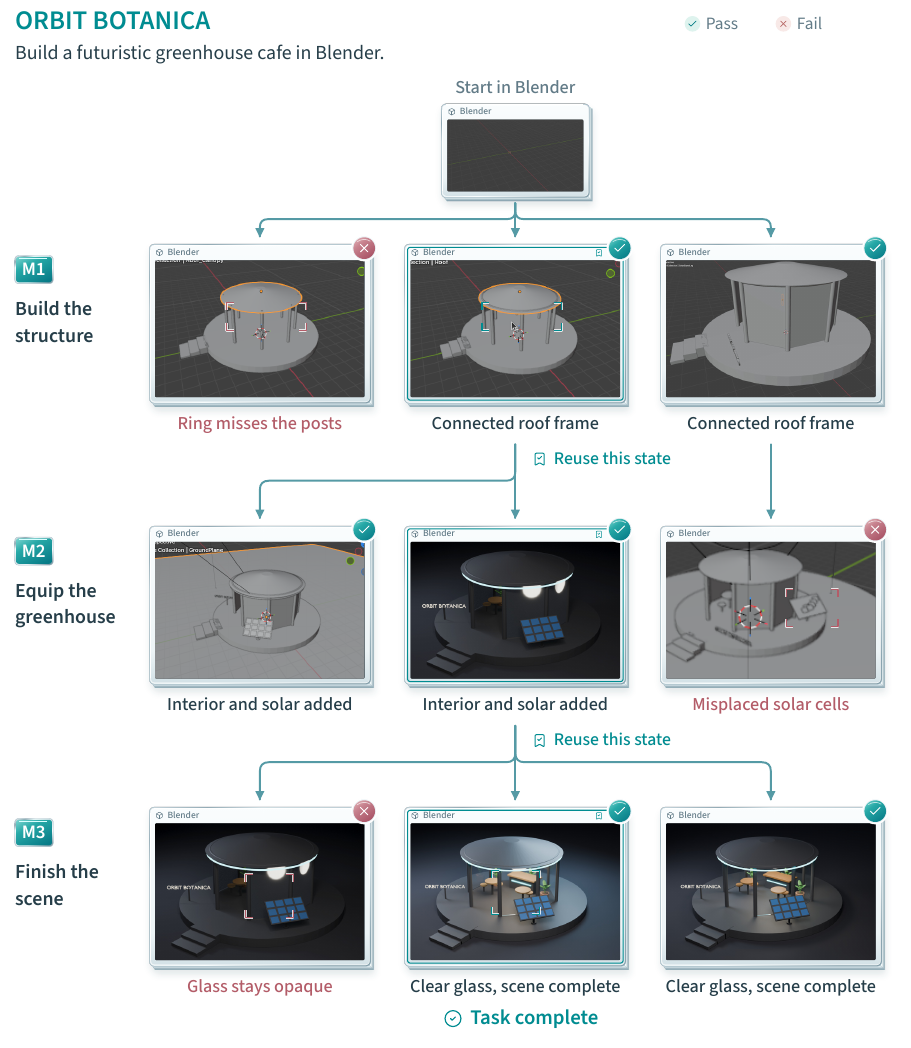}
\caption{\textbf{Milestone-guided construction of ORBIT BOTANICA.}
Three display milestones group the recorded checks. Teal frames mark one
retained path, and arrows preserve ancestry through omitted intermediate
steps. Branches can complete later work early. The equipped checkpoint and
its failed finishing continuation share the unchanged opaque-glass render.
Both successful finishing branches pass the full-task checks. Scene crops
are unretouched, with frames and diagnostic annotations added for presentation.}
\label{fig:milestone-case}
\end{figure}

\Needspace{5\baselineskip}
\paragraph{Progress accumulation in a 3D modeling task.}
Figure~\ref{fig:milestone-case} follows a request to build ORBIT BOTANICA,
a futuristic greenhouse cafe, from an empty Blender scene and deliver an
editable model with its final render. The task combines structural, equipment,
and appearance requirements, so an error in the final render can arise after
much of the construction is complete.

We apply the milestone-guided procedure in Section~\ref{sec:checkpoint-search}
to organize construction into checkable intermediate goals. The figure groups
the recorded checks into three display milestones and shows selected branches.
At M1, the agent builds the platform, frame, and roof. The left branch fails
because its upper ring misses the posts, whereas the other two displayed
branches produce connected frames. Saving these passing scenes with their
agent contexts allows exploration to continue from established structures.

\begin{samepage}
At M2, the continuations add the interior and solar equipment. The right
branch misplaces solar cells and fails the equipment check. The two
continuations from the central M1 checkpoint pass, preserving the completed
structure and equipment for finishing work.
\par\end{samepage}

At M3, the displayed continuations all start from the central equipped
checkpoint. One leaves the glass opaque, obscuring the interior and failing
the finishing requirements. The other two produce clear-glass renders and
pass the full-task checks. Because the equipped checkpoint remains available,
these finishing trials can address the remaining rendering requirements
without rebuilding the frame or furnishing the cafe. The teal frames trace
one retained path through all three milestones. Once its final result passes
the original task verifier, the interaction prefix and successful continuations
along that path form a complete demonstration. Failed siblings remain search
records. This case shows how preserving checked intermediate work lets
collection concentrate further trials on the unresolved part of a long task.

Together, the cases illustrate two ways to carry useful experience forward.
TERRA reuses a procedure learned from earlier execution, while ORBIT BOTANICA
reuses the work already completed in the software environment. Both routes
direct subsequent attempts toward unresolved requirements and retain verified
task trajectories for learning.
Appendix~\ref{app:hardexperience} details checkpoint contents, search budgets,
skill records, and the treatment of complete paths and checkpoint continuations.

\paragraph{Summary.}
The Experience Flywheel supplies the staged learner with both
demonstrations and opportunities for fresh interaction. Curated grounding
examples support control localization, while selected trajectories demonstrate
complete workflows. RSI expands the candidate pool with attempts at difficult tasks.
The same executable tasks can support new rollouts with task-specific
outcome feedback as the policy develops. Section~\ref{sec:training}
explains how the learner uses these complementary resources.
Appendix~\ref{app:corpus} describes data sources, scale, and source-grouped
splits, with construction details in Appendix~\ref{app:data-method-status}.

\section{Staged Learning from Scalable GUI Experience}
\label{sec:training}

The Experience Library assembled by the \experienceflywheel{} provides
quality-selected demonstrations of software work, alongside grounding
examples and executable tasks for further collection
(Section~\ref{sec:data}). The next step is to turn these resources into a
policy that can reason about interfaces and carry out user goals across
desktop and web applications. We design a three-stage learning pipeline
that develops this capability progressively, as outlined in
Figure~\ref{fig:concept}.

\begin{enumerate}[leftmargin=*,itemsep=2pt]
\item \textbf{Broad GUI mid-training} first establishes familiarity with controls,
application functions, and their effects through diverse software experience.
\item \textbf{Verified long-horizon SFT} builds on that foundation and concentrates learning
on high-quality demonstrations that carry dependent work to completion.
\item \textbf{Joint desktop--web agentic reinforcement learning} then lets the policy learn
from its own decisions and their outcomes in interactive environments.
\end{enumerate}
Each stage uses the preceding policy as its starting point while changing
the learning emphasis, from operational breadth to sustained completion and
then outcome-driven improvement.

Table~\ref{tab:stagerecipe} summarizes the resources and priorities of the
three stages. The shared interfaces of the \northcua{} Harness connect
demonstration collection with online execution, so the policy can continue
using the same observation and action conventions as training progresses.
We describe each stage below. Appendix~\ref{app:training} defines the
supervised prediction units and how recorded RL attempts become learning
samples.

\begingroup
\interlinepenalty=10000
For model training, all spatial coordinates, including grounding targets
and coordinate-based action arguments, use a shared \([0,1000]\)
representation. For an image of width \(W\) and height \(H\), a pixel
location \((x,y)\) becomes \((1000x/W,1000y/H)\), giving the model a
consistent coordinate convention across screen resolutions.\par
\endgroup

\begin{table}[!t]
\centering\small
\caption{Learning priorities, data selection, and scale across the three
stages. Supervised scale entries are rounded cohort summaries
(Table~\ref{tab:corpus}). GUI steps count interactions within trajectories.
Grounding examples are counted separately, as are auxiliary evaluator
demonstrations. RL counts describe the prepared task pool.}
\label{tab:stagerecipe}
\begin{tabularx}{\linewidth}{@{}P{0.18\linewidth}P{0.19\linewidth}>{\raggedright\arraybackslash}XP{0.18\linewidth}@{}}
\toprule
Stage & Learning focus & Data and acceptance & Reported scale \\
\midrule
Broad GUI mid-training
& Localization and broad software use
& Grounding, judge-selected task trajectories, and auxiliary evaluator demonstrations
& \(>100{,}000\) trajectories\newline \(>4{,}000{,}000\) GUI steps \\
Verified long-horizon SFT
& Sustained completion of dependent workflows
& Complete trajectories selected for script acceptance, dependencies,
and execution quality
& \(\approx3{,}000\) trajectories\newline \(\approx310{,}000\) GUI steps \\
Joint desktop--web agentic RL
& Improvement through the policy's own execution
& Human-reviewed tasks with outcome variation, followed by fresh
verifier-scored task groups
& \(\approx1{,}500\) desktop tasks\newline \(\approx500\) Web tasks \\
\bottomrule
\end{tabularx}
\end{table}

\subsection{Broad GUI Mid-Training}
\label{sec:midtraining}

Computer-use agents encounter many applications, interface layouts, and
ways of performing similar operations. To work across these settings, a
policy needs experience recognizing controls, interpreting their functions,
and observing how actions change software state. We use broad GUI
mid-training to build this foundation from diverse demonstrations
\citep{wang2025opencua,jung2026procuasft}.

Through the collection routes in the \experienceflywheel{}, the reported
broad task corpus contains over 100,000 trajectories and over 4 million GUI
interaction steps, spanning dozens of applications and hundreds of operation
families. This range exposes the policy to different tasks, materials, and
interfaces within and across applications. We complement the trajectories
with approximately 40,000 single-step grounding examples to train spatial recognition and
control localization directly. The following collection and selection
choices determine what experience enters this mixture.

\paragraph{Broad software coverage.}
We select commonly used desktop applications and organize their functions
into operation families. These profiles guide the task requests and
\casespec{} environments used for experience collection, making it possible
to vary the operations demonstrated within each application.
Sampling across applications and operations, together
with variation in source files and user goals, broadens this coverage.
Representative coverage in the application profiles and workflow templates includes:
\begin{itemize}
\item \emph{Office work.} Calc tasks cover formulas, auditing and
reconciliation, filtering, pivot tables, and print layout. Writer adds
document formatting and annotation, while Impress covers slide composition,
content insertion, styling, and synthesis of source material into presentations.
\item \emph{Visual and media work.} GIMP tasks include cropping, resizing,
annotation, and export. Inkscape adds vector text, color, alignment, and
grouping operations. Image viewers, PDF viewers, and VLC provide content
inspection and playback tasks.
\item \emph{Workspace and information management.} File-manager tasks
cover organization and batch processing. Thunderbird, browsers, and
VS Code add message handling, browser configuration, information lookup,
and GUI-based file editing.
\item \emph{CAD and 3D workflows.} Expanded workflow templates define
model review and part design in FreeCAD and SolveSpace, together with scene
organization in Blender.
\item \emph{Medical-image and anatomy visualization.} 3D Slicer scenarios
cover inspecting image volumes, checking orientation, annotating landmarks,
and saving review scenes.
\end{itemize}
To extend this coverage beyond individual functions, we also use the
long-workflow construction in Section~\ref{sec:scaleintents}. Its dependent
tasks provide experience combining operations, carrying information between
applications, and maintaining progress toward a shared deliverable.
Mid-training therefore includes sustained workflows as part of its broad
foundation, before the next stage concentrates on selected verified examples.

We further enrich web experience with (1) information retrieval, (2) form filling,
(3) human-verification flows, and (4) frontend testing. Retrieval and form filling
represent common everyday uses of websites. Human-verification flows add
visually conditioned interactions encountered during web use, while frontend
testing exercises the effects of controls and the persistence of state changes.
Together, these settings diversify both what users ask and how an agent must
interact, complementing desktop operation with web-specific workflows.
The web scenario library also provides browser-game and sliding-puzzle tasks.
Grounding examples add direct practice locating the requested control,
including specialized controls handled by the documentation-guided route in
Section~\ref{sec:grounding}. Table~\ref{tab:midtraining-scope} summarizes
the coverage of these sources, while Appendix~\ref{app:corpus} gives their
rounded counts by domain.

\begin{table}[!t]
\centering\small
\caption{Representative coverage of broad mid-training sources.
Dataset scale and sampling conventions appear in Appendix~\ref{app:corpus} and
Appendix~\ref{app:supervisedrecipe}.}
\label{tab:midtraining-scope}
\begin{tabularx}{\linewidth}{@{}P{0.25\linewidth}>{\raggedright\arraybackslash}X@{}}
\toprule
Source & Representative coverage \\
\midrule
Desktop base tasks & Office operations, visual/media work,
and workspace/information management \\
Desktop long workflows & Cross-source reasoning, conflicting-information
resolution, and dependent work toward a shared deliverable \\
Web tasks & Information retrieval, form filling,
human-verification flows, and frontend testing \\
Single-step grounding & Locating requested controls in desktop
and web screenshots \\
Evaluator trajectories & Tool-assisted inspection and evidence-based
assessment of task outcomes \\
\bottomrule
\end{tabularx}
\end{table}

\paragraph{Outcome-based trajectory curation.}
\label{sec:experience-selection}
Broad coverage is useful for learning only when the collected behavior
provides a worthwhile demonstration. Large-scale attempts can also leave
tasks incomplete or spend many actions without progress. We therefore use
the agent-as-a-judge evaluator in Section~\ref{sec:evaluator} to inspect each
candidate's requested outcome, trajectory, and resulting artifacts or state
\citep{zhuge2024agentasajudgeevaluateagentsagents}. Its assessment selects
completed work with useful execution behavior for the training corpus.
This selection limits the inclusion of incomplete or low-quality traces
while preserving examples of successful software use across the coverage pool.

Among the trajectories that pass review, selection preserves broad
application and operation coverage. Meaningful exploration and recovery
remain part of this experience when they lead to completed work, giving the
model examples of how actions and subsequent observations inform the next
decision. Outcome-based filtering thus makes quality control part of
building a large and varied training corpus.

\paragraph{Learning localization, interaction, and assessment.}
The selected mixture teaches both where to act and how to continue a task.
For grounding examples, the model predicts the target control's coordinates
from the request and screenshot
\citep{cheng2024seeclick,lu2024omniparser}. Trajectories train the model to
predict each recorded assistant response from the task and the observations
and history available at that decision. Targets include recorded reasoning,
text responses, and action tool calls, while screenshots and tool results
provide context. This response-level objective connects interpreting the
interface, deciding what to do, and issuing the operation.

We also include evaluator trajectories as a separate judgment-task source.
These examples condition the model on the evaluator's inspection context and
train its recorded assessment responses, providing demonstrations of how to
check an outcome and support a judgment with evidence. Actor demonstrations
retain their original decision-time context, separate from these evaluation inputs.
The evaluator therefore serves two distinct purposes in mid-training: its
reports select task demonstrations, and its own trajectories teach assessment.
Appendix~\ref{app:supervisedrecipe} distinguishes the sample types and their
response targets.

Through this broad mixture, mid-training develops familiarity with diverse
GUI functions, action effects, and ways of interpreting results. This provides
the operational foundation for learning more demanding workflows. The next
stage sharpens that foundation using a concentrated set of high-quality,
verified long-horizon demonstrations.

\subsection{Verified Long-Horizon SFT}
\label{sec:longsft}

Mid-training gives the policy broad experience operating software, including
extended workflows. Sustaining correct work over a long and difficult task
requires a more concentrated learning signal: later decisions need to use
earlier findings, preserve constraints, and complete the final deliverable.
We therefore continue from the mid-trained checkpoint with supervised
fine-tuning on selected long-horizon trajectories whose complete outcomes
pass executable task verifiers \citep{yuan2026osworld2,lu2026qwencua}.
The cohort contains approximately 3,000 trajectories and 310,000
GUI interaction steps.

\paragraph{Selecting verified demonstrations.}
We start with tasks and executable checks supplied by Scaling Verifiers
(Section~\ref{sec:verifiers}) and collect successful completions. For
difficult tasks that ordinary attempts do not solve, the skill-guided and
milestone-guided RSI routes in Section~\ref{sec:hardexperience} provide
additional ways to obtain candidate demonstrations. Skills carry useful
operating knowledge between attempts, while retained checkpoints let
exploration build on completed intermediate work. In either route, a
candidate enters this stage only after its final result passes the original
task verifier. Appendix~\ref{app:corpus} lists the source contributions that
remain to be reported separately from this aggregate cohort.

Within this accepted pool, we prioritize difficult workflows with long-range
dependencies while maintaining application diversity. We use length to identify
sustained work, then inspect how later actions use earlier results to assess
its dependencies. We then check whether these demonstrations provide
usable training sequences. Data cleaning rejects unusable records
and traces with malformed or unmatched tool calls or unproductive loops.
This screening retains coherent recovery when the trace shows an error,
its correction, and verified completion. Each retained trace keeps its original
action--observation sequence, preserving the context needed to learn that
recovery. These checks combine programmatic
outcome acceptance with demonstrations of purposeful work across stages.

\paragraph{Learning sustained completion.}
With this selected corpus, we fine-tune the mid-trained policy to predict
the recorded assistant responses from the request and decision-time context.
The objective remains response prediction, but the data now emphasize
maintaining task state, using intermediate results, coordinating work across
applications, and recovering without losing the original goal. For assisted
demonstrations, any supplied skill or milestone guidance remains in the
context under which the recorded response was produced. Representation and
sampling requirements are described in Appendix~\ref{app:supervisedrecipe}.

This stage strengthens sustained task execution on top of the broad
operational foundation. The resulting policy is the starting point for
autonomous rollouts, where its own choices can create states absent from
the demonstrations. Reinforcement learning provides the next learning
signal by connecting those choices to their task outcomes.

\subsection{Joint Desktop--Web Agentic Reinforcement Learning}
\label{sec:rl}

Mid-training and long-horizon SFT provide a policy that can recognize
interfaces and follow demonstrated workflows. To develop its ability further,
we let it learn from autonomous interaction. Its own choices determine which
states it encounters, and task outcomes provide feedback on whether those
choices accomplish the user's goal. We use agentic reinforcement learning
to connect this fresh experience to policy improvement
\citep{qi2024webrl,lai2025computerrl}.

We train a single policy jointly on desktop and web tasks. The shared runtime
in the \northcua{} Harness already expresses both as repeated observation,
reasoning, and tool-based action, making it natural to learn from them through
one training loop. Platform-specific observations and tools retain the
differences needed to operate each interface. Task-specific verifiers then
provide outcome feedback for the work performed in either domain.

\begin{figure}[tbp]
\centering
\includegraphics[width=\linewidth]{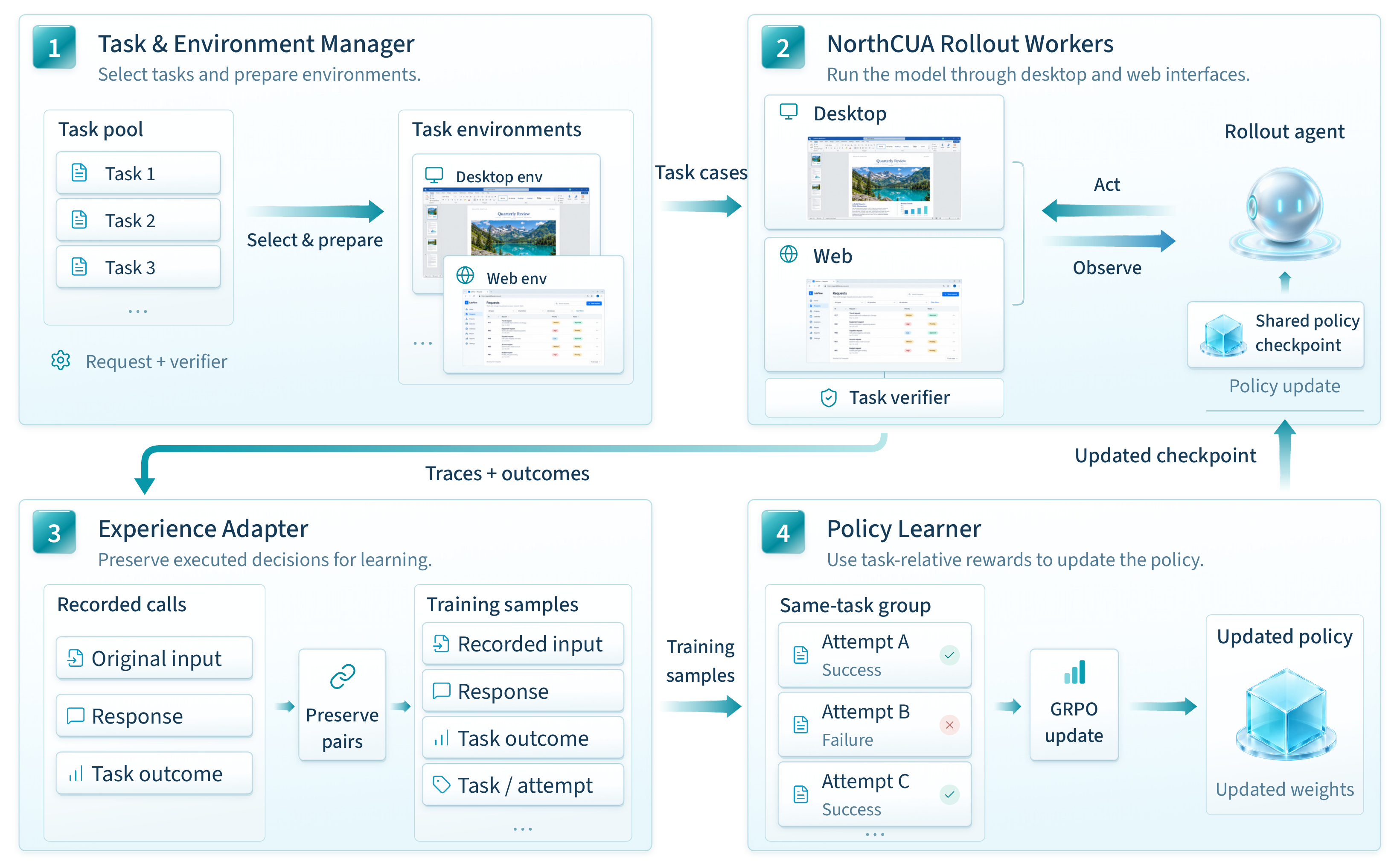}

\caption{\textbf{Joint desktop--web agentic reinforcement learning.}
The manager selects tasks, binds verifiers, and prepares isolated environments.
Workers use a model checkpoint to act through desktop or web interfaces.
Manager-run checks supply outcomes before workspace release. The adapter binds
outcomes to training samples while preserving each recorded input--response
pair and task/attempt identity. The learner uses GRPO to compare valid attempts
within the same task, initial-state, and verifier group, then returns an updated
checkpoint to workers. Worker and learner icons denote execution and training
uses of the same model. GUI scenes and binary outcomes are illustrative.
The example leaves task-group size unspecified.}
\label{fig:rl}
\end{figure}

Figure~\ref{fig:rl} follows an attempt from task preparation to a policy
update. The Task and Environment Manager selects executable cases and
prepares their isolated workspaces and verifiers, giving NorthCUA Rollout
Workers the conditions needed to act with the current policy. Once those
attempts have been recorded and scored, the Experience Adapter pairs the
model calls with their task outcomes to construct training samples.
These task-grouped samples let the Policy Learner compare alternative
attempts with GRPO and update the shared policy
\citep{shao2024deepseekmath,slime_github}. Returning the new weights to the
workers closes the loop, so subsequent interaction uses the updated model.
The following subsections explain how we select useful tasks, obtain
comparable feedback, and preserve executed decisions as that feedback enters
learning.

\subsubsection{Selecting Tasks with Outcome Variation}
\label{sec:rlselection}

GRPO obtains its task-level learning signal by comparing the rewards of
multiple attempts at the same task. An attempt's relative outcome determines
its advantage within that group. If all valid attempts receive the same
terminal reward, their terminal group-relative advantages are zero.
Since GUI rollouts are costly, we first identify tasks on which the SFT
policy produces different outcomes, so initial collection can provide
informative comparisons.

For each model, the Task and Environment Manager starts from human-reviewed
desktop and web tasks associated with its SFT data. For each candidate case,
we collect at least five rollouts under the SFT checkpoint with fixed task
and execution settings. The task verifier scores each attempt's outcome,
and we select cases with variation among these valid verifier scores.
Valid task outcomes determine this selection, while environment and verifier
errors are recorded separately. This focuses RL collection on tasks where
outcome feedback distinguishes alternative executions.

The selected cases form the initial RL task pool. Fresh grouped attempts
supply the subsequent learning data. Appendix~\ref{app:rlpool} describes
the screening protocol and the fields needed to record valid repeats and
the selection rule for each run.

\subsubsection{Joint Rollout and Task-Relative Feedback}
\label{sec:rlgroups}

With the task pool selected, we generate fresh attempts and turn their
outcomes into the comparisons used by GRPO. For each case, the manager first
establishes equivalent starting conditions. Workers then act in those
environments, and the task verifier assesses each completed attempt. The
recorded decisions and outcome are converted into samples for policy learning.

The Task and Environment Manager prepares a separate E2B sandbox for each
attempt in a task group. All members receive the same request, initial files
and application state, verifier, and execution budget. Their sandboxes are
isolated, so one attempt's edits do not change another's starting conditions.
This makes differences in outcomes interpretable as alternative executions
of the same task.

NorthCUA Rollout Workers use the policy to carry out each multi-turn attempt.
Desktop and web execution follow the same runtime loop, using their respective
screen or browser observations and allowed tools. After execution, the manager
runs the task verifier on the resulting workspace. Its valid verdict supplies
the complete attempt's terminal reward. Both successful and unsuccessful
attempts contribute to the comparison, while setup or evaluation errors
follow the separate invalid-attempt path.

The Experience Adapter binds each outcome to the model calls from that
attempt. It reconstructs training samples from the recorded multimodal inputs
and responses, retains behavior-policy likelihoods and versions, and marks
which response tokens are learning targets. Each sample keeps its task and
attempt identity so that the Policy Learner can compute relative advantages
within the correct task group and apply the GRPO update. Desktop and web
groups contribute to the same shared policy through this procedure.

This completes the path from an executable case to outcome-driven learning.
For long GUI attempts, however, converting the record into training samples
requires additional care. The next subsection explains the two rules we use
to preserve both the original decision conditions and the attempt's learning
contribution during that conversion.

\subsubsection{Preserving Executed Decisions for Learning}
\label{sec:rlcontext}

The rollout procedure provides complete attempts and their outcomes, but
the learner must also preserve the conditions under which each decision
was made. Our harness maintains a sliding window of recent screenshots and
can compact earlier history. As a task proceeds, an image may leave the
window or a summary may replace earlier messages. Using only the final
history to train every response would therefore change the evidence
available to some decisions. Preserving the sampled input is also a central
concern in harnessed agentic RL \citep{he2026agentlightningv1}.

\begin{figure}[!t]
\centering
\includegraphics[width=\linewidth]{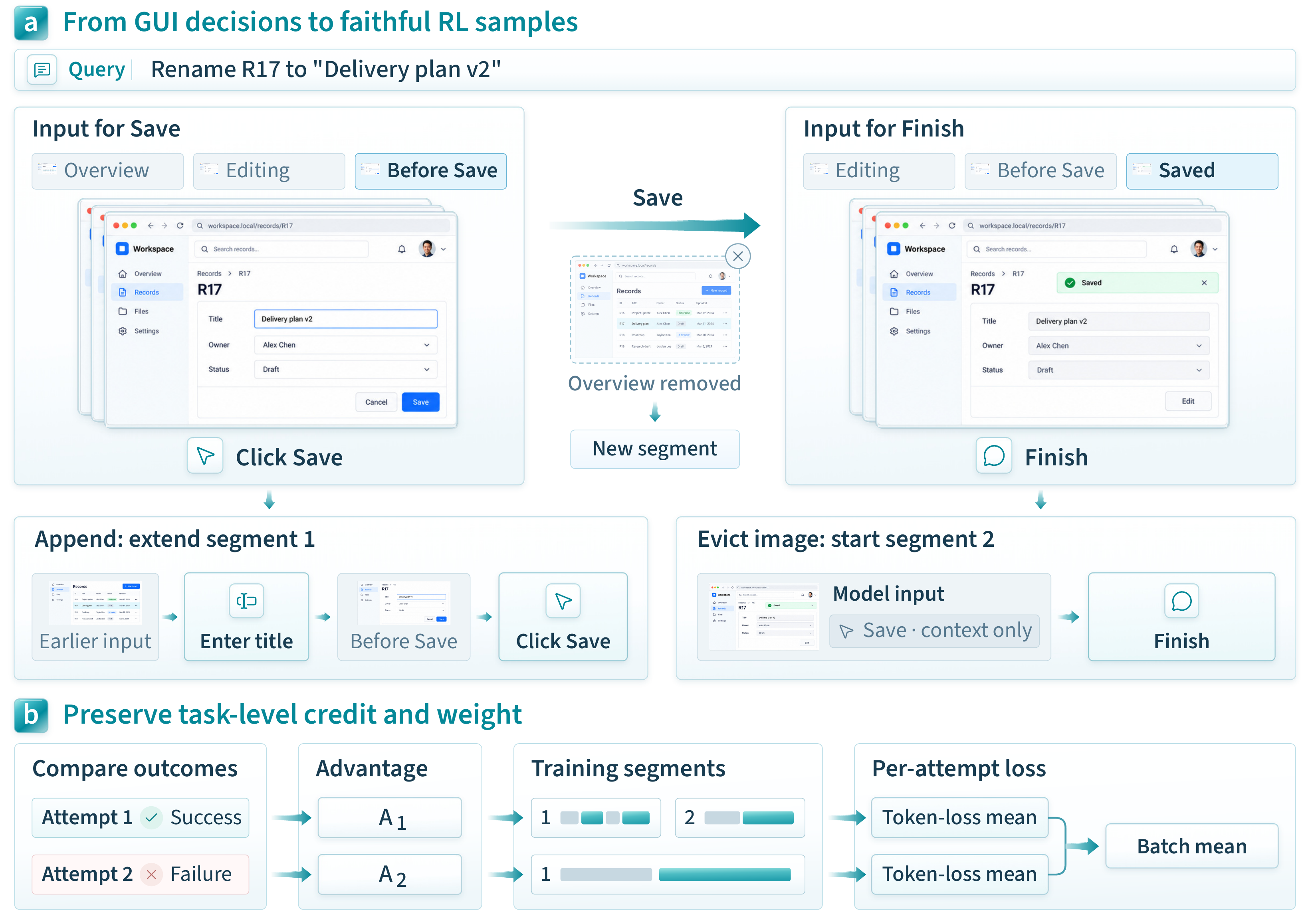}

\caption{\textbf{Context-preserving RL sample construction.}
\textbf{(a)} The earlier input contains Overview and Editing. Appending the
typing result preserves history, so Enter title and Save share one segment.
After the successful Save action, Overview leaves the visual window (cross),
starting a new segment. The highlighted label marks the latest observation.
Gray objects supply context, while teal responses and token spans contribute
to policy loss. \textbf{(b)} Task-group outcomes determine one advantage per
attempt, shared across its segments. In the trajectory-mean configuration,
masked policy losses are pooled and normalized over each attempt's target
tokens before averaging across attempts (Eq.~\ref{eq:memberloss}). Panel (a)
shows one continuous attempt under one behavior-policy version, with earlier
images unchanged until eviction. GUI states, responses, and outcomes are
illustrative. Thumbnails are input excerpts, strip lengths are schematic,
and earlier calls are omitted.}
\label{fig:rlcontext}
\end{figure}

To accommodate this context management, we design the Experience Adapter
around two preservation rules. First, each response is learned with its
recorded decision-time input, starting a new training segment when the
history no longer extends consistently. Second, those segments retain the
reward attribution and learning weight of their original complete attempt.
This matters because frequent context changes can split one attempt into
more segments than another. Figure~\ref{fig:rlcontext} illustrates input
preservation in panel (a) and attempt-level comparison and aggregation in
panel (b). Together, the two rules let the harness manage long histories
without turning a representation change into a different learning problem.

\paragraph{Preserve decision-time inputs.}
Training associates each response with the context that produced it. To
preserve this association through context changes, we capture each model call's
input, delivered response, generation log-probabilities, and behavior-policy
version. The adapter joins consecutive calls only when their text and
ordered image histories form a confirmed prefix extension under the same
version. Image identity includes its content digest, so replacing a clean
frame with the historical action overlay from Section~\ref{sec:visualcontext}
also breaks this prefix. The adapter starts a new segment from the captured
input whenever images are replaced or evicted, a summary rewrites history,
or the behavior version changes.

Earlier responses can still appear as history in a later segment, but they
serve only as context there, following the history-masking principle used
in ScaleCUA \citep{lv2026scalecua}. Response masks assign each learning target to
its original occurrence. Save is supervised when first generated and serves
as context for Finish. This construction lets the learner use the
harness's actual decision conditions while keeping only new responses as
targets. Appendix~\ref{app:training} specifies processor alignment and masks.

\paragraph{Preserve the complete attempt's learning weight.}
Preserving inputs can produce different numbers and lengths of segments
across attempts.
If these segments were treated as separate reward-bearing attempts, more
frequent context changes would alter the group comparison. Averaging segment
losses independently would also give an attempt more weight simply because
it was split more often. Neither change reflects additional task experience.

We therefore retain the complete attempt as the comparison unit. GRPO
computes its relative advantage once, and all of its segments receive that
same value. In Figure~\ref{fig:rlcontext}b, the two segments of attempt 1 share
$A_1$, while attempt 2 receives $A_2$. Loss aggregation then preserves the
same unit. In the supported trajectory-mean configuration, we use the
trajectory-level normalization also adopted in Qwen-CUA \citep{lu2026qwencua}.
We sum the masked policy losses over all target tokens of an attempt, divide by its
total target-token count, and average these per-attempt means:
\begin{equation}
\mathcal{L}_{\mathrm{policy}}
= \frac{1}{N}\sum_{i=1}^{N}
  \frac{\displaystyle\sum_{s\in S_i}\sum_t m_{ist}\ell_{ist}}
       {\displaystyle\max\!\left(1,\sum_{s\in S_i}\sum_t m_{ist}\right)},
\label{eq:memberloss}
\end{equation}
where $N$ is the number of complete attempts, $S_i$ contains the segments
of attempt $i$, $m_{ist}$ selects target tokens, and $\ell_{ist}$ is the
token-level policy loss. All segments of an attempt enter the same optimizer step, so
the denominator covers the complete attempt across microbatches.
Appendix~\ref{app:training} defines the reduction and the run-specific
optimization fields. This normalization prevents segmentation alone from
increasing an attempt's contribution when its target tokens and losses are
unchanged. Combined with decision-time input preservation, it allows one
continuous execution to yield several training segments without creating
extra reward comparisons or increasing the attempt's weight in the policy loss.

\subsubsection{Continuous Rollout and Policy Updates}
\label{sec:rlupdates}

Training from long GUI rollouts is costly even after the samples are
correctly constructed. Overlapping screenshot windows repeat visual
preprocessing and carry large tensors to the learner. Meanwhile, task
durations vary, so waiting for every active task before updating the policy
can delay the next learning cycle. We address these two costs separately:
visual-delivery mechanisms reduce redundant input preparation and transfer,
while asynchronous coordination lets policy updates proceed without
restarting ongoing tasks.

\paragraph{Deliver visual experience efficiently.}
Adjacent decision windows often contain unchanged screenshots. Processing
every window from scratch repeats work on those images. For a supported
per-image processor, the Experience Adapter therefore reuses the
preprocessing tensors of unchanged images within an attempt and processes
only new images. It reconstructs the requested image order under the same
processor settings. Changed images or settings require fresh processing.
The reuse saves CPU preprocessing while retaining every image in the model
input. The neural visual encoder still processes the complete visual input.

Large image tensors also add delivery cost after preprocessing. The
object-store path supports lossless tensor compression, retaining a
compressed payload only when it is smaller. The learner decodes the
microbatch it is about to consume, recovering the same values, shape,
and type. Bounded CPU workers prepare samples, and group admission counts
completed work awaiting delivery so that preparation does not create an
unaccounted queue. These mechanisms target redundant processing and
payload transfer while preserving the sample's contents.
Appendix~\ref{app:visualdelivery} describes the implemented reuse and
transport paths, their configuration, and the checks used to assess sample
equivalence and delivery cost.

\paragraph{Coordinate policy updates asynchronously.}
In asynchronous mode, completed groups can supply the
learner while other workspaces remain active. To publish an update, the
system pauses new model calls, waits for calls already in progress, commits
the new weights, and resumes calls. Ongoing tasks keep their workspace and
completed work, so later decisions can use the updated policy.

The captured behavior version and generation probabilities identify the
policy that produced each response. A version change starts a new segment,
so each policy ratio uses the corresponding behavior probabilities.
For attempts spanning several versions, the clipped objective is an
off-policy approximation. Preserving the recorded inputs keeps the sampled
decisions identifiable, while policy lag affects the resulting update.
Appendix~\ref{app:rlpool} defines policy lag and mixed-version attempts and
distinguishes asynchronous coordination from synchronous collection.
This coordination lets learning continue while
ongoing tasks retain their environments and completed work.

Together, these mechanisms address the cost of supplying visual experience
and coordinating updates, while the preceding preservation rules keep
learning tied to the executed decisions and complete task outcomes.
Section~\ref{sec:stageexperiments} reports the Mid-to-SFT
checkpoint changes. Section~\ref{sec:rlexperiments} then examines joint RL
through reported checkpoint comparisons, experience-delivery measurements,
and a concrete example of task-relative feedback.

\section{Experiments and Analysis}
\label{sec:experiments}

Our evaluation follows the progression from identifying a control to
completing a software task. We first compare grounding, desktop, and Web
performance, then examine how results change through long-horizon SFT and
joint desktop--Web RL. Trace analyses and selected cases connect these
scores to concrete decisions, while an engineering profile measures the
cost of delivering visual experience to the learner. Finally, we study
NorthCUA-Ultrafast to assess how direct action selection changes the
balance between task success and execution latency.

\subsection{Evaluation Setup}
\label{sec:evalsetup}

Computer use requires both finding the right control and maintaining the
user's objective through successive state changes. We select the six
benchmarks in Figure~\ref{fig:capability-overview} to cover these complementary
requirements. ScreenSpot-Pro tests dense professional interfaces, while
OSWorld-G adds functional and spatial instructions. Desktop evaluation
extends these decisions into application workflows, and Web evaluation
covers both user tasks and defect detection. Table~\ref{tab:protocols}
summarizes what each benchmark measures.

\begin{table}[t]\centering\small
\caption{The six-benchmark suite and the capabilities and metrics covered
by each evaluation.}
\label{tab:protocols}
\begin{tabularx}{\linewidth}{P{35mm}XP{32mm}}
\toprule
Benchmark & Coverage & Metric (\%) \\
\midrule
\rowcolor{NexLight}\multicolumn{3}{l}{\textbf{\textcolor{TiffanyDeep}{Grounding}}}\\
ScreenSpot-Pro\newline\citep{li2025screenspotpro} & High-resolution professional software and small controls & Localization accuracy \\
\addlinespace[4pt]
OSWorld-G\newline\citep{xie2025osworldg} & Element semantics, layout, precise targeting, and absent targets & Grounding accuracy \\
\addlinespace[4pt]
\rowcolor{NexLight}\multicolumn{3}{l}{\textbf{\textcolor{TiffanyDeep}{Desktop}}}\\
OSWorld-Verified\newline\citep{xie2025osworldverified} & Application, file, and cross-application workflows & Task success \\
\addlinespace[4pt]
OSWorld~2.0\newline\citep{yuan2026osworld2} & Longer workflows with dependent requirements & Full completion, partial score \\
\addlinespace[4pt]
\rowcolor{NexLight}\multicolumn{3}{l}{\textbf{\textcolor{TiffanyDeep}{Web}}}\\
WebArena-Verified\newline\citep{hattami2025webarenaverified} & Stateful websites and outcome-checked user tasks & Task success \\
\addlinespace[4pt]
WebTestBench\newline\citep{kong2026webtestbench} & Interactive defect detection with a gold checklist & Precision, recall, F1 \\
\bottomrule
\end{tabularx}
\end{table}

\paragraph{Models and checkpoints.}
\begin{samepage}
We evaluate the \name family at three sizes, Nano (9B), Mini (35B), and
Pro (397B), built on Qwen3.5 \citep{qwen2026qwen35}. We call the checkpoint after broad GUI
mid-training \emph{Mid} and the checkpoint after verified long-horizon
supervision \emph{SFT}. Thus, Mid already contains NorthCUA training.
Mini and Pro use nominal 35B/3B and 397B/17B total/active language-model
parameter counts, while Nano is dense. We keep the publicly released
Qwen3.5 checkpoints separate as backbone-family references. The main
grounding results use the long-horizon SFT checkpoints, while the main
desktop, WebArena, and WebTestBench results use the final RL checkpoints.
Ultrafast starts from the pre-RL Nano model. Dashes indicate unreported
metrics. WebTestBench includes models with at least one measured metric.
We report result metrics to one decimal, computing differences before rounding.
\par\end{samepage}

\paragraph{Grounding protocol.}
We evaluate SSP by direct inference on the full screenshot, without
iterative zoom or tool-assisted refinement, using all 1,581 examples.
OSWorld-G uses 564 Refined instructions and normalized 0--1000 click
coordinates. In addition to the main accuracy comparison, repeated
predictions characterize consistency and target coverage.
NorthCUA's OSWorld-G entries use best-complete-run accuracy.
Appendix~\ref{app:benchmarkprotocol} details the sampling settings and
reports three-run means and standard deviations.

\paragraph{Interactive evaluation.}
We use a maximum of 100 steps per task for OSWorld-Verified,
WebArena-Verified, and WebTestBench, and 500 steps for OSWorld~2.0.
OSWorld-Verified uses the officially supported 361-task setting, excluding
eight Google Drive tasks that require separate account configuration.\footnote{\url{https://timothyxxx.github.io/OSWorld/user_guides/run_public_evaluation.html}}
We evaluate OSWorld~2.0 on the full set of 108 tasks and retain binary
completion, requiring a normalized score of 1, and mean partial score
where available. We keep these measures separate because completing
some requirements can still leave a dependent workflow unfinished.
WebArena-Verified uses all 812 tasks and official outcome checks, with
screenshots and structured page references as the agent's observations.
WebTestBench uses the golden-checklist
setting, supplying the reference test items directly to the agent.
This isolates interactive defect detection from checklist generation.
We treat Fail as the positive class. Precision, recall, and F1 are computed
within each task and then averaged separately across tasks, giving each
task equal weight regardless of its checklist length.

\paragraph{Comparison models.}
We select representative GUI specialists and general-purpose models from
original papers and official model cards, including explicitly attributed
evaluations of other providers' models. Source markers distinguish our
evaluations from published scores within each model group. Our OSWorld-G
evaluation includes nine general-purpose baselines. Model-name citations
identify original papers, reports, or official model pages, while
Appendix~\ref{app:benchmarkprotocol} collects score sources and protocol
details, including benchmark versions and action interfaces
\citep{tong2026maigym,zhou2026opengcua}.

\subsection{GUI Grounding Results}
\label{sec:groundingresults}

NorthCUA achieves state-of-the-art grounding results among the systems
compared in Table~\ref{tab:mainresults}, with Pro reaching 93.3\% on SSP
and 91.0\% on OSWorld-G. The two benchmarks test complementary requirements,
from locating small controls in dense professional interfaces to following
functional and spatial instructions. We report the final long-horizon SFT
checkpoints, using direct SSP inference and the Refined OSWorld-G
instructions described in Section~\ref{sec:evalsetup}.

Claude Opus 5.5 is the strongest reference on both benchmarks in this
comparison, scoring 88.1\% on SSP and 89.9\% on OSWorld-G in our
evaluation. Pro exceeds these scores by 5.2 and 1.1 percentage points,
respectively. On SSP, Pro also leads Qwen3.8-Max's reported
84.5\% by 8.8 points. On OSWorld-G, it exceeds Nex-N2.5-Pro's
reported 87.4\% by 3.6 points.

The smaller tiers retain much of this capability. Mini reaches 92.4\%
on SSP and 87.9\% on OSWorld-G with 35B total and 3B active
language-model parameters. Its SSP score is within one percentage point
of Pro. Nano reaches 72.9\% and 83.0\% with 9B parameters, with SSP
accuracy close to GUI-Owl-1.5-32B-Instruct's reported 72.9\%.
To understand what these aggregate scores capture, we next separate
geometric precision, control semantics, and target availability.

\begin{table}[t]\centering\small
\setlength{\tabcolsep}{4pt}
\caption{Grounding accuracy (\%) on ScreenSpot-Pro (SSP) and OSWorld-G
across general-purpose models, GUI specialists, and the three NorthCUA tiers.}
\label{tab:mainresults}
\begin{tabularx}{\linewidth}{X C{25mm} C{18mm} C{24mm}}
\toprule
Model & Parameters & SSP & OSWorld-G \\
\midrule
\rowcolor{NexLight}\multicolumn{4}{l}{\textbf{General-purpose models}}\\
Qwen3.5-35B-A3B$^\dagger$ \citep{qwen2026qwen35} & 35B / 3B & 64.5 & 63.6 \\
Qwen3.5-27B$^\dagger$ \citep{qwen2026qwen35} & 27B & 68.4 & 67.9 \\
Qwen 3.7 Plus$^\dagger$ \citep{qwen2026qwen37plus} & 397B / 17B & 68.9 & 78.2 \\
Qwen3.8-Max$^*$ \citep{qwen2026qwen38} & 2.4T / 95B & 84.5 & 71.6 \\
Seed 2.1 Pro$^\dagger$ \citep{bytedance2026seed21} & Undisclosed & 65.3 & 78.0 \\
Claude Opus 4.8 \citep{anthropic2026opus48systemcard} & Undisclosed & 82.3 & \textendash \\
Claude Opus 5.5$^*$ \citep{anthropic2026opus55} & Undisclosed & 88.1 & 89.9 \\
Claude Fable 5.1$^*$ \citep{anthropic2026fable51systemcard} & Undisclosed & 81.0 & 87.2 \\
GPT-5.6 Sol \citep{openai2026gpt56sol} & Undisclosed & 76.9 & 77.7 \\
Kimi-K3$^*$ \citep{moonshot2026kimik3card} & 2.8T / 104B & 67.7 & 73.8 \\
GPT-6 Astra$^*$ \citep{openai2026gpt6astra} & Undisclosed & 80.0 & 77.0 \\
GPT-6.1 Sol$^*$ \citep{openai2026gpt61sol} & Undisclosed & 81.2 & 73.6 \\
DeepSeek-V4.1-Flash$^*$ \citep{deepseek2026v41flashcard} & 552B / 16B & 60.1 & 60.5 \\
GLM-5.3-Flash$^*$ \citep{zai2026glm53flashcard} & 320B / 18B & 56.7 & 68.8 \\
Gemini 4 Argon$^*$ \citep{deepmind2026gemini4argon} & Undisclosed & 77.6 & 79.1 \\
Nex-N2.5-mini \citep{nexagi2026nexn25mini} & 35B & \textendash & 82.9 \\
Nex-N2.5-Pro \citep{nexagi2026nexn25pro} & 397B & \textendash & 87.4 \\
\midrule
\rowcolor{NexLight}\multicolumn{4}{l}{\textbf{GUI-specialized models}}\\
GUI-Owl-1.5-8B-Instruct \citep{xu2026mobileagent35} & 8B & 71.1 & 69.3 \\
GUI-Owl-1.5-32B-Instruct \citep{xu2026mobileagent35} & 32B & 72.9 & 69.7 \\
UI-Venus-1.5-8B \citep{venusteam2026uivenus15} & 8B & 68.4 & 74.1 \\
UI-Venus-1.5-30B-A3B \citep{venusteam2026uivenus15} & 30B / 3B & 69.6 & 76.4 \\
MAI-UI-32B \citep{zhou2025maiui} & 32B & 67.9 & 73.9 \\
Qwen-UI-Agent-4B \citep{zhou2026qwenuiagent} & 4B & 67.8 & 70.5 \\
Qwen-UI-Agent-35B-A3B \citep{zhou2026qwenuiagent} & 35B / 3B & 76.1 & 74.6 \\
Qwen-UI-Agent-27B \citep{zhou2026qwenuiagent} & 27B & 76.6 & 78.5 \\
\midrule
\rowcolor{NexLight}\multicolumn{4}{l}{\textbf{Our NorthCUA}}\\
\name-Nano & 9B & 72.9 & 83.0 \\
\name-Mini & 35B / 3B & 92.4 & 87.9 \\
\name-Pro & 397B / 17B & \textbf{93.3} & \textbf{91.0} \\
\bottomrule
\end{tabularx}
\par\vspace{3pt}
\begin{minipage}{\linewidth}\footnotesize
$^*$Our evaluation, applying only to OSWorld-G for Qwen3.8-Max.
$^\dagger$Evaluated by the Qwen-UI-Agent authors.
Score sources are listed in Appendix~\ref{app:benchmarkprotocol}. Parameters give reported
total/active counts.
Bold marks column maxima. Dashes indicate unavailable results.

\end{minipage}
\end{table}

\paragraph{Broad coverage and fine localization.}
Mini and Pro both exceed 90\% SSP accuracy in all six application groups.
Pro reaches 95.0\% in CAD and 94.5\% in scientific software.
This broad coverage also includes controls
with little room for a coordinate error. To measure that difficulty
directly, we pair the three models' predictions on all 1,581 examples and
group targets by size, measured as the smaller of the target box's width
and height in the original screenshot (Table~\ref{tab:groundingdiagnostics}).
A target in the smallest group is less than 16 pixels wide or tall.
On these narrow targets, accuracy rises from Nano's 53.4\% to Mini's
85.2\% and Pro's 92.1\%. Pro's 6.8-point gain over Mini is larger
than its 1.2-point gain on targets at least 32 pixels wide and tall.
Its stronger aggregate result therefore also extends to the smallest
target regions in this breakdown.

\begin{table}[t]\centering\small
\setlength{\tabcolsep}{4pt}
\caption{Trace-based grounding accuracy (\%). SSP uses one prediction per
item. Target size is the smaller of the annotated box's width and height,
measured in original-image pixels. OSWorld-G averages
three draws per item. Subgroup definitions and repeated-draw breakdowns
appear in Appendix~\ref{app:benchmarkprotocol}.}
\label{tab:groundingdiagnostics}
\begin{tabularx}{\linewidth}{C{22mm} >{\centering\arraybackslash}X C{12mm} *{3}{C{26mm}}}
\toprule
Benchmark & Target subset & Items & \name-Nano & \name-Mini & \name-Pro\\
\midrule
\multirow{3}{*}{SSP} & $<16$ px & 88 & 53.4 & 85.2 & 92.1\\
& $16$--$<32$ px & 1,079 & 72.4 & 92.1 & 92.4\\
& $\geq32$ px & 414 & 78.3 & 94.7 & 95.9\\
\midrule
\multirow{2}{*}{OSWorld-G} & Visible target & 510 & 84.0 & 88.6 & 91.7\\
& Absent target & 54 & 67.9 & 74.1 & 75.9\\
\bottomrule
\end{tabularx}
\end{table}

\paragraph{Interpreting specialized controls.}
\begin{samepage}
A small target must also be matched to the user's intended function.
Pro reaches 90.2\% on SSP's icon targets, followed by Mini at 88.6\%
and Nano at 55.0\%.
The difference persists within the most populous size band of
16--$<32$ pixels, where Pro and Mini both locate 89.0\% of the 418 icon
targets and Nano locates 53.1\%. Their corresponding text-target scores
are closer at 94.6\%, 94.1\%, and 84.6\%, respectively.
Even after this coarse size grouping, the gap between Nano and the larger
models remains greater for icon targets than for text targets.
\par\end{samepage}

\begin{samepage}
This distinction helps explain why
our data construction addresses both target geometry and control meaning.
The broad route retains the image coordinates, while documentation supplies
functional descriptions for specialized controls.
\par\end{samepage}

\begin{figure}[t]
\centering
\includegraphics[width=\linewidth]{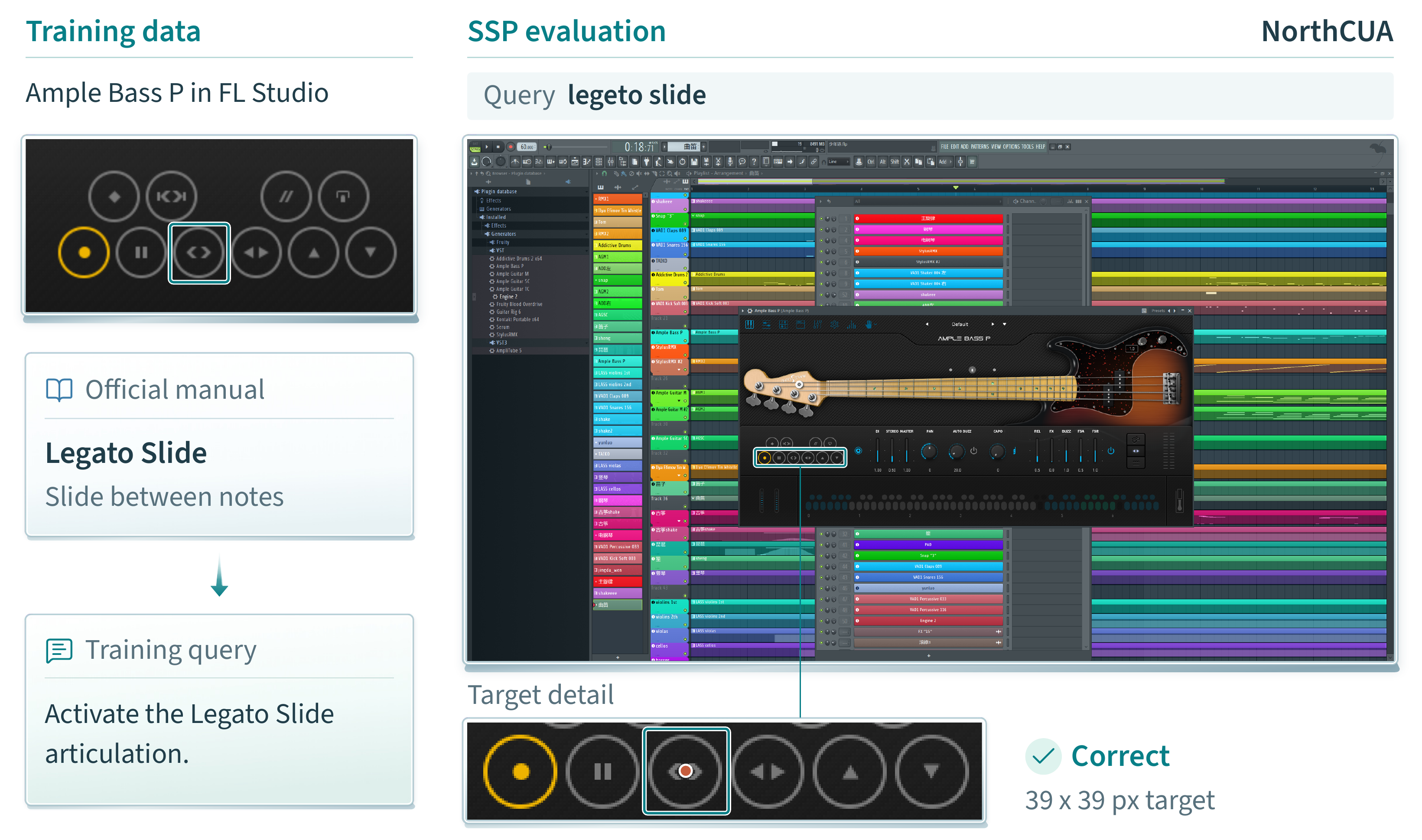}
\caption{\textbf{Grounding a documented control.}
Left, an actual documentation-grounded annotation pairs an Ample Bass P
control with its manual-defined articulation and a generated training query.
Right, NorthCUA localizes this control in an SSP screenshot with a different
surrounding GUI state, following the recorded query \emph{legeto slide}.
The target's local appearance and position match the construction view.
In the enlarged detail, the teal box marks the reference target and the orange
point is the recorded prediction. Enlargements and annotations are reader aids. Documentation is
used in data construction, not supplied during direct evaluation.}
\label{fig:eval-grounding}
\end{figure}

Figure~\ref{fig:eval-grounding} illustrates this construction with the Legato Slide control in
Ample Bass P, a virtual instrument hosted in FL Studio. Its articulation
buttons encode different playing techniques through small symbols, so the
icon alone gives little indication of the requested function.

The construction record links this control to the manual's Legato Slide
description \citep{amplesoundabpmanual} and generates the instruction
\emph{Activate the Legato Slide articulation}, retaining the original target
location.

This documentation-grounded package was used in Nano and Mini training.
In the SSP example, NorthCUA correctly locates the documented control
from the full screenshot and query alone, without access to the manual
during inference.

\paragraph{Target availability and repeated predictions.}
Reliable grounding also requires recognizing when a requested target is
absent. OSWorld-G tests this distinction alongside visible-target localization.
Across three draws, Pro averages 91.7\% on the 510 visible-target items,
compared with Mini's 88.6\%. Its correct-refusal rate reaches 75.9\% on
the 54 absent-target items, compared with Mini's 74.1\%
(Table~\ref{tab:groundingdiagnostics}). The larger gain is therefore in
locating visible controls, while recognizing when to decline a location
request remains harder across all three tiers. This distinction matters
when a predicted coordinate becomes an agent's next action.

Repeated sampling then reveals whether correct predictions persist across
draws. Pro is correct in all three draws on 489 of 564 items, compared
with 469 for Mini and 445 for Nano. Its pass@3 reaches 94.0\%, covering
530 items, while its three complete runs range from 89.4\% to 91.0\%.
Pro also succeeds at least once on 36 of Mini's 53 persistently missed
items, including 26 that Pro gets right in every draw. Thus, its stronger
aggregate result accompanies both more stable successes and coverage of
additional difficult items.

\subsection{Desktop Agent Results}
\label{sec:desktopresults}

Grounding accuracy is useful only if the policy can sustain correct decisions
until the requested software state is reached.
Tables~\ref{tab:desktopresults} and~\ref{tab:osworld2results}
therefore compare complete application workflows and longer dependent tasks.
NorthCUA-Pro reaches 88.6\% on OSWorld-Verified, the highest score among
the compared systems. Mini reaches 85.3\% with 35B total parameters,
while the 9B Nano reaches 78.9\%. These final RL results extend the grounding
comparison from selecting individual controls to completing software tasks.

\begin{table}[t]\centering\small
\setlength{\tabcolsep}{4pt}
\caption{OSWorld-Verified results (\%). NorthCUA uses RL checkpoints on 361 tasks.
Reference protocols, including partial-credit variants, are detailed in
Appendix~\ref{app:benchmarkprotocol}. An asterisk marks our evaluation.
Bold marks the best available NorthCUA result.}
\label{tab:desktopresults}
\begin{tabularx}{\linewidth}{X C{30mm} C{27mm}}
\toprule
Model & Parameters & Score\\
\midrule
\rowcolor{NexLight}\multicolumn{3}{l}{\textbf{General-purpose models}}\\
Qwen3.5-9B \citep{qwen359bcard} & 9B & 41.8\\
Qwen3.5-35B-A3B \citep{qwen3535bcard} & 35B / 3B & 54.5\\
Qwen3.5-397B-A17B \citep{qwen35397bcard} & 397B / 17B & 62.2\\
Qwen3.8-Max \citep{qwen2026qwen38} & 2.4T / 95B & 86.1\\
Seed 2.1 Pro \citep{bytedance2026seed21} & Undisclosed & 78.8\\
Kimi-K3 (max) \citep{kimiteam2026kimik3} & 2.8T / 104B & 84.8\\
MiniMax M3 \citep{minimax2026m3card} & $\sim$428B / 23B & 75.2\\
GLM-5.3-Flash \citep{zai2026glm53flashcard} & 320B / 18B & 84.7\\
DeepSeek-V4.1-Flash$^*$ \citep{deepseek2026v41flashcard} & 552B / 16B & 71.7\\
DeepSeek-V4-Flash-Vision \citep{deepseek2026v4flashvision} & Undisclosed & 76.7\\
GPT-5.6 Sol \citep{openai2026gpt56sol} & Undisclosed & 83.2\\
Claude Opus 5 [1m] \citep{anthropic2026opus5systemcard} & Undisclosed & 83.4\\
Claude Fable 5 [1m] \citep{anthropic2026fable5systemcard} & Undisclosed & 86.0\\
Gemini 3.6 Flash \citep{deepmind2026gemini36flash} & Undisclosed & 83.0\\
Nex-N2.5-mini \citep{nexagi2026nexn25mini} & 35B & 71.2\\
Nex-N2.5-Pro \citep{nexagi2026nexn25pro} & 397B & 82.2\\
\midrule
\rowcolor{NexLight}\multicolumn{3}{l}{\textbf{GUI-specialized models}}\\
OpenCUA-72B \citep{wang2025opencua} & 72B & 45.0\\
GUI-Owl-1.5-8B-Instruct \citep{xu2026mobileagent35} & 8B & 52.3\\
GUI-Owl-1.5-32B-Instruct \citep{xu2026mobileagent35} & 32B & 56.5\\
UI-Venus-2-9B \citep{venusteam2026uivenus2} & 9B & 70.8\\
UI-Venus-2-27B \citep{venusteam2026uivenus2} & 27B & 80.5\\
ScaleCUA-Qwen3.5-9B \citep{lv2026scalecua} & 9B & 68.7\\
MAI-Gym-9B \citep{tong2026maigym} & 9B & 76.2\\
Qwen-UI-Agent-27B \citep{zhou2026qwenuiagent} & 27B & 79.5\\
Qwen-CUA \citep{lu2026qwencua} & 397B / 17B & 86.2\\
Qwen-CUA-Max \citep{lu2026qwencua} & $>1$T & 87.6\\
\midrule
\rowcolor{NexLight}\multicolumn{3}{l}{\textbf{Our NorthCUA}}\\
\name-Nano & 9B & 78.9\\
\name-Mini & 35B / 3B & 85.3\\
\name-Pro & 397B / 17B & \textbf{88.6}\\
\bottomrule
\end{tabularx}
\end{table}

Pro exceeds Qwen-CUA-Max's reported 87.6\% by 1.0 percentage points
and Qwen3.8-Max's 86.1\% by 2.5 points, using 397B total parameters
against their reported trillion-scale sizes. At a smaller scale, Mini
exceeds Nex-N2.5-Pro's 82.2\% by 3.1 points, while Nano exceeds
MAI-Gym-9B's 76.2\% by 2.7 points. The comparison covers both general-purpose and GUI-specialized
systems under the source-specific protocols in
Appendix~\ref{app:benchmarkprotocol}. Section~\ref{sec:stageexperiments}
examines how performance changes within the NorthCUA training pipeline.

OSWorld~2.0 extends the comparison to longer workflows with dependent
requirements (Table~\ref{tab:osworld2results}). Binary completion tests
whether all requirements are met, while partial credit records intermediate
progress. Both measures increase from Nano through Mini to Pro in the
final RL checkpoints. Pro reaches 30.6\% binary completion and 66.8\%
partial credit on the full 108-task set. Within the 108-task reference
block, these scores exceed GPT-5.6 Sol by 3.2 and 4.1 percentage points,
respectively, under the listed evaluation protocols. The result combines
higher earned credit with a higher rate of fully completed workflows.
Section~\ref{sec:stageexperiments} examines how both measures develop
through the NorthCUA training pipeline.

\begin{table}[t]\centering\small
\setlength{\tabcolsep}{4pt}
\caption{OSWorld~2.0 results (\%). Binary requires full completion, while
Partial measures earned task credit. NorthCUA uses all 108 tasks.
Reference blocks identify reported task scope, with release dates,
budgets, and scoring details in Appendix~\ref{app:benchmarkprotocol}.
An asterisk marks a baseline evaluated by us. Dashes denote unreported results.}
\label{tab:osworld2results}
\begin{tabularx}{\linewidth}{X C{30mm} C{22mm} C{22mm}}
\toprule
Model & Parameters & Binary & Partial\\
\midrule
\rowcolor{NexLight}\multicolumn{4}{l}{\textbf{Reference models, 108 tasks}}\\
GPT-5.5 \citep{openai2026gpt55} & Undisclosed & 13.0 & 49.5\\
GPT-5.6 Sol \citep{openai2026gpt56sol} & Undisclosed & 27.3 & 62.7\\
Claude Opus 4.8 \citep{anthropic2026opus48systemcard} & Undisclosed & 20.6 & 54.8\\
Claude Opus 5 \citep{anthropic2026opus5systemcard} & Undisclosed & 31.4 & 68.3\\
Claude Fable 5 \citep{anthropic2026fable5systemcard} & Undisclosed & 36.1 & 72.9\\
Claude Fable 5.1 \citep{anthropic2026fable51systemcard} & Undisclosed & 41.7 & 77.9\\
GLM-5.3-Flash$^*$ \citep{zai2026glm53flashcard} & 320B / 18B & 14.8 & 48.7\\
MiniMax M3 \citep{minimax2026m3card} & $\sim$428B / 23B & 4.6 & 22.3\\
UI-Venus-2-9B \citep{venusteam2026uivenus2} & 9B & 0.0 & 7.5\\
UI-Venus-2-27B \citep{venusteam2026uivenus2} & 27B & 2.8 & 13.2\\
MAI-Gym-9B \citep{tong2026maigym} & 9B & 3.7 & 11.3\\
\midrule
\rowcolor{NexLight}\multicolumn{4}{l}{\textbf{Published references, task scope unspecified}}\\
Qwen3.8-Max \citep{qwen2026qwen38} & 2.4T / 95B & 19.4 & 46.7\\
Qwen-UI-Agent-27B \citep{zhou2026qwenuiagent} & 27B & 13.9 & 40.0\\
Qwen-CUA \citep{lu2026qwencua} & 397B / 17B & 18.5 & 48.4\\
Qwen-CUA-Max \citep{lu2026qwencua} & $>1$T & 21.2 & 53.3\\
Kimi-K3 \citep{kimiteam2026kimik3} & 2.8T / 104B & \textendash & 58.3\\
MiniMax M3.1 Flash \citep{minimax2026m31flash} & Undisclosed & \textendash & 61.7\\
DeepSeek-V4.1-Flash \citep{deepseek2026v41flashcard} & 552B / 16B & \textendash & 54.3\\
Gemini 3.8 Flash \citep{deepmind2026gemini38flash} & Undisclosed & \textendash & 59.0\\
Nex-N2.5-mini \citep{nexagi2026nexn25mini} & 35B & \textendash & 30.5\\
Nex-N2.5-Pro \citep{nexagi2026nexn25pro} & 397B & \textendash & 56.4\\
\midrule
\rowcolor{NexLight}\multicolumn{4}{l}{\textbf{Our NorthCUA}, 108 tasks}\\
\name-Nano & 9B & 13.0 & 45.6\\
\name-Mini & 35B / 3B & 18.5 & 50.1\\
\name-Pro & 397B / 17B & \textbf{30.6} & \textbf{66.8}\\
\bottomrule
\end{tabularx}
\end{table}

\subsection{Web Agent Results}
\label{sec:webresults}

\begin{samepage}
We next test whether accurate interaction supports complete requests on
stateful websites. NorthCUA-Pro's RL checkpoint reaches 73.9\% on
WebArena-Verified, the highest score in Table~\ref{tab:webmainresults}.
Mini reaches 70.4\%, while Nano reaches 67.2\%, also after RL. We compare these
results with published references and six additional general-purpose
models evaluated by us. We then use the available trajectory exports to
examine which website workflows are completed and where difficulties remain.
\par\end{samepage}

\begin{table}[t]\centering\small
\caption{Task success (\%) on WebArena-Verified across reference
systems and NorthCUA models.}
\label{tab:webmainresults}
\setlength{\tabcolsep}{4pt}
\begin{tabularx}{\linewidth}{X C{30mm} C{27mm}}
\toprule
Model & Parameters & Success \\
\midrule
\rowcolor{NexLight}\multicolumn{3}{l}{\textbf{General-purpose models}}\\
Qwen3.7-Plus \citep{qwen2026qwen37plus} & 397B / 17B & 55.3\\
Qwen3.8-Max \citep{qwen2026qwen38} & 2.4T / 95B & 66.8\\
Claude Opus 4.8 \citep{anthropic2026opus48systemcard} & Undisclosed & 67.9\\
Claude Fable 5 \citep{anthropic2026fable5systemcard} & Undisclosed & 71.3\\
Gemini 3.1 Pro \citep{deepmind2026gemini31pro} & Undisclosed & 64.3\\
GPT-5.6 Sol \citep{openai2026gpt56sol} & Undisclosed & 69.7\\
Kimi-K3 \citep{kimiteam2026kimik3} & 2.8T / 104B & 71.6\\
DeepSeek-V4-Flash-Vision \citep{deepseek2026v4flashvision} & Undisclosed & 62.3\\
GPT-6 Astra$^*$ \citep{openai2026gpt6astra} & Undisclosed & 71.6\\
GPT-6.1 Sol$^*$ \citep{openai2026gpt61sol} & Undisclosed & 69.7\\
Claude Opus 5.5$^*$ \citep{anthropic2026opus55} & Undisclosed & 70.3\\
Gemini 4 Argon$^*$ \citep{deepmind2026gemini4argon} & Undisclosed & 65.8\\
GLM-5.3-Flash$^*$ \citep{zai2026glm53flashcard} & 320B / 18B & 65.8\\
DeepSeek-V4.1-Flash$^*$ \citep{deepseek2026v41flashcard} & 552B / 16B & 67.5\\
Nex-N2.5-mini \citep{nexagi2026nexn25mini} & 35B & 63.4\\
Nex-N2.5-Pro \citep{nexagi2026nexn25pro} & 397B & 67.6\\
\midrule
\rowcolor{NexLight}\multicolumn{3}{l}{\textbf{Our NorthCUA}}\\
\name-Nano & 9B & 67.2\\
\name-Mini & 35B / 3B & 70.4\\
\name-Pro & 397B / 17B & \textbf{73.9}\\
\bottomrule
\end{tabularx}
\par\vspace{3pt}
\begin{minipage}{\linewidth}\footnotesize
$^*$Our evaluation. All NorthCUA entries use RL checkpoints.
Bold marks the highest NorthCUA score. Appendix~\ref{app:benchmarkprotocol}
records score sources, evaluation settings, and task-level breakdowns.
\end{minipage}
\end{table}

\begin{samepage}
Pro's score is 2.3 points above the joint-highest reference score of
71.6\%, reported for Kimi-K3 and obtained by GPT-6 Astra in our evaluation.
Mini reaches 70.4\%
with 35B parameters, exceeding Qwen3.8-Max's reported 66.8\% and
Nex-N2.5-Pro's 67.6\%. Nano's 67.2\% provides a 9B alternative.
\par\end{samepage}

\begin{figure}[t]
\centering
\includegraphics[width=\linewidth]{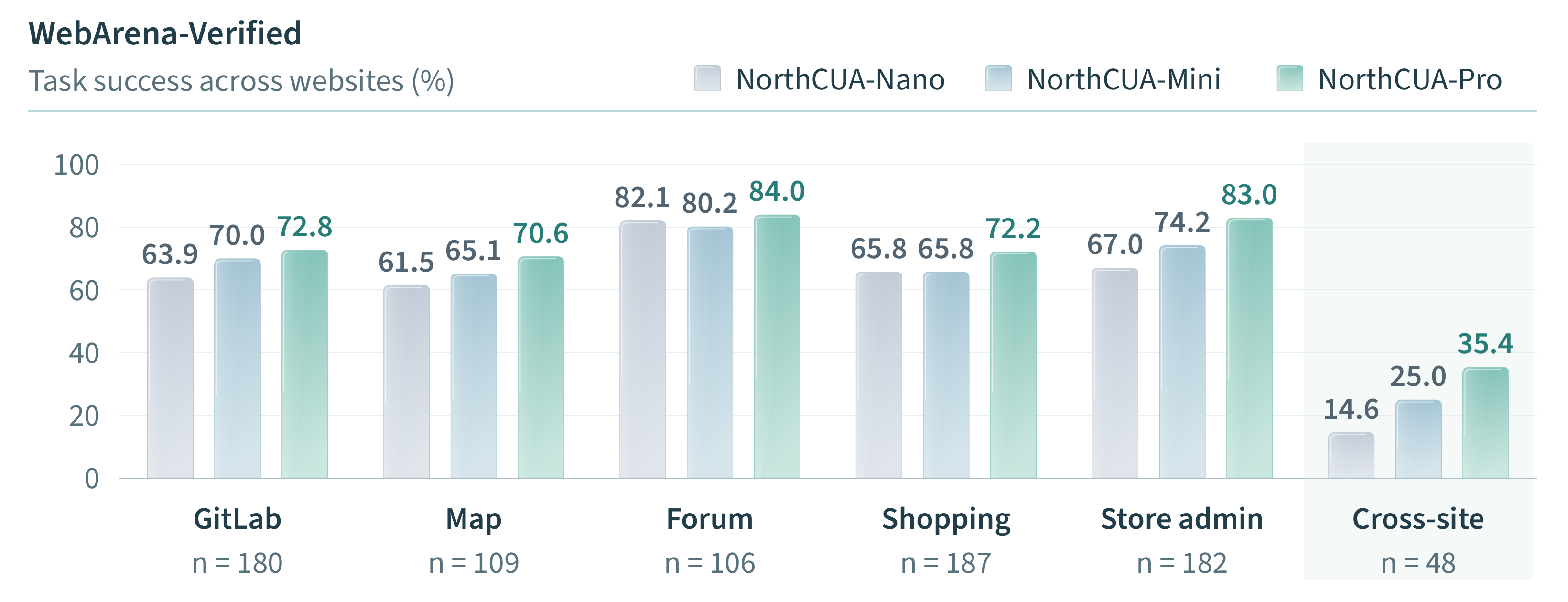}
\caption{\textbf{Web success by website scope.}
The six groups partition all 812 WebArena-Verified tasks using retained
attempts. Nano and Mini use SFT checkpoints, while Pro uses RL.
Pro exceeds 70\% in every single-site group and completes 17 of 48
cross-site tasks. Appendix~\ref{app:benchmarkprotocol} gives the
selection and group details.}
\label{fig:webbreakdown}
\end{figure}

\paragraph{Coverage across websites.}
\begin{samepage}
Figure~\ref{fig:webbreakdown} examines trajectories across six website
groups, using Nano and Mini after SFT and Pro after RL.
Pro exceeds 70\% success in each of the five single-site groups, reaching
84.0\% on forums and 83.0\% in store administration. This coverage spans
consumer-facing websites and operational interfaces with editable records.
\par\end{samepage}

\begin{samepage}
Cross-site tasks remain more demanding. Pro completes 17 of 48 tasks
(35.4\%), compared with 583 of 764 single-site tasks (76.3\%).
These workflows require carrying information or requirements between
websites, making them a useful test of the sustained execution targeted
by our long-horizon experience pipeline. The gap identifies cross-site
coordination as a continuing challenge despite broad single-site coverage.
\par\end{samepage}

\begin{samepage}
A complementary breakdown by requested outcome shows that Pro completes
274 of 374 state-changing tasks (73.3\%), alongside 245 retrieval tasks
and 81 navigation tasks. Thus, the result includes workflows that must
leave a correct persistent site state, beyond locating or reading a page.
Mini completes 253 state-changing and 223 retrieval tasks, compared with
Nano's 240 and 204. Its additional completions are concentrated in these
two categories, while navigation counts are similar across the two models.
Appendix~\ref{app:benchmarkprotocol} gives the group counts and the
decision-length distributions for all three checkpoints.
\par\end{samepage}

\paragraph{Execution length and local setbacks.}
\begin{samepage}
The Pro export contains 14,604 model decisions across its 812 retained
attempts. Successful attempts have a median of ten decisions. Of these,
24 contain at least 50 decisions, and the longest reaches
93. This range shows completion across both short interactions and
extended workflows within the same benchmark.
\par\end{samepage}

\begin{samepage}
For Pro, unsuccessful attempts have a median of 15.5 decisions and contribute
38.4\% of the exported decisions despite accounting for 26.1\% of
tasks. Mini shows a similar concentration of work in failed tasks, whose
median is 18 decisions compared with 10 for successes. Its 260 failures
account for 50.6\% of the 15,224 recorded decisions. This difference
between successful and unsuccessful execution motivates our emphasis on
feedback-guided correction during long workflows. Counts include terminal
responses, with the record boundaries defined in
Appendix~\ref{app:benchmarkprotocol}.
\par\end{samepage}

\begin{minipage}{\linewidth}
Section~\ref{sec:evalcases} illustrates such a correction in a Pro
trajectory. The agent uses interface feedback to recover from an
unavailable option while preserving the user's date constraint.
\par\end{minipage}

\paragraph{Web testing.}
\begin{samepage}
Completing a user request and detecting defects require different judgments.
We therefore evaluate defect detection on WebTestBench with the gold
checklist supplied to each agent \citep{kong2026webtestbench}.
Given these test items, the agent must exercise the application and
distinguish genuine defects from correctly implemented behavior.
Table~\ref{tab:webtestresults} reports task-averaged precision, recall,
and F1. NorthCUA-Pro achieves the highest F1 at 60.3\%, exceeding the
strongest comparison model, GPT-6 Astra at 58.2\%, by 2.1 percentage
points. At 35B, NorthCUA-Mini reaches 54.4\% F1, improving on the
same-size Nex-N2.5-mini by 5.8 points. Its precision and recall both
increase by 6.3 points in this comparison, indicating gains in both
the correctness of reported defects and the fraction of defects found.
\par\end{samepage}

\begin{samepage}
The separate metrics clarify these detection profiles. NorthCUA-Pro also
achieves the highest precision at 67.8\%, alongside 63.7\% recall.
GPT-6 Astra combines the highest recall of 74.6\% with 55.1\%
precision. GPT-5.6 Sol shows a similar emphasis on recall, at 70.3\%
against 54.8\% precision, with an F1 of 56.1\%.
These differences distinguish finding a larger fraction of
defects from making more accurate defect reports. We compute F1 separately
for each task and report the mean of those per-task scores.
The development case in Figure~\ref{fig:webtestcase}, separate from the
scored WebTestBench evaluation, illustrates how such findings can guide
a test--repair--retest cycle. Appendix~\ref{app:web} describes the
browser-evidence qualification protocol.
\par\end{samepage}

\begin{table}[t]\centering\small
\setlength{\tabcolsep}{4pt}
\caption{WebTestBench defect detection with a gold checklist.
All scores (\%) are task-level macro averages from our 100-step evaluation.}
\label{tab:webtestresults}
\begin{tabularx}{\linewidth}{X C{23mm} *{3}{C{17mm}}}
\toprule
Model & Parameters & Precision & Recall & F1 \\
\midrule
\rowcolor{NexLight}\multicolumn{5}{l}{\textbf{General-purpose models}}\\
Nex-N2.5-mini \citep{nexagi2026nexn25mini} & 35B & 54.7 & 51.4 & 48.6\\
Nex-N2.5-Pro \citep{nexagi2026nexn25pro} & 397B & 59.5 & 55.8 & 52.8\\
Kimi-K3 \citep{moonshot2026kimik3card} & 2.8T / 104B & 67.3 & 48.2 & 52.4\\
GPT-5.6 Sol \citep{openai2026gpt56sol} & Undisclosed & 54.8 & 70.3 & 56.1\\
GPT-6 Astra \citep{openai2026gpt6astra} & Undisclosed & 55.1 & \textbf{74.6} & 58.2\\
GPT-6.1 Sol \citep{openai2026gpt61sol} & Undisclosed & 53.0 & 73.8 & 56.8\\
GLM-5.3-Flash \citep{zai2026glm53flashcard} & 320B / 18B & 60.6 & 54.0 & 53.1\\
Qwen3.8-Max \citep{qwen2026qwen38} & 2.4T / 95B & 59.6 & 54.2 & 52.3\\
DeepSeek-V4.1-Flash \citep{deepseek2026v41flashcard} & 552B / 16B & 59.7 & 48.9 & 50.1\\
\midrule
\rowcolor{NexLight}\multicolumn{5}{l}{\textbf{Our NorthCUA}}\\
\name-Nano & 9B & 53.6 & 53.7 & 48.0\\
\name-Mini & 35B / 3B & 61.0 & 57.7 & 54.4\\
\name-Pro & 397B / 17B & \textbf{67.8} & 63.7 & \textbf{60.3}\\
\bottomrule
\end{tabularx}
\par\vspace{3pt}
\begin{minipage}{\linewidth}\footnotesize
Bold marks the best reported value in each column.
Each metric is averaged over tasks independently.
\end{minipage}
\end{table}

\subsection{Progress Across Training Stages}
\label{sec:stageexperiments}

\begin{samepage}
The final-checkpoint comparisons summarize benchmark performance. We next
examine how the reported scores develop through training, beginning with
the transition from broad mid-training to verified long-horizon SFT
(Table~\ref{tab:stages}). Desktop success increases
across all three tiers, by 7.4 points for Nano, 15.0 for Mini, and
10.0 for Pro. Pro reaches 85.9\% after this transition, while Mini
shows the largest absolute desktop gain.
\par\end{samepage}

\begin{samepage}
The Web results show a different distribution of gains. Nano improves
by 9.2 points and Mini by 2.6, narrowing their gap from 10.5 to
3.8 points. Pro improves from 69.1\% to 69.6\%, a smaller gain of
0.5 points. The largest gains therefore depend on both model size and
task domain. Mini gains most on desktop tasks, while Nano gains most on
Web tasks. These checkpoint comparisons measure the complete
training transition, including data selection and additional optimization.
Appendix~\ref{app:benchmarkprotocol} records the available evaluation
settings and aggregation rules for the stage comparisons.
\par\end{samepage}

The longer OSWorld~2.0 tasks expose the gap between partial progress and
complete workflows at the Mid stage. Across the three tiers, partial
scores range from 24.2\% to 30.6\%, while binary completion ranges
from 0.9\% to 9.3\% (Table~\ref{tab:stages}). Long-horizon SFT increases
partial scores by 16.7, 20.7, and 33.0 points for Nano, Mini, and Pro,
respectively. Binary completion also increases in every tier, with Pro
rising from 9.3\% to 27.3\%. The gains therefore include more fully
completed tasks as well as higher earned credit. These SFT checkpoints
provide the starting points for the joint-RL comparison in
Table~\ref{tab:rlresults}, allowing the two training transitions to be
examined separately.

\begin{table}[t]\centering\small
\caption{Results (\%) before and after long-horizon SFT across three
interactive benchmarks. OSWorld~2.0 separates full-task Binary completion
from Partial credit.
Table~\ref{tab:rlresults} continues the comparison from SFT to RL.}
\label{tab:stages}
\setlength{\tabcolsep}{4pt}
\renewcommand{\arraystretch}{1.24}
\newcommand{\stagevalue}[1]{\cellcolor{NexLight}\textbf{#1}}
\begin{tabularx}{\linewidth}{P{30mm}C{18mm}*{4}{>{\centering\arraybackslash}X}}
\toprule
\textbf{Model} & \textbf{Stage} & \makecell{\textbf{OSWorld-}\\\textbf{Verified}} & \multicolumn{2}{c}{\textbf{OSWorld~2.0}} & \makecell{\textbf{WebArena-}\\\textbf{Verified}}\\
\cmidrule(lr){3-3}\cmidrule(lr){4-5}\cmidrule(lr){6-6}
& & \textcolor{NexSubtitle}{Success} & \textcolor{NexSubtitle}{Binary} & \textcolor{NexSubtitle}{Partial} & \textcolor{NexSubtitle}{Success}\\
\midrule
\multirow{2}{*}{\name-Nano} & \textcolor{NexSubtitle}{Mid} & 68.7 & 0.9 & 24.2 & 54.9\\
& \cellcolor{NexLight}\textcolor{TiffanyDeep}{\textbf{SFT}} & \stagevalue{76.1} & \stagevalue{12.0} & \stagevalue{40.9} & \stagevalue{64.2}\\
\arrayrulecolor{NexRule}\cmidrule(lr){1-6}\arrayrulecolor{NexInk}
\multirow{2}{*}{\name-Mini} & \textcolor{NexSubtitle}{Mid} & 68.3 & 2.8 & 25.6 & 65.4\\
& \cellcolor{NexLight}\textcolor{TiffanyDeep}{\textbf{SFT}} & \stagevalue{83.3} & \stagevalue{18.5} & \stagevalue{46.3} & \stagevalue{68.0}\\
\arrayrulecolor{NexRule}\cmidrule(lr){1-6}\arrayrulecolor{NexInk}
\multirow{2}{*}{\name-Pro} & \textcolor{NexSubtitle}{Mid} & 75.8 & 9.3 & 30.6 & 69.1\\
& \cellcolor{NexLight}\textcolor{TiffanyDeep}{\textbf{SFT}} & \stagevalue{85.9} & \stagevalue{27.3} & \stagevalue{63.6} & \stagevalue{69.6}\\
\bottomrule
\end{tabularx}
\end{table}

\FloatBarrier
\begin{figure}[!t]
\centering
\includegraphics[width=\linewidth]{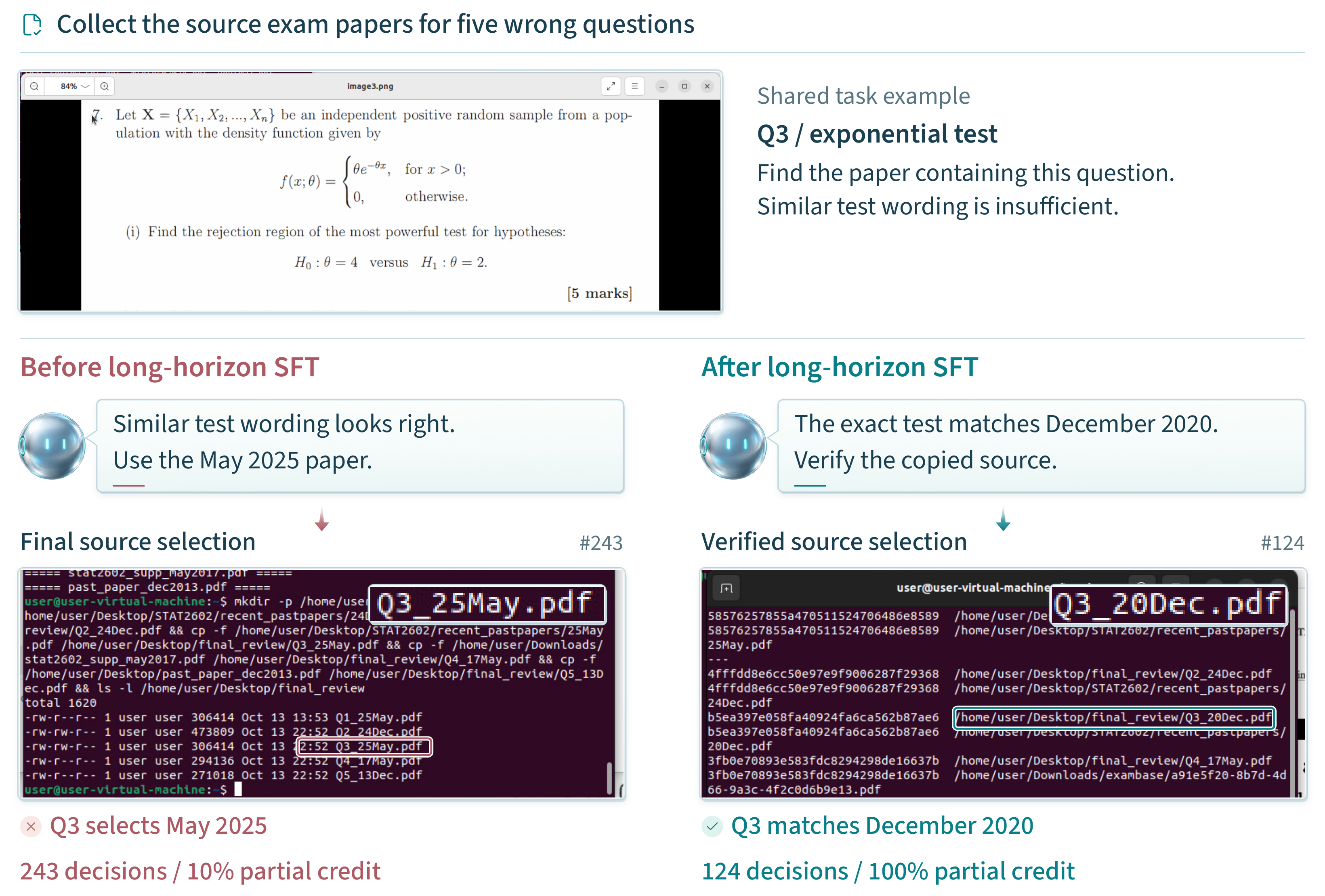}
\caption{Source-paper selection before and after long-horizon SFT.
A shared question anchors the two equal-width result views. Agent callouts
paraphrase recorded selection and checking decisions. Highlighted filenames
show the different Q3 sources, while scores cover each complete attempt.
Screenshots are cropped observations before the indexed decisions, with
enlarged filenames.
The runs use different dataset revisions and OCR settings.}
\label{fig:os2sftcase}
\end{figure}

\paragraph{From similar questions to verified source papers.}
\begin{samepage}
Figure~\ref{fig:os2sftcase} compares the same OSWorld~2.0 request in two
\name-Mini runs before and after long-horizon SFT, focusing on how the
agent selects and verifies source documents.
A student has collected
five question images and asks the agent to find their source exam papers.
Similar statistical topics recur across sessions, so recognizing the topic
alone can lead to a convincing but incorrect match. The third question,
shown at the top of the figure, asks for an exponential-model test with
specific null and alternative parameters.
\par\end{samepage}

The earlier run reads the question images and searches the candidate papers.
It treats related exponential-test wording as confirmation and assigns Q3
to May 2025. Its final folder contains all five named PDFs, but the complete
attempt receives 10\% partial credit. The later run matches Q3's specific
test to December 2020 and checks that the copied file agrees with its source.
This more discriminating search also appears elsewhere in the task. For Q5,
it rejects a 15-mark candidate that differs from the requested 22-mark
question, continues into Trash, and recovers the matching original paper.
The resulting attempt receives full credit after 124 decisions, compared
with 243 in the earlier run.

The contrast concerns what counts as completion. The earlier run assembles
all five named PDFs while accepting an incorrect source match. The
later run carries the question-specific criteria through search, rejection
and final file checking. These observed behaviors illustrate the sustained
verification targeted by long-horizon training. Appendix~\ref{app:os2cases}
gives the execution settings.
\FloatBarrier

\subsection{Joint Desktop--Web Reinforcement Learning}
\label{sec:rlexperiments}

Long-horizon SFT teaches the policy to follow verified demonstrations.
RL adds feedback on the states reached by the policy's own actions, allowing
successful and unsuccessful attempts at the same task to inform the next
update. We examine this transition across desktop and Web execution under
the shared training interface in Section~\ref{sec:rl}. The prepared RL task
pool contains approximately 1,500 desktop tasks and 500 Web tasks. Each
model starts from its long-horizon SFT checkpoint and learns from fresh
attempts on these tasks.

\paragraph{Task completion across platforms.}
\begin{samepage}
Table~\ref{tab:rlresults} compares each long-horizon SFT checkpoint
(Before RL) with its subsequent joint-training checkpoint (After RL).
Within each tier, the task set, tools, observation settings, and execution
budget stay fixed, allowing the comparison to track performance through
the transition from supervised learning to joint RL.
OSWorld~2.0 retains Binary and Partial separately to
distinguish complete workflows from progress on individual requirements.
\par\end{samepage}

\begin{samepage}
After joint RL, reported success increases on both platforms at all three
model sizes. OSWorld-Verified success rises by 2.9 points
for Nano, 2.0 for Mini, and 2.8 for Pro. The corresponding WebArena-Verified
gains are 3.0, 2.5, and 4.3 points. Pro reaches 88.6\% and 73.9\%
on the two benchmarks, respectively. The desktop gains are similar across
tiers, whereas Pro has the largest reported Web increase.
\par\end{samepage}

\begin{samepage}
OSWorld~2.0 distinguishes gains in earned credit from gains in full-task
completion. Partial scores increase by 4.7, 3.8, and 3.2 points for
Nano, Mini, and Pro, respectively. Nano and Pro also improve binary
completion by 0.9 and 3.3 points, with Pro reaching 30.6\% binary
completion and 66.8\% partial credit. Mini's binary score stays at
18.5\% as its partial score rises from 46.3\% to 50.1\%.
Together, the two metrics show where the final
checkpoints improve requirement satisfaction and where that improvement
also appears in complete workflows.
\par\end{samepage}

The remaining analyses address two different parts of the learning
process. We first measure how the runtime delivers long visual records
without changing their contents, then examine how task rewards distinguish
two superficially similar interaction paths.

\begin{table}[t]\centering\small
\setlength{\tabcolsep}{4pt}
\caption{Before- and after-RL results on three interactive benchmarks (\%).
OSWorld-Verified uses 361 tasks, OSWorld~2.0 uses 108 tasks, and
WebArena-Verified uses 812 tasks. Scores retain their reported aggregation
rules (Appendix~\ref{app:benchmarkprotocol}).}
\label{tab:rlresults}
\begin{tabularx}{\linewidth}{P{27mm}C{21mm}*{4}{>{\centering\arraybackslash}X}}
\toprule
Model & Stage & \multicolumn{3}{c}{Desktop} & Web \\
\cmidrule(lr){3-5}\cmidrule(lr){6-6}
& & \makecell{OSWorld-\\Verified} & \multicolumn{2}{c}{OSWorld~2.0} & \makecell{WebArena-\\Verified}\\
\cmidrule(lr){4-5}
& & Success & Binary & Partial & Success\\
\midrule
\rowcolor{NexLight}\multicolumn{6}{l}{\textbf{Our NorthCUA}}\\
\multirow{2}{*}{\name-Nano} & Before RL & 76.1 & 12.0 & 40.9 & 64.2\\
& After RL & 78.9 & 13.0 & 45.6 & 67.2\\
\addlinespace[4pt]
\multirow{2}{*}{\name-Mini} & Before RL & 83.3 & 18.5 & 46.3 & 68.0\\
& After RL & 85.3 & 18.5 & 50.1 & 70.4\\
\addlinespace[4pt]
\multirow{2}{*}{\name-Pro} & Before RL & 85.9 & 27.3 & 63.6 & 69.6\\
& After RL & 88.6 & 30.6 & 66.8 & 73.9\\
\bottomrule
\end{tabularx}
\end{table}

\Needspace{7\baselineskip}
\paragraph{Delivering long visual experience efficiently.}
The learning interface preserves the screenshots seen at each decision.
As a result, long attempts can contribute repeated image windows across
training segments. We use a separate archived 9B desktop run for the
engineering profile and trajectory example below. To evaluate the lossless
delivery path in Section~\ref{sec:rlupdates}, we examine 28 batches.
Pairing the raw and encoded counters for each batch holds the
underlying image tensors fixed while measuring their representation size.

\begin{figure}[t]
\centering
\includegraphics[width=\linewidth]{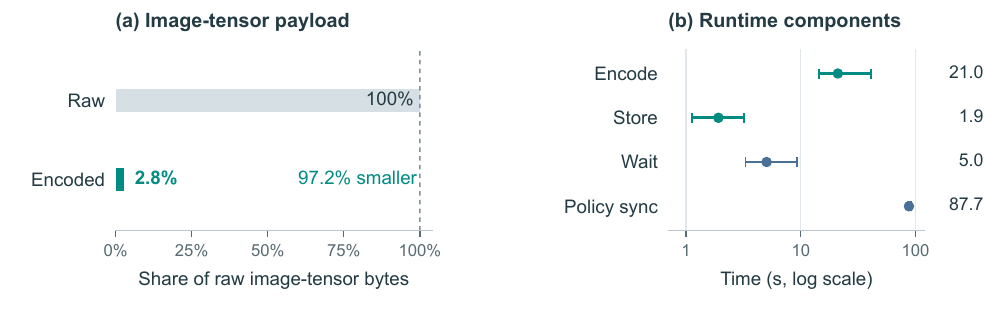}
\caption{\textbf{Desktop visual-experience delivery profiling.}
The profile uses a separate desktop workload.
(a) Encoded image-tensor bytes as a percentage of raw bytes, calculated
from the summed paired counters across 28 batches. The dashed reference
is the original representation size. (b) Median and interquartile range
of component timings for the same update range. Policy synchronization
includes the in-flight wait, so the displayed timings are not additive.
Payload sizes exclude non-image fields and transport overhead.}
\label{fig:rl-engineering}
\end{figure}

\begin{samepage}
Across this workload, encoding reduces the summed image-tensor payload
by 97.2\%, leaving 2.8\% of the original representation size. The
median raw and encoded batch sizes are 110.9 and 2.7 GiB, respectively.
Encoding preserves all screenshots and their values.
The counters cover eligible image tensors, including repeated
segment occurrences, and exclude non-image fields and transport overhead.
\par\end{samepage}

\begin{figure}[t]
\centering
\includegraphics[width=\linewidth]{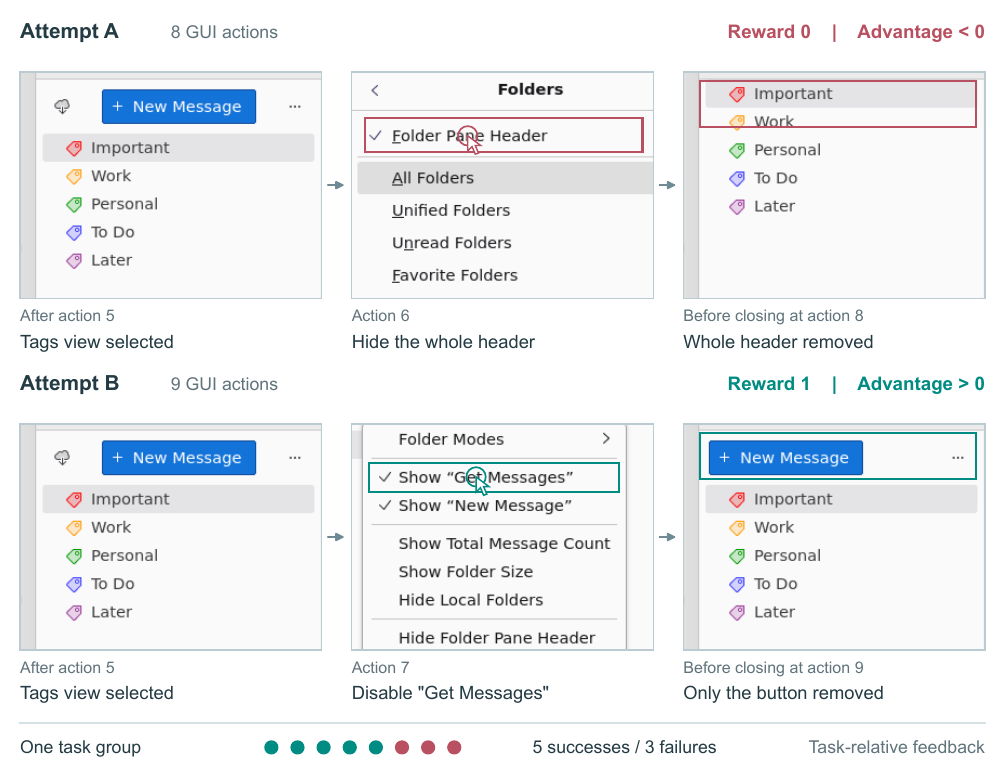}
\caption{\textbf{Executed trajectories with contrasting task rewards.}
Both Thunderbird attempts reach Tags view. A hides the entire header,
while B disables Get Messages specifically. Native screenshot crops show
executed click positions in the middle panels and the state before
application closure on the right. A's first two panels crop one observation.
B opens the header menu between those panels. The archived configuration verifier
returns rewards 0 and 1. Both attempts belong to one eight-member GRPO
group containing five successes and three failures, which determines
their advantage signs. They span an asynchronous policy update.
This is a within-group reward comparison, not a before/after checkpoint
evaluation.}
\label{fig:rl-reward-case}
\end{figure}

\begin{samepage}
Figure~\ref{fig:rl-engineering}(b) reports delivery and policy-synchronization
timings. Image encoding takes a median of 21.0 seconds, while
object-store put calls take 1.9 seconds. The recorded policy-sync lifecycle
takes 87.7 seconds and includes the 5.0-second in-flight call wait,
weight export, rollout-engine reload, and receipt work. The in-flight wait
is nested within the synchronization lifecycle, so these component timings
are non-additive. They characterize delivery and synchronization costs,
while the payload comparison measures the reduction in representation size
for the same visual experience.
\par\end{samepage}

\paragraph{From interface choices to learning feedback.}
\begin{samepage}
Figure~\ref{fig:rl-reward-case} examines what the terminal reward distinguishes
in an actual task. The user asks the agent to switch Thunderbird to Tags
view, hide the Get Messages button, and close the application to save
the settings. This requires selecting the intended control and preserving
its configuration after the application exits.
\par\end{samepage}

\begin{samepage}
The two attempts follow the same first five actions and reach Tags view.
Their next choices differ in scope. Attempt A disables the entire Folder
Pane Header, removing all its controls from view. Attempt B opens the
header's own menu and unchecks Show Get Messages, leaving the remaining
header controls visible. Both then close Thunderbird and report completion.
The task verifier checks the saved Tags mode and the specific button's
hidden setting, returning reward 0 for A and reward 1 for B. Thus,
the reward distinguishes the saved button configuration from hiding its
entire parent header.
\par\end{samepage}

\Needspace{8\baselineskip}
The complete group contains five successful and three unsuccessful
attempts. Under the group-relative rule in Eq.~\ref{eq:grpo}, the successful
attempts receive positive advantages and the unsuccessful attempts receive
negative advantages, although their task reward is zero. These advantages
carry the verifier's distinction between saved configurations into the
policy update. The recorded attempts also cross a
policy-publication boundary. The learning interface in
Section~\ref{sec:rlcontext} preserves each segment's input and behavior
version while keeping its attempt reward and group identity.
Checkpoint-level improvement is evaluated separately in Table~\ref{tab:rlresults}.

\FloatBarrier
\Needspace{9\baselineskip}
\subsection{An Evaluated Browser Recovery Case}
\label{sec:evalcases}

\begin{figure}[t]
\centering
\includegraphics[width=\linewidth]{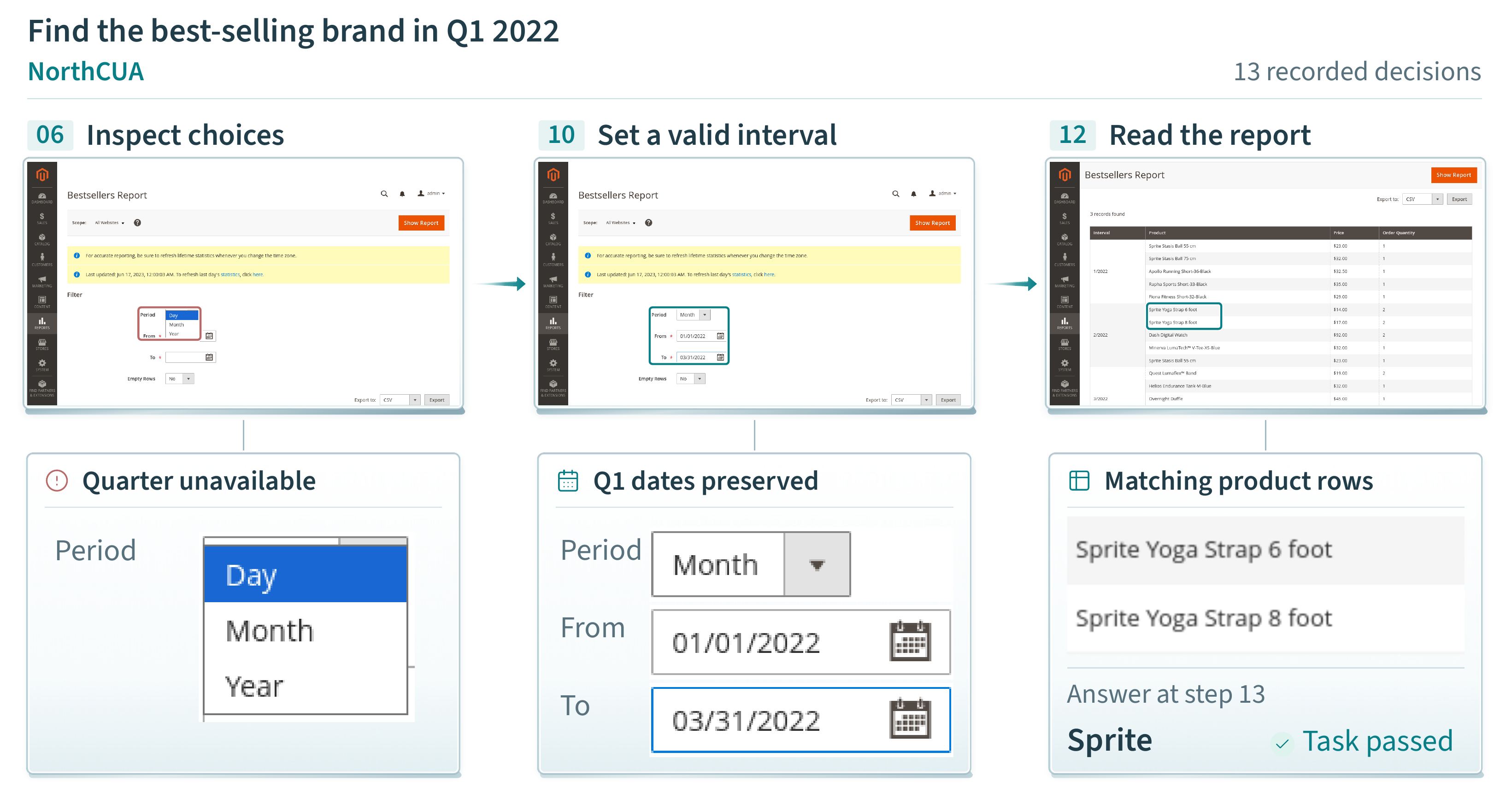}
\caption{\textbf{Recovering from an unavailable browser action.}
In a retained Pro evaluation attempt, NorthCUA first attempts to select
an unavailable Quarter period. It inspects the dropdown, switches to Month
with the exact Q1 date range, and reads the resulting table before answering
Sprite. The saved evaluator marks the task successful. Numbered panels
show observations before decisions 6, 10, and 12 of a 13-decision attempt,
with intermediate actions omitted. Detail views enlarge original control
crops, with field labels arranged for readability. The answer at decision~13
and the external evaluator's success marker are annotations outside the webpage.}
\label{fig:eval-web}
\end{figure}

\begin{samepage}
Aggregate scores summarize completion, but a recorded interaction can
show how the policy handles a mismatch between its plan and the interface.
In the Pro episode in Figure~\ref{fig:eval-web}, NorthCUA is asked to identify the best-selling
brand in the first quarter of 2022. Reaching the answer requires navigating
to the appropriate report, applying the correct time interval, and reading
the returned product rows.
\par\end{samepage}

\begin{samepage}
The first attempted period selection requests
an unavailable Quarter option and times out. To identify the mismatch,
the agent opens the dropdown. At decision~6, it observes that only Day,
Month, and Year are available. It then uses Month for aggregation and
enters January~1 and March~31 as the date bounds, preserving the requested
quarter through the two date fields. By decision~10, the visible form
contains the intended interval and is ready to run.
\par\end{samepage}

After running the report, the agent scrolls through its rows. The state
at decision~12 shows two Sprite products among the highest-quantity
entries. The agent checks the remaining rows before returning Sprite, and the
saved task evaluator records success. The episode contains 13 model
decisions, including the terminal response. Using the date fields with the
available Month aggregation mode lets the agent correct its execution plan
while preserving the requested quarter.

\Needspace{11\baselineskip}
\subsection{NorthCUA-Ultrafast}
\label{sec:fastweb}

Long computer-use tasks accumulate the latency of many successive model
decisions. Some decisions require open-ended content, but others reduce to
choosing an operation and a control already present on the page. Predicting
these finite choices directly could remove the decoding cost of routine
actions. Jev provides a relevant interface for this purpose by answering
structured choice questions without generating their answers token by token
\citep{almeida2026jev}. We adapt this idea to \name-Nano through
\name-Ultrafast, which combines a choice head for finite decisions with the
original model's generation path for open-ended outputs.

\begin{figure}[!t]
\centering
\includegraphics[width=\linewidth]{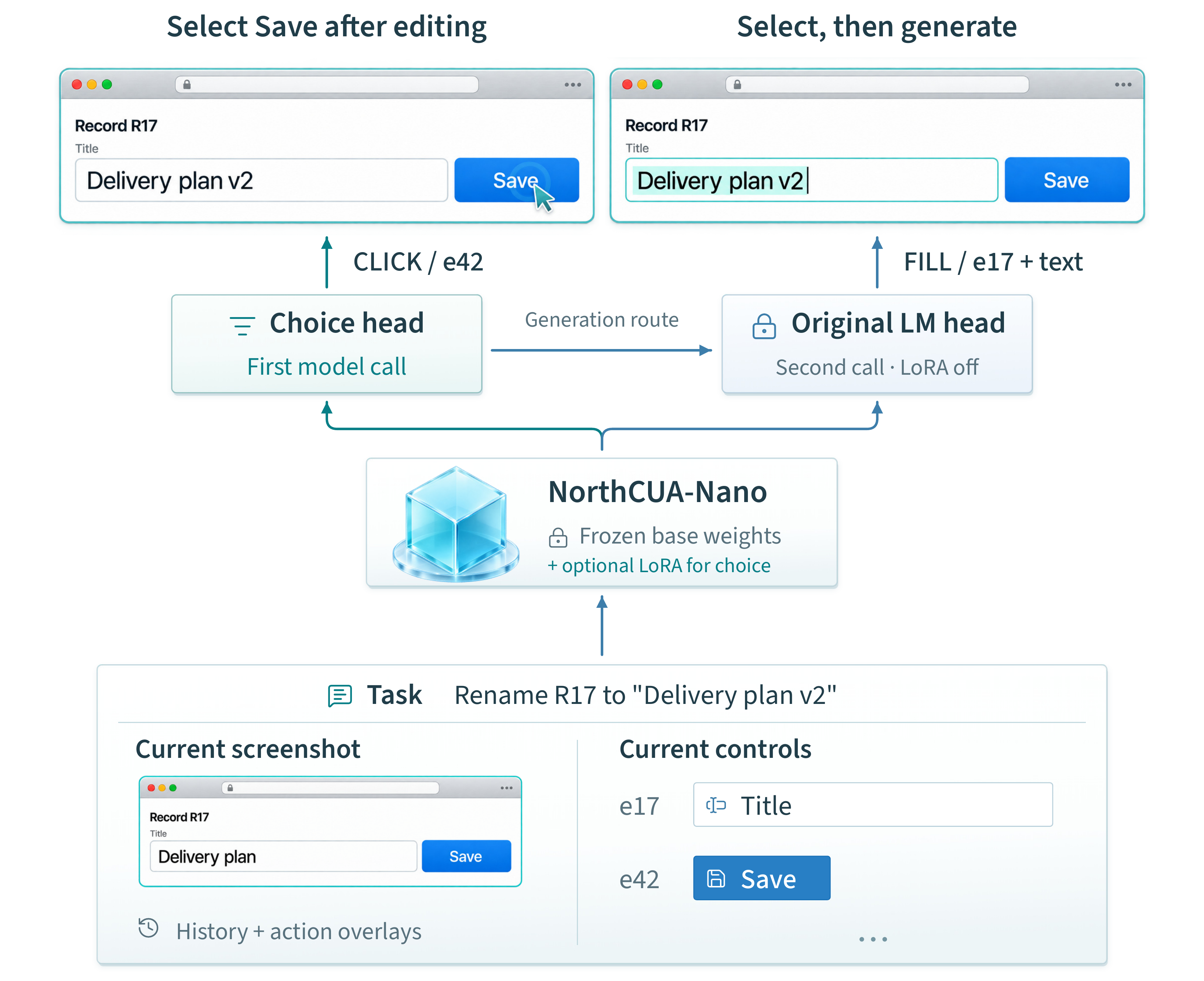}
\caption{\textbf{Choice-based Web execution.}
The input shows the title before editing. Generated GUI views illustrate
filling the new title (right) and a subsequent Save click (left).
At each decision, current controls support operation and target selection.
A direct choice executes without decoding, while a generation route triggers
a second full-model call with the original prompt and a fresh cache.
Optional LoRA adapters are active only for classification.
Route and target checks are omitted.}
\label{fig:webchoice}
\end{figure}

Figure~\ref{fig:webchoice} illustrates the resulting division of work.
The classifier receives the task and recorded interaction history, including
page observations, screenshots, and action overlays. Candidate controls are
bound to the latest valid page observation for execution. One backbone forward pass supplies the
features used to score the operation and its applicable targets. A selected
click, hover, or checkbox change can then be converted directly into a tool
call.

\begin{samepage}
Filling a field additionally requires text, so it triggers a new call
using the original model prompt and generation head. Other tool actions and
final responses use the same generation path. The executor validates the
returned tool and arguments, and uncertain finite choices can fall back to
native generation. This preserves the original output interface while
allowing routine finite actions to bypass decoding.
Appendix~\ref{app:ultrafastonline} specifies the routing and validation profiles.
\par\end{samepage}

\begin{samepage}
We study how much adaptation this choice interface requires, starting from
the long-horizon-SFT \name-Nano checkpoint before joint RL. Following
Valen's head-training approach \citep{liu2026valen}, \emph{Head-only} freezes
all original weights and trains a 4.21M-parameter choice head, about 0.045\%
of the complete multimodal checkpoint. \emph{Head + LoRA} also trains
low-rank updates in the language layers \citep{hu2022lora}, bringing the
trainable total to 47.49M parameters. These adapters are enabled for
classification and disabled for native generation, which keeps the original
generator common to both variants. Both use about 226,000 training decision
states, with fixed validation and test sets of 5,203 and 6,087 states.
Supervision covers the recorded operation and its required target.
Appendix~\ref{app:ultrafast} gives the readout, training objective, and recipe.
\par\end{samepage}

\begin{samepage}
We first evaluate whether the choice interface retains the recorded
decisions. Head-only and Head + LoRA reach validation route-and-target
accuracies of 86.1\% and 88.2\%, respectively. On the held-out test set,
Head-only reaches 85.4\% route-and-target accuracy against Full generation's
85.0\%, while their full-action exact-match scores are 64.0\% and 64.6\%.
Generated parameters account for much of the remaining full-action error.
The LoRA result is a validation comparison.
Appendix~\ref{app:ultrafast} gives the separate metrics and checkpoint-selection details.
\par\end{samepage}

\begin{samepage}
To assess complete task execution, we compare the three policies on all
812 WebArena-Verified tasks, deploying each inference replica on a single
GPU. Table~\ref{tab:webchoice} reports success and measured execution cost
under the definitions in Appendix~\ref{app:ultrafastonline}. Success uses
the full benchmark, while timing uses a fixed 320-task
subset selected from tasks jointly successful in the timing snapshot. Decoding bypass measures
the fraction of dispatched actions produced directly by the choice head.
\par\end{samepage}

\begin{table}[!t]
\centering\small
\setlength{\tabcolsep}{3pt}
\caption{Online quality and execution cost of NorthCUA-Ultrafast.
Success uses all 812 tasks. Mean task time uses the fixed 320-task
success-selected subset. Decoding bypass uses dispatched actions. Appendix~\ref{app:ultrafastonline}
defines the populations and deployment settings.}
\label{tab:webchoice}
\begin{tabularx}{\linewidth}{X C{24mm} C{30mm} C{28mm}}
\toprule
Method & Task & Mean task & Decoding \\
 & success & time & bypass \\
 & (\%) & (s) & (\%) \\
\midrule
Full generation & \textbf{64.2} & 616.8 & 0.0 \\
\rowcolor{NexLight}
Ultrafast, Head-only & 59.2 & 288.8 & \textbf{40.2} \\
Ultrafast, Head + LoRA & 60.6 & \textbf{266.8} & 31.1 \\
\bottomrule
\end{tabularx}
\end{table}

Head-only completes 481 tasks using only the choice head for adaptation.
Head + LoRA completes 492 (60.6\%), gaining 11 tasks or 1.4 percentage
points. With about 0.5\% trainable parameters, LoRA retains 94.4\% of
Full generation's 64.2\% success rate, leaving a gap of 3.6 percentage
points.

\begin{samepage}
The direct-action path is used frequently during execution. Head-only and
Head + LoRA bypass generation for 40.2\% and 31.1\% of recorded actions,
respectively. We next examine the task-level cost of these execution
policies on the fixed success-selected subset.
\par\end{samepage}

Mean task time falls from 616.8\,s to 288.8\,s for Head-only and
266.8\,s for Head + LoRA. On the fixed timing subset, this is a
factor-of-$2.3$ reduction in mean task time for Head + LoRA.
Its shorter task time despite a lower
bypass rate also shows why direct-action frequency alone is insufficient
to characterize execution efficiency. Each replica uses one GPU.
Appendix~\ref{app:ultrafastonline} gives the decoding and serving settings
for these measured execution costs.
The following case
illustrates how fast selection and flexible generation cooperate within
one successful task.

\begin{figure}[!t]
\centering
\includegraphics[width=\linewidth]{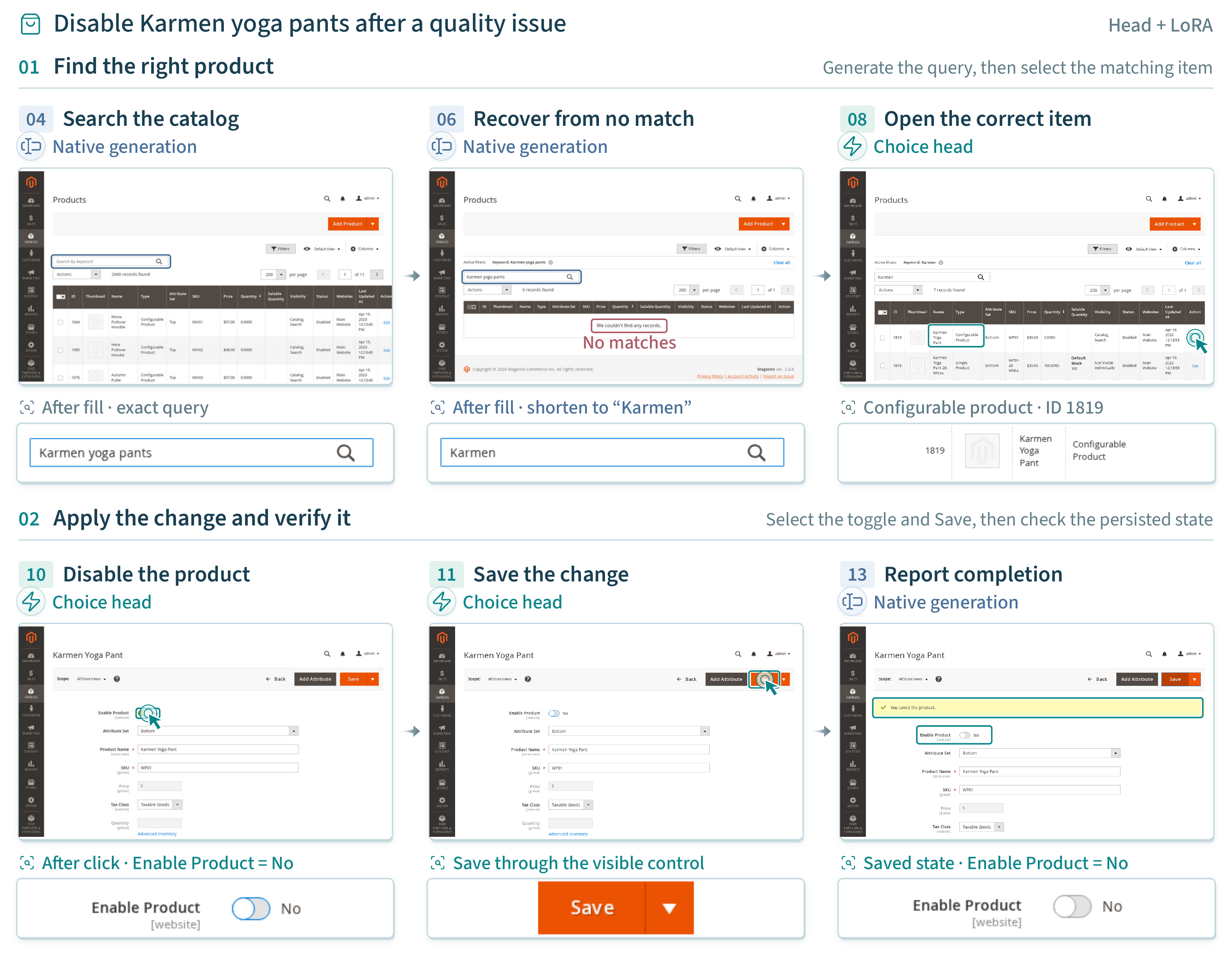}
\caption{\textbf{Ultrafast product-update case.}
Selected Head + LoRA decisions from one WebArena-Verified attempt.
Lightning and text-entry icons distinguish direct choice from native
generation. Numbers retain the original decision indices, with intermediate
steps omitted. Insets enlarge recorded details, with post-action states labeled.}
\label{fig:ultrafast-case}
\end{figure}

\paragraph{From search recovery to a saved product update.}
Figure~\ref{fig:ultrafast-case} follows a Head + LoRA attempt at a
WebArena-Verified task that asks the agent to disable Karmen yoga pants
because of quality issues. The request requires finding the correct catalog
entry and saving its disabled state. Searching for the full phrase returns
no records. The native generator responds to this observation by shortening
the query to \emph{Karmen}, after which the search exposes the configurable
\emph{Karmen Yoga Pant} and its individual variants.

The choice path opens the configurable product, switches
\emph{Enable Product} off, and selects \emph{Save}, all without decoding
action text. The final page shows the saved disabled state. The independent
evaluator verifies the corresponding product-1819 save request and awards
full credit.

Across the complete attempt, six of twelve executed browser actions bypass
generation. The other actions handle navigation, query text, and waiting,
and a thirteenth decision generates the final response. This execution
illustrates the intended division of work. Generation remains available when the search needs
revision, while discrete selections carry the recovered task through to a
verified update. The aggregate bypass rates in Table~\ref{tab:webchoice}
measure how often this direct execution path is used across the benchmark.

\section{Related Work}
\label{sec:related}

NorthCUA connects computer-use learning with the tasks and execution systems
that supply its experience. We review these foundations in three parts.
We first discuss policies and learning methods, then the construction of
tasks and experience, and finally the harnesses and infrastructure that
support their collection and use in training.

\paragraph{Computer-use policies and learning.}
Accurate GUI actions depend on linking instructions to screen elements.
SeeClick studies grounding pre-training \citep{cheng2024seeclick}.
GUI-Actor develops attention-based pointing, while GUI-Cursor refines cursor
movements through visual feedback \citep{wu2025guiactor,zhao2025learning}.
End-to-end policies build on this capability to coordinate actions over time.
UI-TARS integrates perception, reasoning, and action prediction, and OpenCUA
develops reflective supervision from human demonstrations
\citep{qin2025uitars,wang2025opencua}.
Learning from outcomes offers a further source of improvement. ComputerRL
scales live-desktop RL \citep{lai2025computerrl}, while VEM learns offline
value estimates to guide environment-free policy exploration
\citep{zheng2025vem}. UI-TARS-2 and Qwen-CUA connect staged training with
interactive learning \citep{wang2025uitars2,lu2026qwencua}.
For long trajectories, learning methods also specify how history and outcome
feedback are carried across training segments.
ScaleCUA masks reused history and shares rollout rewards and advantages
across segments \citep{lv2026scalecua}. Qwen-CUA preserves episode credit
across slices and normalizes each trajectory's objective over active policy
tokens \citep{lu2026qwencua}.
NorthCUA combines broad GUI mid-training, verified long-horizon SFT, and joint
desktop--web RL. Our emphasis is on integrating decision-time input
preservation and complete-attempt weighting with the Harness and
experience-production pipeline.

\paragraph{Scalable tasks and experience.}
Scaling workflow learning creates a corresponding demand for executable tasks
and reliable outcomes.
OS-Genesis derives tasks from exploratory interaction
\citep{sun2024osgenesis}, while GUI-360 combines sourced requests, environment
templates, execution, and filtering for Windows data collection
\citep{mu2025gui}. ProCUA-SFT grounds tasks in real materials and checks
preconditions before rollout \citep{jung2026procuasft}.
For outcome-driven learning, CUA-Gym and Qwen-UI-Agent synthesize compatible
tasks, initial states, and verifiers \citep{wang2026cuagym,zhou2026qwenuiagent}.
WEFT extends this whole-system view to tool-use post-training, using
execution feedback to revise interacting components and fresh rollouts
to evaluate each revision \citep{mao2026weft}.
NorthCUA organizes construction around available software and content, sourced
requests, or verified seed tasks. Each route produces an executable
\mbox{\casespec}. A separate grounding route uses application documentation
to identify control functions and construct screenshot--query--location
supervision.

Hard tasks also motivate retaining procedural knowledge across executions.
Work2Skill turns synchronized human GUI demonstrations into reusable skills
through on-demand inspection of action, visual, and available accessibility
evidence \citep{huang2026work2skill}. WebXSkill extracts executable
procedures with language guidance from trajectories
\citep{wang2026webxskill}. Search complements procedural guidance by considering
alternative decisions. Computer-Using World Model simulates UI transitions for action
selection \citep{guan2026computer}, and Everything of Thoughts combines
learned guidance with tree search for reasoning \citep{ding2024everything}.
Our RSI collection retains validated textual skills and real
workspace checkpoints, so later attempts can reuse procedures or resume
partial progress before full-task verification.

\paragraph{Agent harnesses and learning infrastructure.}
Such collection and learning require an execution layer that preserves both
software state and training evidence. UFO introduces hierarchical Windows
interaction \citep{zhang2025ufo}. UFO2 adds application-specialized agents and
GUI--API integration \citep{zhang2025ufo2}, and UFO3 extends orchestration to
heterogeneous devices through a task DAG \citep{zhang2025ufo3}.
For reuse across training systems, Agent Lightning v1.0 separates harness
execution from RL optimization \citep{he2026agentlightningv1}.
Orchard provides shared sandbox lifecycle services \citep{peng2026orchard},
while Polar reconstructs token-faithful training trajectories from recorded
model calls \citep{xu2026polar}.
NorthCUA focuses on integrating these system responsibilities for stateful
visual interaction. Its shared desktop/web runtime, live outcome inspection,
and parallel lifecycle service support experience collection. The Experience
Adapter then preserves the recorded decisions when converting online
attempts into learning samples, connecting execution and assessment to
policy training through the same task and outcome interfaces.

\section{Limitations and Responsible Use}
\label{sec:limitations}

\paragraph{Generalization and coverage.}
NorthCUA operates across diverse applications, while unfamiliar software
and changing interface conventions remain challenging. Documentation-guided
grounding expands the functional knowledge available during data construction,
and skill- and milestone-guided collection revisit workflows that ordinary
attempts leave unfinished. Extending this coverage to new software and
maintaining reliable execution as applications evolve remain important
directions for the experience pipeline.

\paragraph{Feedback and optimization.}
Executable verifiers and live-workspace inspection tie learning feedback
to the work actually produced. Feedback quality depends on predicate
coverage and the calibration of model judgments for open-ended content
\citep{zheng2023judge}. Controlled verifier probes and evidence-linked
judgments help check both. Long workflows
also concentrate many decisions into a terminal outcome. More informative
intermediate feedback could improve credit assignment and help agents
reach correct results with fewer unnecessary actions.

\paragraph{Responsible deployment.}
Isolated workspaces, explicit tool profiles, and retained execution records
provide controls for deployment. Operating real software also requires
authorized access, least-privilege credentials, and confirmation for
consequential actions. External pages and documents can carry indirect
prompt injections \citep{debenedetti2024agentdojo}, motivating adversarial
testing appropriate to each application. Released materials should undergo
licensing, privacy, and credential review alongside the operational checks.

\section{Conclusion}

We presented \name, an integrated system for developing computer-use agents
through experience at scale. By connecting the \northcua Harness,
\experienceflywheel, and Staged Learning through shared tasks and outcome
checks, the system supports learning from both verified demonstrations and
fresh policy interaction. Improved policies can return to the same
infrastructure to collect further experience. Recursive self-improvement
extends collection to difficult workflows by preserving useful procedures
and verified milestones, allowing later attempts to build on earlier
progress.

Across six benchmarks, the Nano, Mini, and Pro family achieves
state-of-the-art results among open-source models in GUI grounding,
desktop execution, and Web interaction. At 397B parameters, Pro outperforms
the trillion-parameter general-purpose baselines on both grounding
benchmarks, OSWorld-Verified, and WebArena-Verified. Stage comparisons
further show improvements in desktop and Web execution across all three
model sizes following long-horizon SFT and joint RL. Alongside these gains
in task completion, \name-Ultrafast explores a complementary route to
faster decision-making.

We release the three models, the full Harness code, and this technical
report so that the work can be extended at both the policy and system
levels. Together, they provide an open foundation for adapting agents to
new software and developing better ways to collect and learn from
experience. The broader goal is to make reliable software execution a
general AI capability, allowing users to delegate complete workflows from
intent to verifiable outcome.

\par\addvspace{17pt}
\Needspace{15\baselineskip}
\begin{tcolorbox}[enhanced,frame hidden,boxrule=0pt,
  colback=NexLight,coltext=NexInk,arc=1mm,outer arc=1mm,
  boxsep=0pt,left=12pt,right=12pt,top=11pt,bottom=11pt,
  before skip=0pt,after skip=5pt]
\phantomsection
\pdfbookmark[1]{NorthCUA Team}{northcua-team}
\label{sec:northcuateam}
{\sffamily\bfseries\fontsize{11.5}{15}\selectfont NorthCUA Team\par}
\vspace{7pt}
\newcommand{\teammark}[1]{\textsuperscript{\color{TiffanyDeep}#1}}
{\rmfamily\mdseries\fontsize{10.5}{15}\selectfont\raggedright
\mbox{Chaoyun Zhang\teammark{\ensuremath{\bigstar\mkern2mu\blacklozenge}}},
\mbox{Bingxuan Li\teammark{\ensuremath{\bigstar\mkern2mu\blacklozenge}}},
\mbox{Pinxue Guo\teammark{\ensuremath{\bigstar\mkern2mu\blacklozenge}}},
\mbox{Yangfan Gao\teammark{\ensuremath{\blacklozenge}}},
\mbox{Zikun Cai\teammark{\ensuremath{\blacklozenge}}},
\mbox{Tianci Xie\teammark{\ensuremath{\blacklozenge}}},
\mbox{Hui Li\teammark{\ensuremath{\blacklozenge}}},
\mbox{Yuqi Huang\teammark{\ensuremath{\blacklozenge}}},
\mbox{Yifei Cao\teammark{\ensuremath{\blacklozenge}}},
\mbox{Bo Mao}, \mbox{Tian Tang}, \mbox{Qichang Zheng},
\mbox{Tongtong Liu}, \mbox{Xiaoxing Wang}, \mbox{Jinxiu Liu},
\mbox{Lizhi Lin}, \mbox{Zhou Shao}, \mbox{Maosen Zhou},
\mbox{Miao Zheng}, \mbox{Shuo Zhang}, \mbox{Yangji He},
\mbox{Kepeng Lei}, \mbox{Ning Zhao}, \mbox{Wei Shen},
\mbox{Zhenjiang Jin}, \mbox{Xinyue Zhang}, \mbox{Peng Sun},
\mbox{Zhenhua Han}, \mbox{Rui Zheng},
\mbox{Hang Yan}\par}
\vspace{9pt}
{\rmfamily\mdseries\fontsize{9.2}{12}\selectfont\color{NexSubtitle}
\teammark{\ensuremath{\bigstar}}\,Equal contribution\hspace{1.8em}
\teammark{\ensuremath{\blacklozenge}}\,Core contributors\par}
\end{tcolorbox}

\begingroup
\setlength{\bibsep}{1.4pt}
\bibliography{references}
\bibliographystyle{colm2026_conference}
\endgroup
\clearpage\appendix
\renewcommand{\sectionfloatboundary}{}
\section{Implementation and Reproducibility}
\label{app:implementation}

The executable task, its workspace, and the resulting learning samples share
an attempt identity. \casespec binds the request to its materials, initial
state, and execution settings. The environment service manages the
workspace, the \northcua{} Harness records interaction, and the configured
verifier or evaluator establishes the outcome. The Experience Adapter
retains the policy's decision-time inputs when constructing samples, and
the learner associates those samples with their attempt-level feedback.

\subsection{Information and Permission Boundaries}

Construction, execution, evaluation, and learning use different information.
The generator can inspect construction metadata, while the acting policy
receives the request and permitted task inputs. Host-only acceptance
criteria remain with evaluation. Tool capability checks control permitted
operations, while input construction and sample extraction separately control
which information each role receives.

Actor and evaluator records remain separate, preserving the actor's
submitted result and decision-time inputs during inspection. Actor imitation
excludes privileged evaluation calls. Evaluator demonstrations used in
mid-training are separate judgment tasks with their own inspection context
(Appendix~\ref{app:supervisedrecipe}). During ordinary execution, the actor
receives the task request and permitted inputs, while construction dependency
graphs remain construction metadata. Runtime profiles,
live-inspection rules, and resource ownership are detailed in
Appendices~\ref{app:runtimeconfig}, \ref{app:evaluatorprotocol},
and~\ref{app:batchprotocol}.

\section{Grounding Data Details}
\label{app:grounding}

Section~\ref{sec:grounding} describes visual, structural, and
documentation-grounded labeling. The common annotation unit links an
unannotated screenshot, a visible control, its target geometry, and one or
more queries. Documentation supplies functional evidence, while the GUI
image supplies the target location.

\begin{table}[htbp]\centering\small
\caption{Annotation records for tracing a grounding label from source
evidence to the original screenshot.}
\label{tab:groundingprocedure}
\begin{tabularx}{\linewidth}{P{34mm}X}
\toprule
Record & Contents and alignment rule \\
\midrule
GUI target & Original image dimensions, target point or region, and any
crop/resize transform. Predictions and labels refer to the same coordinate
system. \\
Documentation match & Application and available version evidence, manual
source, candidate illustrations and function descriptions, and the
context supporting the selected match. \\
Query variants & Function-, name-, appearance-, or position-based queries
linked to the same visible control. Variants remain grouped with their
source target. \\
Candidate disposition & Accepted, rejected, or unresolved matches, keeping
semantic ambiguity separate from geometric invalidity. \\
\bottomrule
\end{tabularx}

\end{table}

Crop and resize transforms preserve correspondence with the original
screen. A point target lies inside the actionable region, without an
occluding foreground element. A menu-item target requires the item to be
visible in the captured menu state. Queries requiring unprovided hidden
state are excluded. Diagnostic boxes and action annotations are kept
separate from the unannotated grounding input.

Documentation matching checks the control's appearance, nearby controls,
visible application state, and the manual's applicability. Ambiguous or
version-incompatible matches remain outside the accepted set. Label checks
distinguish a wrong function from a spatial miss, even when both refer to
similar-looking controls. We count source images, target controls, and
query variants separately, grouping paraphrases with their original target.
Source grouping and split rules are shared with Appendix~\ref{app:corpus}.

\section{Task Construction Details}
\label{app:generation}

The construction routes in Section~\ref{sec:data} bind a request, usable
materials, an initial workspace, and acceptance conditions into an
executable task. A material record combines file format, bounded content
summaries, and provenance. After selection, the workspace builder copies
the inputs and records their content hashes and initialization procedure.
The task distinguishes read-only inputs, distractors, mutable targets, and
required outputs.

For dependent work, the plan connects producers to the artifacts or
services consumed by later stages. Structural checks reject unknown
predecessors, cycles, disconnected stages, incompatible references, and
required outputs without production stages. Host-only evaluation artifacts
remain with the evaluator.
These constraints define valid dependencies
while leaving the actor free to choose GUI actions within the task's
prerequisites. The following details cover environment preparation,
assisted collection, and verifier test states.

\subsection{Environment Construction and Binding}
\label{app:data-method-status}

The request-conditioned desktop builder first filters candidate materials
by format and application requirements, then ranks them by content.
Request refinement uses the selected inputs to make references concrete
while preserving the source goal and constraints. A mismatch between the
request and available materials sends the task back to construction. Content summaries support
selection without replacing inspection of the files being packaged.

EnvAgent finalizes the workspace against a current report of file evidence
and query alignment. Material changes require renewed inspection before
finalization. The compiler retains packaged paths, checksums, environment
configuration, and startup actions, with ownership and permissions that
make the initialized files accessible to the desktop user. The specified
live-validation procedure additionally checks GUI opening and required
operations in an isolated attempt. Collection begins from a fresh
initialization, preserving the specified starting state for the actor.

Dependency checks distinguish data flow from required workflow order.
For example, the office plan in Figure~\ref{fig:longhorizonconstruction}
uses content from A and B in its final PDF, while the record update in C
has an independent acceptance condition. The C-to-D edge imposes order
without making the record a source of PDF content. A designated terminal
artifact therefore need not be the task's only required result.

Reference-based web construction specifies both the visible interface and
the state supporting its interactions. Its environment description records
the reference source, supported pages and operations, observed versus
generated content, backend schema and seed state, reset procedure, and
external dependencies. Rendering, navigation, persistence, initialization,
and task feasibility are checked separately against this description.
It defines the reconstructed pages and behaviors used for task collection.
Appendix~\ref{app:corpus} summarizes the resulting data sources and scale.

\subsection{Hard-Task Collection Protocol}
\label{app:hardexperience}

The two collection routes in Section~\ref{sec:hardexperience} retain
different products between attempts: validated operating guidance and
recoverable execution progress. Selection starts in the training partition
after ordinary attempts expose a useful obstacle. Invalid environments,
ambiguous requirements, and unreliable verifiers are repaired or excluded
before further exploration. The task's final acceptance conditions remain
fixed as candidate skills and intermediate milestones evolve.

A skill record identifies its task family, applicability, guidance,
version, source attempts, validation cases, and known failures. Validation
uses a frozen candidate, separate from the attempts that produced it.
Task-specific assistance, reuse within a family, and transfer to a new
application remain distinct uses of the record.

For milestone search, a planning agent may inspect the request and verifier
to define dependent subtask--check pairs. Actor-visible guidance uses only
permitted task information. Hidden predicate constants and expected
answers stay in evaluation context. A checkpoint binds the environment
snapshot to its parent attempt and action prefix, prompt-visible history,
summary, notes, skill and milestone guidance, model/tool configuration,
and remaining budget. Restoration re-establishes device bindings and obtains
fresh observations. External service state needs its own isolation and
restoration mechanism.

A candidate actor state is retained before state-changing inspection.
Such inspection uses an evaluation copy or separate branch and receives
its own action record. Actor continuations resume from the retained
pre-inspection state. The checkpoint record retains enough ancestry to reconstruct the
selected path and the information available at each decision.

Search configuration specifies candidate counts, state selection,
diversity and retention limits, uncertainty handling, and stopping rules.
Its budget includes failed attempts, skill validation, snapshot/branch
operations, inspection, and live-resource costs. A selected recoverable
failure retains its unresolved milestone label until its continuation
repairs that requirement. Local checks guide expansion, while the original task
verifier determines full-task acceptance.

Training extraction follows the selected actor path. Search-controller
decisions, snapshot operations, and failed sibling branches are excluded
from actor targets. Supplied skills and milestone guidance remain in the
context that conditioned those targets. Cross-branch knowledge transfers
are recorded at their point of use. Deleting this guidance would change
the decision inputs, so an unguided demonstration requires recollection.

A complete assisted path, a checkpoint-conditioned continuation, and a
verified subtask from an incomplete attempt are different sample types.
Unassisted completion is evaluated in a fresh-start run using the original
task inputs. Their counting units and
source-family separation follow
Appendix~\ref{app:corpus}.

\subsection{Controlled Verifier States}

Verifier tests hold the intended obligations fixed while varying the
submitted state. An incomplete initial state is a negative control, a
correct result is a positive control, a near miss isolates one violation,
and a valid alternative tests sensitivity to presentation. Each test retains
the initial and final state, protected inputs, acceptance conditions, and
the expected verdict. Testing score composition with stipulated predicate
values is separate from executing the predicates on actual artifacts.

Figure~\ref{fig:verification} gives a concrete regional-report example.
Expected gross sales are North 420 and South 380, totaling 800. Acceptance
also requires complete regional content, template compliance, and unchanged
protected source files. The two accepted presentations satisfy all of those
conditions, including the regional content and protected-file requirements.

\paperfigure{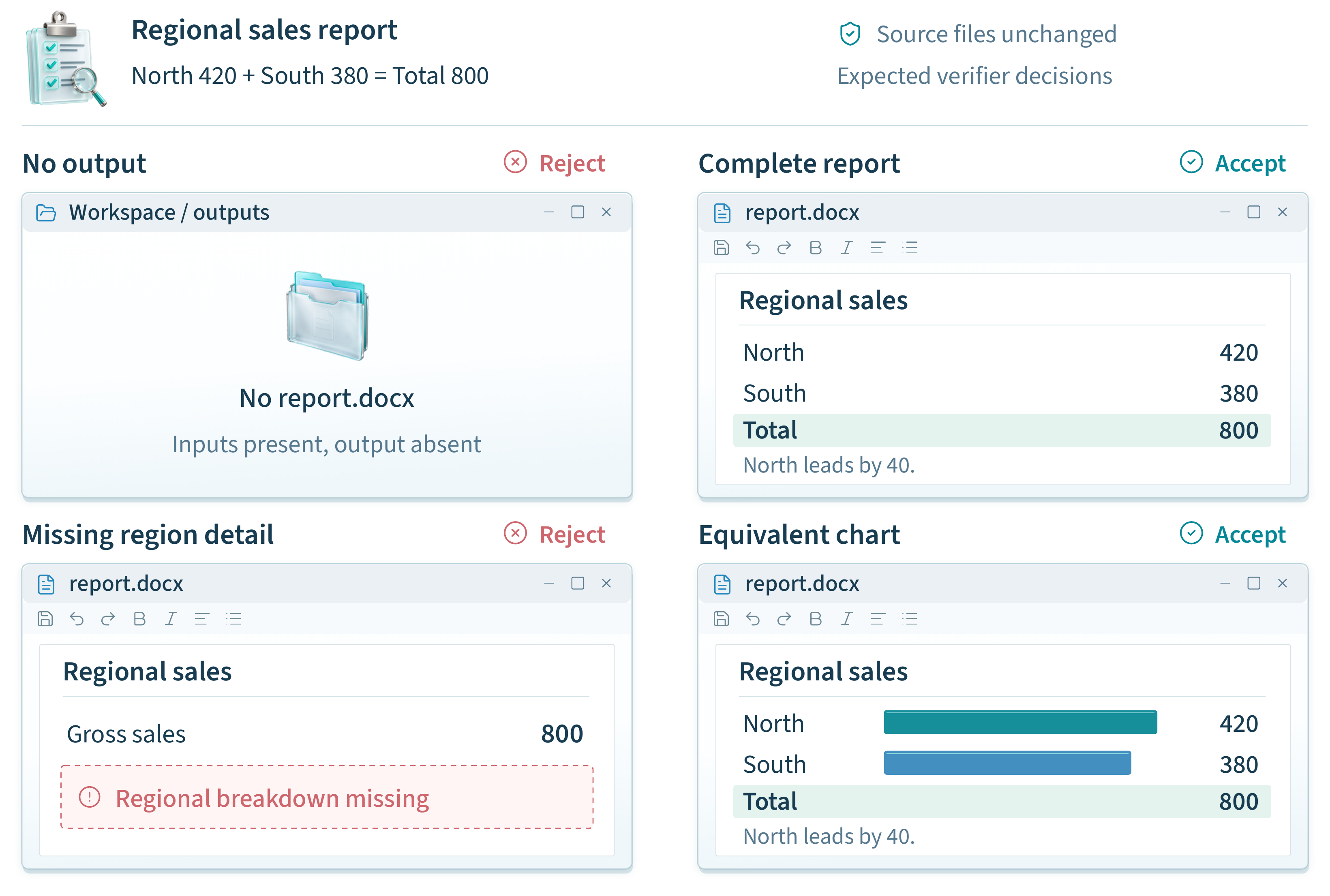}{fig:verification}{\textbf{Four states of the same report task.}
Missing output and missing regional detail should be rejected.
A complete table and an equivalent chart should be accepted.
Both accepted examples are stipulated to satisfy the full contract, including
regional discussion, template compliance, and unchanged source files.
The GUI excerpts and inspection asset are illustrative.
The dashed red region annotates missing content.
Accept/Reject labels show the expected verdicts for these illustrative states.}

\section{Training Data Sources and Scale}
\label{app:corpus}

The dataset records distinguish source materials, constructed tasks,
execution attempts, accepted trajectories, extracted samples, and training
consumption. A task can produce several attempts, and a trajectory can
produce several prediction units. Grounding images, target controls, and
query variants use their own units. Table~\ref{tab:corpus} summarizes the
supervised data by stage and domain using rounded counts.

\begin{table}[htbp]\centering\small
\caption{Approximate supervised data scale. Counts are rounded independently.
SFT domain step totals and the web mean are estimated from the rounded
query counts, total steps, and desktop mean. Grounding steps denote
coordinate targets and are separate from trajectory interactions.}
\label{tab:corpus}
\begin{tabularx}{\linewidth}{@{}>{\raggedright\arraybackslash}X
>{\centering\arraybackslash}P{29mm}
>{\centering\arraybackslash}P{32mm}
>{\centering\arraybackslash}P{31mm}@{}}
\toprule
Source & Queries & Steps & Mean steps/query \\
\midrule
\rowcolor{NexLight}\multicolumn{4}{l}{\textbf{Broad GUI mid-training}} \\
Desktop & \(\approx100{,}000\) & \(\approx4{,}200{,}000\) & \(\approx40\) \\
Web & \(\approx10{,}000\) & \(\approx280{,}000\) & \(\approx30\) \\
Grounding & \(\approx40{,}000\) & \(\approx40{,}000\) & \(1\) \\
\rowcolor{NexLight}\multicolumn{4}{l}{\textbf{Verified long-horizon SFT}} \\
Desktop & \(\approx2{,}000\) & \(\approx240{,}000\) & \(\approx120\) \\
Web & \(\approx1{,}000\) & \(\approx70{,}000\) & \(\approx70\) \\
\bottomrule
\end{tabularx}

\end{table}

Together, the desktop and web sources provide over 100,000 trajectories
and over 4 million GUI interactions for mid-training. The long-horizon
SFT cohort contains approximately 3,000 queries and 310,000 interaction steps.

These rounded counts summarize source cohorts by stage and domain before
model-specific sampling and repeated consumption.
Auxiliary evaluator demonstrations are accounted for separately from
task-execution trajectories.

A sample's lineage connects source and task family, environment and fixture
identity, collection policy, permitted tools, outcome check, and extraction
version. A consumed training sample additionally identifies its training
shard and stage. Actor episodes and evaluator judgment episodes retain
separate roles and permissions. Complete assisted paths, checkpoint
continuations, and verified subtask demonstrations also retain distinct
sample types.

Related screenshots, documents, material sets, task seeds, source sites,
repository/issue families, and derived skills are grouped before splitting.
Source-policy disclosure, licensing, and access permissions apply to each
supervised source.

\section{Model Identities and Learning Details}
\label{app:training}

The Nano, Mini, and Pro tiers use the Qwen3.5-9B, Qwen3.5-35B-A3B,
and Qwen3.5-397B-A17B backbone variants, respectively
\citep{qwen2026qwen35}. Table~\ref{tab:trainingconfig} lists the backbone
identities and their language-model parameter counts. The mixture-of-experts
variants have distinct total and
activated counts.

\begin{table}[htbp]\centering\small
\caption{Model-family identities and language-model parameter counts
following Qwen3.5 \citep{qwen2026qwen35}.}
\label{tab:trainingconfig}
\begin{tabularx}{\linewidth}{P{42mm}XXX}
\toprule
Field & NorthCUA-Nano & NorthCUA-Mini & NorthCUA-Pro \\
\midrule
Backbone variant & Qwen3.5-9B & \makecell[l]{Qwen3.5-\\35B-A3B} &
\makecell[l]{Qwen3.5-\\397B-A17B} \\
Language-model parameters & 9B, dense &
\makecell[l]{35B total\\3B activated} &
\makecell[l]{397B total\\17B activated} \\
\bottomrule
\end{tabularx}

\end{table}

The results use \emph{Mid} for broad GUI mid-training and \emph{SFT} for
verified long-horizon supervision. Mid
is already a NorthCUA checkpoint, distinct from an upstream pre-trained
release. The supervised stages differ in their source mixture and
selection criteria. We first describe the supervised prediction units,
then explain how task-group feedback and recorded RL attempts become
learning samples.

\subsection{Supervised Representation and Sampling}
\label{app:supervisedrecipe}

The supervised stages share response prediction but use different sources
and selection rules. An extraction map \(\Phi\) converts a grounding
example, task trajectory, or evaluator episode into prediction units
\(x\sim\mathcal D\). A unit can be a decision or a context-bounded slice.
For a binary target mask \(m_t\) with \(\sum_t m_t>0\), the conditional
objective is
\begin{equation}
 \mathcal L_{\rm sup}=-\mathbb E_{x\sim\mathcal D}
 \left[\frac{1}{\sum_t m_t}\sum_t m_t
 \log p_\theta(x_t\mid x_{<t})\right].
 \label{eq:sup}
\end{equation}
Here \(x_{<t}\) includes the recorded multimodal input at that decision.
The expectation is over extracted units after source sampling, so an
episode's weight depends jointly on extraction, sampling, and the
within-unit mean.

Grounding examples target coordinates. Actor trajectories target recorded
assistant responses, including reasoning, text, and tool calls according
to the target mask. Evaluator episodes are separate judgment tasks targeting
inspection and assessment responses from their own permitted context.
Grounding, actor trajectories, and evaluator examples enter mid-training.
Long-horizon SFT uses task trajectories. The three sources have separate
counts and sampling weights.

An actor input contains the request and the observations, tool results,
summaries, notes, and guidance available at that decision. Post-execution
judgments and hidden acceptance values belong to the evaluation context.
A post-action screenshot or action overlay is available only to later
decisions. Actor-issued note calls remain in the response stream with the
mask specified by the recipe. Construction-agent and privileged evaluator
calls are outside actor targets.

Sampling determines how source and episode lengths influence learning.
Uniform decision sampling, uniform trajectory sampling, and token-weighted
sampling give different weights to long episodes. The recipe therefore
distinguishes extraction, source/domain weighting, packing, target
deduplication, and loss reduction. A response repeated as context has one
target occurrence. Assisted traces retain the skill or milestone guidance
that conditioned them. These supervised choices are specified separately
from online RL assembly.

Selection precedes sampling. Broad trajectories come from judge-selected
completed tasks. Long-horizon trajectories require original full-task
script acceptance followed by execution-quality and dependency checks.
Skill- and milestone-assisted trajectories undergo the same checks.
Cleaning rejects unusable records. Retained episodes preserve
action--observation alignment, including coherent recoveries.
Source attribution and cohort overlap follow
Appendix~\ref{app:corpus}.

\subsection{Task Groups and Terminal Feedback}
\label{app:rlpool}

The initial RL pool follows Section~\ref{sec:rlselection}: human-reviewed
desktop and web tasks are screened with at least five attempts under each
model's SFT checkpoint and fixed execution settings. Selection uses
variation among valid verifier scores, with environment and verifier errors
recorded separately from task outcomes. Fresh grouped attempts on the
selected pool provide the learning data.

Members of a group attempt the same case package from equivalent initial
states under the same verifier and budget. A task-script result can be a
finite numeric reward or an interpretable Boolean/status verdict, with
Boolean fallback mapped to zero or one. Reward ranges and preprocessing
remain task-specific. Script errors, timeouts, nonzero exits, and
uninterpretable results follow the invalid-evaluation path. Valid negative
verdicts remain learning outcomes.

Retries retain logical member identity under the configured retry policy.
Complete-group checks prevent missing members from silently changing the
comparison. Domain sampling weights specify scheduled work, while completed,
admitted, and consumed members describe successive stages of actual use.
The logical attempt remains the unit of group membership, independently
of infrastructure retries or the number of extracted training segments.

For \(G>1\) valid members, terminal advantages are
\begin{equation}
 A_i=\frac{R_i-\bar R}{s_R+\epsilon},\qquad
 \bar R=\frac{1}{G}\sum_{i=1}^{G}R_i,\qquad
 s_R^2=\frac{1}{G-1}\sum_{i=1}^{G}(R_i-\bar R)^2,
 \label{eq:grpo}
\end{equation}
where \(R_i\) is the attempt's terminal reward and \(\epsilon=10^{-6}\).
The baseline is computed within the task group, independently of other
tasks or domains. Identical rewards produce zero terminal advantages.
Each attempt contributes one advantage shared by its segments. The
trajectory-mean configuration uses the full attempt's target-token
denominator in Eq.~\ref{eq:memberloss}.

Synchronous and asynchronous execution differ in when work can overlap
policy updates. For asynchronous runs, policy lag is the learner version
at consumption minus the behavior version of each call. Lag and the
fraction of mixed-version attempts describe the executed configuration.

\subsection{Clipped Policy Objective}

Let \(b(s)\) denote segment \(s\)'s captured behavior version, and let
\(h_{ist}\) contain its multimodal context and generated prefix before
token \(x_{ist}\). The learner uses the PPO-style clipped surrogate
\citep{schulman2017ppo} with the member advantage from Eq.~\ref{eq:grpo}:
\begin{equation}
 \begin{aligned}
 \rho_{ist}&=\exp\!\left(\log\pi_\theta(x_{ist}\mid h_{ist})
 -\log\pi_{b(s)}(x_{ist}\mid h_{ist})\right),\\
 \ell_{ist}^{\rm clip}&=-\min\{\rho_{ist}A_i,
 \operatorname{clip}(\rho_{ist},1-\delta_-,1+\delta_+)A_i\}.
 \end{aligned}
\end{equation}
The likelihoods use the temperature and recorded-support convention in
Appendix~\ref{app:likelihood}. The parameters \(\delta_-\) and
\(\delta_+\) define the lower and upper clipping widths, respectively.
In the trajectory-mean configuration, Eq.~\ref{eq:memberloss} aggregates
token losses within each complete attempt before averaging attempts.

A behavior version is constant within a segment, but an asynchronous
attempt can span several versions. Captured likelihoods identify the
denominator for each response. For attempts spanning policy versions,
the clipped update is an off-policy approximation under policy lag and
changing state distributions. Terminal feedback assigns one attempt-level
advantage across the episode. Local process rewards and distillation are
separate extensions to this terminal-reward recipe.

\subsection{Visual Experience Delivery}
\label{app:visualdelivery}

The delivery mechanisms in Section~\ref{sec:rlupdates} address repeated
CPU image preparation and visual-tensor transfer. They preserve the
learner's inputs, unlike semantic compaction, which changes those inputs.
Lossless tensor encoding reduces transfer volume. CPU preprocessing reuse
is a separate mechanism whose enablement depends on processor compatibility.

The preprocessing-reuse path supports the Hugging Face Qwen2-VL processor's
independent per-image outputs. Within a member, it retains
\texttt{pixel\_values} and \texttt{image\_grid\_thw} under fixed processor
settings, reconstructs the requested image order, and processes new
content. A clean-to-overlay replacement or a configuration change requires
fresh processing. Unsupported processors or incompatible output structures
use ordinary processing. The cache stores CPU tensors, while neural visual
encoding remains part of the model forward pass.

Owned per-image storage avoids retaining an obsolete batch through a view.
The member's window is cleared on exit, and state publication remains scoped
to that member during cancellation. Tensors already staged for prompt projection
can be reused during training materialization under the same content and
processor identity.

The selected transport path uses Zstandard level~1 on contiguous CPU image
tensor bytes after data-parallel partitioning. An encoding replaces the
original only when smaller. A failed encoding leaves the original
available. The consumer checks the expected byte count, decodes its selected
microbatch, and transfers tensors to the accelerator without retaining a
decoded full-batch cache. Dtype, shape, resolution, and floating-point
values remain unchanged.

Payload accounting includes delivery-pending groups as well as reserved,
active, and ready work. Admission limits are expressed as numbers of groups.
Successful compact object-store transfer allows the producer to
release visual references, while consumption records keep sample identities
without those payloads. The data-parallel delivery barrier remains in place.

Equivalence checks compare prompt tokens, image order, preprocessing
tensors, dtype, shape, masks, rewards, and behavior likelihoods on the same
captured calls. Full-update comparison additionally fixes model, optimizer,
random state, and partition schedule. Component measurements cover
preprocessing, codec, transfer, and memory costs. The complete optimizer
interval is a separate measurement boundary.

\subsection{Sequence Integrity and Update Boundaries}
\label{app:learningprotocol}

Before assembling an actor RL segment, the adapter checks the
delivered-response receipt, policy/tokenizer/template/processor identities,
call order, and response digests. It admits main-agent calls only.
Terminal Controller evidence confirms delivery by linking ordered execution
steps and their artifact digests to the captured responses. Task success
is established separately by the outcome check.

Calls can be appended when prompt tokens and ordered media bindings
confirm a prefix extension under the same behavior version. Media identity
includes a content digest. Replacing a clean frame with an action overlay,
evicting an image, rewriting history, or changing behavior version therefore
starts a new segment from the newly captured input. Responses repeated as
history provide context and are masked out of the new segment's targets.
Their original target occurrences remain available for learning.
Processor-specific visual-position projection preserves captured image
content and order.

Segments from the same logical attempt share its identity and reward
provenance. The adapter checks group completeness before computing
advantages. In the trajectory-mean configuration, the learner sums target
token counts over the full attempt before partitioning. Every segment uses
this denominator within the same optimizer step. Partition invariance holds
when conditioning, masks, and token losses stay fixed.

Weight publication pauses new model calls, drains calls already in
progress, commits the new weights, and resumes calls with the new behavior
version. The tasks themselves can remain active in their workspaces.
After abnormal recovery, execution restarts from committed checkpoint and
cursor state, rebuilding warm queues and unfinished rollouts from that
committed state.

\subsection{Sampling Likelihood Convention}
\label{app:likelihood}

The policy ratio requires compatible learner and behavior likelihoods.
For temperature \(T>0\) and the recorded behavior support \(S_b(h)\),
the likelihood convention is
\[
 p_\theta^{(T,S_b)}(x\mid h)=
 \frac{\mathbf{1}\{x\in S_b(h)\}\exp(z_\theta(x\mid h)/T)}
 {\sum_{u\in S_b(h)}\exp(z_\theta(u\mid h)/T)}.
\]
Here \(h\) contains the captured context and generated prefix, and
\(z_\theta\) denotes the learner's logits. With top-p equal to one and no
other vocabulary restriction, \(S_b(h)\) is the full vocabulary. For
nucleus sampling \citep{holtzman2020nucleus}, replay uses the recorded token
identities and offsets at each target position rather than constructing
a new nucleus from current learner logits. Temperature, support capture,
and replay settings belong to the run configuration. Entropy
regularization has a separately configured support mask.

\subsection{NorthCUA-Ultrafast Adaptation and Evaluation}
\label{app:ultrafast}

This section details the choice-based execution mode introduced in
Section~\ref{sec:fastweb}. Its training unit is a decision state extracted
from a WebAgent trajectory. The original causal prefix supplies the input,
and the following recorded action supplies the label. The classifier and
native generator use this same underlying evidence through different prompts.

\subsubsection{Multimodal Inputs and Action Routing}

The classification prompt retains the task, preceding actions, tool results,
screenshots, and action overlays. It replaces the original system instruction
with a finite-choice instruction and appends the current operation and target
candidates. Native generation instead uses the original messages, tool
schemas, and chat template, keeping the original action prompt separate
from classification-specific instructions and selections.

Targets come from the latest valid set-of-mark observation. We exclude
explicitly disabled elements and restrict filling to input-like roles and
checkbox changes to checkbox roles. At most 150 elements are retained per
observation, with the number omitted recorded. A finite operation is offered
when its tool and eligible targets are available. Selection operates over
this retained candidate set.
Table~\ref{tab:ultrafastroutes} gives the seven routes. The operation and
each available target type form separate variable-size choice questions.
Only the chosen operation's target is used for execution.

\begin{table}[!htbp]\centering\small
\caption{Ultrafast routing and completion. Current references are resolved
through the latest observation. Completion describes strict routing.
The original runner performs tool execution and checks the final response format.}
\label{tab:ultrafastroutes}
\begin{tabularx}{\linewidth}{P{24mm}P{23mm}X}
\toprule
Route & Target choice & Completion \\
\midrule
CLICK & Element & Construct a click call directly. \\
HOVER & Element & Construct a hover call directly. \\
CHECK & Checkbox & Construct a check call directly. \\
UNCHECK & Checkbox & Construct an uncheck call directly. \\
FILL & Input field & Generate the original fill action and require its field
to agree with the selected target. \\
NATIVE & None & Generate a complete offered tool call and its arguments. \\
RESPOND & None & Generate the original structured final response. \\
\bottomrule
\end{tabularx}
\end{table}

In the offline full-action evaluation below, generation starts with a fresh
input encoding and cache, with any LoRA adapters disabled. Decoding is greedy
with thinking disabled and at most
4,096 new tokens. A valid result requires model termination and the relevant
tool schema or response structure. Strict routing rejects FILL target disagreements.
NATIVE accepts a valid offered tool, including a finite-action tool if the
original generator selects one. This matters when comparing route labels
with complete actions below.

\subsubsection{Readout and Training}

The last hidden layer supplies four vectors for each candidate. These are
the mean context representation before the appended choice block, the mean
question-instruction representation, the mean candidate representation, and
the final decision-marker representation. Each vector has dimension 4,096.
A shared head normalizes and projects these roles to width 512, adds learned
role embeddings, and applies two Mixer blocks. Each block mixes the four
roles through a $4\!\rightarrow\!16\!\rightarrow\!4$ MLP and the channels
through a $512\!\rightarrow\!1024\!\rightarrow\!512$ MLP. A scalar readout
from the updated candidate role supplies its score. Softmax is applied within
each question, and argmax selects its answer. The Mixer operates within each
candidate's four roles, while the backbone has already encoded the full
candidate list.

For state $i$, let $Q_i$ contain the supervised operation question and,
when required, the corresponding target question. With correct candidate
$y_{iq}$, score $z_{iqk}$, and candidate set $C_{iq}$, the objective for a
minibatch of decision states $\mathcal{B}$ is
\begin{equation}
\mathcal{L}_{\mathrm{choice}}=
\frac{1}{|\mathcal{B}|}\sum_{i\in\mathcal{B}}
\frac{1}{|Q_i|}\sum_{q\in Q_i}
-\log\frac{\exp z_{iqy_{iq}}}{\sum_{k\in C_{iq}}\exp z_{iqk}}.
\label{eq:ultrafastchoice}
\end{equation}
Averaging within each state keeps states with an operation and target
equally weighted with operation-only states. Other target questions receive
no label, and the original output tokens receive no generation loss.

The dataset contains 225,710 training, 5,203 validation, and 6,087 test states
from 25 trajectory collections. Normalized task goals define groups, and
the fixed holdouts remain unchanged when training sources are expanded.
Filtering removes stale or missing targets, conflicting labels, invalid
actions, and known target-tool errors. The native multimodal prefixes are
rejoined to their source records before training. Both input paths are
checked against a 73,728-token budget after image-token expansion, without
truncation. Grouping removes identified request duplicates. Semantically
related paraphrases and task templates may still occur across groups.

\begin{table}[!htbp]\centering\small
\caption{Ultrafast adaptation settings. Original model weights remain frozen
in both variants. Parameter percentages use the complete 9.410B-parameter
multimodal checkpoint as the denominator.}
\label{tab:ultrafastconfig}
\begin{tabularx}{\linewidth}{P{49mm}XX}
\toprule
Setting & Head-only & Head + LoRA \\
\midrule
Trainable parameters & 4.214M (0.0448\%) & 47.492M (0.5047\%) \\
Choice head & Width 512, two blocks & Same \\
LoRA & None & Rank 16, scale 2, dropout 0 \\
Adapter scope & None & All 248 language-layer linear modules \\
Optimizer & AdamW & AdamW \\
Learning rate & $10^{-4}$, constant & Same \\
Global batch & 128 states & Same \\
Precision & BF16 backbone, FP32 head & BF16 backbone, FP32 head/adapters \\
Gradient clipping & Global norm 1.0 & Same \\
Seed & 42 & 42 \\
Completed updates & 63,504 (36 epochs) & 24,521 (13 epochs plus part of 14) \\
Selected checkpoint & Epoch 22 & Epoch 1 \\
\bottomrule
\end{tabularx}
\end{table}

AdamW uses $\beta=(0.9,0.999)$, $\epsilon=10^{-8}$, and weight decay 0.01,
with no learning-rate warmup or schedule. Candidates retain deterministic
ID order, without shuffling or class reweighting. Head + LoRA adds 43.278M
adapter parameters to the same 4.214M head. The adapters remain unmerged so
that generation can disable them. The two variants share data, logical
batch size, and seed, but use different completed budgets. Checkpoints are
selected by full-validation joint accuracy, then loss and earlier epoch.

\subsubsection{Offline Decision and Action Results}

Route-and-target joint accuracy requires all labeled questions for a state
to be correct. It excludes generated text and other open-ended parameters.
Full-action exact matching instead compares the tool name and complete
arguments, or the final response object, with the reference action. It
normalizes reference IDs to strings, then compares tool names, parameter
values, and response text exactly. A NATIVE route can still generate the
correct finite-action tool call, so full-action matching can credit a
decision whose route label differs from the reference.

\begin{table}[!htbp]\centering\small
\caption{Offline Ultrafast results in percent. Validation and test contain
5,203 and 6,087 states. Original generated actions are projected into the same
route labels. Missing results are left as dashes.}
\label{tab:ultrafastoffline}
\begin{tabularx}{\linewidth}{X C{28mm}C{28mm}C{32mm}}
\toprule
Policy & Validation joint & Test joint & Test action exact \\
\midrule
Full generation & -- & 85.0 & 64.6 \\
\rowcolor{NexLight}
Head-only & 86.1 & 85.4 & 64.0 \\
Head + LoRA & 88.2 & -- & -- \\
\bottomrule
\end{tabularx}
\end{table}

Head-only correctly selects the route and target in 5,196 test states,
compared with 5,172 for projected full generation. Complete-action matching
is lower for the hybrid policy, at 3,897 versus 3,933 states. In the paired
comparison, 3,757 states match under both policies, 140 only under Head-only,
and 176 only under full generation.

\Needspace{5\baselineskip}
The hybrid gains 46 exact matches among
2,576 CLICK states, while losing 25 among 906 FILL states and 30 among 2,087
NATIVE states. Losses on generation-dependent FILL and NATIVE actions
exceed the gains on CLICK.

The diagnosis is clearest among the 1,403 states with a correct Head-only route
but an incorrect hybrid action. Of these, 935 have mismatched parameters for
the same tool, 359 have a tool/final-output kind mismatch, 18 differ in final
JSON fields, and 91 are rejected by route or target consistency checks.
These are reference-matching outcomes, distinct from environment success.
The paired test reuses one saved native generation per state for both
policies, including states whose hybrid action bypasses generation. Its
execution time covers this shared-generation offline procedure.

Validation selects epoch 1 for Head + LoRA and epoch 22 for Head-only.
At the same completed epoch 13, their joint accuracies are 86.9\% and
85.5\%, respectively. This provides an equal-epoch comparison,
alongside the selected checkpoints in Table~\ref{tab:ultrafastoffline}.
All results come from one training seed, and states can share a task.
Earlier text-only experiments use a different input protocol and are
excluded from this multimodal comparison.

\subsubsection{Online Task Completion and Timing}
\label{app:ultrafastonline}

The online comparison adapts the long-horizon-SFT NorthCUA-Nano checkpoint
before joint RL and retains the original WebAgent runner. Each inference
replica occupies one GPU. Full generation sends every decision through the
native generation path. Both Ultrafast variants and the case in
Figure~\ref{fig:ultrafast-case} use the generation-preferred profile with a
finite-choice confidence threshold of 0.8. A direct action requires both
its operation and target probabilities
to meet this threshold. Lower-confidence selections, missing or stale
current references, and unsupported observation surfaces defer to native
generation. When generation is used, its complete valid action is accepted
without replacing its arguments with the classifier's selection. The
gateway records any disagreement. This differs from the strict agreement
checks used in the offline comparison above.

\paragraph{Task outcomes and matched execution records.}
The evaluator assigns success to 521, 481, and 492 of the 812 tasks for
Full generation, Head-only, and Head + LoRA, respectively. Failures,
timeouts, and evaluation errors remain in the denominator. The latest
Head-only and LoRA scores come from rescoring existing trajectories.
Their efficiency statistics retain the same execution records.

Task time uses a fixed matched subset of 320 tasks,
selected from 343 tasks jointly successful in the timing snapshot.
The remaining 23 lack complete timing or action-dispatch evidence for at
least one policy. This fixed set supports comparisons on the same tasks
and defines the population for the conditional means. Rescoring
updates the success totals while leaving this timing population unchanged.

\paragraph{Step and action boundaries.}
A step is one decision round, including a final response. Classification
followed by generation or a length retry remains one step. An action counts
a tool call dispatched to its handler, including calls that subsequently
fail during execution. A step may issue multiple actions or none, and a
final response is not a tool action. Only the final selected attempt is
counted.

Decoding bypass counts recorded dispatched actions produced directly by
the choice head without invoking native generation. The fractions are
6,182/15,385 for Head-only and 4,248/13,652 for Head + LoRA, or 40.2\%
and 31.1\%. These denominators cover available actions from successful and
failed tasks, separately from the 320-task timing set. Incomplete action records affect
12 and eight tasks, respectively. Missing actions are excluded from both
counts. Full generation has zero bypass by definition.

\paragraph{Measured time and deployment.}
Task time spans the agent execution loop
through terminal observation and result persistence. It includes decisions,
browser operations, and observations, while excluding environment
provisioning, browser teardown, and evaluator scoring.
The measured mean for Full generation is 616.8\,s on the matched task set.

Full generation uses temperature 0.7, top-$p$ 0.95, top-$k$ 40, high
reasoning effort, and up to 16,384 output tokens. The hybrid runs combine
temperature-zero attempts with temperature-one, top-$p$-one length retries.
Each replica occupies one GPU, with run-specific replica counts, concurrency,
and browser placement. The reported task times characterize these
evaluation deployments.

\section{Web Repair and Testing Protocol}
\label{app:web}

\subsection{Qualifying Browser Evidence}
\label{sec:webscenario}

The main-text case in Figure~\ref{fig:webtestcase} records browser testing
and code repair during local-page development. To decide whether an
observation justifies a product change, the proposed evaluation protocol
first checks that the relevant test actions completed. A mismatch may
otherwise originate in the test specification, the agent's actions, or
the environment
\citep{yang2024swebenchmultimodal,wan2026tddev}.

Figure~\ref{fig:webloop} uses a title-persistence requirement to make this
decision concrete. Record R17 should retain \emph{Delivery plan v2} after
reload. The Web Agent completes the edit and save, observes the new title,
then reloads the record and finds the old title. Together, these observations
support a candidate persistence defect. The report pairs the expected and
observed values with the completed actions, tested build, reset state, and
supporting artifacts, giving the Coding Agent a specific discrepancy to
investigate.

\paperfigure{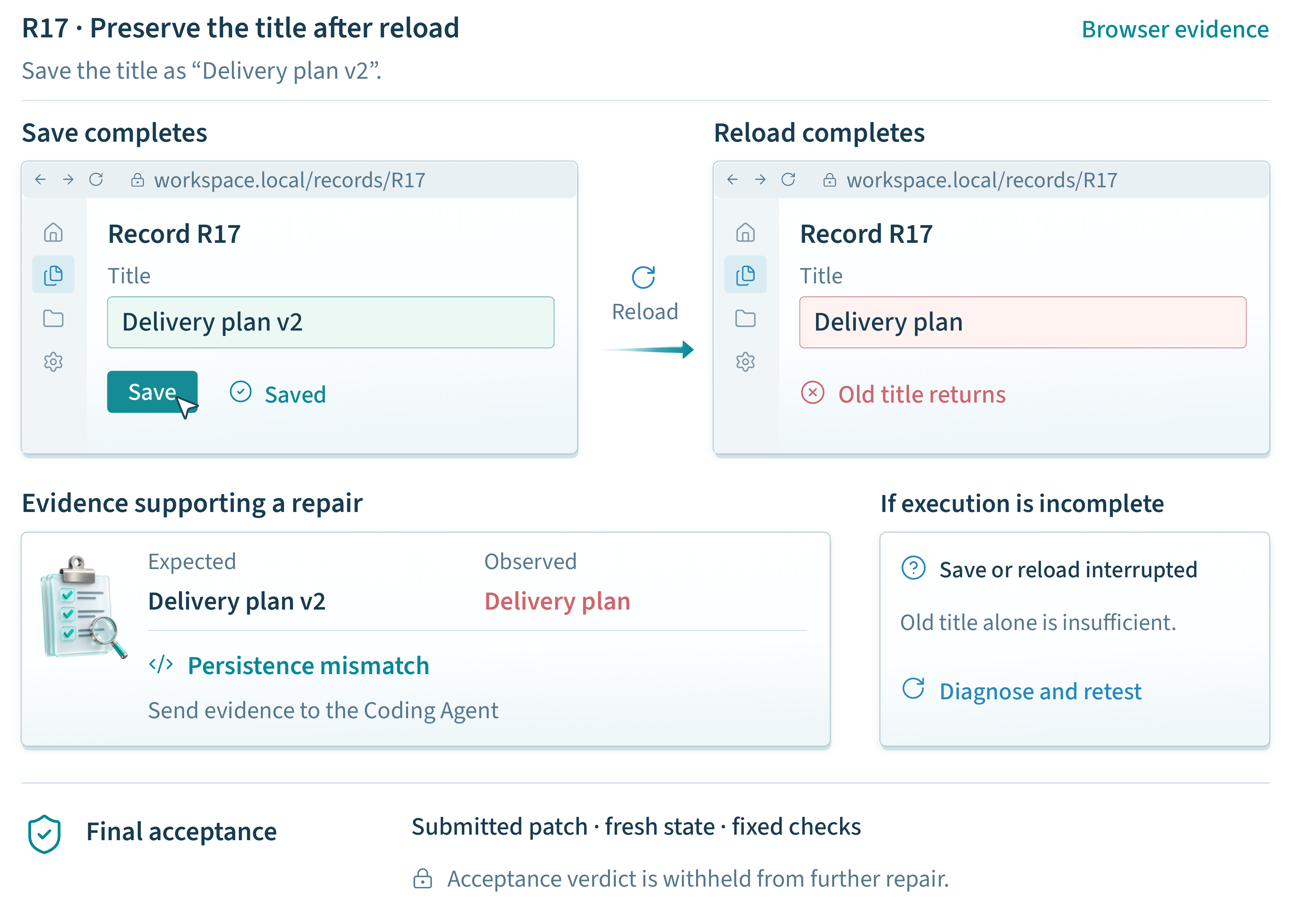}{fig:webloop}{\textbf{A persistence example for evidence qualification.}
The new title appears after saving but reverts after a completed reload.
The report preserves this mismatch for repair. An interrupted save or reload
requires diagnosis and retesting before assigning a product defect.
Final acceptance uses fixed checks in a fresh state outside the repair loop.
GUI states and inspection materials illustrate the proposed protocol.}

The same old title has a different interpretation if Save could not be
activated or reload did not complete. In that case, the test has not
established that the product lost a saved value. The proposed adjudication
rule therefore sends unresolved execution failures to diagnosis or bounded
retesting. Test steps may be repaired and exploratory checks added while
the issue's acceptance goals and regression requirements stay fixed.

\Needspace{5\baselineskip}
Repair feedback and final acceptance also serve different purposes.
Feedback helps the agents revise a candidate. Final acceptance evaluates
the submitted patch from a protected clean state, using checks the repair
agents cannot change. Its verdict is withheld from further revision, so
acceptance tests whether the fix survives beyond the session used to develop it.

Appendix~\ref{app:webassignment} details the common assignments, resets,
and denominators needed to compare feedback conditions.

These controls support the proposed comparison of browser-feedback conditions
\citep{wang2026asukabench}. Inclusion of an episode in model training follows
the separate source and consumption accounting in Appendix~\ref{app:corpus}.

\section{NorthCUA Case Studies}
\label{app:showcase-cases}

The following cases trace complex software tasks from an initial workspace
or source records to editable deliverables and native task outcomes. They
illustrate sustained GUI interaction across geometry, data analysis, visual
composition, circuit simulation, 3D modeling, and gameplay. The recorded step
counts include GUI actions,
observations, preparation, checks, and retries. The release packages preserve the full
traces, infrastructure repairs, native outputs, and acceptance records.

\subsection{Ruby Heart}
\label{app:showcase-heart}

\begin{figure}[!t]
  \centering
  \includegraphics[width=\linewidth]{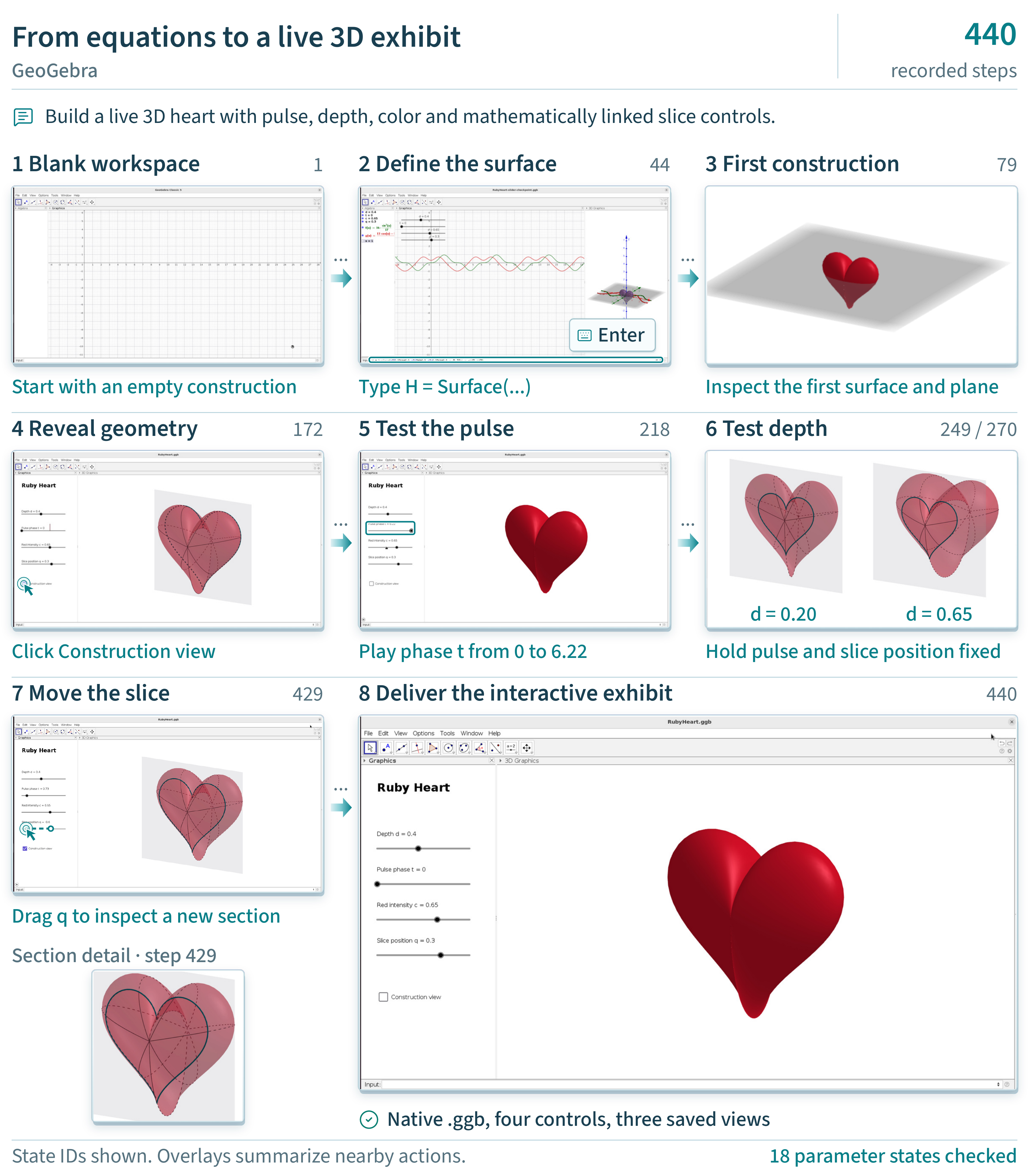}
  \caption{\textbf{NorthCUA case study in GeoGebra.} The trajectory connects
  expression entry to geometry, playback, and parameter checks. The first
  construction is enlarged from state 79. Panel 6 compares depths 0.20 and
  0.65 at fixed pulse phase 0 and slice position $-0.8$. The section detail
  enlarges state 429. Numbers identify captured states in row order.
  Editorial overlays summarize nearby actions, with intervening actions omitted.}
  \label{fig:showcase-heart}
\end{figure}

The task is to turn supplied equations into a GeoGebra heart that can pulse,
change depth and color, and reveal a moving cross-section. The difficulty is
keeping the cutting plane and section linked to the changing surface.
Figure~\ref{fig:showcase-heart} begins with an empty workspace. The agent
enters native expressions to build a first construction, then arranges the
geometry and four controls into an interactive exhibit. Construction view
reveals the auxiliary geometry linked to the surface for inspection.

The agent then exercises these dependencies through playback and direct
manipulation. Holding pulse phase and slice position fixed, it checks two
depth settings that visibly change the surface and its section. It also
reopens the saved file and later drags the slice to another position.
The trace covers 18 combinations of
depth, slice position, and pulse phase, connecting the visible shape to its
underlying parameters. The 440-step record ends with the final saved
\texttt{.ggb} and three separate view images. The editable construction
supports continued interaction with the geometry established during modeling.

\subsection{Silicon Atlas}
\label{app:showcase-atlas}

\begin{figure}[!t]
  \centering
  \includegraphics[width=\linewidth]{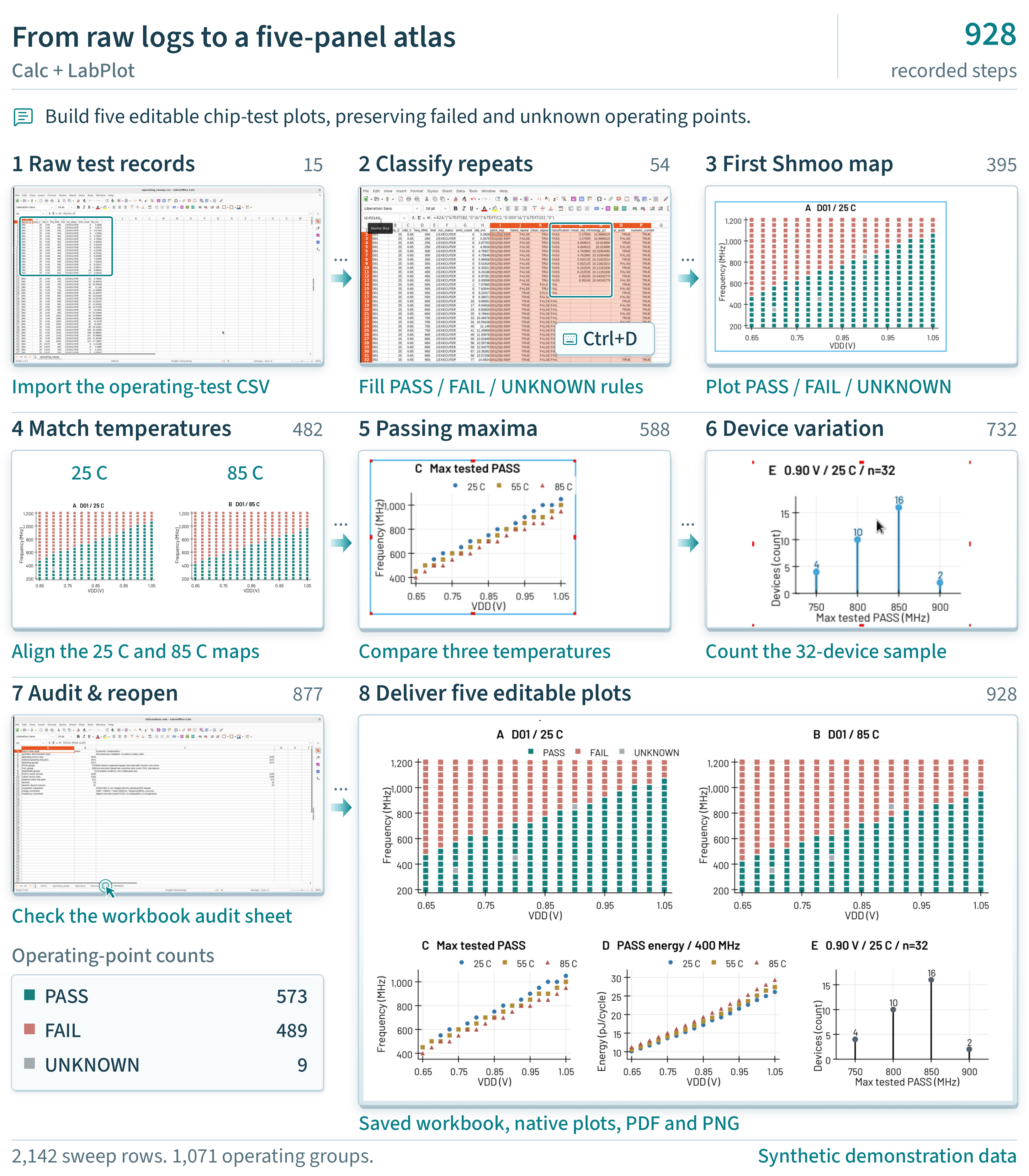}
  \caption{\textbf{NorthCUA case study in Calc and LabPlot.} Repeat-aware
  classification supplies the temperature maps and derived plots. Focused
  crops distinguish temperature comparison, passing-frequency maxima, and
  32-device counts before the complete atlas. The count inset transcribes
  the audit totals at state 877. Numbers identify captured states, and
  editorial overlays summarize nearby actions. Intervening actions are omitted.
  The data describe a synthetic demonstration.}
  \label{fig:showcase-atlas}
\end{figure}

This task asks for five editable plots from synthetic chip-test records.
The challenge is preserving measurement meaning across spreadsheet analysis
and plotting. Recorded failures take precedence. Otherwise, incomplete tests
remain UNKNOWN, while missing current measurements stay missing. In
Figure~\ref{fig:showcase-atlas}, the agent imports 2,142 sweep rows into Calc
and fills formulas that classify 1,071 operating groups. A passing group
requires two clean expected repeats, and energy is computed only for
passing points. These rules govern the subsequent plots.

The classified table first feeds a Shmoo map of voltage and frequency in
LabPlot. Matching axes on a second temperature map make the operating regions
directly comparable. The agent next plots the maximum tested passing
frequencies and passing-point energy. A separate 32-device sample supplies
the final count plot, completing the atlas. The agent then reopens
the workbook to check counts and units, repairs crowded labels, and exports
the revised figure. The audit retains 573 passing, 489 failing, and nine
unknown groups. After 928 recorded steps, the workbook and plotting project
preserve the calculations and editable plots alongside the PDF and PNG outputs.

\subsection{NOVA Presentation}
\label{app:showcase-nova}

\begin{figure}[!t]
  \centering
  \includegraphics[width=\linewidth]{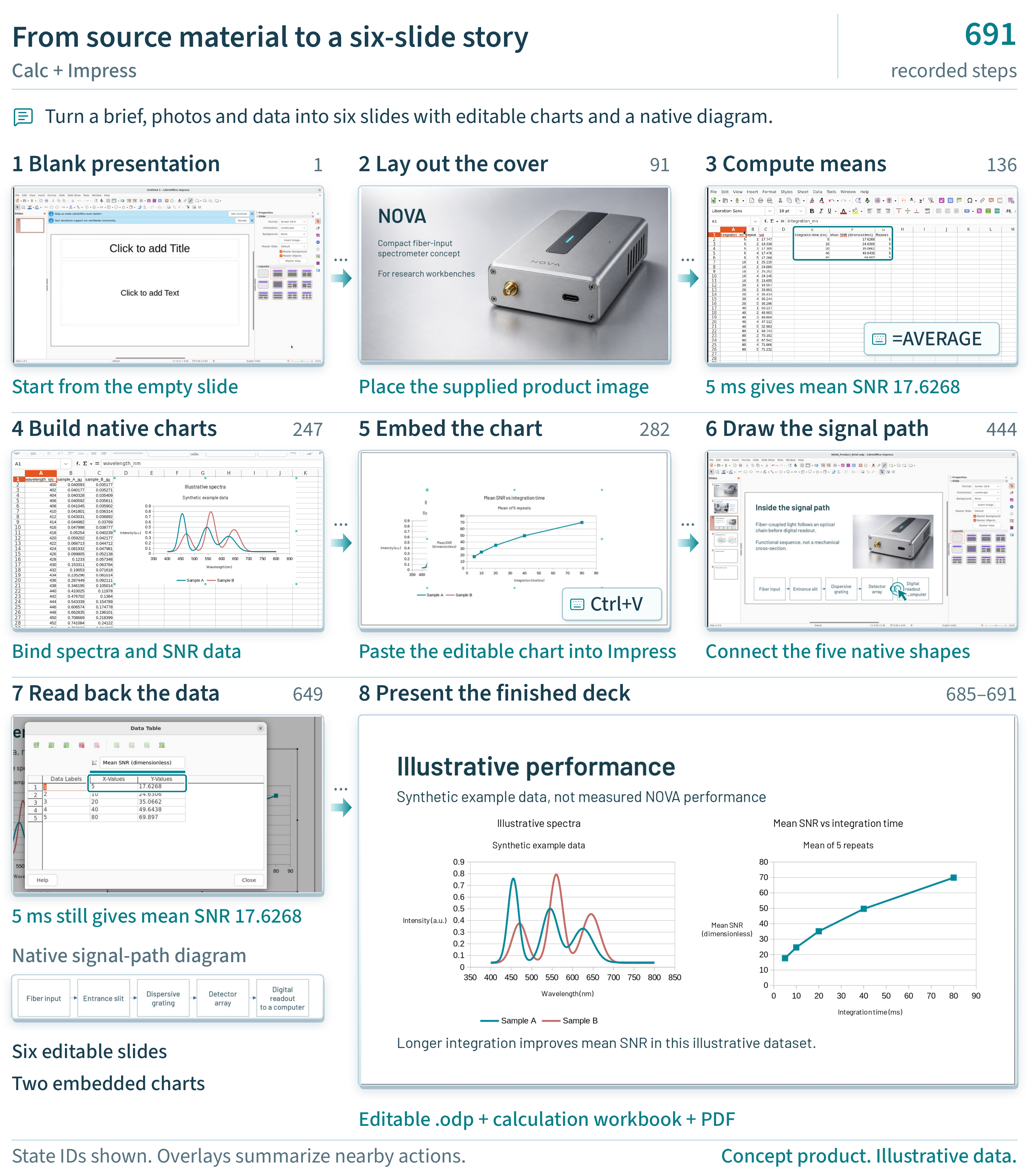}
  \caption{\textbf{NorthCUA case study in Calc and Impress.} Selected states
  follow calculation, chart transfer, diagram construction, and data readback.
  The chart at state 282 has just been inserted and is later resized and
  positioned. The lower-left inset enlarges the native diagram at state 687,
  while the large view shows the finished evidence slide at state 688.
  Enlarged chart and data-table views connect the mean at state 136 to its
  embedded value at state 649. Numbers identify captured states.
  Editorial overlays summarize nearby actions.}
  \label{fig:showcase-nova}
\end{figure}

The NOVA task asks for a six-slide presentation from a product brief,
supplied images, and numerical data. The challenge is preserving accurate,
editable scientific content while building a coherent visual story.
Figure~\ref{fig:showcase-nova} starts with an empty presentation and a first
cover. To support the performance slide, the agent computes mean
signal-to-noise ratios from five repeats in Calc. The resulting table and
supplied spectra feed native charts, which it embeds in Impress. Editable
shapes and connectors then form the five-stage signal-path diagram.

The agent reads back the embedded chart data to check that application
transfer preserved the calculation. The figure shows the same mean of
17.6268 at 5 ms in both views, making this check directly inspectable.
A presentation-mode inspection covers all six slides before the 691-step
record ends with an editable \texttt{.odp}, the calculation workbook, and
a matching PDF. The task thus preserves a concrete numerical result through
calculation, chart transfer, and final presentation. NOVA is a fictional
product concept with illustrative imagery and numerical data.

\FloatBarrier
\subsection{CMOS Inverter Analysis}
\label{app:showcase-virtuoso}

\begin{figure}[!ht]
  \centering
  \includegraphics[width=\linewidth]{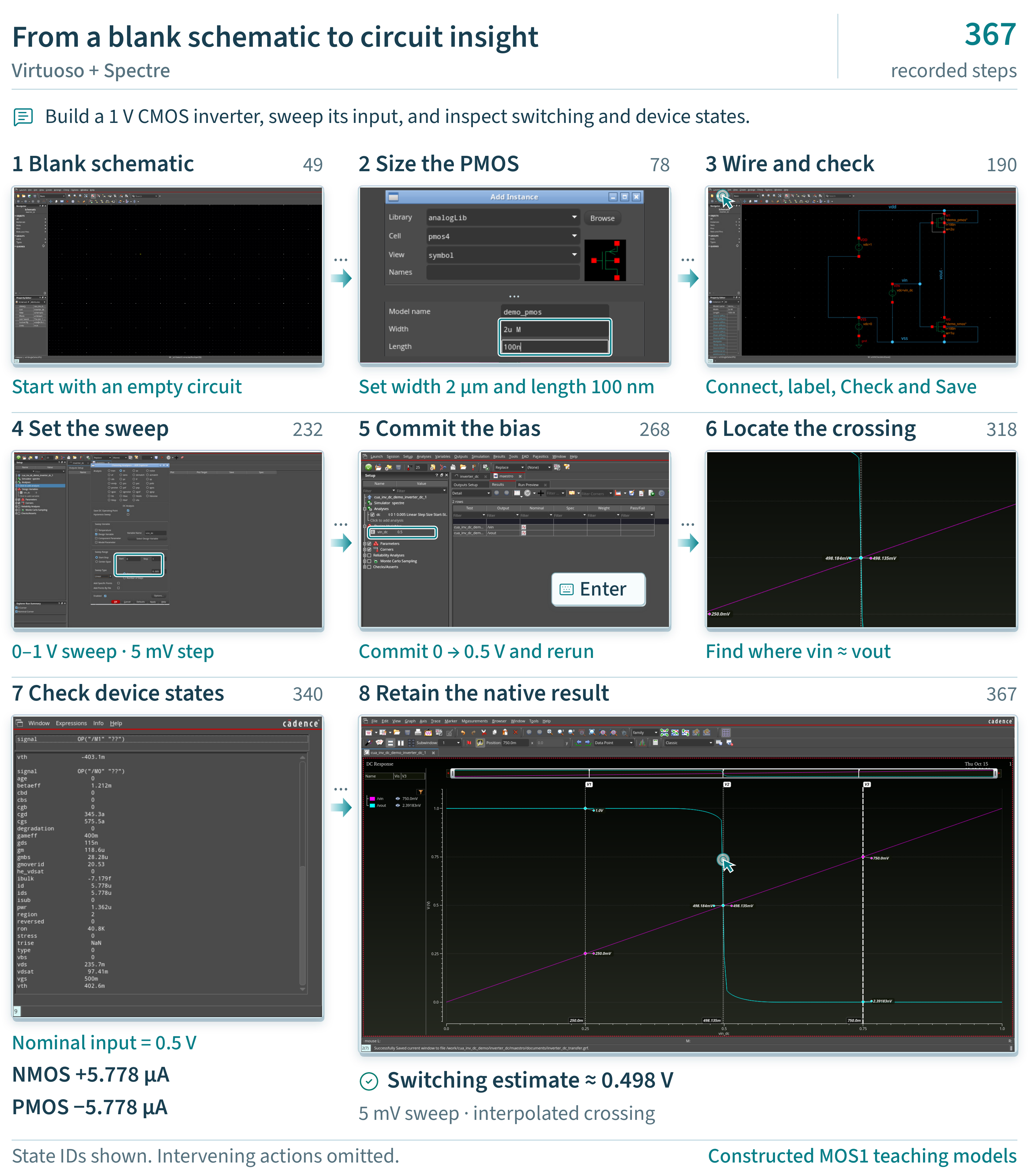}
  \caption{\textbf{NorthCUA case study in Virtuoso and Spectre.} Panel 2 pairs
  device identity and size fields from the same native dialog at state 78.
  Panel 6 enlarges the crossing at state 318, estimated by interpolation
  between the 0.495 and 0.500 V sweep samples. Current annotations transcribe
  the PMOS and NMOS reports at states 335 and 340, respectively, both at
  nominal input 0.5 V. Numbers identify captured states. Editorial overlays
  summarize nearby actions, with intervening actions omitted.}
  \label{fig:showcase-virtuoso}
\end{figure}

The task asks for a 1 V CMOS inverter, its DC transfer curve, and transistor
operating-point checks in Virtuoso and Spectre. The difficulty is carrying
the intended device and analysis settings from the schematic into the
simulator, then connecting numerical readouts to the conditions that produced
them. Figure~\ref{fig:showcase-virtuoso} begins with an empty schematic.
The agent sizes the two transistors, with the PMOS width and length fields
shown in the figure. It then wires and labels the circuit and checks
the completed design before configuring a 0--1 V input sweep with a 5 mV step.
An uncommitted input edit then exposes a subtle GUI-state mismatch. The first
sweep completes successfully, but the saved nominal input remains at 0 V.
The agent returns to ADE, enters 0.5 V, commits it with Enter, and reruns the
analysis so that the operating-point report uses the intended bias.

With the corrected configuration, the agent inspects the transfer curve in
ViVA and locates the crossing of input and output near 0.498 V. This is an
interpolated estimate from the 5 mV sweep. It separately reads both device
reports at the nominal 0.5 V input, where the drain currents are
$+5.778\,\mu$A and $-5.778\,\mu$A under the simulator's sign convention.
The 367-request record ends with the native schematic, simulator setup,
waveform state, and operating-point evidence retained for inspection.
The case shows how a sustained GUI workflow can carry a circuit through
construction, configuration repair, and numerical interpretation while
preserving the engineering artifacts behind the result. The supplied
MOS1 models are constructed teaching models.

\FloatBarrier

\begin{figure}[!t]
  \centering
  \includegraphics[width=\linewidth]{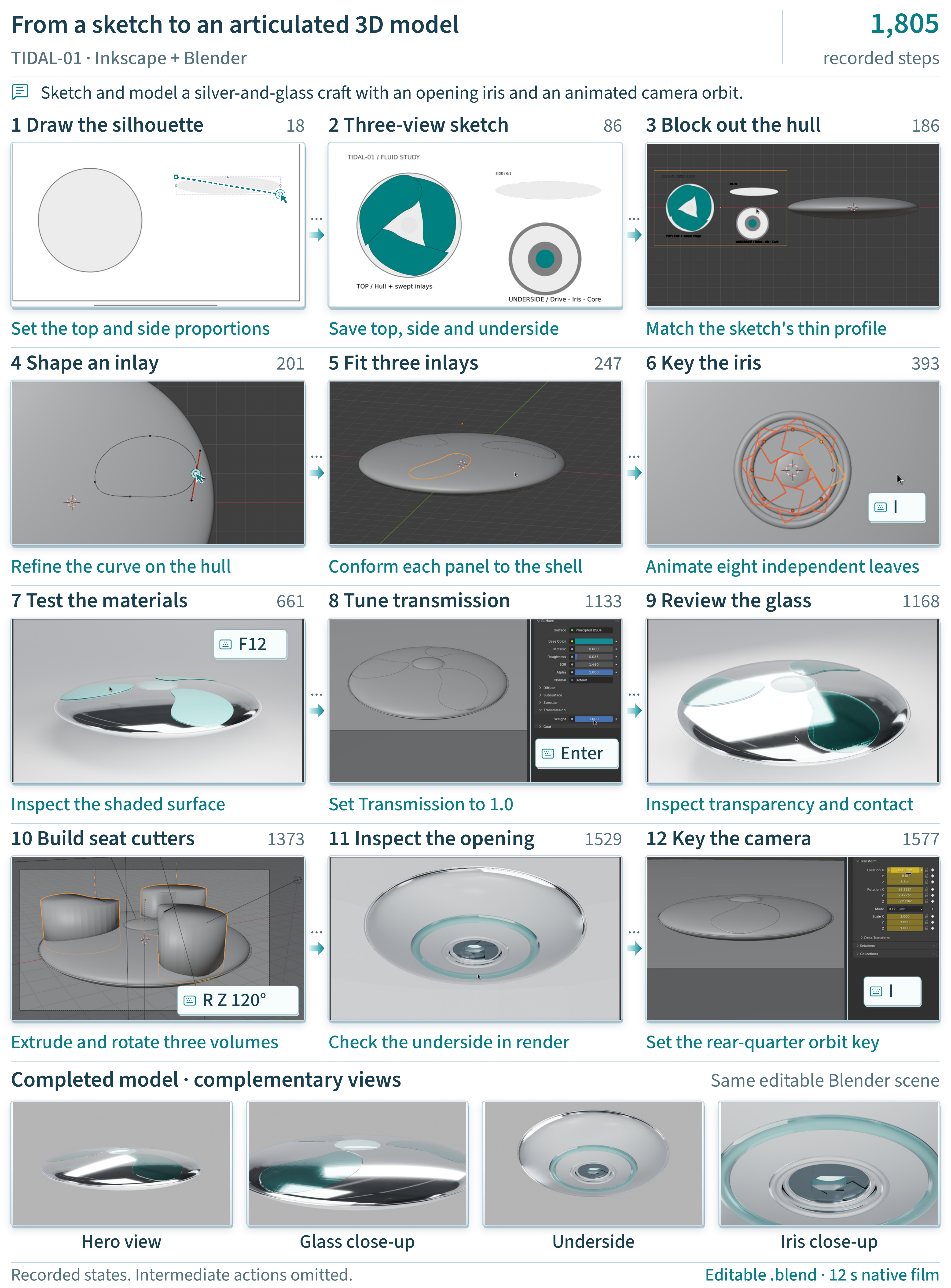}
  \caption{\textbf{NorthCUA case study in Inkscape and Blender.} Twelve selected
  states trace drawing, modeling, material revision, and camera keying. Panel 2
  is the exported sketch. Panels 8 and 12 pair viewport and settings crops
  from their respective states. Editorial cues summarize nearby actions.
  The compact final row combines native hero, glass-detail, underside, and
  iris-detail renders from the same scene. Intermediate actions are omitted.}
  \label{fig:showcase-tidal}
\end{figure}

\subsection{TIDAL-01 Modeling and Animation}
\label{app:showcase-tidal}

TIDAL-01 asks for a silver-and-glass craft with a thin hull, swept inlays,
and an opening iris beneath the body. The challenge is carrying this visual
design from a two-dimensional drawing into editable geometry whose separate
parts can be animated and inspected from different angles.
Figure~\ref{fig:showcase-tidal} begins with the top and side outlines in
Inkscape, then shows the completed three-view sketch. The agent places this
drawing beside a Blender blockout to match the thin side profile. It then
shapes a swept curve on the hull and fits
three inlays to its surface. To animate the underside opening, it assembles
the iris from separately pivoted leaves and keys their opening poses. The close view of
the eight-leaf assembly makes this change from static shape to articulated
geometry visible.

An earlier render still shows pale, opaque inlays. The agent sets glass
transmission to 1.0, then inspects a later render to assess transparency and
contact with the hull. To refine the panel seats, it constructs three
matching cutter volumes by extruding and rotating the inlay outlines.
After further geometry work, an underside render exposes the optical ring
and open iris. With these parts established,
it keys a tracked camera arc that moves from the upper hull to the underside
and returns to a rear-quarter view. The final strip combines whole-object
views with glass and iris close-ups from the same editable scene, connecting
the sketch's silhouette and underside layout to the rendered object. Across 1,805 recorded
steps, including observations, checks, and retries, the workflow produces an
original SVG, a packed \texttt{.blend}, and a 12-second native animation with
288 frames at 24 fps. Reopening the packed scene reports no missing files.
The case demonstrates a sustained 3D-authoring workflow in which geometric
construction, material revision, and camera control contribute to one
reusable animated asset.

\FloatBarrier

\subsection{Zuma Gameplay}
\label{app:showcase-zuma}

\begin{figure}[!t]
  \centering
  \includegraphics[width=\linewidth]{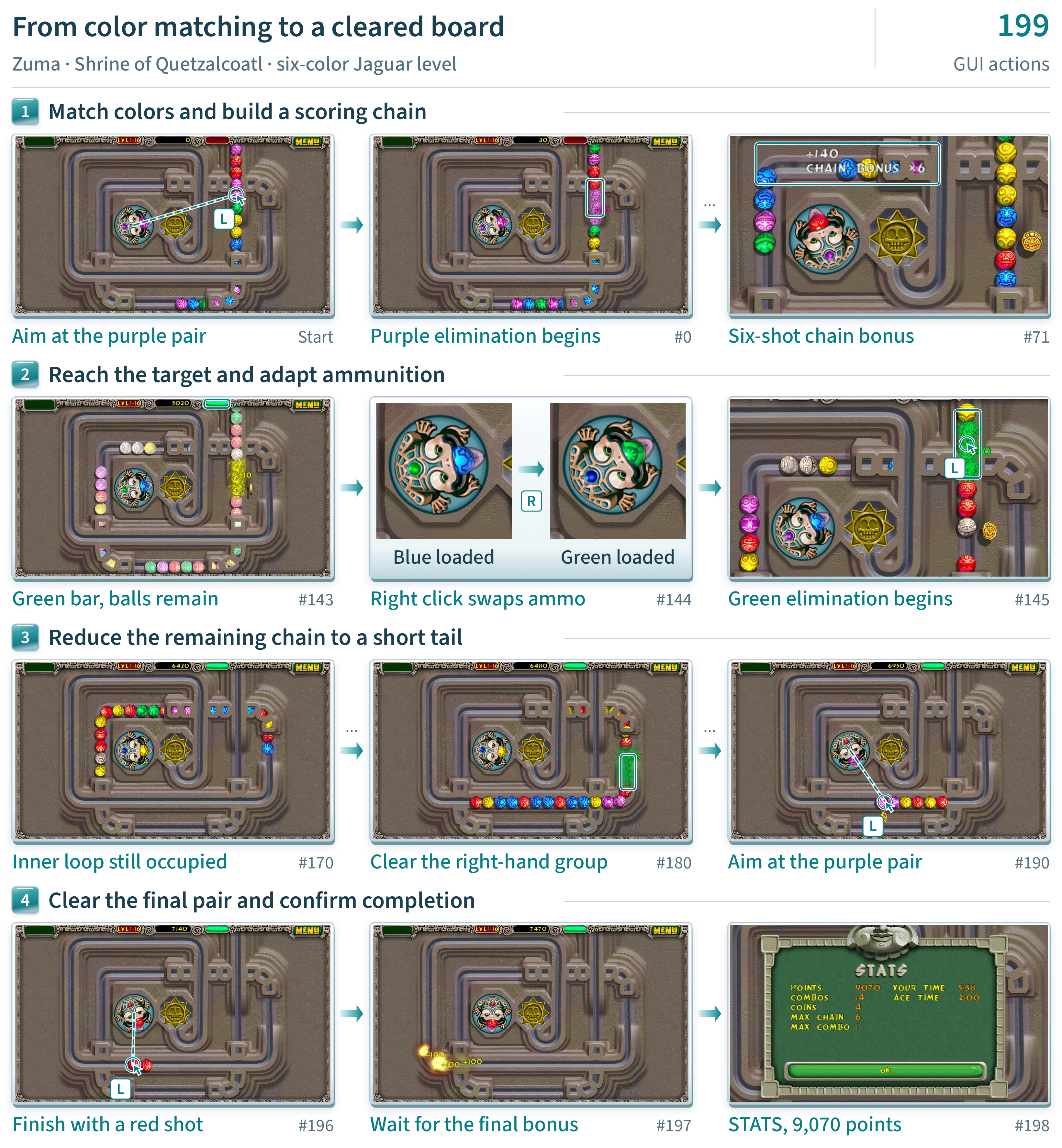}
  \caption{\textbf{Screenshot-driven Zuma gameplay.} Panels follow row order.
  \# denotes zero-based after-action IDs. Editorial aim guides precede actions 0, 191,
  and 197. Paired ammunition crops straddle action 144. L/R denotes mouse
  buttons. Intervening actions are omitted. The game pauses between actions.}
  \label{fig:showcase-zuma}
\end{figure}

Turns and tunnels change the accessible targets in this six-color Jaguar-level
task. Figure~\ref{fig:showcase-zuma} begins with a purple match and progresses
to a six-shot chain bonus. At 5,020 points, the progress bar turns green with
balls still in play. The agent swaps blue ammunition for green, clears the
exposed pair, and works through the bend to a short bottom-lane tail. It
eliminates the last purple and red groups, waits for the completion bonus,
and reaches native STATS with 9,070 points after 199 GUI actions.

\section{Runtime and Admission Protocols}

This appendix defines the runtime choices needed to interpret an
interaction record, including the observations presented to the policy, the tools
available to each role, environment ownership, and demonstration selection.
The learning-side segment and update rules are consolidated in
Appendix~\ref{app:learningprotocol}.

\subsection{Runtime Configuration}
\label{app:runtimeconfig}

A context profile determines which retained interactions enter each model
call. The sliding visual profile counts the latest clean frame within its
image budget and keeps textual action/result records after image eviction.
Other profiles replay visual history within a compaction epoch. Text
windows, image budgets, and semantic compaction are separate settings.
The generic context manager defaults to off. Shadow mode records decisions
without applying them, and enforce mode applies its configured policy.

Historical overlays mark issued actions on pre-action images, while the
latest clean frame shows the currently observed state. Text-only tools,
including notes, do not add visual frames. A later application change can
require waiting or a fresh observation. After a timeout or failed refresh,
state inspection precedes retry. Web element references belong to the
observed page state, and failed refreshes invalidate affected aliases.

Compaction retains a structured account of the goal, constraints, supported
facts, completed work, failures, current state, open items, and artifact
references. Summaries and notes carry their temporal provenance and remain
distinguishable from observations. The summary prompt, protected recent
interactions, trigger/target budgets, and failure policy determine the
resulting input. Note categories are fact, correction, precaution,
constraint, decision, and checkpoint. Compatibility names map
\texttt{mistake} to correction and \texttt{plan}/\texttt{other} to
checkpoint. The exposed schema, replacement behavior, and prompt budget
control which notes remain visible.

\paragraph{Execution scopes and tool interfaces.}
\label{app:executioninterfaces}

A session binds the agent to its environment and resources. A round tracks
one request and execution budget, and a step records one policy-decision
and action-execution iteration. A step can contain several tool calls,
so model decisions, tool calls, and GUI interactions are different counts.
The dispatcher validates declared arguments before invoking local handlers
or connected MCP tools and returns results to the requesting step.
Appendix~\ref{app:actionspaces} specifies the model-facing interfaces.
Runtime probes use a separate permission profile from actor tools.

\paragraph{Evaluator profiles and attribution.}
\label{app:evaluatorprotocol}

Evaluation binds a task and rubric to an evidence source, inspection
permissions, budget, and report schema. The rubric-generation template
requests three to seven criteria. The Agent Kit desktop entry path can use
a request-level fallback, while the web profile supplies eight fixed
criteria. The selected profile determines which criteria, weights,
applicability rules, and reward conversion apply. Report-format validation
checks the report's structure.

Artifact tools inspect retained records. Live tools additionally require
authorization and a dispatcher bound to the evaluated workspace.
Artifact-only replay has no live device access. Read-only probes are
preferred. Shell inspection uses an allowlist, and state-changing GUI or
browser inspection requires explicit permission. The workspace remains
reserved through required live checks.

Actor credit depends on the state before evaluator intervention. Terminal
evidence and evaluator actions are therefore recorded separately.
Inspection checks the submitted artifact or page state. A potentially mutating
check needs a procedure that preserves the original state, such as a
prepared evaluation copy. When that basis cannot be preserved, assessment
uses retained evidence or leaves the condition unresolved. Probe failure,
conflicting evidence, and an exhausted inspection budget are distinct from
a supported task-failure verdict. When evaluator protection is enabled,
recognized evaluation failures remain unresolved without replacing the
actor's terminal status.

\subsection{Action Spaces and Parameter Schemas}
\label{app:actionspaces}

The desktop and web instances in Section~\ref{sec:platformagents} expose
different tools through the same dispatcher. The tables below summarize
their model-facing contracts and the evaluator interfaces in
Section~\ref{sec:evaluator}. Fields with neither a default nor a question mark are required.
A question mark marks an optional field with no supplied value, and
\texttt{T[]} denotes an array. Empty arguments are written as \texttt{\{\}}.
These are tool signatures, independent of provider-specific call envelopes.
Runtime-only probes and environment-setup commands are outside the acting
agents' action spaces.

\subsubsection{Desktop Agent}
\label{app:desktopactions}

Desktop actions address the current screen and focused application.
Coordinates follow the run's declared coordinate policy, which maps the
model's image coordinates to device coordinates. A drag point has the
schema \texttt{\{x: number, y: number\}}. The \texttt{button} argument
accepts \texttt{left}, \texttt{right}, or \texttt{middle}.

\begingroup
\small\sffamily
\setlength{\tabcolsep}{4pt}
\renewcommand{\arraystretch}{1.18}
\begin{longtable}{@{}P{0.24\linewidth}P{0.42\linewidth}P{\dimexpr0.34\linewidth-4\tabcolsep\relax}@{}}
\caption{Desktop agent action space. Defaults and argument constraints are shown explicitly.}
\label{tab:desktopactions}\\
\toprule
\textbf{Tool} & \textbf{Parameter schema} & \textbf{Operation or constraint} \\
\midrule
\endfirsthead
\multicolumn{3}{l}{\textcolor{TiffanyDeep}{\tablename~\thetable\ (continued)}}\\
\toprule
\textbf{Tool} & \textbf{Parameter schema} & \textbf{Operation or constraint} \\
\midrule
\endhead
\midrule\multicolumn{3}{r}{Continued on next page}\\\endfoot
\bottomrule\endlastfoot
\path{click}
& \texttt{x, y: number}\newline\texttt{button: string = left}
& Click at the specified point. \\
\midrule[0.25pt]
\path{double_click}
& \texttt{x, y: number}\newline\texttt{button: string = left}
& Double click at the specified point. \\
\midrule[0.25pt]
\path{move}
& \texttt{x, y: number}
& Move the pointer without clicking. \\
\midrule[0.25pt]
\path{drag}
& \texttt{path: Point[]}\newline\texttt{button: string = left}
& Follow an ordered path containing at least two points. \\
\midrule[0.25pt]
\path{scroll}
& \texttt{x?, y?: number}\newline\path{delta_x}\texttt{: number = 0}\newline\path{delta_y}\texttt{: number = 0}
& Scroll at the pointer or a supplied location. Positive vertical deltas scroll up, negative deltas down. \\
\midrule[0.25pt]
\path{type}
& \texttt{text: string}\newline\path{target_kind}\texttt{: enum = auto}
& Enter literal text into the focused control. Target kind is \texttt{auto}, \texttt{editable}, or \texttt{terminal}. \\
\midrule[0.25pt]
\path{keypress}
& \texttt{key?: string}\newline\texttt{keys?: string[]}
& Supply exactly one field. Array entries are ordered key-press units, with \texttt{+} joining a chord. \\
\midrule[0.25pt]
\path{wait}
& \texttt{ms: integer = 1000}
& Pause for the given milliseconds, then obtain a fresh screenshot. \\
\midrule[0.25pt]
\path{screenshot}
& \path{all_screens}\texttt{: boolean = false}
& Capture the primary display, or the virtual desktop when enabled. \\
\midrule[0.25pt]
\path{note}
& \texttt{text: string}\newline\texttt{reason: string = ""}\newline\texttt{kind: enum = fact}
& Preserve working memory without GUI input. Text is at most 8,000 characters and reason at most 2,000. \\
\end{longtable}
\endgroup

The note categories are \texttt{fact}, \texttt{mistake}, \texttt{precaution},
\texttt{plan}, \texttt{constraint}, \texttt{decision}, and \texttt{other}.
These exposed names map to the working-memory categories described in
Appendix~\ref{app:runtimeconfig}. The same note fields apply to
\texttt{web\_note} below.

\subsubsection{Web Agent}
\label{app:webactions}

Web actions can use element references from the latest page observation
or viewport-relative CSS coordinates. In Table~\ref{tab:webactions},
\texttt{Ref} is a string or integer element label. References are refreshed
with page state, while \texttt{tab\_ref} identifies a browser tab.
Point actions use nonnegative CSS coordinates measured from the current
viewport's top-left corner. These differ from full-page or resized-image
coordinates.

\begingroup
\small\sffamily
\setlength{\tabcolsep}{4pt}
\renewcommand{\arraystretch}{1.18}
\begin{longtable}{@{}P{0.29\linewidth}P{0.39\linewidth}P{\dimexpr0.32\linewidth-4\tabcolsep\relax}@{}}
\caption{Web agent action space. Each row gives one tool's arguments and operation.}
\label{tab:webactions}\\
\toprule\textbf{Tool} & \textbf{Parameter schema} & \textbf{Operation or constraint} \\
\midrule\endfirsthead
\multicolumn{3}{l}{\textcolor{TiffanyDeep}{\tablename~\thetable\ (continued)}}\\
\toprule\textbf{Tool} & \textbf{Parameter schema} & \textbf{Operation or constraint} \\
\midrule\endhead
\midrule\multicolumn{3}{r}{Continued on next page}\\\endfoot
\bottomrule\endlastfoot
\path{web_open} & \texttt{url: string} & Navigate after URL-policy validation. \\
\midrule[0.25pt]
\path{web_snapshot}
& \texttt{\{\}} & Read the current page structure. \\
\midrule[0.25pt]
\path{web_get_url}
& \texttt{\{\}} & Read the active tab's URL. \\
\midrule[0.25pt]
\path{web_list_tabs}
& \texttt{\{\}} & List open browser tabs. \\
\midrule[0.25pt]
\path{web_click_ref}
& \texttt{ref: Ref} & Click the referenced element. \\
\midrule[0.25pt]
\path{web_hover_ref}
& \texttt{ref: Ref} & Hover over the referenced element. \\
\midrule[0.25pt]
\path{web_check_ref}
& \texttt{ref: Ref} & Set the referenced checkbox to checked. \\
\midrule[0.25pt]
\path{web_uncheck_ref}
& \texttt{ref: Ref} & Set the referenced checkbox to unchecked. \\
\midrule[0.25pt]
\path{web_get_text}
& \texttt{ref: Ref} & Read text from the referenced element. \\
\midrule[0.25pt]
\path{web_fill_ref}
& \texttt{ref: Ref}\newline\texttt{text: string}
& Replace the content of an editable element. \\
\midrule[0.25pt]
\path{web_select_option_ref}
& \texttt{ref: Ref}\newline\texttt{values: string[]}
& Select one or more nonempty HTML option values. \\
\midrule[0.25pt]
\path{web_drag_ref}
& \path{from_ref}\texttt{: Ref}\newline\path{to_ref}\texttt{: Ref}
& Drag between two referenced elements. \\
\midrule[0.25pt]
\path{web_click_point}
& \texttt{x, y: number}\newline\texttt{button: enum = left}
& Click at a viewport point. Button is \texttt{left}, \texttt{right}, or \texttt{middle}. \\
\midrule[0.25pt]
\path{web_double_click_point}
& \texttt{x, y: number}\newline\texttt{button: enum = left}
& Double click at a viewport point. Button is \texttt{left}, \texttt{right}, or \texttt{middle}. \\
\midrule[0.25pt]
\path{web_drag_point}
& \path{from_x}\texttt{, }\path{from_y}\texttt{: number}\newline\path{to_x}\texttt{, }\path{to_y}\texttt{: number}
& Drag between viewport points using the left button. \\
\midrule[0.25pt]
\path{web_type_text} & \texttt{text: string} & Type into the focused element. \\
\midrule[0.25pt]
\path{web_press} & \texttt{key: string} & Send a browser key or supported key expression to the focused element. \\
\midrule[0.25pt]
\path{web_scroll}
& \texttt{direction: enum}\newline\texttt{amount: integer = 600}
& Direction is \texttt{up}, \texttt{down}, \texttt{left}, or \texttt{right}. Amount is a positive CSS-pixel distance. \\
\midrule[0.25pt]
\path{web_wait}
& \path{duration_ms}\texttt{?: integer}\newline\path{load_state}\texttt{?: enum}\newline\texttt{text?: string}\newline\path{url_pattern}\texttt{?: string}\newline\texttt{selector?: string}
& Choose exactly one target. Duration is nonnegative. Load state is \texttt{load}, \texttt{domcontentloaded}, or \texttt{networkidle}. \\
\midrule[0.25pt]
\path{web_screenshot}
& \texttt{annotate: boolean = false}
& Capture the viewport. Annotation adds current Set-of-Mark labels. \\
\midrule[0.25pt]
\path{web_new_tab}
& \texttt{url?: string}\newline\texttt{label?: string}
& Open a tab, optionally with a URL and stable label. \\
\midrule[0.25pt]
\path{web_close_tab}
& \path{tab_ref}\texttt{?: string}
& Close the specified tab, or the active tab when omitted. \\
\midrule[0.25pt]
\path{web_focus_tab}
& \path{tab_ref}\texttt{: string}
& Focus the specified browser tab. \\
\midrule[0.25pt]
\path{web_note}
& Same fields as \path{note} in Table~\ref{tab:desktopactions}.
& Preserve nonempty text without interacting with the page. \\
\end{longtable}
\endgroup

\subsubsection{Evaluator}
\label{app:evaluatoractions}

The evaluator retrieves recorded evidence and, when authorized, inspects the
evaluated workspace through live tools. Table~\ref{tab:evaluatortools}
separates these interfaces. The active desktop or web profile selects the
applicable tools. Evidence references identify saved artifacts, independently
of the web agent's element references. Retrieval tools with
\texttt{detail\_level} use \texttt{standard} by default and accept
\texttt{debug} for expanded records.

\begingroup
\small\sffamily
\setlength{\tabcolsep}{4pt}
\renewcommand{\arraystretch}{1.18}
\begin{longtable}{@{}P{0.32\linewidth}P{0.37\linewidth}P{\dimexpr0.31\linewidth-4\tabcolsep\relax}@{}}
\caption{Evaluator inspection interfaces. Live tools require access to the evaluated workspace and the corresponding permissions.}
\label{tab:evaluatortools}\\
\toprule\textbf{Tool} & \textbf{Parameter schema} & \textbf{Purpose or scope} \\
\midrule\endfirsthead
\multicolumn{3}{l}{\textcolor{TiffanyDeep}{\tablename~\thetable\ (continued)}}\\
\toprule\textbf{Tool} & \textbf{Parameter schema} & \textbf{Purpose or scope} \\
\midrule\endhead
\midrule\multicolumn{3}{r}{Continued on next page}\\\endfoot
\bottomrule\endlastfoot
\multicolumn{3}{@{}l}{\textbf{Recorded evidence}}\\*
\path{read_round_bundle}
& \texttt{section?: string}\newline\path{detail_level}\texttt{: string = standard}
& Read a compact bundle or a selected section, such as trajectory or final state. \\
\midrule[0.25pt]
\path{read_step}
& \path{session_step}\texttt{: integer}\newline\path{detail_level}\texttt{: string = standard}
& Retrieve one recorded step. \\
\midrule[0.25pt]
\path{read_trace}
& \path{session_step}\texttt{: integer}\newline\path{detail_level}\texttt{: string = standard}
& Alias of \path{read_step}. \\
\midrule[0.25pt]
\path{query_artifacts} & \texttt{kind?: string} & List evidence, optionally by kind. \\
\midrule[0.25pt]
\path{query_runtime_events} & \texttt{query?: string} & Search runtime event summaries. \\
\midrule[0.25pt]
\path{query_tool_results}
& \path{tool_name}\texttt{?: string}\newline\path{session_step}\texttt{?: integer}\newline\path{detail_level}\texttt{: string = standard}
& Retrieve results by tool or step. \\
\midrule[0.25pt]
\path{query_a11y_snapshot}
& \path{evidence_ref}\texttt{?: string}\newline\path{session_step}\texttt{?: integer}\newline\path{detail_level}\texttt{: string = standard}
& Read saved desktop accessibility evidence. \\
\midrule[0.25pt]
\path{query_dom_snapshot}
& \path{evidence_ref}\texttt{?: string}\newline\path{session_step}\texttt{?: integer}\newline\path{detail_level}\texttt{: string = standard}
& Read saved web DOM evidence. \\
\midrule[0.25pt]
\path{query_web_page_state}
& \path{session_step}\texttt{: integer}\newline\path{detail_level}\texttt{: string = standard}
& Read a recorded URL, title, and page-state summary. \\
\midrule[0.25pt]
\path{query_som_annotation}
& \path{evidence_ref}\texttt{?: string}\newline\path{session_step}\texttt{?: integer}\newline\path{include_geometry}\texttt{: boolean = false}\newline\path{ref_filter}\texttt{?: string[]}\newline\path{detail_level}\texttt{: string = standard}
& Read saved Set-of-Mark metadata. In standard detail, geometry requires the flag and a reference filter. \\
\midrule[0.25pt]
\path{query_web_terminal_verification}
& \path{session_step}\texttt{?: integer}\newline\path{detail_level}\texttt{: string = standard}
& Retrieve terminal browser-verification evidence. \\
\midrule[0.25pt]
\path{compare_images}
& \path{left_ref}\texttt{: string}\newline\path{right_ref}\texttt{: string}
& Compare two evidence images using deterministic similarity metrics. \\
\midrule[0.25pt]
\path{extract_image_text_summary}
& \path{evidence_ref}\texttt{: string}
& Extract an OCR summary when the profile includes this tool. \\
\midrule[0.25pt]
\multicolumn{3}{@{}l}{\textbf{Authorized live inspection}}\\*
\path{client_screenshot}
& \path{all_screens}\texttt{: boolean = false}
& Capture the evaluated desktop. \\
\midrule[0.25pt]
\path{gui_inspect}
& \path{remove_empty}\texttt{: boolean = true}\newline\path{refresh_app_windows}\texttt{: boolean = true}
& Inspect desktop window targets. The refresh field is retained for compatibility. \\
\midrule[0.25pt]
\path{client_bash_readonly}
& \texttt{cmd: string}
& Run a nonempty command under the read-only command allowlist. \\
\midrule[0.25pt]
\path{client_browser_snapshot}
& \texttt{\{\}}
& Read the evaluated browser's page structure. \\
\midrule[0.25pt]
\path{client_browser_get_url}
& \texttt{\{\}}
& Read the evaluated browser's current URL. \\
\midrule[0.25pt]
\path{client_browser_tabs}
& \texttt{\{\}}
& List tabs in the evaluated browser session. \\
\midrule[0.25pt]
\path{client_browser_get_text}
& \texttt{ref: Ref}
& Read a current browser element. \\
\midrule[0.25pt]
\path{client_browser_screenshot}
& \texttt{annotate: boolean = false}
& Capture the evaluated browser viewport. \\
\end{longtable}
\endgroup

Authorized desktop GUI inspection reuses the schemas in
Table~\ref{tab:desktopactions} through \path{client_gui_screenshot},
\path{client_gui_click}, \path{client_gui_type},
\path{client_gui_keypress}, and \path{client_gui_wait}.
Additional browser operations use the matching \path{client_browser_}
wrappers and schemas from Table~\ref{tab:webactions}, excluding
\texttt{web\_note}. State-changing inspection requires explicit permission
and follows the attribution protocol in Appendix~\ref{app:evaluatorprotocol}.

Evaluation ends with \texttt{final\_report}, whose required fields appear in
Table~\ref{tab:evaluatorreport}. Optional structured fields carry
criterion scores, diagnoses, side effects, safety and efficiency assessments,
and evidence-linked divergence information. The web report additionally
records the final URL and title, verification flags, a blocker category,
and critical evidence references.

\begingroup
\small\sffamily
\setlength{\tabcolsep}{4pt}
\renewcommand{\arraystretch}{1.18}
\begin{longtable}{@{}P{0.34\linewidth}P{\dimexpr0.66\linewidth-2\tabcolsep\relax}@{}}
\caption{Required fields of the evaluator's \texttt{final\_report} argument object. Platform-specific and diagnostic fields extend this shared core.}
\label{tab:evaluatorreport}\\
\toprule\textbf{Field} & \textbf{Type and allowed values}\\\midrule\endfirsthead
\multicolumn{2}{l}{\textcolor{TiffanyDeep}{\tablename~\thetable\ (continued)}}\\
\toprule\textbf{Field} & \textbf{Type and allowed values}\\\midrule\endhead
\midrule\multicolumn{2}{r}{Continued on next page}\\\endfoot
\bottomrule\endlastfoot
\path{overall_verdict} & Enum \texttt{success}, \texttt{partial}, \texttt{failure}, or \texttt{uncertain}.\\
\midrule[0.25pt]
\path{outcome_label} & Enum \texttt{success}, \texttt{failure}, or \texttt{uncertain}.\\
\midrule[0.25pt]
\path{process_rating} & Integer from 1 to 5 rating execution quality.\\
\midrule[0.25pt]
\path{completion_rating} & Integer from 1 to 5 rating task completion.\\
\midrule[0.25pt]
\path{confidence_rating} & Integer from 1 to 5 rating confidence in the assessment.\\
\midrule[0.25pt]
\path{controllability} & Enum \texttt{controllable}, \texttt{uncontrollable}, \texttt{mixed}, or \texttt{unknown}.\\
\midrule[0.25pt]
\path{reason} & String giving the evidence-grounded rationale.\\
\end{longtable}
\endgroup

\subsection{Batch Collection and Resource Ownership}
\label{app:batchprotocol}

Task preparation binds \casespec to its inputs, initialization procedure,
application/tool profile, budgets, and optional post-run checks.
Placement considers configured evaluator requirements as well as actor
requirements. Resumed work uses the original task and configuration
identity. Supported context-managed settings are retained in batch-local
snapshots, while external dependencies keep their own version records.

Fresh single-use sandboxes isolate new collection attempts. Task-specific
fixtures also cover external accounts and service state. An optional
prepared queue, disabled by library default, performs initialization ahead
of execution. Reservations are task/attempt-specific and assigned once.
On-demand acquisition is a separate fallback. Prepared queues and warm
pools have different allocation and reuse rules. An externally owned
environment retains its own termination authority.

Creation, preparation, execution, submission, and status-query limits
govern different resources. Assignment checks the remaining provider
lifetime against the task timeout and release margin. Attempt identifiers
link scheduler state to controller state, and reconciliation checks an
existing attempt before starting another. A scheduler claim and a device
execution lease represent distinct ownership decisions.

Live evaluation and environment-resident output capture finish before
owned resources are released. Once a sandbox is destroyed, replay is limited
to retained artifacts. Unconfirmed cleanup remains unresolved and continues
to incur resource cost until destruction is confirmed. This lifecycle
distinguishes transport disconnection from environment termination.

\begin{table}[htbp]\centering\small
\caption{Shared execution interfaces with different collection and training
lifecycle owners.}
\label{tab:harness-learning-paths}
\begin{tabularx}{\linewidth}{P{31mm}XX}
\toprule
Responsibility & Supervised collection & Online RL \\
\midrule
Acting policy & Configured demonstration policy & Current policy being optimized \\
Scheduling & Standalone batch service & Training-side group orchestration \\
Environment lifetime & Owned sandboxes released after required checks and capture &
External orchestrator retains release authority \\
Learning output & Selected trajectories and role-specific demonstrations &
Attempt outcomes and captured calls for task groups \\
\bottomrule
\end{tabularx}

\end{table}

\subsection{Cohort Admission}
\label{app:acceptance}

Outcome-verified completion demonstrations require actor-only extraction,
valid action/context alignment, required artifacts, full-task acceptance,
and the cohort's quality checks.
Broad judge-filtered mid-training uses its separately declared selector,
and evaluator demonstrations are separate judgment-task samples
(Appendix~\ref{app:supervisedrecipe}). Online RL retains valid negative
outcomes within task groups.

\begin{table}[htbp]\centering\small
\caption{Selection distinguishes task outcomes from missing or invalid
evaluation and defines completion-demonstration eligibility.}
\begin{tabularx}{\linewidth}{P{49mm}X}
\toprule
Attempt evidence & Completion-demonstration disposition \\
\midrule
Verified complete with quality checks & Eligible. Retain source and outcome evidence. \\
Supported infeasibility or uncontrollable blocker & Excluded. Retain separately if used. \\
Partial or valid failed outcome & Excluded. Retain as failure/diagnostic experience if declared. \\
Uncertain verdict or conflicting evidence & Unresolved. Assessment is required. \\
Required evaluation error or missing outcome evidence & Not admitted as a verified completion \\
\bottomrule
\end{tabularx}

\end{table}

\FloatBarrier
\Needspace{11\baselineskip}
\section{Benchmark Evaluation Details}
\label{app:benchmarkprotocol}

\begin{samepage}
Section~\ref{sec:evalsetup} introduces the six benchmarks. This appendix
collects the metric definitions, task sets, and implementation settings
used to interpret the results, together with the sources of published
baseline scores.
\par\end{samepage}

\begin{samepage}
\paragraph{Comparable tasks and interfaces.}

Interactive evaluation uses the benchmark's scorer independently of the
harness evaluator used for collection. OSWorld-Verified uses the supported
361-task setting without the eight Google Drive tasks. Our OSWorld~2.0
binary-completion and partial scores both use the full 108-task set.
Grounding records identify the OSWorld-G instruction variant and
absent-target convention, and map crop/resize predictions back to
annotation coordinates.
\par\end{samepage}

WebArena-Verified retains the responses and network traces required by its
scorer. Our WebTestBench evaluation supplies the gold checklist directly,
following the benchmark's oracle setting. Agents return Pass/Fail judgments, with Fail
as the positive class. Precision, recall, and F1 measure defect detection.
The gold checklist fixes the inspection scope for defect detection.
We first compute precision, recall, and F1 from the checklist judgments
within each task, then average each metric over all evaluated tasks.
For task-level scores $P_t,R_t,F_{1,t}\in[0,1]$ across $N$ tasks,
the reported percentages are
\begin{equation}
\overline{P}=\frac{100}{N}\sum_{t=1}^{N}P_t,\qquad
\overline{R}=\frac{100}{N}\sum_{t=1}^{N}R_t,\qquad
\overline{F}_1=\frac{100}{N}\sum_{t=1}^{N}F_{1,t}.
\label{eq:webtest-macro}
\end{equation}
This task-level macro average gives equal weight to tasks with different
checklist lengths. It differs from pooling all checklist judgments into
one confusion matrix. In particular, $\overline{F}_1$ averages the
per-task F1 scores and generally differs from the harmonic mean of
$\overline{P}$ and $\overline{R}$. Table~\ref{tab:webtestresults} reports
all three metrics for each included model and omits models without
measured results.
The NorthCUA Nano, Mini, and Pro results all use final checkpoints after
joint desktop--web RL.

The source notes below record the task scope, tool permissions, budgets,
and scoring conventions of published baselines. Parameter columns distinguish
total from activated language-model counts. Within each NorthCUA tier,
stage comparisons use the declared benchmark tasks, tools, observations,
and execution budgets.

\paragraph{Repeated predictions.}
Table~\ref{tab:groundingrepeats} summarizes repeated predictions on the
same grounding examples, including run-to-run variation and pass@3
\citep{dror2018significance}. The accompanying task-level breakdowns explain
coverage and execution behavior.

\subsection{Grounding Scores and Repeated Sampling}

The direct SSP evaluation produces one coordinate prediction from a full
screenshot. Table~\ref{tab:sspdomains} breaks down the final-model evaluation records
by application group and target type. Every domain uses the same sample
set for Nano, Mini, and Pro. This makes the coverage differences visible without
changing the task distribution between models.

\begin{table}[t]\centering\small
\setlength{\tabcolsep}{4pt}
\caption{SSP accuracy (\%) from per-example evaluation records.
The application groups form one partition, and target type forms a second.}
\label{tab:sspdomains}
\begin{tabularx}{\linewidth}{X C{18mm} *{3}{C{28mm}}}
\toprule
SSP group & Examples & \name-Nano & \name-Mini & \name-Pro\\
\midrule
Office & 230 & 82.6 & 93.0 & 93.0\\
Creative & 341 & 71.3 & 90.6 & 91.5\\
Operating systems & 196 & 68.9 & 93.9 & 93.9\\
Development & 299 & 72.9 & 93.7 & 92.6\\
Scientific & 254 & 71.7 & 92.1 & 94.5\\
CAD & 261 & 70.5 & 92.0 & 95.0\\
\midrule
Text targets & 977 & 83.9 & 94.8 & 95.2\\
Icon targets & 604 & 55.0 & 88.6 & 90.2\\
\midrule
\rowcolor{NexLight}Overall & 1,581 & 72.9 & 92.4 & 93.3\\
\bottomrule
\end{tabularx}
\end{table}

We further stratify SSP by target size and type
(Table~\ref{tab:sspsizetype}). Target size is the smaller of the
annotated box's width and height,
$\min(x_2-x_1,y_2-y_1)$ in the original image coordinates. Fixed boundaries
at 16 and 32 pixels define the three exploratory size bins. Predictions
are paired across all three models by the exact image filename and query.
All 1,581 identities are unique and matched, with identical reference
boxes and target-type labels. The bins measure geometric extent in the
original screenshot before model preprocessing.

\begin{table}[t]\centering\small
\setlength{\tabcolsep}{4pt}
\caption{SSP accuracy (\%) by annotated target extent and type.
All six rows are disjoint and together cover the full SSP evaluation.}
\label{tab:sspsizetype}
\begin{tabularx}{\linewidth}{>{\centering\arraybackslash}X C{20mm} C{15mm} *{3}{C{26mm}}}
\toprule
Target size & Target type & Items & \name-Nano & \name-Mini & \name-Pro\\
\midrule
\multirow{2}{*}{$<16$ px} & Text & 31 & 58.1 & 87.1 & 90.3\\
& Icon & 57 & 50.9 & 84.2 & 93.0\\
\addlinespace[3pt]
\multirow{2}{*}{$16$--$<32$ px} & Text & 661 & 84.6 & 94.1 & 94.6\\
& Icon & 418 & 53.1 & 89.0 & 89.0\\
\addlinespace[3pt]
\multirow{2}{*}{$\geq32$ px} & Text & 285 & 85.3 & 97.2 & 97.2\\
& Icon & 129 & 62.8 & 89.2 & 93.0\\
\bottomrule
\end{tabularx}
\end{table}

\begin{samepage}
The paired Nano--Mini predictions show 1,086 items correct for both models, 375 correct
only for Mini, 66 correct only for Nano, and 54 missed by both. Mini locates
375 of Nano's 429 misses, but each model also has unique successes.
For Mini and Pro, 1,373 items are correct for both, 102 only for Pro,
88 only for Mini, and 18 for neither. Pro's net gain of 14 correct items
therefore coexists with distinct successes from each model.
The paired counts distinguish shared errors from checkpoint-specific successes.
\par\end{samepage}

\begin{samepage}
Pro's remaining 106 errors all produce valid coordinates outside the target
box. Icons account for 59 errors, including 46 in the 16--$<32$ pixel band.
That band contains 418 icon targets, of which 11.0\% are missed, and
contributes 43.4\% of Pro's errors overall. On icons smaller than 16 pixels,
Pro reaches 93.0\% accuracy. Across application groups, Pro gains most over Mini in CAD
(3.1 points) and scientific software (2.4 points), while development
software is 1.0 point lower.
\par\end{samepage}

OSWorld-G Refined uses temperature 0.7, top-$p$ 0.95, top-$k$ 40, and a
1,024-token output limit. Table~\ref{tab:groundingrepeats} reports the mean
and sample standard deviation over three complete runs, alongside the best
complete-run accuracy used in the main comparison. Pass@3 counts an item
as correct when at least one of its three predictions succeeds.

\begin{table}[t]\centering\small
\setlength{\tabcolsep}{4pt}
\caption{Grounding summaries (\%). OSWorld-G Refined SD is the sample standard
deviation across three complete runs.
SSP accuracies and subgroup results are recomputed from one-prediction
records for all three models. SSP pass@3 values use benchmark-level summaries.}
\label{tab:groundingrepeats}
\begin{tabularx}{\linewidth}{X C{18mm} C{18mm} C{29mm} C{19mm} C{19mm}}
\toprule
Model & \multicolumn{2}{c}{SSP} & \multicolumn{3}{c}{OSWorld-G}\\
\cmidrule(lr){2-3}\cmidrule(lr){4-6}
& Accuracy & Pass@3 & Mean $\pm$ SD & Best run & Pass@3\\
\midrule
\name-Nano & 72.9 & 74.6 & $82.5\pm0.77$ & 83.0 & 86.0\\
\rowcolor{NexLight}\name-Mini & 92.4 & 93.6 & $87.2\pm0.71$ & 87.9 & 90.6\\
\name-Pro & 93.3 & 95.6 & $90.2\pm0.80$ & 91.0 & 94.0 \\
\bottomrule
\end{tabularx}
\end{table}

On SSP, Pro raises single-run accuracy by 0.9 percentage points over
Mini and reaches 95.6\% pass@3, 2.3 points above its single-run result.

\begin{samepage}
The Refined records contain 564 unique item IDs in each run, with matched
annotations across all three models and all three draws. Of these, 510 have
visible targets, comprising 470 rectangular and 40 polygonal reference
regions. The other 54 require an absent-target refusal. Visible-target
accuracy uses the corresponding region test, and refusal accuracy uses
the benchmark's designated no-target output.
\par\end{samepage}

\begin{samepage}
All three tiers share the grounding system prompt, sampling settings, and
coordinate scoring. Nano and Mini use a grounding-specific click schema
that describes normalized coordinates and refusal, requires
\texttt{x}, \texttt{y}, and \texttt{button}, and enumerates button choices.
Pro uses the frozen NorthCUA desktop click schema, whose descriptions
refer to desktop interaction and whose button argument is optional.
The reported tier differences therefore describe these evaluated model
and interface configurations.
\par\end{samepage}

Table~\ref{tab:groundingstability} reports how often each item succeeds.
Nano's three complete runs contain 467, 460, and 468 correct predictions,
while Mini's contain 488, 496, and 492, and Pro's contain 504, 509, and 513.
All runs have zero request errors and unparsed predictions. Pro's second
run includes one separately completed missing item, yielding the same
564 unique IDs as the other runs. This completion contributes one prediction
for that item to run~2 and adds no extra draw to pass@3.

\begin{samepage}
Pro's 34 persistently missed items contribute 102 of its 166 incorrect
predictions (61.5\%). Its remaining 41 items vary across draws, while
489 succeed in all three. Mini and Pro share 17 persistent misses, and
16 items remain persistently incorrect across all three model tiers.
These counts show that Pro reduces the recurring error set while leaving
a smaller common subset unresolved. The 1,692 predictions are repeated
measurements of 564 items across 251 screenshots.
\par\end{samepage}

\begin{table}[t]\centering\small
\setlength{\tabcolsep}{4pt}
\caption{Per-item success counts across three OSWorld-G Refined draws.
Each row sums to 564 items, including absent-target cases.}
\label{tab:groundingstability}
\begin{tabularx}{\linewidth}{X *{4}{C{24mm}}}
\toprule
Model & 0/3 correct & 1/3 correct & 2/3 correct & 3/3 correct\\
\midrule
\name-Nano & 79 & 20 & 20 & 445\\
\rowcolor{NexLight}\name-Mini & 53 & 15 & 27 & 469\\
\name-Pro & 34 & 23 & 18 & 489\\
\bottomrule
\end{tabularx}
\end{table}

The improvement also spans reference geometries
(Table~\ref{tab:osggeometry}). On rectangular targets, Pro averages
91.1\%, compared with Mini's 87.9\%. Polygonal regions are already
high-performing for Mini at 97.5\%, and Pro reaches 98.3\%.
The geometry labels identify the benchmark's scoring regions, not
separate semantic task categories. Together with the absent-target
analysis in the main text, they identify refusal decisions as a larger
remaining gap than polygonal targeting.

\begin{table}[t]\centering\small
\setlength{\tabcolsep}{4pt}
\caption{Mean OSWorld-G accuracy (\%) across three draws by reference
geometry. These disjoint groups cover all 564 items.}
\label{tab:osggeometry}
\begin{tabularx}{\linewidth}{X C{16mm} *{3}{C{28mm}}}
\toprule
Reference region & Items & \name-Nano & \name-Mini & \name-Pro\\
\midrule
Rectangle & 470 & 83.1 & 87.9 & 91.1\\
Polygon & 40 & 95.0 & 97.5 & 98.3\\
Absent target & 54 & 67.9 & 74.1 & 75.9\\
\bottomrule
\end{tabularx}
\end{table}

\subsection{Interactive Results and Trajectory Breakdowns}

\begin{samepage}
The main WebArena-Verified table reports final RL checkpoints at 67.2\%
for Nano, 70.4\% for Mini, and 73.9\% for Pro. Pro's score comes from
one complete evaluation of all 812 tasks.
The trajectory breakdowns below cover Nano and Mini after SFT and Pro
after RL, with 521, 552, and 600 successful tasks, respectively. Mini uses
the second evaluation round, matching Figure~\ref{fig:webbreakdown}.
\par\end{samepage}

The earlier SFT evaluations additionally report pass@3 of 69.3\% for
Nano and 73.9\% for Mini. Figure~\ref{fig:webbreakdown} includes Mini's SFT
website-level breakdown alongside Nano SFT and Pro RL.

\begin{samepage}
Table~\ref{tab:webgroups} groups the 812 tasks by website scope and
requested outcome. Website groups use the task's site set, pooling
multi-site combinations into one group. Outcome types follow the
evaluator's expected category. Each grouping therefore partitions the
same task set and reveals a different aspect of task completion.
\par\end{samepage}

\begin{table}[t]\centering\small
\setlength{\tabcolsep}{4pt}
\caption{WebArena-Verified task counts for Nano SFT, Mini SFT, and Pro RL.
Website scope and expected task type each partition the same 812 tasks.
Forum denotes Reddit, and state change denotes the evaluator's
\texttt{MUTATE} category.}
\label{tab:webgroups}
\renewcommand{\arraystretch}{1.24}
\begin{tabularx}{\linewidth}{X C{14mm} *{3}{C{27mm}}}
\toprule
Task group & Tasks & \makecell{NorthCUA-Nano\\passes} & \makecell{NorthCUA-Mini\\passes} & \makecell{NorthCUA-Pro\\passes}\\
\midrule
\rowcolor{NexLight}\multicolumn{5}{l}{\textbf{Website scope}}\\
GitLab only & 180 & 115 & 126 & 131\\
Map only & 109 & 67 & 71 & 77\\
Forum only & 106 & 87 & 85 & 89\\
Shopping only & 187 & 123 & 123 & 135\\
Store administration only & 182 & 122 & 135 & 151\\
Cross-site & 48 & 7 & 12 & 17\\
\addlinespace[3pt]
\rowcolor{NexLight}\multicolumn{5}{l}{\textbf{Expected task type}}\\
State change & 374 & 240 & 253 & 274\\
Retrieval & 325 & 204 & 223 & 245\\
Navigation & 113 & 77 & 76 & 81\\
\midrule
Overall & 812 & 521 & 552 & 600\\
\bottomrule
\end{tabularx}
\end{table}
\FloatBarrier

Mini's additional completions over Nano concentrate in state-changing
and retrieval tasks, with 13 and 19 more successes, respectively, while
their navigation counts are nearly identical. Pro reaches 75.4\% on
retrieval, 73.3\% on state change, and 71.7\% on navigation.
Cross-site work is more demanding for all three checkpoints, with
7, 12, and 17 successes out of 48 tasks. In comparison, Pro completes
76.3\% of the 764 single-site tasks.

Mini's records show how this difference relates to execution length within
one checkpoint. It completes 540 of 764 single-site tasks (70.7\%) and
12 of 48 cross-site tasks (25.0\%). Their median lengths are 10 and
27.5 decisions, respectively. Even among successful tasks, the cross-site
median is 17 decisions, compared with nine for single-site work. These
tasks combine lower completion with longer interactions, making them a
useful focus for experience that trains the agent to carry constraints
and intermediate results between websites.

\begin{samepage}
Table~\ref{tab:weblengths} summarizes decision counts by evaluator outcome.
Each record contributes its current assistant response once, including
the final response. Nano has 811 traced tasks and one untraced failure,
while Mini and Pro each have 812. All 812 tasks enter each model's
success-rate denominator. Quartiles use linear interpolation at
$(n-1)p$ for percentile fraction $p$.
\par\end{samepage}

\begin{samepage}
Successful tasks have similar median lengths of 9, 10, and 10 decisions
for Nano, Mini, and Pro. Failed tasks have longer medians of 19, 18, and
15.5 decisions and contribute 58.5\%, 50.6\%, and 38.4\% of their
recorded decisions, respectively. For Mini, this is 7,698 of 15,224
decisions across 260 failures. Separating outcomes thus exposes the
disproportionate execution devoted to unsuccessful tasks. At the same
time, Pro completes 24 tasks with at least 50 decisions, including a
93-decision success, showing that extended execution can also support
successful completion.
\par\end{samepage}

\clearpage
\begin{table}[t]\centering\small
\setlength{\tabcolsep}{4pt}
\caption{Decision records by task outcome for Nano SFT, Mini SFT, and Pro RL.
Counts include terminal responses. Nano uses the archived record boundary,
which includes interrupted execution, while Mini and Pro use one
retained attempt per task.}
\label{tab:weblengths}
\renewcommand{\arraystretch}{1.24}
\begin{tabularx}{\linewidth}{X C{22mm} C{17mm} C{19mm} C{19mm} C{25mm}}
\toprule
Model & Outcome & Traces & Mean & Median & Q25--Q75\\
\midrule
\multirow{2}{*}{\name-Nano} & Passed & 521 & 13.4 & 9 & 7--16\\
& Failed & 290 & 33.9 & 19 & 10--44\\
\addlinespace[3pt]
\multirow{2}{*}{\name-Mini} & Passed & 552 & 13.6 & 10 & 7--16\\
& Failed & 260 & 29.6 & 18 & 10--36.3\\
\addlinespace[3pt]
\multirow{2}{*}{\name-Pro} & Passed & 600 & 15.0 & 10 & 7--17.3\\
& Failed & 212 & 26.5 & 15.5 & 10--33.3\\
\bottomrule
\end{tabularx}
\end{table}

Mini likewise completes 13 tasks with at least 50 decisions, with its
longest success taking 75. Across all three checkpoints, successful-task
means exceed their medians, reflecting a tail of extended but productive
execution. This profile favors training examples that retain successful
long trajectories alongside concise workflows. In our learning pipeline,
outcome-based selection retains successful long workflows for supervision,
while RL also uses unsuccessful attempts to provide negative feedback.

The browser case in Figure~\ref{fig:eval-web} is task~1 in the Pro export.
Its 13 decisions include an unavailable-option error, inspection of the
actual dropdown choices, and successful completion after revising the
report settings. The screenshots show how inspecting the available
choices resolves the error and lets the agent finish the report query.

The desktop checkpoint summaries cover 361 OSWorld-Verified tasks with a
100-step limit and the full 108-task OSWorld~2.0 set with a 500-step limit.
The OSWorld-Verified main table uses the RL checkpoints at 78.9\%,
85.3\%, and 88.6\% for Nano, Mini, and Pro. Tables~\ref{tab:stages}
and~\ref{tab:rlresults} retain the earlier checkpoints for stage comparisons.
On OSWorld~2.0, the final RL checkpoints have Binary/Partial scores of
13.0/45.6 for Nano, 18.5/50.1 for
Mini, and 30.6/66.8 for Pro. The Mid pairs are 0.9/24.2, 2.8/25.6,
and 9.3/30.6, respectively. The long-horizon SFT checkpoints score
12.0/40.9, 18.5/46.3, and 27.3/63.6 in the same tier order.
These SFT checkpoints serve as the Before-RL baselines in
Table~\ref{tab:rlresults}.
The binary and partial scores describe full completion and intermediate
progress on the same tasks.

\subsection{External Comparison Settings}

Table~\ref{tab:externalsettings} summarizes the evaluation settings of
published specialist and general-purpose models. Scores retain their
reported task scopes, metrics, and aggregation rules.

\paragraph{Additional OSWorld-Verified references.}
Table~\ref{tab:desktopresults} extends the comparison across model
generations and specialist families. OpenCUA-72B's 45.0\% is the
100-step, three-run mean in its final conference paper
\citep{wang2025opencua}. GUI-Owl-1.5 uses the 8B and 32B Instruct
checkpoints \citep{xu2026mobileagent35}, while UI-Venus-2 contributes
its 9B and 27B computer-use models \citep{venusteam2026uivenus2}.
These are distinct from UI-Venus-1.5 in the grounding comparison.
MiniMax M3's 75.2\% and Seed 2.1 Pro's 78.8\% follow the
Qwen-UI-Agent comparison. Qwen3.8-Max's 86.1\% and GPT-5.6 Sol's
83.2\% follow the Qwen3.8
release \citep{zhou2026qwenuiagent,qwen2026qwen38}.
The Nex-N2.5 release adds 76.7\% for DeepSeek-V4-Flash-Vision on
OSWorld-Verified. It also reports WebArena-Verified success of 71.6\%
for Kimi-K3 and 62.3\% for DeepSeek-V4-Flash-Vision
\citep{nexagi2026nexn25pro}. We retain the release's DeepSeek model label.

The benchmark authors separately report 86.0\% for Claude Fable 5 [1m]
on 358 scored tasks and 83.4\% for Claude Opus 5 [1m] on 360, both
with 100 steps \citep{xlang2026osworldverifiedresults}.
Google's Gemini 3.6 Flash card reports 83.0\%, using a five-run mean with 100 steps
\citep{deepmind2026gemini36flash}.

GLM-5.3-Flash's 84.7\% is an independent public evaluation by
\citet{hojuna2026glm53osworld}, with released code and trajectories.
It reports 305.0 earned credits over 360 tasks after excluding one
grader failure from the non-Google-Drive set. The run uses screenshot
observations, PyAutoGUI, a 100-step budget, and no task retries.
The main-table citation identifies the model, while the source above
identifies this independent evaluation.

Our DeepSeek-V4.1-Flash evaluation yields 71.7\% on the 361-task
OSWorld-Verified setting with a 100-step limit. Its starred entry in
Table~\ref{tab:desktopresults} is separate from the published
DeepSeek-V4-Flash-Vision result.

\paragraph{OSWorld~2.0 task scopes.}
Our GLM-5.3-Flash evaluation uses the full 108-task set and a 500-step
budget, yielding 14.8\% binary completion and 48.7\% partial credit.
The star in Table~\ref{tab:osworld2results} identifies this measurement
as our evaluation, with the model citation identifying the tested system.
For GPT-5.6 Sol and Claude Opus 5, Table~\ref{tab:osworld2results} uses the
official full 108-task v2026.08.08 results, evaluated with batched
actions, maximum effort, and 500 steps \citep{xlang2026osworld2leaderboard}.
GPT-5.6 Sol averages 27.3\% binary and 62.7\% partial over two runs.
Claude Opus 5 averages 31.4\% and 68.3\% over seven runs. These references
match our full-set task scope, while their execution settings and aggregation
remain source-specific. OpenAI's 65.7\% GPT-5.6 Sol result uses the 82-task
offline subset \citep{openai2026gpt6astra}. MiniMax's release comparison
reports 70.6\% for Claude Opus 5 without identifying its task subset
\citep{minimax2026m31flash}. Both alternative scores are omitted from the
main table in favor of the full-set pairs. The earlier GPT-5.5 and
Claude Opus 4.8 pairs use the official June full-set release, also at
500 steps with batched actions. UI-Venus-2 uses 108 tasks with a
shorter 150-step budget \citep{venusteam2026uivenus2}.

Claude Fable 5 and Fable 5.1 report binary/partial pairs of
36.1\%/72.9\% and 41.7\%/77.9\%, respectively, on 108 tasks
\citep{anthropic2026fable51systemcard}. Binary corresponds to the source's
strict pass rate. Both are five-run means at maximum effort with a 500-step
budget and 1080p observations. They use the August 2026 task release with
subsequent task, setup, and grading fixes, so their scores differ in
protocol from earlier releases. Safeguard interventions count as failures.

MiniMax's comparison also reports partial scores of 61.7\% for MiniMax
M3.1 Flash, 54.3\% for DeepSeek-V4.1-Flash, and 58.3\% for Kimi-K3
\citep{minimax2026m31flash}. These remain additional published references,
with task versions and subsets unspecified. Corresponding binary results
are unavailable for these reported settings and remain dashes. We retain
DeepSeek-V4.1-Flash without transferring its score to the later 1017 variant.
DeepSeek's size entry uses 552B backbone parameters and 16B activated
during decoding. Its model card separately reports 8B activated during
prefill and 196B of additional Engram memory
\citep{deepseek2026v41flashcard}.

Other reports provide both metrics without an explicit task count.
Qwen3.8-Max reports 19.4\% binary and 46.7\% partial, while Qwen-CUA,
Qwen-CUA-Max, and Qwen-UI-Agent retain their published pairs
\citep{qwen2026qwen38,lu2026qwencua,zhou2026qwenuiagent}.
Gemini 3.8 Flash's 59.0\% partial score is the maximum of three
500-step benchmark runs on the pre-August-patch setting
\citep{deepmind2026gemini38flash}.

The later Opus 5.5 system card reports 48.7\% binary and 81.8\% partial
for Opus 5.5, alongside 42.8\%/80.7\% for Fable 5.1
\citep{anthropic2026opus55systemcard}. These five-run means use 108 tasks
and a 500-step budget, but update the task assets to September 10, 2026
and retain screenshots until server-side context compaction. The release
announcement labels this revision OSWorld~2.1 \citep{anthropic2026opus55}.
We keep these later-revision results separate from the OSWorld~2.0 main
table because both the task files and harness configuration changed.

\paragraph{Grounding and browser references.}
For grounding, the dagger-marked general-purpose rows in
Table~\ref{tab:mainresults} are measurements by the Qwen-UI-Agent authors.
Its SSP table separates direct inference from zoom-assisted results
\citep{zhou2026qwenuiagent}.
GUI-Owl-1.5 uses the Instruct checkpoints, and UI-Venus-1.5 uses the final
unified models. We take their refined OSWorld-G results from
the respective model papers \citep{xu2026mobileagent35,venusteam2026uivenus15}.
MAI-UI's refined score is its 73.9\% no-zoom result, with the 75.0\%
zoom-assisted result excluded \citep{zhou2025maiui}. Claude Opus 4.8's
82.3\% SSP result is a five-run mean using adaptive thinking at maximum
effort without Python tools \citep{anthropic2026opus48systemcard}.
GPT-5.6 Sol's 76.9\% SSP accuracy is OpenAI's reported no-tools result
\citep{openai2026gpt6astra}. Its 77.7\% OSWorld-G score comes from
the Nex-N2.5 comparison \citep{nexagi2026nexn25pro}.
We confirmed that this result uses Refined instructions.
Qwen3.8-Max's 84.5\% SSP score follows its August release report
\citep{qwen2026qwen38}, while its 71.6\% OSWorld-G score comes from
our Refined evaluation. Its star in Table~\ref{tab:mainresults} therefore
applies only to OSWorld-G. The published references retain each source's
inference settings and aggregation.
Nex-N2.5-mini and Pro contribute reported Refined OSWorld-G accuracies of
82.9\% and 87.4\%, respectively
\citep{nexagi2026nexn25mini,nexagi2026nexn25pro}.

The other starred rows in Table~\ref{tab:mainresults} contain our single-run
Refined evaluations of Claude Opus 5.5, Claude Fable 5.1, Kimi-K3,
GPT-6 Astra, GPT-6.1 Sol,
DeepSeek-V4.1-Flash, GLM-5.3-Flash, and Gemini 4 Argon.
Their model citations identify the evaluated systems,
while the stars identify the measurement source.
The stars also identify our SSP evaluations of these same models.

\begin{table}[t]\centering\small
\caption{Evaluation settings of published comparison models.}
\label{tab:externalsettings}
\begin{tabularx}{\linewidth}{P{35mm}X}
\toprule
Reference & Comparison setting\\
\midrule
Qwen-UI-Agent \citep{zhou2026qwenuiagent}
& SSP uses the no-zoom row. OSWorld-G uses refined instructions. Desktop
evaluation allows CLI and batched actions, with 361 OSWorld-Verified tasks.
The source table labels 79.5 as success rate, while its text describes
partial progress.\\
\addlinespace[4pt]
Qwen-CUA / Qwen-CUA-Max \citep{lu2026qwencua}
& The reported 86.2 is earned credit on 360 tasks with a 100-step limit.
Max reports 87.6 under the same protocol. Their native interface uses
screenshots and keyboard/mouse actions. Max discloses over 1T total
parameters without an activated count.\\
\addlinespace[4pt]
MAI-Gym-9B \citep{tong2026maigym}
& GUI and CLI actions. The paper lists 361 OSWorld-Verified tasks with
100 steps and 108 OSWorld~2.0 tasks with 500 steps. It evaluates three
trials per task, without a complete aggregation rule in the main table.\\
\addlinespace[4pt]
ScaleCUA-Qwen3.5-9B \citep{lv2026scalecua}
& Four-run mean with a 50-step budget. The official reproduction setup
uses the 361-task non-Google-Drive list. The original paper calls the
benchmark OSWorld.\\
\addlinespace[4pt]
Nex-N2.5-mini / Pro \citep{nexagi2026nexn25mini,nexagi2026nexn25pro}
& These general-purpose models have 35B and 397B total parameters, respectively.
Provider-reported computer/browser scores use the NorthCUA harness family.
The release reports OSWorld~2.0 partial credit of 30.5/56.4 and
WebArena-Verified success of 63.4/67.6 for mini/Pro. Corresponding binary
scores are unavailable.\\
\addlinespace[4pt]
MiniMax M3 \citep{minimax2026m3card}
& Benchmark-author OSWorld~2.0 evaluation on 108 tasks with 500 steps,
single-action calls, and thinking enabled \citep{yuan2026osworld2}.\\
\bottomrule
\end{tabularx}
\end{table}

\begin{samepage}
For browser specialists, Qwen-UI-Agent reports original-WebArena success
of 73.6\% for 27B and 69.2\% for 35B-A3B, with author corrections to some
reference answers and evaluators \citep{zhou2026qwenuiagent}. Qwen-CUA
reports 64.2\% on WebArena \citep{lu2026qwencua}. These results use the
original WebArena setting.
Likewise, OpenGCUA evaluates an 82-task offline OSWorld~2.0 subset and
varies the harness around existing models \citep{zhou2026opengcua}.
Its comparison measures the effects of harness design around those models.
\par\end{samepage}

\begin{samepage}
In Table~\ref{tab:webmainresults}, the Qwen3.7-Plus, Qwen3.8-Max,
Claude Opus 4.8, Claude Fable 5, Gemini 3.1 Pro, and GPT-5.6 Sol scores
come from the August Qwen3.8-Max release report \citep{qwen2026qwen38}.
It uses WebArena's official grader within
the OSWorld scaffold. Model variants and scores follow that release,
including Claude Fable 5 and GPT-5.6 Sol.
Qwen3.8-Max's 2.4T total and 95B active parameters describe its language-model
backbone \citep{qwen2026qwen38card}. The evaluated hosted model supports
vision, while the downloadable Qwen3.8-2.4T-A95B checkpoint is text-only.
\par\end{samepage}

The starred GPT-6 Astra, GPT-6.1 Sol, Claude Opus 5.5, Gemini 4 Argon,
GLM-5.3-Flash, and DeepSeek-V4.1-Flash
entries are our WebArena-Verified evaluations. Their citations identify
the evaluated models, while the stars distinguish these measurements from
the published reference scores. The main table reports task success,
with stage-specific trace accounting above. The Mid-to-SFT
comparison appears in Section~\ref{sec:stageexperiments}.

\subsection{Web Repair Evaluation Protocol}
\label{app:webassignment}

The recorded case in Figure~\ref{fig:webtestcase} illustrates browser feedback
and code repair. A controlled repair comparison additionally fixes the issue
set, starting patch, Coding Agent, test generator, adjudicator, budgets, and
protected acceptance conditions. Role-specific information remains separate
even when roles share model weights.

Each assigned issue begins with the nominated defect reproduced.
Before a new condition or patch is tested, browser storage, account/backend
data, service-worker state, and external fixtures are restored. Intentional
stateful test dependencies remain explicit. Reports bind the build,
reset state, test item, and attempt. Only evidence-supported product
discrepancies justify a repair.

Acceptance uses the common assigned issue set as its denominator, retaining
tester noncompletion. Infrastructure-invalid attempts follow the same
bounded rerun rule, and unresolved issues remain in the results.
Final acceptance runs the submitted patch from a protected clean state
outside the repair loop, with its outputs withheld from further revision.
\FloatBarrier
\clearpage
\section{OSWorld 2.0 Execution Cases}
\label{app:os2cases}

These cases follow NorthCUA through selected decisions in complete desktop
tasks. Counts refer to model decisions, each of which can issue multiple
tools. Agent callouts summarize recorded decisions for readability.

The pair in Figure~\ref{fig:os2sftcase} has identical requests but different
dataset revisions. The earlier run records unavailable OCR, while the later
run uses it. The case follows source selection and verification in each
execution.

\subsection{Preserving a Delivery Constraint Across Long Contexts}
\label{app:os2context}

\begin{figure}[!htbp]
\centering
\includegraphics[width=\linewidth]{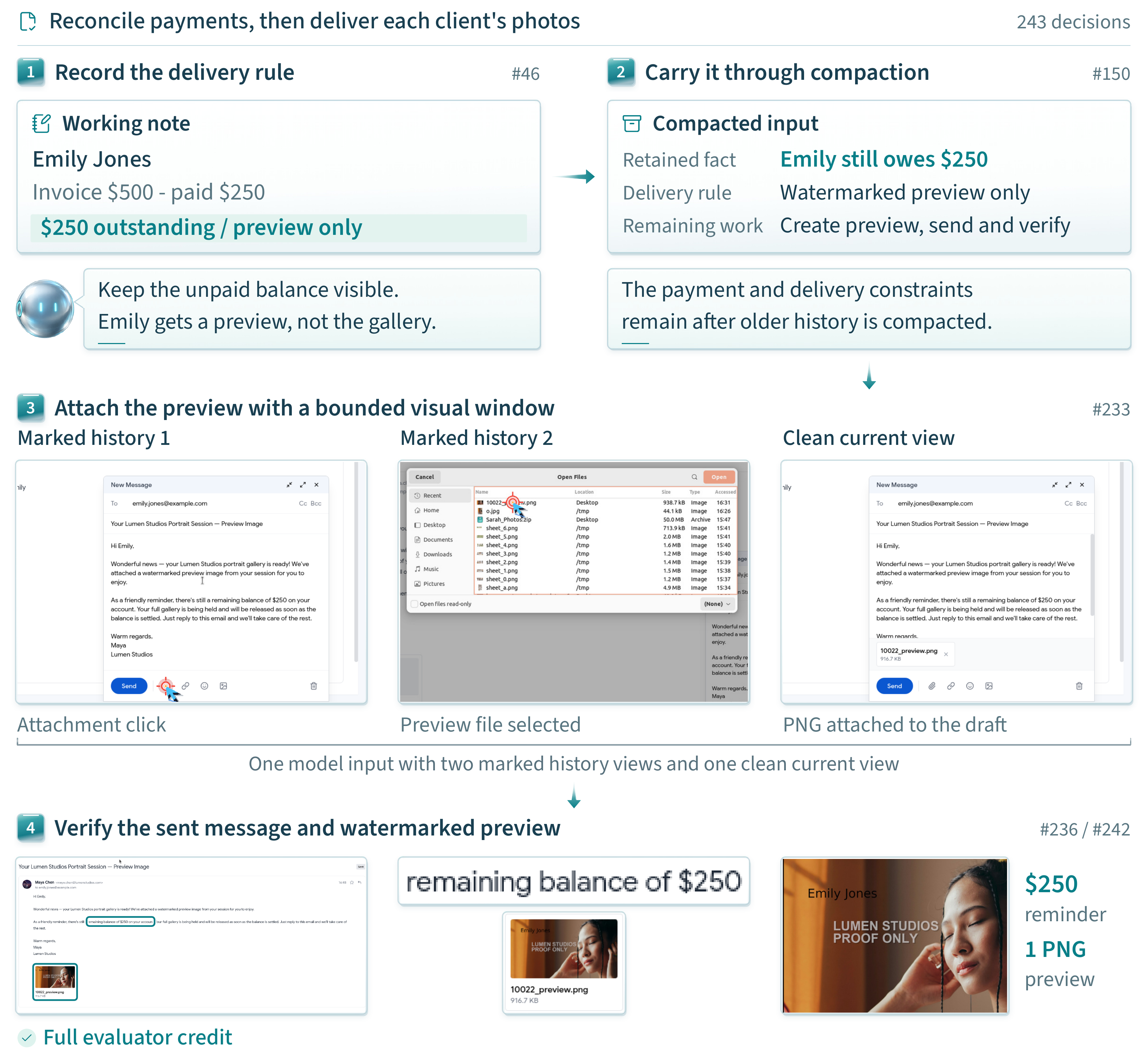}
\caption{A payment constraint retained through a long-context task.
Stages follow note 46, compacted input 150, the three image inputs at
decision 233, and delivery checks. Historical action marks are preserved.
Text artifacts are abridged and the agent callout summarizes a recorded
decision. Screenshot crops precede the indexed decisions, with enlarged details.}
\label{fig:os2contextcase}
\end{figure}

The photo-delivery task in Figure~\ref{fig:os2contextcase} requires the agent
to reconcile invoices against bank payments, identify two clients' photos,
and send different deliveries. Sarah has paid her \$800 invoice in full and
should receive her complete gallery. Emily has paid \$250 of a \$500 invoice
and should receive a watermarked solo-indoor preview with a balance reminder.
The difficulty is carrying this distinction through image preparation and
email composition after the original banking interaction leaves recent
context.

After exact-name payment matching, the agent records Emily's outstanding
\$250 and preview-only rule at decision 46. The compacted input at decision
150 retains both and the unfinished delivery checks, carrying the payment
constraint beyond the earlier interaction history.

At decision 233, two action-marked historical screenshots identify the
attachment click and selected PNG, while the clean current frame shows the
file attached to Emily's draft. The agent sends the message and checks
its sent copy at decision 236, confirming the preview and \$250 reminder.
A final image inspection confirms the watermark, and the 243-decision
attempt receives full credit. Notes and compaction preserve the constraint,
while the bounded image window exposes the current attachment state.

\FloatBarrier
\subsection{Using a Terminal Through the GUI}
\label{app:os2terminal}

\begin{figure}[!htbp]
\centering
\includegraphics[width=\linewidth]{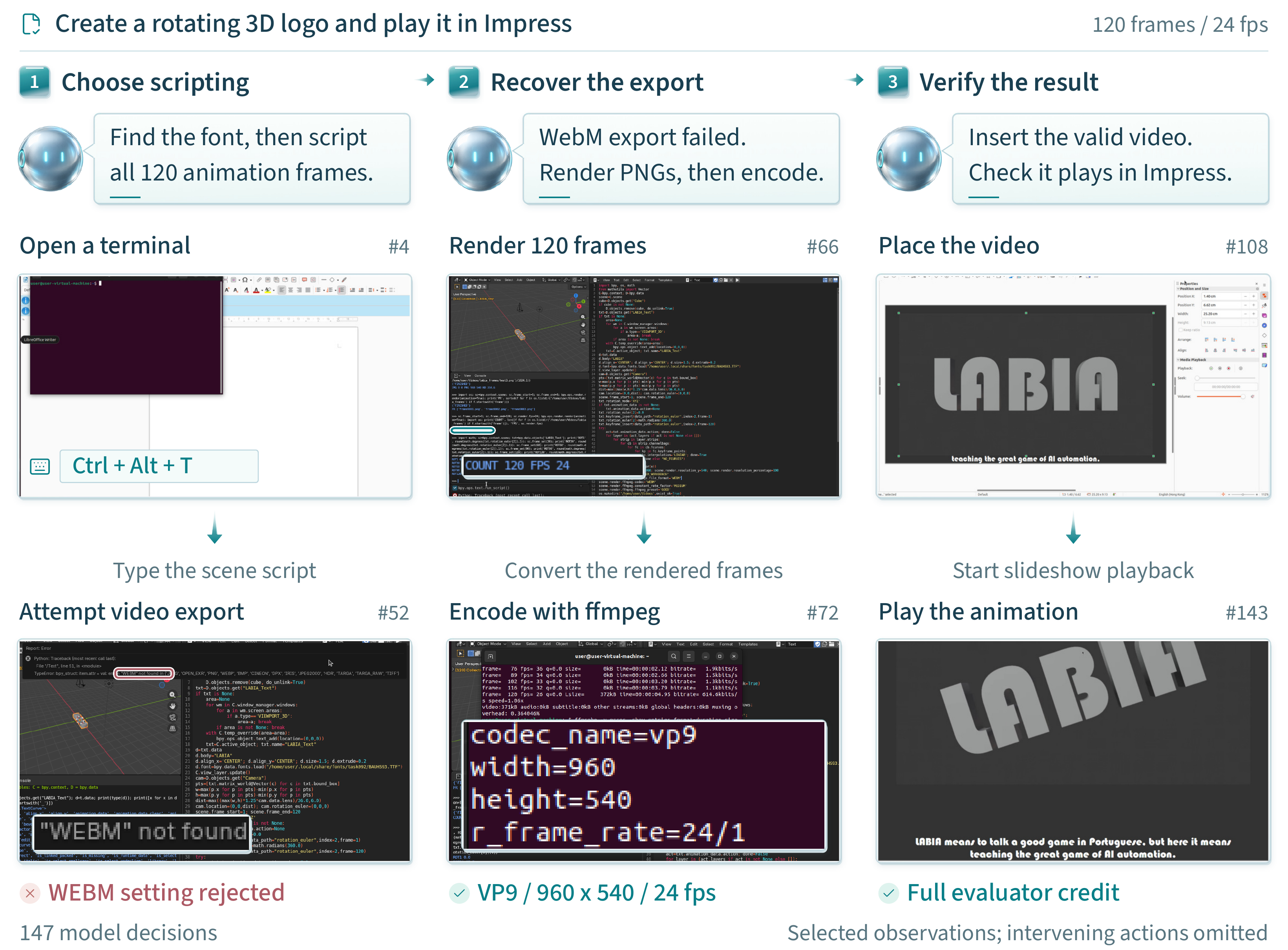}
\caption{GUI-mediated scripting and recovery in an animation task.
Each numbered phase reads top to bottom. Callouts paraphrase recorded
decisions. Screenshots show cropped pre-decision observations with enlarged
diagnostic details. Intermediate interactions are omitted.}
\label{fig:os2terminalcase}
\end{figure}

Figure~\ref{fig:os2terminalcase} follows a 3D course-logo task requiring
Bauhaus 93 text, a 120-frame rotation, and a 960\,$\times$\,540 WebM video
at 24 fps inside Impress. To control scene settings and automate repeated
rendering, the agent opens a terminal through GUI keypresses and types
commands. Its exposed tools are desktop actions and notes, and the task
permits terminal and application-console use.

After opening a terminal at decision 3 to identify fonts, the agent builds
the scene in Blender's scripting workspace. When direct video export
rejects \texttt{WEBM} and \texttt{FFMPEG} settings, it renders 120 PNG frames
and encodes them with terminal-based \texttt{ffmpeg} using VP9 at 24 fps,
preserving the 960\,$\times$\,540 resolution. Separating rendering from
encoding lets it complete the repeated frame processing programmatically.

After checking video properties, the agent positions the video in Impress
and starts the slideshow. Observations 143 and 144 show changing logo
orientations. The task receives full credit after 147 decisions and 253
tool calls, including four notes. GUI-accessible scripting handles the
repetitive work, and GUI playback verifies the result.
\FloatBarrier

\end{document}